\pdfoutput=1
\documentclass[sigconf]{acmart}

\usepackage{tikz}
\usepackage{pgfplots}
\usepackage{graphicx}
\usetikzlibrary{pgfplots.groupplots}

\def\BACSDATA{
(2005.019,30)
(2005.038,30)
(2005.058,31)
(2005.077,43)
(2005.096,53)
(2005.115,77)
(2005.135,54)
(2005.154,64)
(2005.173,57)
(2005.192,129)
(2005.212,81)
(2005.231,51)
(2005.25,98)
(2005.269,59)
(2005.288,84)
(2005.308,62)
(2005.327,120)
(2005.346,81)
(2005.365,103)
(2005.385,86)
(2005.404,125)
(2005.423,145)
(2005.442,129)
(2005.462,94)
(2005.481,94)
(2005.5,153)
(2005.519,38)
(2005.538,67)
(2005.558,11)
(2005.577,7)
(2005.596,21)
(2005.615,6)
(2005.635,1)
(2005.654,0)
(2005.673,2)
(2005.692,9)
(2005.712,1)
(2005.731,6)
(2005.75,3)
(2005.769,6)
(2005.788,5)
(2005.808,15)
(2005.827,26)
(2005.846,33)
(2005.865,35)
(2005.885,58)
(2005.904,36)
(2005.923,65)
(2005.942,40)
(2005.962,77)
(2005.981,44)
(2006.0,21)
(2006.019,60)
(2006.038,46)
(2006.058,29)
(2006.077,41)
(2006.096,31)
(2006.115,42)
(2006.135,70)
(2006.154,48)
(2006.173,52)
(2006.192,53)
(2006.212,59)
(2006.231,109)
(2006.25,69)
(2006.269,97)
(2006.288,55)
(2006.308,75)
(2006.327,97)
(2006.346,86)
(2006.365,59)
(2006.385,49)
(2006.404,34)
(2006.423,118)
(2006.442,55)
(2006.462,80)
(2006.481,65)
(2006.5,56)
(2006.519,25)
(2006.538,25)
(2006.558,16)
(2006.577,7)
(2006.596,3)
(2006.615,1)
(2006.635,4)
(2006.654,0)
(2006.673,0)
(2006.692,2)
(2006.712,1)
(2006.731,0)
(2006.75,4)
(2006.769,3)
(2006.788,2)
(2006.808,9)
(2006.827,10)
(2006.846,18)
(2006.865,19)
(2006.885,39)
(2006.904,14)
(2006.923,26)
(2006.942,28)
(2006.962,10)
(2006.981,44)
(2007.0,24)
(2007.019,24)
(2007.038,56)
(2007.058,46)
(2007.077,46)
(2007.096,55)
(2007.115,39)
(2007.135,73)
(2007.154,37)
(2007.173,40)
(2007.192,77)
(2007.212,22)
(2007.231,89)
(2007.25,61)
(2007.269,69)
(2007.288,59)
(2007.308,63)
(2007.327,60)
(2007.346,66)
(2007.365,91)
(2007.385,145)
(2007.404,75)
(2007.423,91)
(2007.442,79)
(2007.462,65)
(2007.481,47)
(2007.5,67)
(2007.519,23)
(2007.538,16)
(2007.558,19)
(2007.577,19)
(2007.596,6)
(2007.615,5)
(2007.635,6)
(2007.654,1)
(2007.673,2)
(2007.692,4)
(2007.712,2)
(2007.731,2)
(2007.75,4)
(2007.769,6)
(2007.788,12)
(2007.808,20)
(2007.827,34)
(2007.846,9)
(2007.865,48)
(2007.885,34)
(2007.904,29)
(2007.923,40)
(2007.942,36)
(2007.962,52)
(2007.981,38)
(2008.0,31)
(2008.019,57)
(2008.038,34)
(2008.058,95)
(2008.077,71)
(2008.096,48)
(2008.115,18)
(2008.135,17)
(2008.154,52)
(2008.173,26)
(2008.192,70)
(2008.212,32)
(2008.231,50)
(2008.25,18)
(2008.269,44)
(2008.288,65)
(2008.308,51)
(2008.327,134)
(2008.346,18)
(2008.365,69)
(2008.385,111)
(2008.404,70)
(2008.423,81)
(2008.442,34)
(2008.462,51)
(2008.481,31)
(2008.5,34)
(2008.519,15)
(2008.538,17)
(2008.558,9)
(2008.577,10)
(2008.596,5)
(2008.615,7)
(2008.635,5)
(2008.654,1)
(2008.673,2)
(2008.692,2)
(2008.712,8)
(2008.731,5)
(2008.75,9)
(2008.769,9)
(2008.788,34)
(2008.808,20)
(2008.827,0)
(2008.846,72)
(2008.865,25)
(2008.885,19)
(2008.904,35)
(2008.923,17)
(2008.942,23)
(2008.962,14)
(2008.981,20)
(2009.0,3)
(2009.019,1)
(2009.038,58)
(2009.058,66)
(2009.077,40)
(2009.096,44)
(2009.115,25)
(2009.135,42)
(2009.154,64)
(2009.173,55)
(2009.192,46)
(2009.212,51)
(2009.231,38)
(2009.25,46)
(2009.269,41)
(2009.288,53)
(2009.308,42)
(2009.327,152)
(2009.346,30)
(2009.365,37)
(2009.385,72)
(2009.404,46)
(2009.423,70)
(2009.442,50)
(2009.462,40)
(2009.481,52)
(2009.5,20)
(2009.519,42)
(2009.538,0)
(2009.558,25)
(2009.577,23)
(2009.596,4)
(2009.615,5)
(2009.635,0)
(2009.654,0)
(2009.673,4)
(2009.692,0)
(2009.712,0)
(2009.731,2)
(2009.75,1)
(2009.769,6)
(2009.788,9)
(2009.808,3)
(2009.827,10)
(2009.846,12)
(2009.865,22)
(2009.885,4)
(2009.904,9)
(2009.923,25)
(2009.942,17)
(2009.962,43)
(2009.981,39)
(2010.0,3)
(2010.019,19)
(2010.038,50)
(2010.058,52)
(2010.077,20)
(2010.096,52)
(2010.115,26)
(2010.135,82)
(2010.154,87)
(2010.173,102)
(2010.192,118)
(2010.212,86)
(2010.231,96)
(2010.25,110)
(2010.269,88)
(2010.288,108)
(2010.308,96)
(2010.327,238)
(2010.346,117)
(2010.365,167)
(2010.385,62)
(2010.404,160)
(2010.423,83)
(2010.442,135)
(2010.462,94)
(2010.481,57)
(2010.5,31)
(2010.519,43)
(2010.538,48)
(2010.558,16)
(2010.577,6)
(2010.596,6)
(2010.615,3)
(2010.635,4)
(2010.654,4)
(2010.673,0)
(2010.692,2)
(2010.712,1)
(2010.731,1)
(2010.75,13)
(2010.769,4)
(2010.788,10)
(2010.808,31)
(2010.827,3)
(2010.846,30)
(2010.865,2)
(2010.885,25)
(2010.904,24)
(2010.923,14)
(2010.942,49)
(2010.962,49)
(2010.981,54)
(2011.0,24)
(2011.019,20)
(2011.038,35)
(2011.058,41)
(2011.077,15)
(2011.096,28)
(2011.115,23)
(2011.135,31)
(2011.154,17)
(2011.173,32)
(2011.192,46)
(2011.212,45)
(2011.231,24)
(2011.25,29)
(2011.269,19)
(2011.288,34)
(2011.308,48)
(2011.327,28)
(2011.346,45)
(2011.365,87)
(2011.385,71)
(2011.404,101)
(2011.423,71)
(2011.442,86)
(2011.462,64)
(2011.481,46)
(2011.5,46)
(2011.519,51)
(2011.538,42)
(2011.558,16)
(2011.577,10)
(2011.596,8)
(2011.615,5)
(2011.635,3)
(2011.654,0)
(2011.673,1)
(2011.692,1)
(2011.712,0)
(2011.731,2)
(2011.75,1)
(2011.769,1)
(2011.788,3)
(2011.808,4)
(2011.827,9)
(2011.846,11)
(2011.865,11)
(2011.885,11)
(2011.904,22)
(2011.923,42)
(2011.942,28)
(2011.962,63)
(2011.981,86)
(2012.0,41)
(2012.019,32)
(2012.038,140)
(2012.058,86)
(2012.077,41)
(2012.096,44)
(2012.115,61)
(2012.135,38)
(2012.154,121)
(2012.173,66)
(2012.192,69)
(2012.212,47)
(2012.231,16)
(2012.25,119)
(2012.269,49)
(2012.288,67)
(2012.308,50)
(2012.327,90)
(2012.346,31)
(2012.365,26)
(2012.385,77)
(2012.404,77)
(2012.423,49)
(2012.442,43)
(2012.462,127)
(2012.481,53)
(2012.5,34)
(2012.519,26)
(2012.538,53)
(2012.558,40)
(2012.577,39)
(2012.596,11)
(2012.615,7)
(2012.635,0)
(2012.654,6)
(2012.673,4)
(2012.692,0)
(2012.712,0)
(2012.731,8)
(2012.75,5)
(2012.769,0)
(2012.788,12)
(2012.808,0)
(2012.827,7)
(2012.846,8)
(2012.865,1)
(2012.885,9)
(2012.904,4)
(2012.923,3)
(2012.942,6)
(2012.962,8)
(2012.981,18)
(2013.0,18)
(2013.019,6)
(2013.038,7)
(2013.058,8)
(2013.077,28)
(2013.096,5)
(2013.115,20)
(2013.135,6)
(2013.154,33)
(2013.173,13)
(2013.192,47)
(2013.212,11)
(2013.231,39)
(2013.25,41)
(2013.269,12)
(2013.288,22)
(2013.308,21)
(2013.327,38)
(2013.346,31)
(2013.365,50)
(2013.385,34)
(2013.404,40)
(2013.423,46)
(2013.442,55)
(2013.462,12)
(2013.481,35)
(2013.5,5)
(2013.519,37)
(2013.538,22)
(2013.558,28)
(2013.577,22)
(2013.596,11)
(2013.615,8)
(2013.635,16)
(2013.654,0)
(2013.673,3)
(2013.692,9)
(2013.712,0)
(2013.731,1)
(2013.75,3)
(2013.769,5)
(2013.788,9)
(2013.808,0)
(2013.827,1)
(2013.846,10)
(2013.865,1)
(2013.885,8)
(2013.904,14)
(2013.923,8)
(2013.942,3)
(2013.962,14)
(2013.981,30)
(2014.0,35)
(2014.019,0)
(2014.038,1)
(2014.058,4)
(2014.077,41)
(2014.096,72)
(2014.115,41)
(2014.135,14)
(2014.154,1)
(2014.173,72)
(2014.192,36)
(2014.212,51)
(2014.231,77)
(2014.25,21)
(2014.269,34)
(2014.288,68)
(2014.308,157)
(2014.327,38)
(2014.346,25)
(2014.365,5)
(2014.385,40)
(2014.404,274)
(2014.423,48)
(2014.442,80)
(2014.462,85)
(2014.481,0)
(2014.5,177)
(2014.519,84)
(2014.538,25)
(2014.558,17)
(2014.577,25)
(2014.596,10)
(2014.615,12)
(2014.635,8)
(2014.654,2)
(2014.673,1)
(2014.692,1)
(2014.712,0)
(2014.731,0)
(2014.75,0)
(2014.769,1)
(2014.788,2)
(2014.808,0)
(2014.827,0)
(2014.846,3)
(2014.865,1)
(2014.885,7)
(2014.904,10)
(2014.923,26)
(2014.942,14)
(2014.962,41)
(2014.981,31)
(2015.0,15)
(2015.019,8)
(2015.038,49)
}

\def\BARANYADATA{
(2005.019,79)
(2005.038,60)
(2005.058,44)
(2005.077,49)
(2005.096,78)
(2005.115,76)
(2005.135,103)
(2005.154,74)
(2005.173,86)
(2005.192,81)
(2005.212,59)
(2005.231,74)
(2005.25,63)
(2005.269,83)
(2005.288,53)
(2005.308,74)
(2005.327,50)
(2005.346,66)
(2005.365,81)
(2005.385,46)
(2005.404,99)
(2005.423,62)
(2005.442,98)
(2005.462,30)
(2005.481,56)
(2005.5,40)
(2005.519,18)
(2005.538,15)
(2005.558,12)
(2005.577,12)
(2005.596,4)
(2005.615,4)
(2005.635,1)
(2005.654,0)
(2005.673,1)
(2005.692,3)
(2005.712,1)
(2005.731,3)
(2005.75,2)
(2005.769,4)
(2005.788,3)
(2005.808,8)
(2005.827,9)
(2005.846,15)
(2005.865,12)
(2005.885,21)
(2005.904,9)
(2005.923,13)
(2005.942,28)
(2005.962,32)
(2005.981,24)
(2006.0,33)
(2006.019,42)
(2006.038,56)
(2006.058,8)
(2006.077,34)
(2006.096,35)
(2006.115,23)
(2006.135,25)
(2006.154,50)
(2006.173,51)
(2006.192,58)
(2006.212,62)
(2006.231,78)
(2006.25,48)
(2006.269,66)
(2006.288,30)
(2006.308,57)
(2006.327,139)
(2006.346,56)
(2006.365,68)
(2006.385,38)
(2006.404,72)
(2006.423,78)
(2006.442,60)
(2006.462,60)
(2006.481,51)
(2006.5,63)
(2006.519,70)
(2006.538,21)
(2006.558,54)
(2006.577,19)
(2006.596,5)
(2006.615,16)
(2006.635,1)
(2006.654,5)
(2006.673,1)
(2006.692,5)
(2006.712,3)
(2006.731,7)
(2006.75,3)
(2006.769,6)
(2006.788,1)
(2006.808,4)
(2006.827,3)
(2006.846,9)
(2006.865,16)
(2006.885,15)
(2006.904,24)
(2006.923,17)
(2006.942,24)
(2006.962,52)
(2006.981,53)
(2007.0,21)
(2007.019,39)
(2007.038,63)
(2007.058,38)
(2007.077,50)
(2007.096,55)
(2007.115,49)
(2007.135,58)
(2007.154,62)
(2007.173,79)
(2007.192,53)
(2007.212,65)
(2007.231,109)
(2007.25,90)
(2007.269,68)
(2007.288,89)
(2007.308,121)
(2007.327,41)
(2007.346,60)
(2007.365,137)
(2007.385,118)
(2007.404,141)
(2007.423,38)
(2007.442,116)
(2007.462,75)
(2007.481,73)
(2007.5,34)
(2007.519,25)
(2007.538,21)
(2007.558,8)
(2007.577,17)
(2007.596,9)
(2007.615,4)
(2007.635,1)
(2007.654,1)
(2007.673,1)
(2007.692,4)
(2007.712,1)
(2007.731,4)
(2007.75,11)
(2007.769,5)
(2007.788,13)
(2007.808,10)
(2007.827,14)
(2007.846,9)
(2007.865,20)
(2007.885,22)
(2007.904,21)
(2007.923,50)
(2007.942,34)
(2007.962,27)
(2007.981,36)
(2008.0,15)
(2008.019,41)
(2008.038,68)
(2008.058,76)
(2008.077,50)
(2008.096,20)
(2008.115,24)
(2008.135,33)
(2008.154,39)
(2008.173,46)
(2008.192,42)
(2008.212,50)
(2008.231,31)
(2008.25,78)
(2008.269,64)
(2008.288,78)
(2008.308,42)
(2008.327,60)
(2008.346,19)
(2008.365,93)
(2008.385,63)
(2008.404,103)
(2008.423,77)
(2008.442,145)
(2008.462,96)
(2008.481,69)
(2008.5,35)
(2008.519,68)
(2008.538,14)
(2008.558,8)
(2008.577,17)
(2008.596,4)
(2008.615,2)
(2008.635,1)
(2008.654,0)
(2008.673,2)
(2008.692,1)
(2008.712,0)
(2008.731,0)
(2008.75,1)
(2008.769,4)
(2008.788,2)
(2008.808,4)
(2008.827,3)
(2008.846,7)
(2008.865,9)
(2008.885,13)
(2008.904,8)
(2008.923,29)
(2008.942,17)
(2008.962,20)
(2008.981,42)
(2009.0,20)
(2009.019,18)
(2009.038,42)
(2009.058,23)
(2009.077,49)
(2009.096,62)
(2009.115,41)
(2009.135,57)
(2009.154,47)
(2009.173,94)
(2009.192,73)
(2009.212,59)
(2009.231,52)
(2009.25,22)
(2009.269,113)
(2009.288,45)
(2009.308,68)
(2009.327,25)
(2009.346,46)
(2009.365,70)
(2009.385,39)
(2009.404,47)
(2009.423,20)
(2009.442,45)
(2009.462,22)
(2009.481,55)
(2009.5,41)
(2009.519,17)
(2009.538,26)
(2009.558,6)
(2009.577,3)
(2009.596,24)
(2009.615,10)
(2009.635,2)
(2009.654,1)
(2009.673,1)
(2009.692,1)
(2009.712,2)
(2009.731,0)
(2009.75,0)
(2009.769,2)
(2009.788,5)
(2009.808,6)
(2009.827,10)
(2009.846,4)
(2009.865,1)
(2009.885,3)
(2009.904,8)
(2009.923,9)
(2009.942,31)
(2009.962,38)
(2009.981,46)
(2010.0,50)
(2010.019,28)
(2010.038,37)
(2010.058,23)
(2010.077,56)
(2010.096,103)
(2010.115,50)
(2010.135,47)
(2010.154,127)
(2010.173,83)
(2010.192,101)
(2010.212,68)
(2010.231,135)
(2010.25,123)
(2010.269,112)
(2010.288,80)
(2010.308,140)
(2010.327,106)
(2010.346,116)
(2010.365,96)
(2010.385,115)
(2010.404,102)
(2010.423,17)
(2010.442,84)
(2010.462,74)
(2010.481,66)
(2010.5,42)
(2010.519,25)
(2010.538,44)
(2010.558,34)
(2010.577,16)
(2010.596,19)
(2010.615,3)
(2010.635,2)
(2010.654,7)
(2010.673,1)
(2010.692,0)
(2010.712,1)
(2010.731,2)
(2010.75,0)
(2010.769,10)
(2010.788,5)
(2010.808,9)
(2010.827,23)
(2010.846,43)
(2010.865,16)
(2010.885,44)
(2010.904,28)
(2010.923,29)
(2010.942,18)
(2010.962,52)
(2010.981,45)
(2011.0,36)
(2011.019,41)
(2011.038,35)
(2011.058,38)
(2011.077,18)
(2011.096,51)
(2011.115,31)
(2011.135,28)
(2011.154,53)
(2011.173,46)
(2011.192,37)
(2011.212,43)
(2011.231,51)
(2011.25,58)
(2011.269,38)
(2011.288,37)
(2011.308,53)
(2011.327,41)
(2011.346,28)
(2011.365,49)
(2011.385,17)
(2011.404,55)
(2011.423,8)
(2011.442,62)
(2011.462,29)
(2011.481,38)
(2011.5,21)
(2011.519,23)
(2011.538,10)
(2011.558,3)
(2011.577,1)
(2011.596,11)
(2011.615,2)
(2011.635,0)
(2011.654,4)
(2011.673,1)
(2011.692,0)
(2011.712,1)
(2011.731,0)
(2011.75,3)
(2011.769,5)
(2011.788,5)
(2011.808,7)
(2011.827,9)
(2011.846,2)
(2011.865,10)
(2011.885,2)
(2011.904,12)
(2011.923,16)
(2011.942,16)
(2011.962,30)
(2011.981,40)
(2012.0,42)
(2012.019,52)
(2012.038,36)
(2012.058,41)
(2012.077,91)
(2012.096,48)
(2012.115,18)
(2012.135,29)
(2012.154,37)
(2012.173,20)
(2012.192,37)
(2012.212,43)
(2012.231,6)
(2012.25,38)
(2012.269,28)
(2012.288,21)
(2012.308,14)
(2012.327,27)
(2012.346,24)
(2012.365,2)
(2012.385,11)
(2012.404,16)
(2012.423,11)
(2012.442,2)
(2012.462,16)
(2012.481,7)
(2012.5,7)
(2012.519,11)
(2012.538,14)
(2012.558,7)
(2012.577,11)
(2012.596,2)
(2012.615,6)
(2012.635,2)
(2012.654,0)
(2012.673,0)
(2012.692,0)
(2012.712,4)
(2012.731,1)
(2012.75,3)
(2012.769,2)
(2012.788,12)
(2012.808,12)
(2012.827,11)
(2012.846,11)
(2012.865,3)
(2012.885,9)
(2012.904,12)
(2012.923,8)
(2012.942,14)
(2012.962,9)
(2012.981,18)
(2013.0,37)
(2013.019,11)
(2013.038,47)
(2013.058,34)
(2013.077,54)
(2013.096,22)
(2013.115,34)
(2013.135,194)
(2013.154,66)
(2013.173,29)
(2013.192,185)
(2013.212,51)
(2013.231,81)
(2013.25,68)
(2013.269,110)
(2013.288,38)
(2013.308,27)
(2013.327,27)
(2013.346,54)
(2013.365,54)
(2013.385,80)
(2013.404,115)
(2013.423,44)
(2013.442,67)
(2013.462,58)
(2013.481,47)
(2013.5,105)
(2013.519,106)
(2013.538,35)
(2013.558,30)
(2013.577,11)
(2013.596,36)
(2013.615,84)
(2013.635,10)
(2013.654,0)
(2013.673,0)
(2013.692,1)
(2013.712,1)
(2013.731,1)
(2013.75,0)
(2013.769,3)
(2013.788,2)
(2013.808,22)
(2013.827,15)
(2013.846,26)
(2013.865,9)
(2013.885,18)
(2013.904,5)
(2013.923,24)
(2013.942,14)
(2013.962,20)
(2013.981,30)
(2014.0,44)
(2014.019,3)
(2014.038,27)
(2014.058,49)
(2014.077,50)
(2014.096,19)
(2014.115,24)
(2014.135,43)
(2014.154,31)
(2014.173,29)
(2014.192,40)
(2014.212,53)
(2014.231,48)
(2014.25,49)
(2014.269,60)
(2014.288,51)
(2014.308,21)
(2014.327,31)
(2014.346,21)
(2014.365,10)
(2014.385,67)
(2014.404,35)
(2014.423,23)
(2014.442,35)
(2014.462,17)
(2014.481,4)
(2014.5,27)
(2014.519,20)
(2014.538,19)
(2014.558,27)
(2014.577,4)
(2014.596,7)
(2014.615,0)
(2014.635,5)
(2014.654,2)
(2014.673,4)
(2014.692,3)
(2014.712,0)
(2014.731,2)
(2014.75,4)
(2014.769,4)
(2014.788,4)
(2014.808,10)
(2014.827,1)
(2014.846,12)
(2014.865,7)
(2014.885,4)
(2014.904,5)
(2014.923,15)
(2014.942,15)
(2014.962,12)
(2014.981,39)
(2015.0,7)
(2015.019,23)
(2015.038,42)
}

\def\BEKESDATA{
(2005.019,173)
(2005.038,92)
(2005.058,86)
(2005.077,126)
(2005.096,87)
(2005.115,152)
(2005.135,192)
(2005.154,174)
(2005.173,171)
(2005.192,217)
(2005.212,243)
(2005.231,271)
(2005.25,119)
(2005.269,130)
(2005.288,111)
(2005.308,101)
(2005.327,126)
(2005.346,86)
(2005.365,116)
(2005.385,94)
(2005.404,139)
(2005.423,85)
(2005.442,100)
(2005.462,54)
(2005.481,56)
(2005.5,32)
(2005.519,26)
(2005.538,13)
(2005.558,12)
(2005.577,10)
(2005.596,1)
(2005.615,8)
(2005.635,1)
(2005.654,6)
(2005.673,1)
(2005.692,2)
(2005.712,0)
(2005.731,5)
(2005.75,1)
(2005.769,10)
(2005.788,7)
(2005.808,7)
(2005.827,19)
(2005.846,4)
(2005.865,11)
(2005.885,5)
(2005.904,14)
(2005.923,13)
(2005.942,4)
(2005.962,6)
(2005.981,17)
(2006.0,8)
(2006.019,20)
(2006.038,15)
(2006.058,7)
(2006.077,15)
(2006.096,9)
(2006.115,23)
(2006.135,16)
(2006.154,38)
(2006.173,16)
(2006.192,32)
(2006.212,40)
(2006.231,35)
(2006.25,26)
(2006.269,41)
(2006.288,41)
(2006.308,75)
(2006.327,44)
(2006.346,52)
(2006.365,31)
(2006.385,39)
(2006.404,14)
(2006.423,106)
(2006.442,23)
(2006.462,47)
(2006.481,23)
(2006.5,43)
(2006.519,5)
(2006.538,6)
(2006.558,12)
(2006.577,4)
(2006.596,4)
(2006.615,4)
(2006.635,4)
(2006.654,8)
(2006.673,2)
(2006.692,4)
(2006.712,5)
(2006.731,1)
(2006.75,5)
(2006.769,11)
(2006.788,3)
(2006.808,6)
(2006.827,7)
(2006.846,8)
(2006.865,21)
(2006.885,18)
(2006.904,20)
(2006.923,25)
(2006.942,24)
(2006.962,34)
(2006.981,33)
(2007.0,46)
(2007.019,76)
(2007.038,35)
(2007.058,93)
(2007.077,125)
(2007.096,155)
(2007.115,75)
(2007.135,114)
(2007.154,117)
(2007.173,87)
(2007.192,86)
(2007.212,64)
(2007.231,140)
(2007.25,163)
(2007.269,66)
(2007.288,58)
(2007.308,84)
(2007.327,111)
(2007.346,23)
(2007.365,53)
(2007.385,61)
(2007.404,87)
(2007.423,76)
(2007.442,62)
(2007.462,80)
(2007.481,18)
(2007.5,16)
(2007.519,29)
(2007.538,27)
(2007.558,90)
(2007.577,2)
(2007.596,3)
(2007.615,2)
(2007.635,0)
(2007.654,3)
(2007.673,1)
(2007.692,2)
(2007.712,2)
(2007.731,1)
(2007.75,2)
(2007.769,11)
(2007.788,12)
(2007.808,19)
(2007.827,13)
(2007.846,20)
(2007.865,37)
(2007.885,41)
(2007.904,37)
(2007.923,37)
(2007.942,33)
(2007.962,30)
(2007.981,60)
(2008.0,3)
(2008.019,138)
(2008.038,86)
(2008.058,57)
(2008.077,94)
(2008.096,129)
(2008.115,50)
(2008.135,12)
(2008.154,9)
(2008.173,8)
(2008.192,21)
(2008.212,105)
(2008.231,34)
(2008.25,18)
(2008.269,71)
(2008.288,37)
(2008.308,58)
(2008.327,47)
(2008.346,31)
(2008.365,57)
(2008.385,35)
(2008.404,59)
(2008.423,24)
(2008.442,66)
(2008.462,12)
(2008.481,30)
(2008.5,11)
(2008.519,0)
(2008.538,18)
(2008.558,13)
(2008.577,7)
(2008.596,0)
(2008.615,2)
(2008.635,6)
(2008.654,1)
(2008.673,1)
(2008.692,4)
(2008.712,2)
(2008.731,1)
(2008.75,3)
(2008.769,0)
(2008.788,10)
(2008.808,16)
(2008.827,3)
(2008.846,11)
(2008.865,4)
(2008.885,24)
(2008.904,4)
(2008.923,34)
(2008.942,19)
(2008.962,42)
(2008.981,37)
(2009.0,20)
(2009.019,8)
(2009.038,57)
(2009.058,44)
(2009.077,9)
(2009.096,53)
(2009.115,23)
(2009.135,25)
(2009.154,57)
(2009.173,75)
(2009.192,35)
(2009.212,64)
(2009.231,65)
(2009.25,34)
(2009.269,29)
(2009.288,16)
(2009.308,39)
(2009.327,74)
(2009.346,38)
(2009.365,59)
(2009.385,56)
(2009.404,68)
(2009.423,47)
(2009.442,46)
(2009.462,55)
(2009.481,59)
(2009.5,39)
(2009.519,15)
(2009.538,14)
(2009.558,21)
(2009.577,4)
(2009.596,19)
(2009.615,3)
(2009.635,4)
(2009.654,3)
(2009.673,3)
(2009.692,5)
(2009.712,1)
(2009.731,2)
(2009.75,1)
(2009.769,0)
(2009.788,1)
(2009.808,3)
(2009.827,3)
(2009.846,13)
(2009.865,24)
(2009.885,10)
(2009.904,13)
(2009.923,14)
(2009.942,20)
(2009.962,11)
(2009.981,12)
(2010.0,11)
(2010.019,13)
(2010.038,6)
(2010.058,11)
(2010.077,11)
(2010.096,3)
(2010.115,12)
(2010.135,12)
(2010.154,9)
(2010.173,13)
(2010.192,15)
(2010.212,13)
(2010.231,17)
(2010.25,16)
(2010.269,11)
(2010.288,17)
(2010.308,28)
(2010.327,17)
(2010.346,18)
(2010.365,4)
(2010.385,27)
(2010.404,25)
(2010.423,17)
(2010.442,11)
(2010.462,8)
(2010.481,25)
(2010.5,17)
(2010.519,12)
(2010.538,21)
(2010.558,8)
(2010.577,4)
(2010.596,2)
(2010.615,2)
(2010.635,0)
(2010.654,2)
(2010.673,5)
(2010.692,1)
(2010.712,1)
(2010.731,1)
(2010.75,2)
(2010.769,3)
(2010.788,5)
(2010.808,5)
(2010.827,18)
(2010.846,45)
(2010.865,12)
(2010.885,36)
(2010.904,30)
(2010.923,61)
(2010.942,86)
(2010.962,24)
(2010.981,46)
(2011.0,78)
(2011.019,28)
(2011.038,26)
(2011.058,149)
(2011.077,95)
(2011.096,34)
(2011.115,76)
(2011.135,47)
(2011.154,76)
(2011.173,102)
(2011.192,44)
(2011.212,55)
(2011.231,37)
(2011.25,43)
(2011.269,54)
(2011.288,42)
(2011.308,38)
(2011.327,37)
(2011.346,73)
(2011.365,33)
(2011.385,53)
(2011.404,31)
(2011.423,49)
(2011.442,33)
(2011.462,45)
(2011.481,47)
(2011.5,37)
(2011.519,7)
(2011.538,7)
(2011.558,5)
(2011.577,2)
(2011.596,0)
(2011.615,0)
(2011.635,1)
(2011.654,0)
(2011.673,2)
(2011.692,1)
(2011.712,1)
(2011.731,1)
(2011.75,2)
(2011.769,2)
(2011.788,0)
(2011.808,2)
(2011.827,7)
(2011.846,1)
(2011.865,3)
(2011.885,1)
(2011.904,0)
(2011.923,28)
(2011.942,5)
(2011.962,12)
(2011.981,19)
(2012.0,11)
(2012.019,1)
(2012.038,12)
(2012.058,11)
(2012.077,11)
(2012.096,16)
(2012.115,8)
(2012.135,16)
(2012.154,7)
(2012.173,4)
(2012.192,3)
(2012.212,1)
(2012.231,7)
(2012.25,6)
(2012.269,3)
(2012.288,3)
(2012.308,2)
(2012.327,15)
(2012.346,5)
(2012.365,2)
(2012.385,8)
(2012.404,14)
(2012.423,21)
(2012.442,8)
(2012.462,3)
(2012.481,3)
(2012.5,9)
(2012.519,3)
(2012.538,5)
(2012.558,2)
(2012.577,12)
(2012.596,9)
(2012.615,20)
(2012.635,6)
(2012.654,9)
(2012.673,0)
(2012.692,1)
(2012.712,6)
(2012.731,1)
(2012.75,0)
(2012.769,2)
(2012.788,0)
(2012.808,3)
(2012.827,10)
(2012.846,30)
(2012.865,21)
(2012.885,32)
(2012.904,34)
(2012.923,25)
(2012.942,59)
(2012.962,24)
(2012.981,54)
(2013.0,22)
(2013.019,25)
(2013.038,35)
(2013.058,61)
(2013.077,22)
(2013.096,20)
(2013.115,46)
(2013.135,23)
(2013.154,55)
(2013.173,32)
(2013.192,39)
(2013.212,27)
(2013.231,15)
(2013.25,65)
(2013.269,50)
(2013.288,8)
(2013.308,13)
(2013.327,14)
(2013.346,30)
(2013.365,43)
(2013.385,8)
(2013.404,4)
(2013.423,5)
(2013.442,8)
(2013.462,6)
(2013.481,28)
(2013.5,7)
(2013.519,12)
(2013.538,70)
(2013.558,14)
(2013.577,1)
(2013.596,8)
(2013.615,5)
(2013.635,3)
(2013.654,2)
(2013.673,0)
(2013.692,3)
(2013.712,0)
(2013.731,1)
(2013.75,1)
(2013.769,1)
(2013.788,8)
(2013.808,0)
(2013.827,1)
(2013.846,7)
(2013.865,11)
(2013.885,4)
(2013.904,11)
(2013.923,8)
(2013.942,19)
(2013.962,10)
(2013.981,11)
(2014.0,23)
(2014.019,0)
(2014.038,1)
(2014.058,4)
(2014.077,4)
(2014.096,13)
(2014.115,24)
(2014.135,14)
(2014.154,26)
(2014.173,21)
(2014.192,30)
(2014.212,41)
(2014.231,20)
(2014.25,50)
(2014.269,58)
(2014.288,54)
(2014.308,26)
(2014.327,56)
(2014.346,6)
(2014.365,0)
(2014.385,59)
(2014.404,62)
(2014.423,9)
(2014.442,74)
(2014.462,27)
(2014.481,8)
(2014.5,11)
(2014.519,8)
(2014.538,4)
(2014.558,5)
(2014.577,3)
(2014.596,2)
(2014.615,7)
(2014.635,0)
(2014.654,4)
(2014.673,1)
(2014.692,7)
(2014.712,0)
(2014.731,0)
(2014.75,1)
(2014.769,1)
(2014.788,0)
(2014.808,1)
(2014.827,1)
(2014.846,26)
(2014.865,2)
(2014.885,0)
(2014.904,0)
(2014.923,7)
(2014.942,2)
(2014.962,6)
(2014.981,10)
(2015.0,0)
(2015.019,0)
(2015.038,32)
}

\def\BORSODDATA{
(2005.019,169)
(2005.038,200)
(2005.058,93)
(2005.077,46)
(2005.096,103)
(2005.115,189)
(2005.135,148)
(2005.154,140)
(2005.173,90)
(2005.192,167)
(2005.212,99)
(2005.231,215)
(2005.25,51)
(2005.269,152)
(2005.288,103)
(2005.308,84)
(2005.327,111)
(2005.346,48)
(2005.365,129)
(2005.385,135)
(2005.404,84)
(2005.423,93)
(2005.442,104)
(2005.462,61)
(2005.481,147)
(2005.5,85)
(2005.519,69)
(2005.538,36)
(2005.558,29)
(2005.577,17)
(2005.596,6)
(2005.615,1)
(2005.635,4)
(2005.654,5)
(2005.673,2)
(2005.692,19)
(2005.712,1)
(2005.731,19)
(2005.75,28)
(2005.769,34)
(2005.788,60)
(2005.808,46)
(2005.827,60)
(2005.846,12)
(2005.865,100)
(2005.885,51)
(2005.904,45)
(2005.923,81)
(2005.942,40)
(2005.962,91)
(2005.981,75)
(2006.0,50)
(2006.019,89)
(2006.038,92)
(2006.058,50)
(2006.077,103)
(2006.096,68)
(2006.115,119)
(2006.135,108)
(2006.154,119)
(2006.173,112)
(2006.192,89)
(2006.212,61)
(2006.231,82)
(2006.25,68)
(2006.269,68)
(2006.288,62)
(2006.308,23)
(2006.327,97)
(2006.346,44)
(2006.365,44)
(2006.385,49)
(2006.404,22)
(2006.423,59)
(2006.442,15)
(2006.462,15)
(2006.481,65)
(2006.5,8)
(2006.519,50)
(2006.538,28)
(2006.558,17)
(2006.577,8)
(2006.596,4)
(2006.615,6)
(2006.635,5)
(2006.654,3)
(2006.673,2)
(2006.692,1)
(2006.712,4)
(2006.731,3)
(2006.75,12)
(2006.769,11)
(2006.788,16)
(2006.808,24)
(2006.827,16)
(2006.846,6)
(2006.865,39)
(2006.885,33)
(2006.904,39)
(2006.923,10)
(2006.942,37)
(2006.962,22)
(2006.981,31)
(2007.0,15)
(2007.019,25)
(2007.038,61)
(2007.058,160)
(2007.077,44)
(2007.096,71)
(2007.115,125)
(2007.135,52)
(2007.154,144)
(2007.173,96)
(2007.192,107)
(2007.212,110)
(2007.231,72)
(2007.25,96)
(2007.269,138)
(2007.288,114)
(2007.308,284)
(2007.327,157)
(2007.346,88)
(2007.365,23)
(2007.385,93)
(2007.404,174)
(2007.423,82)
(2007.442,161)
(2007.462,129)
(2007.481,34)
(2007.5,68)
(2007.519,80)
(2007.538,33)
(2007.558,33)
(2007.577,30)
(2007.596,14)
(2007.615,7)
(2007.635,21)
(2007.654,4)
(2007.673,8)
(2007.692,14)
(2007.712,3)
(2007.731,5)
(2007.75,8)
(2007.769,11)
(2007.788,38)
(2007.808,50)
(2007.827,20)
(2007.846,25)
(2007.865,63)
(2007.885,49)
(2007.904,70)
(2007.923,74)
(2007.942,54)
(2007.962,51)
(2007.981,82)
(2008.0,34)
(2008.019,126)
(2008.038,143)
(2008.058,202)
(2008.077,209)
(2008.096,61)
(2008.115,104)
(2008.135,62)
(2008.154,66)
(2008.173,76)
(2008.192,85)
(2008.212,78)
(2008.231,55)
(2008.25,127)
(2008.269,64)
(2008.288,42)
(2008.308,119)
(2008.327,101)
(2008.346,36)
(2008.365,108)
(2008.385,64)
(2008.404,91)
(2008.423,55)
(2008.442,127)
(2008.462,70)
(2008.481,56)
(2008.5,69)
(2008.519,21)
(2008.538,13)
(2008.558,36)
(2008.577,18)
(2008.596,8)
(2008.615,11)
(2008.635,13)
(2008.654,4)
(2008.673,6)
(2008.692,2)
(2008.712,4)
(2008.731,6)
(2008.75,7)
(2008.769,21)
(2008.788,33)
(2008.808,53)
(2008.827,3)
(2008.846,41)
(2008.865,45)
(2008.885,52)
(2008.904,39)
(2008.923,56)
(2008.942,34)
(2008.962,108)
(2008.981,65)
(2009.0,14)
(2009.019,32)
(2009.038,69)
(2009.058,355)
(2009.077,112)
(2009.096,100)
(2009.115,71)
(2009.135,64)
(2009.154,74)
(2009.173,71)
(2009.192,97)
(2009.212,125)
(2009.231,67)
(2009.25,65)
(2009.269,55)
(2009.288,74)
(2009.308,219)
(2009.327,156)
(2009.346,73)
(2009.365,96)
(2009.385,114)
(2009.404,69)
(2009.423,123)
(2009.442,99)
(2009.462,76)
(2009.481,69)
(2009.5,36)
(2009.519,20)
(2009.538,87)
(2009.558,35)
(2009.577,6)
(2009.596,21)
(2009.615,6)
(2009.635,13)
(2009.654,4)
(2009.673,2)
(2009.692,2)
(2009.712,6)
(2009.731,6)
(2009.75,6)
(2009.769,1)
(2009.788,10)
(2009.808,3)
(2009.827,11)
(2009.846,9)
(2009.865,31)
(2009.885,22)
(2009.904,16)
(2009.923,37)
(2009.942,34)
(2009.962,69)
(2009.981,17)
(2010.0,28)
(2010.019,55)
(2010.038,43)
(2010.058,28)
(2010.077,33)
(2010.096,42)
(2010.115,59)
(2010.135,42)
(2010.154,93)
(2010.173,63)
(2010.192,35)
(2010.212,54)
(2010.231,37)
(2010.25,145)
(2010.269,85)
(2010.288,52)
(2010.308,105)
(2010.327,79)
(2010.346,40)
(2010.365,82)
(2010.385,78)
(2010.404,65)
(2010.423,54)
(2010.442,68)
(2010.462,48)
(2010.481,44)
(2010.5,70)
(2010.519,24)
(2010.538,51)
(2010.558,26)
(2010.577,8)
(2010.596,12)
(2010.615,6)
(2010.635,7)
(2010.654,5)
(2010.673,13)
(2010.692,6)
(2010.712,1)
(2010.731,4)
(2010.75,2)
(2010.769,1)
(2010.788,4)
(2010.808,8)
(2010.827,9)
(2010.846,22)
(2010.865,7)
(2010.885,32)
(2010.904,25)
(2010.923,35)
(2010.942,15)
(2010.962,25)
(2010.981,20)
(2011.0,53)
(2011.019,36)
(2011.038,25)
(2011.058,93)
(2011.077,66)
(2011.096,65)
(2011.115,61)
(2011.135,79)
(2011.154,63)
(2011.173,164)
(2011.192,93)
(2011.212,173)
(2011.231,95)
(2011.25,192)
(2011.269,144)
(2011.288,171)
(2011.308,109)
(2011.327,72)
(2011.346,75)
(2011.365,158)
(2011.385,101)
(2011.404,109)
(2011.423,132)
(2011.442,127)
(2011.462,118)
(2011.481,64)
(2011.5,89)
(2011.519,36)
(2011.538,31)
(2011.558,35)
(2011.577,27)
(2011.596,26)
(2011.615,8)
(2011.635,5)
(2011.654,4)
(2011.673,5)
(2011.692,10)
(2011.712,8)
(2011.731,2)
(2011.75,8)
(2011.769,8)
(2011.788,11)
(2011.808,15)
(2011.827,12)
(2011.846,10)
(2011.865,25)
(2011.885,10)
(2011.904,7)
(2011.923,41)
(2011.942,52)
(2011.962,44)
(2011.981,31)
(2012.0,86)
(2012.019,81)
(2012.038,77)
(2012.058,248)
(2012.077,36)
(2012.096,218)
(2012.115,58)
(2012.135,55)
(2012.154,113)
(2012.173,122)
(2012.192,59)
(2012.212,66)
(2012.231,46)
(2012.25,124)
(2012.269,74)
(2012.288,73)
(2012.308,8)
(2012.327,134)
(2012.346,88)
(2012.365,33)
(2012.385,81)
(2012.404,26)
(2012.423,26)
(2012.442,150)
(2012.462,128)
(2012.481,105)
(2012.5,161)
(2012.519,127)
(2012.538,34)
(2012.558,75)
(2012.577,28)
(2012.596,19)
(2012.615,8)
(2012.635,4)
(2012.654,14)
(2012.673,2)
(2012.692,3)
(2012.712,6)
(2012.731,11)
(2012.75,7)
(2012.769,5)
(2012.788,17)
(2012.808,13)
(2012.827,19)
(2012.846,18)
(2012.865,44)
(2012.885,13)
(2012.904,56)
(2012.923,55)
(2012.942,84)
(2012.962,20)
(2012.981,13)
(2013.0,122)
(2013.019,2)
(2013.038,77)
(2013.058,156)
(2013.077,50)
(2013.096,47)
(2013.115,84)
(2013.135,90)
(2013.154,63)
(2013.173,45)
(2013.192,84)
(2013.212,53)
(2013.231,66)
(2013.25,77)
(2013.269,107)
(2013.288,40)
(2013.308,109)
(2013.327,83)
(2013.346,55)
(2013.365,39)
(2013.385,139)
(2013.404,85)
(2013.423,52)
(2013.442,59)
(2013.462,37)
(2013.481,75)
(2013.5,42)
(2013.519,67)
(2013.538,15)
(2013.558,54)
(2013.577,81)
(2013.596,17)
(2013.615,7)
(2013.635,20)
(2013.654,3)
(2013.673,2)
(2013.692,6)
(2013.712,8)
(2013.731,13)
(2013.75,3)
(2013.769,3)
(2013.788,9)
(2013.808,4)
(2013.827,17)
(2013.846,31)
(2013.865,13)
(2013.885,16)
(2013.904,71)
(2013.923,37)
(2013.942,80)
(2013.962,44)
(2013.981,81)
(2014.0,84)
(2014.019,3)
(2014.038,49)
(2014.058,205)
(2014.077,73)
(2014.096,40)
(2014.115,76)
(2014.135,66)
(2014.154,76)
(2014.173,114)
(2014.192,104)
(2014.212,103)
(2014.231,121)
(2014.25,134)
(2014.269,185)
(2014.288,72)
(2014.308,109)
(2014.327,107)
(2014.346,22)
(2014.365,0)
(2014.385,76)
(2014.404,109)
(2014.423,133)
(2014.442,79)
(2014.462,158)
(2014.481,17)
(2014.5,63)
(2014.519,106)
(2014.538,7)
(2014.558,81)
(2014.577,14)
(2014.596,18)
(2014.615,22)
(2014.635,7)
(2014.654,3)
(2014.673,0)
(2014.692,1)
(2014.712,1)
(2014.731,3)
(2014.75,1)
(2014.769,2)
(2014.788,3)
(2014.808,3)
(2014.827,4)
(2014.846,9)
(2014.865,13)
(2014.885,16)
(2014.904,29)
(2014.923,14)
(2014.942,3)
(2014.962,39)
(2014.981,34)
(2015.0,0)
(2015.019,11)
(2015.038,38)
}

\def\BUDAPESTDATA{
(2005.019,168)
(2005.038,157)
(2005.058,96)
(2005.077,163)
(2005.096,122)
(2005.115,174)
(2005.135,153)
(2005.154,115)
(2005.173,119)
(2005.192,114)
(2005.212,127)
(2005.231,135)
(2005.25,116)
(2005.269,132)
(2005.288,129)
(2005.308,113)
(2005.327,114)
(2005.346,98)
(2005.365,140)
(2005.385,120)
(2005.404,173)
(2005.423,135)
(2005.442,143)
(2005.462,139)
(2005.481,155)
(2005.5,98)
(2005.519,87)
(2005.538,110)
(2005.558,70)
(2005.577,43)
(2005.596,37)
(2005.615,30)
(2005.635,21)
(2005.654,22)
(2005.673,15)
(2005.692,11)
(2005.712,7)
(2005.731,12)
(2005.75,5)
(2005.769,17)
(2005.788,11)
(2005.808,27)
(2005.827,23)
(2005.846,30)
(2005.865,48)
(2005.885,37)
(2005.904,32)
(2005.923,49)
(2005.942,67)
(2005.962,106)
(2005.981,83)
(2006.0,114)
(2006.019,129)
(2006.038,189)
(2006.058,79)
(2006.077,163)
(2006.096,110)
(2006.115,158)
(2006.135,182)
(2006.154,136)
(2006.173,207)
(2006.192,251)
(2006.212,141)
(2006.231,193)
(2006.25,169)
(2006.269,141)
(2006.288,193)
(2006.308,359)
(2006.327,268)
(2006.346,168)
(2006.365,338)
(2006.385,186)
(2006.404,266)
(2006.423,240)
(2006.442,162)
(2006.462,264)
(2006.481,133)
(2006.5,201)
(2006.519,236)
(2006.538,111)
(2006.558,76)
(2006.577,52)
(2006.596,30)
(2006.615,30)
(2006.635,74)
(2006.654,45)
(2006.673,41)
(2006.692,20)
(2006.712,15)
(2006.731,11)
(2006.75,10)
(2006.769,9)
(2006.788,35)
(2006.808,20)
(2006.827,19)
(2006.846,51)
(2006.865,53)
(2006.885,77)
(2006.904,62)
(2006.923,93)
(2006.942,78)
(2006.962,82)
(2006.981,84)
(2007.0,58)
(2007.019,114)
(2007.038,145)
(2007.058,93)
(2007.077,129)
(2007.096,114)
(2007.115,86)
(2007.135,127)
(2007.154,106)
(2007.173,211)
(2007.192,229)
(2007.212,145)
(2007.231,210)
(2007.25,267)
(2007.269,166)
(2007.288,168)
(2007.308,350)
(2007.327,224)
(2007.346,136)
(2007.365,278)
(2007.385,161)
(2007.404,385)
(2007.423,230)
(2007.442,284)
(2007.462,210)
(2007.481,125)
(2007.5,104)
(2007.519,140)
(2007.538,99)
(2007.558,35)
(2007.577,14)
(2007.596,34)
(2007.615,7)
(2007.635,7)
(2007.654,4)
(2007.673,4)
(2007.692,10)
(2007.712,6)
(2007.731,6)
(2007.75,12)
(2007.769,20)
(2007.788,14)
(2007.808,33)
(2007.827,20)
(2007.846,25)
(2007.865,93)
(2007.885,72)
(2007.904,54)
(2007.923,175)
(2007.942,126)
(2007.962,60)
(2007.981,193)
(2008.0,68)
(2008.019,183)
(2008.038,156)
(2008.058,479)
(2008.077,143)
(2008.096,193)
(2008.115,105)
(2008.135,136)
(2008.154,91)
(2008.173,150)
(2008.192,82)
(2008.212,64)
(2008.231,100)
(2008.25,95)
(2008.269,136)
(2008.288,80)
(2008.308,122)
(2008.327,117)
(2008.346,84)
(2008.365,133)
(2008.385,62)
(2008.404,211)
(2008.423,143)
(2008.442,128)
(2008.462,105)
(2008.481,140)
(2008.5,51)
(2008.519,70)
(2008.538,46)
(2008.558,17)
(2008.577,21)
(2008.596,51)
(2008.615,22)
(2008.635,11)
(2008.654,3)
(2008.673,10)
(2008.692,4)
(2008.712,3)
(2008.731,9)
(2008.75,14)
(2008.769,12)
(2008.788,22)
(2008.808,43)
(2008.827,27)
(2008.846,67)
(2008.865,78)
(2008.885,61)
(2008.904,62)
(2008.923,40)
(2008.942,98)
(2008.962,110)
(2008.981,142)
(2009.0,53)
(2009.019,39)
(2009.038,304)
(2009.058,300)
(2009.077,115)
(2009.096,176)
(2009.115,176)
(2009.135,157)
(2009.154,188)
(2009.173,128)
(2009.192,165)
(2009.212,129)
(2009.231,197)
(2009.25,129)
(2009.269,217)
(2009.288,109)
(2009.308,212)
(2009.327,265)
(2009.346,149)
(2009.365,143)
(2009.385,165)
(2009.404,198)
(2009.423,197)
(2009.442,211)
(2009.462,177)
(2009.481,168)
(2009.5,110)
(2009.519,65)
(2009.538,104)
(2009.558,85)
(2009.577,54)
(2009.596,18)
(2009.615,16)
(2009.635,14)
(2009.654,7)
(2009.673,27)
(2009.692,16)
(2009.712,11)
(2009.731,2)
(2009.75,4)
(2009.769,7)
(2009.788,15)
(2009.808,30)
(2009.827,31)
(2009.846,25)
(2009.865,50)
(2009.885,37)
(2009.904,59)
(2009.923,58)
(2009.942,42)
(2009.962,91)
(2009.981,86)
(2010.0,41)
(2010.019,47)
(2010.038,166)
(2010.058,113)
(2010.077,62)
(2010.096,82)
(2010.115,56)
(2010.135,46)
(2010.154,48)
(2010.173,124)
(2010.192,74)
(2010.212,116)
(2010.231,120)
(2010.25,101)
(2010.269,121)
(2010.288,160)
(2010.308,199)
(2010.327,147)
(2010.346,130)
(2010.365,175)
(2010.385,156)
(2010.404,185)
(2010.423,123)
(2010.442,151)
(2010.462,150)
(2010.481,169)
(2010.5,154)
(2010.519,68)
(2010.538,93)
(2010.558,74)
(2010.577,56)
(2010.596,35)
(2010.615,24)
(2010.635,11)
(2010.654,35)
(2010.673,11)
(2010.692,22)
(2010.712,8)
(2010.731,5)
(2010.75,23)
(2010.769,27)
(2010.788,33)
(2010.808,74)
(2010.827,57)
(2010.846,95)
(2010.865,73)
(2010.885,171)
(2010.904,206)
(2010.923,144)
(2010.942,155)
(2010.962,191)
(2010.981,125)
(2011.0,157)
(2011.019,121)
(2011.038,59)
(2011.058,333)
(2011.077,145)
(2011.096,208)
(2011.115,174)
(2011.135,128)
(2011.154,190)
(2011.173,228)
(2011.192,149)
(2011.212,181)
(2011.231,131)
(2011.25,227)
(2011.269,158)
(2011.288,168)
(2011.308,181)
(2011.327,138)
(2011.346,129)
(2011.365,161)
(2011.385,121)
(2011.404,128)
(2011.423,150)
(2011.442,146)
(2011.462,115)
(2011.481,120)
(2011.5,140)
(2011.519,73)
(2011.538,91)
(2011.558,71)
(2011.577,61)
(2011.596,19)
(2011.615,16)
(2011.635,11)
(2011.654,17)
(2011.673,14)
(2011.692,13)
(2011.712,22)
(2011.731,42)
(2011.75,4)
(2011.769,18)
(2011.788,6)
(2011.808,10)
(2011.827,60)
(2011.846,103)
(2011.865,31)
(2011.885,103)
(2011.904,44)
(2011.923,91)
(2011.942,83)
(2011.962,111)
(2011.981,91)
(2012.0,163)
(2012.019,28)
(2012.038,106)
(2012.058,291)
(2012.077,185)
(2012.096,197)
(2012.115,147)
(2012.135,192)
(2012.154,190)
(2012.173,147)
(2012.192,131)
(2012.212,149)
(2012.231,99)
(2012.25,233)
(2012.269,146)
(2012.288,182)
(2012.308,68)
(2012.327,180)
(2012.346,142)
(2012.365,80)
(2012.385,194)
(2012.404,286)
(2012.423,129)
(2012.442,93)
(2012.462,199)
(2012.481,120)
(2012.5,151)
(2012.519,88)
(2012.538,87)
(2012.558,67)
(2012.577,52)
(2012.596,65)
(2012.615,6)
(2012.635,64)
(2012.654,18)
(2012.673,3)
(2012.692,0)
(2012.712,13)
(2012.731,9)
(2012.75,4)
(2012.769,8)
(2012.788,12)
(2012.808,13)
(2012.827,31)
(2012.846,35)
(2012.865,8)
(2012.885,73)
(2012.904,77)
(2012.923,21)
(2012.942,104)
(2012.962,105)
(2012.981,92)
(2013.0,192)
(2013.019,2)
(2013.038,67)
(2013.058,283)
(2013.077,135)
(2013.096,129)
(2013.115,203)
(2013.135,152)
(2013.154,107)
(2013.173,185)
(2013.192,110)
(2013.212,138)
(2013.231,100)
(2013.25,228)
(2013.269,130)
(2013.288,137)
(2013.308,235)
(2013.327,138)
(2013.346,166)
(2013.365,122)
(2013.385,149)
(2013.404,124)
(2013.423,187)
(2013.442,195)
(2013.462,130)
(2013.481,126)
(2013.5,154)
(2013.519,149)
(2013.538,106)
(2013.558,151)
(2013.577,100)
(2013.596,91)
(2013.615,46)
(2013.635,38)
(2013.654,13)
(2013.673,22)
(2013.692,33)
(2013.712,9)
(2013.731,14)
(2013.75,48)
(2013.769,7)
(2013.788,18)
(2013.808,6)
(2013.827,21)
(2013.846,30)
(2013.865,21)
(2013.885,79)
(2013.904,64)
(2013.923,55)
(2013.942,18)
(2013.962,101)
(2013.981,63)
(2014.0,59)
(2014.019,6)
(2014.038,26)
(2014.058,150)
(2014.077,80)
(2014.096,82)
(2014.115,58)
(2014.135,84)
(2014.154,59)
(2014.173,60)
(2014.192,116)
(2014.212,99)
(2014.231,112)
(2014.25,197)
(2014.269,116)
(2014.288,215)
(2014.308,82)
(2014.327,155)
(2014.346,80)
(2014.365,13)
(2014.385,70)
(2014.404,391)
(2014.423,178)
(2014.442,124)
(2014.462,235)
(2014.481,88)
(2014.5,187)
(2014.519,140)
(2014.538,111)
(2014.558,72)
(2014.577,75)
(2014.596,44)
(2014.615,40)
(2014.635,37)
(2014.654,10)
(2014.673,6)
(2014.692,7)
(2014.712,5)
(2014.731,10)
(2014.75,11)
(2014.769,16)
(2014.788,20)
(2014.808,17)
(2014.827,21)
(2014.846,34)
(2014.865,28)
(2014.885,44)
(2014.904,33)
(2014.923,85)
(2014.942,16)
(2014.962,95)
(2014.981,43)
(2015.0,35)
(2015.019,30)
(2015.038,259)
}

\def\CSONGRADDATA{
(2005.019,42)
(2005.038,53)
(2005.058,30)
(2005.077,39)
(2005.096,34)
(2005.115,26)
(2005.135,65)
(2005.154,56)
(2005.173,65)
(2005.192,64)
(2005.212,81)
(2005.231,48)
(2005.25,48)
(2005.269,54)
(2005.288,41)
(2005.308,66)
(2005.327,48)
(2005.346,40)
(2005.365,51)
(2005.385,25)
(2005.404,37)
(2005.423,27)
(2005.442,48)
(2005.462,30)
(2005.481,24)
(2005.5,13)
(2005.519,32)
(2005.538,20)
(2005.558,8)
(2005.577,6)
(2005.596,6)
(2005.615,6)
(2005.635,3)
(2005.654,0)
(2005.673,3)
(2005.692,3)
(2005.712,1)
(2005.731,3)
(2005.75,1)
(2005.769,1)
(2005.788,0)
(2005.808,8)
(2005.827,18)
(2005.846,12)
(2005.865,19)
(2005.885,9)
(2005.904,24)
(2005.923,16)
(2005.942,28)
(2005.962,37)
(2005.981,9)
(2006.0,3)
(2006.019,48)
(2006.038,100)
(2006.058,26)
(2006.077,44)
(2006.096,48)
(2006.115,40)
(2006.135,121)
(2006.154,102)
(2006.173,85)
(2006.192,118)
(2006.212,85)
(2006.231,91)
(2006.25,131)
(2006.269,91)
(2006.288,152)
(2006.308,108)
(2006.327,165)
(2006.346,107)
(2006.365,82)
(2006.385,68)
(2006.404,149)
(2006.423,109)
(2006.442,118)
(2006.462,99)
(2006.481,51)
(2006.5,26)
(2006.519,27)
(2006.538,32)
(2006.558,3)
(2006.577,16)
(2006.596,11)
(2006.615,9)
(2006.635,1)
(2006.654,2)
(2006.673,6)
(2006.692,5)
(2006.712,0)
(2006.731,3)
(2006.75,1)
(2006.769,9)
(2006.788,4)
(2006.808,11)
(2006.827,19)
(2006.846,7)
(2006.865,20)
(2006.885,17)
(2006.904,18)
(2006.923,7)
(2006.942,39)
(2006.962,31)
(2006.981,11)
(2007.0,30)
(2007.019,28)
(2007.038,98)
(2007.058,21)
(2007.077,38)
(2007.096,48)
(2007.115,22)
(2007.135,26)
(2007.154,89)
(2007.173,40)
(2007.192,67)
(2007.212,25)
(2007.231,105)
(2007.25,89)
(2007.269,21)
(2007.288,77)
(2007.308,86)
(2007.327,29)
(2007.346,46)
(2007.365,70)
(2007.385,103)
(2007.404,90)
(2007.423,30)
(2007.442,58)
(2007.462,47)
(2007.481,61)
(2007.5,39)
(2007.519,21)
(2007.538,3)
(2007.558,5)
(2007.577,15)
(2007.596,14)
(2007.615,3)
(2007.635,2)
(2007.654,1)
(2007.673,1)
(2007.692,0)
(2007.712,0)
(2007.731,0)
(2007.75,6)
(2007.769,2)
(2007.788,6)
(2007.808,6)
(2007.827,7)
(2007.846,0)
(2007.865,9)
(2007.885,10)
(2007.904,1)
(2007.923,24)
(2007.942,10)
(2007.962,2)
(2007.981,4)
(2008.0,2)
(2008.019,7)
(2008.038,15)
(2008.058,31)
(2008.077,11)
(2008.096,15)
(2008.115,27)
(2008.135,20)
(2008.154,15)
(2008.173,26)
(2008.192,26)
(2008.212,18)
(2008.231,31)
(2008.25,15)
(2008.269,48)
(2008.288,8)
(2008.308,41)
(2008.327,90)
(2008.346,14)
(2008.365,41)
(2008.385,4)
(2008.404,109)
(2008.423,89)
(2008.442,53)
(2008.462,23)
(2008.481,43)
(2008.5,43)
(2008.519,32)
(2008.538,11)
(2008.558,16)
(2008.577,11)
(2008.596,9)
(2008.615,1)
(2008.635,1)
(2008.654,1)
(2008.673,1)
(2008.692,1)
(2008.712,0)
(2008.731,6)
(2008.75,3)
(2008.769,5)
(2008.788,25)
(2008.808,17)
(2008.827,17)
(2008.846,17)
(2008.865,16)
(2008.885,22)
(2008.904,32)
(2008.923,27)
(2008.942,19)
(2008.962,14)
(2008.981,49)
(2009.0,13)
(2009.019,0)
(2009.038,57)
(2009.058,79)
(2009.077,31)
(2009.096,46)
(2009.115,70)
(2009.135,67)
(2009.154,77)
(2009.173,42)
(2009.192,65)
(2009.212,45)
(2009.231,87)
(2009.25,148)
(2009.269,199)
(2009.288,102)
(2009.308,111)
(2009.327,173)
(2009.346,52)
(2009.365,108)
(2009.385,146)
(2009.404,38)
(2009.423,93)
(2009.442,48)
(2009.462,115)
(2009.481,42)
(2009.5,62)
(2009.519,36)
(2009.538,62)
(2009.558,54)
(2009.577,34)
(2009.596,4)
(2009.615,4)
(2009.635,3)
(2009.654,3)
(2009.673,1)
(2009.692,6)
(2009.712,3)
(2009.731,1)
(2009.75,4)
(2009.769,15)
(2009.788,7)
(2009.808,33)
(2009.827,31)
(2009.846,11)
(2009.865,36)
(2009.885,10)
(2009.904,44)
(2009.923,21)
(2009.942,8)
(2009.962,21)
(2009.981,17)
(2010.0,4)
(2010.019,8)
(2010.038,53)
(2010.058,23)
(2010.077,29)
(2010.096,29)
(2010.115,4)
(2010.135,17)
(2010.154,24)
(2010.173,32)
(2010.192,12)
(2010.212,17)
(2010.231,52)
(2010.25,41)
(2010.269,55)
(2010.288,11)
(2010.308,61)
(2010.327,39)
(2010.346,19)
(2010.365,28)
(2010.385,58)
(2010.404,17)
(2010.423,2)
(2010.442,49)
(2010.462,36)
(2010.481,33)
(2010.5,32)
(2010.519,29)
(2010.538,13)
(2010.558,11)
(2010.577,3)
(2010.596,16)
(2010.615,6)
(2010.635,4)
(2010.654,15)
(2010.673,0)
(2010.692,1)
(2010.712,0)
(2010.731,2)
(2010.75,1)
(2010.769,8)
(2010.788,4)
(2010.808,10)
(2010.827,5)
(2010.846,8)
(2010.865,8)
(2010.885,20)
(2010.904,3)
(2010.923,31)
(2010.942,37)
(2010.962,26)
(2010.981,44)
(2011.0,28)
(2011.019,13)
(2011.038,16)
(2011.058,72)
(2011.077,24)
(2011.096,65)
(2011.115,24)
(2011.135,42)
(2011.154,58)
(2011.173,72)
(2011.192,53)
(2011.212,10)
(2011.231,36)
(2011.25,22)
(2011.269,150)
(2011.288,133)
(2011.308,78)
(2011.327,14)
(2011.346,25)
(2011.365,54)
(2011.385,71)
(2011.404,95)
(2011.423,74)
(2011.442,39)
(2011.462,48)
(2011.481,52)
(2011.5,24)
(2011.519,95)
(2011.538,27)
(2011.558,17)
(2011.577,8)
(2011.596,1)
(2011.615,4)
(2011.635,28)
(2011.654,13)
(2011.673,23)
(2011.692,2)
(2011.712,1)
(2011.731,2)
(2011.75,3)
(2011.769,4)
(2011.788,32)
(2011.808,2)
(2011.827,16)
(2011.846,71)
(2011.865,8)
(2011.885,71)
(2011.904,88)
(2011.923,50)
(2011.942,24)
(2011.962,5)
(2011.981,15)
(2012.0,7)
(2012.019,27)
(2012.038,20)
(2012.058,20)
(2012.077,2)
(2012.096,157)
(2012.115,53)
(2012.135,37)
(2012.154,41)
(2012.173,29)
(2012.192,42)
(2012.212,17)
(2012.231,9)
(2012.25,74)
(2012.269,40)
(2012.288,81)
(2012.308,48)
(2012.327,39)
(2012.346,41)
(2012.365,3)
(2012.385,40)
(2012.404,34)
(2012.423,18)
(2012.442,9)
(2012.462,22)
(2012.481,17)
(2012.5,21)
(2012.519,7)
(2012.538,10)
(2012.558,6)
(2012.577,7)
(2012.596,3)
(2012.615,1)
(2012.635,0)
(2012.654,12)
(2012.673,0)
(2012.692,0)
(2012.712,0)
(2012.731,0)
(2012.75,1)
(2012.769,0)
(2012.788,0)
(2012.808,0)
(2012.827,16)
(2012.846,8)
(2012.865,0)
(2012.885,19)
(2012.904,31)
(2012.923,8)
(2012.942,1)
(2012.962,28)
(2012.981,47)
(2013.0,8)
(2013.019,24)
(2013.038,0)
(2013.058,57)
(2013.077,1)
(2013.096,47)
(2013.115,67)
(2013.135,55)
(2013.154,35)
(2013.173,12)
(2013.192,49)
(2013.212,9)
(2013.231,1)
(2013.25,27)
(2013.269,8)
(2013.288,57)
(2013.308,8)
(2013.327,10)
(2013.346,18)
(2013.365,1)
(2013.385,9)
(2013.404,70)
(2013.423,64)
(2013.442,10)
(2013.462,1)
(2013.481,12)
(2013.5,37)
(2013.519,78)
(2013.538,13)
(2013.558,19)
(2013.577,1)
(2013.596,2)
(2013.615,35)
(2013.635,0)
(2013.654,41)
(2013.673,16)
(2013.692,3)
(2013.712,11)
(2013.731,0)
(2013.75,1)
(2013.769,2)
(2013.788,1)
(2013.808,9)
(2013.827,9)
(2013.846,19)
(2013.865,28)
(2013.885,15)
(2013.904,22)
(2013.923,2)
(2013.942,10)
(2013.962,20)
(2013.981,16)
(2014.0,59)
(2014.019,0)
(2014.038,0)
(2014.058,48)
(2014.077,69)
(2014.096,12)
(2014.115,67)
(2014.135,47)
(2014.154,77)
(2014.173,73)
(2014.192,56)
(2014.212,61)
(2014.231,1)
(2014.25,112)
(2014.269,54)
(2014.288,144)
(2014.308,61)
(2014.327,57)
(2014.346,21)
(2014.365,3)
(2014.385,44)
(2014.404,20)
(2014.423,30)
(2014.442,14)
(2014.462,72)
(2014.481,29)
(2014.5,35)
(2014.519,4)
(2014.538,31)
(2014.558,8)
(2014.577,17)
(2014.596,2)
(2014.615,93)
(2014.635,57)
(2014.654,25)
(2014.673,34)
(2014.692,0)
(2014.712,1)
(2014.731,0)
(2014.75,0)
(2014.769,7)
(2014.788,3)
(2014.808,1)
(2014.827,0)
(2014.846,1)
(2014.865,2)
(2014.885,0)
(2014.904,3)
(2014.923,3)
(2014.942,0)
(2014.962,0)
(2014.981,3)
(2015.0,0)
(2015.019,4)
(2015.038,15)
}

\def\FEJERDATA{
(2005.019,136)
(2005.038,51)
(2005.058,93)
(2005.077,52)
(2005.096,95)
(2005.115,74)
(2005.135,100)
(2005.154,111)
(2005.173,118)
(2005.192,93)
(2005.212,72)
(2005.231,115)
(2005.25,51)
(2005.269,78)
(2005.288,58)
(2005.308,83)
(2005.327,77)
(2005.346,59)
(2005.365,89)
(2005.385,93)
(2005.404,84)
(2005.423,75)
(2005.442,73)
(2005.462,56)
(2005.481,77)
(2005.5,31)
(2005.519,33)
(2005.538,24)
(2005.558,13)
(2005.577,13)
(2005.596,9)
(2005.615,1)
(2005.635,2)
(2005.654,4)
(2005.673,1)
(2005.692,1)
(2005.712,2)
(2005.731,0)
(2005.75,4)
(2005.769,10)
(2005.788,16)
(2005.808,25)
(2005.827,25)
(2005.846,30)
(2005.865,13)
(2005.885,18)
(2005.904,15)
(2005.923,28)
(2005.942,22)
(2005.962,30)
(2005.981,17)
(2006.0,18)
(2006.019,36)
(2006.038,27)
(2006.058,27)
(2006.077,15)
(2006.096,29)
(2006.115,15)
(2006.135,22)
(2006.154,42)
(2006.173,44)
(2006.192,38)
(2006.212,42)
(2006.231,44)
(2006.25,58)
(2006.269,56)
(2006.288,64)
(2006.308,29)
(2006.327,103)
(2006.346,26)
(2006.365,57)
(2006.385,39)
(2006.404,66)
(2006.423,26)
(2006.442,30)
(2006.462,47)
(2006.481,32)
(2006.5,39)
(2006.519,22)
(2006.538,8)
(2006.558,6)
(2006.577,2)
(2006.596,0)
(2006.615,4)
(2006.635,1)
(2006.654,2)
(2006.673,2)
(2006.692,2)
(2006.712,0)
(2006.731,6)
(2006.75,2)
(2006.769,6)
(2006.788,0)
(2006.808,7)
(2006.827,25)
(2006.846,20)
(2006.865,37)
(2006.885,24)
(2006.904,50)
(2006.923,61)
(2006.942,101)
(2006.962,79)
(2006.981,40)
(2007.0,43)
(2007.019,97)
(2007.038,66)
(2007.058,26)
(2007.077,116)
(2007.096,54)
(2007.115,36)
(2007.135,36)
(2007.154,90)
(2007.173,49)
(2007.192,93)
(2007.212,31)
(2007.231,88)
(2007.25,85)
(2007.269,66)
(2007.288,57)
(2007.308,78)
(2007.327,72)
(2007.346,20)
(2007.365,59)
(2007.385,46)
(2007.404,63)
(2007.423,36)
(2007.442,61)
(2007.462,13)
(2007.481,13)
(2007.5,29)
(2007.519,10)
(2007.538,6)
(2007.558,5)
(2007.577,3)
(2007.596,0)
(2007.615,0)
(2007.635,2)
(2007.654,0)
(2007.673,2)
(2007.692,3)
(2007.712,3)
(2007.731,5)
(2007.75,9)
(2007.769,12)
(2007.788,16)
(2007.808,13)
(2007.827,6)
(2007.846,4)
(2007.865,26)
(2007.885,25)
(2007.904,17)
(2007.923,44)
(2007.942,46)
(2007.962,71)
(2007.981,40)
(2008.0,59)
(2008.019,42)
(2008.038,124)
(2008.058,76)
(2008.077,37)
(2008.096,55)
(2008.115,43)
(2008.135,73)
(2008.154,102)
(2008.173,105)
(2008.192,73)
(2008.212,84)
(2008.231,84)
(2008.25,104)
(2008.269,130)
(2008.288,122)
(2008.308,101)
(2008.327,164)
(2008.346,49)
(2008.365,123)
(2008.385,55)
(2008.404,98)
(2008.423,134)
(2008.442,63)
(2008.462,78)
(2008.481,43)
(2008.5,45)
(2008.519,30)
(2008.538,16)
(2008.558,8)
(2008.577,3)
(2008.596,14)
(2008.615,4)
(2008.635,5)
(2008.654,1)
(2008.673,0)
(2008.692,2)
(2008.712,1)
(2008.731,1)
(2008.75,2)
(2008.769,1)
(2008.788,4)
(2008.808,15)
(2008.827,6)
(2008.846,8)
(2008.865,5)
(2008.885,4)
(2008.904,8)
(2008.923,6)
(2008.942,9)
(2008.962,17)
(2008.981,14)
(2009.0,2)
(2009.019,2)
(2009.038,18)
(2009.058,18)
(2009.077,13)
(2009.096,5)
(2009.115,14)
(2009.135,11)
(2009.154,8)
(2009.173,34)
(2009.192,3)
(2009.212,13)
(2009.231,16)
(2009.25,34)
(2009.269,31)
(2009.288,54)
(2009.308,37)
(2009.327,61)
(2009.346,22)
(2009.365,45)
(2009.385,35)
(2009.404,49)
(2009.423,48)
(2009.442,57)
(2009.462,28)
(2009.481,21)
(2009.5,36)
(2009.519,12)
(2009.538,11)
(2009.558,12)
(2009.577,3)
(2009.596,13)
(2009.615,0)
(2009.635,5)
(2009.654,0)
(2009.673,1)
(2009.692,1)
(2009.712,3)
(2009.731,3)
(2009.75,6)
(2009.769,4)
(2009.788,10)
(2009.808,10)
(2009.827,10)
(2009.846,10)
(2009.865,23)
(2009.885,17)
(2009.904,15)
(2009.923,46)
(2009.942,19)
(2009.962,38)
(2009.981,21)
(2010.0,15)
(2010.019,34)
(2010.038,83)
(2010.058,33)
(2010.077,40)
(2010.096,48)
(2010.115,46)
(2010.135,34)
(2010.154,31)
(2010.173,37)
(2010.192,24)
(2010.212,61)
(2010.231,51)
(2010.25,72)
(2010.269,63)
(2010.288,34)
(2010.308,101)
(2010.327,75)
(2010.346,55)
(2010.365,78)
(2010.385,56)
(2010.404,65)
(2010.423,33)
(2010.442,99)
(2010.462,46)
(2010.481,77)
(2010.5,72)
(2010.519,36)
(2010.538,29)
(2010.558,49)
(2010.577,13)
(2010.596,3)
(2010.615,3)
(2010.635,10)
(2010.654,4)
(2010.673,1)
(2010.692,2)
(2010.712,1)
(2010.731,3)
(2010.75,2)
(2010.769,0)
(2010.788,2)
(2010.808,1)
(2010.827,5)
(2010.846,4)
(2010.865,7)
(2010.885,5)
(2010.904,17)
(2010.923,12)
(2010.942,24)
(2010.962,6)
(2010.981,10)
(2011.0,24)
(2011.019,35)
(2011.038,30)
(2011.058,26)
(2011.077,29)
(2011.096,22)
(2011.115,20)
(2011.135,64)
(2011.154,41)
(2011.173,54)
(2011.192,69)
(2011.212,57)
(2011.231,61)
(2011.25,73)
(2011.269,96)
(2011.288,67)
(2011.308,114)
(2011.327,54)
(2011.346,51)
(2011.365,92)
(2011.385,58)
(2011.404,71)
(2011.423,42)
(2011.442,73)
(2011.462,56)
(2011.481,42)
(2011.5,53)
(2011.519,22)
(2011.538,35)
(2011.558,21)
(2011.577,15)
(2011.596,6)
(2011.615,4)
(2011.635,5)
(2011.654,0)
(2011.673,2)
(2011.692,0)
(2011.712,0)
(2011.731,0)
(2011.75,1)
(2011.769,7)
(2011.788,10)
(2011.808,9)
(2011.827,17)
(2011.846,16)
(2011.865,2)
(2011.885,16)
(2011.904,17)
(2011.923,17)
(2011.942,12)
(2011.962,12)
(2011.981,40)
(2012.0,20)
(2012.019,68)
(2012.038,36)
(2012.058,101)
(2012.077,21)
(2012.096,46)
(2012.115,63)
(2012.135,56)
(2012.154,92)
(2012.173,64)
(2012.192,104)
(2012.212,60)
(2012.231,68)
(2012.25,116)
(2012.269,83)
(2012.288,64)
(2012.308,41)
(2012.327,79)
(2012.346,79)
(2012.365,43)
(2012.385,104)
(2012.404,67)
(2012.423,119)
(2012.442,60)
(2012.462,64)
(2012.481,52)
(2012.5,72)
(2012.519,23)
(2012.538,61)
(2012.558,16)
(2012.577,18)
(2012.596,2)
(2012.615,5)
(2012.635,3)
(2012.654,3)
(2012.673,1)
(2012.692,0)
(2012.712,4)
(2012.731,0)
(2012.75,3)
(2012.769,0)
(2012.788,18)
(2012.808,5)
(2012.827,30)
(2012.846,12)
(2012.865,3)
(2012.885,38)
(2012.904,12)
(2012.923,16)
(2012.942,25)
(2012.962,14)
(2012.981,45)
(2013.0,19)
(2013.019,15)
(2013.038,15)
(2013.058,92)
(2013.077,22)
(2013.096,36)
(2013.115,29)
(2013.135,27)
(2013.154,24)
(2013.173,38)
(2013.192,18)
(2013.212,21)
(2013.231,29)
(2013.25,23)
(2013.269,10)
(2013.288,11)
(2013.308,23)
(2013.327,17)
(2013.346,33)
(2013.365,34)
(2013.385,30)
(2013.404,24)
(2013.423,40)
(2013.442,31)
(2013.462,20)
(2013.481,31)
(2013.5,8)
(2013.519,32)
(2013.538,17)
(2013.558,11)
(2013.577,6)
(2013.596,1)
(2013.615,3)
(2013.635,5)
(2013.654,0)
(2013.673,2)
(2013.692,0)
(2013.712,3)
(2013.731,0)
(2013.75,1)
(2013.769,11)
(2013.788,1)
(2013.808,12)
(2013.827,3)
(2013.846,25)
(2013.865,8)
(2013.885,14)
(2013.904,48)
(2013.923,13)
(2013.942,32)
(2013.962,7)
(2013.981,38)
(2014.0,37)
(2014.019,1)
(2014.038,37)
(2014.058,47)
(2014.077,61)
(2014.096,44)
(2014.115,24)
(2014.135,28)
(2014.154,18)
(2014.173,71)
(2014.192,23)
(2014.212,78)
(2014.231,25)
(2014.25,34)
(2014.269,54)
(2014.288,46)
(2014.308,41)
(2014.327,47)
(2014.346,1)
(2014.365,0)
(2014.385,88)
(2014.404,42)
(2014.423,22)
(2014.442,74)
(2014.462,20)
(2014.481,7)
(2014.5,60)
(2014.519,14)
(2014.538,15)
(2014.558,16)
(2014.577,11)
(2014.596,1)
(2014.615,3)
(2014.635,4)
(2014.654,3)
(2014.673,6)
(2014.692,1)
(2014.712,3)
(2014.731,3)
(2014.75,2)
(2014.769,3)
(2014.788,7)
(2014.808,18)
(2014.827,20)
(2014.846,22)
(2014.865,13)
(2014.885,18)
(2014.904,20)
(2014.923,28)
(2014.942,9)
(2014.962,16)
(2014.981,2)
(2015.0,7)
(2015.019,1)
(2015.038,11)
}

\def\GYORDATA{
(2005.019,120)
(2005.038,70)
(2005.058,84)
(2005.077,114)
(2005.096,131)
(2005.115,181)
(2005.135,118)
(2005.154,175)
(2005.173,105)
(2005.192,154)
(2005.212,107)
(2005.231,148)
(2005.25,92)
(2005.269,128)
(2005.288,91)
(2005.308,111)
(2005.327,119)
(2005.346,131)
(2005.365,146)
(2005.385,123)
(2005.404,122)
(2005.423,123)
(2005.442,128)
(2005.462,79)
(2005.481,93)
(2005.5,62)
(2005.519,66)
(2005.538,41)
(2005.558,35)
(2005.577,16)
(2005.596,9)
(2005.615,1)
(2005.635,12)
(2005.654,8)
(2005.673,7)
(2005.692,6)
(2005.712,2)
(2005.731,1)
(2005.75,2)
(2005.769,1)
(2005.788,5)
(2005.808,6)
(2005.827,8)
(2005.846,18)
(2005.865,20)
(2005.885,19)
(2005.904,27)
(2005.923,27)
(2005.942,34)
(2005.962,38)
(2005.981,25)
(2006.0,35)
(2006.019,47)
(2006.038,31)
(2006.058,28)
(2006.077,22)
(2006.096,35)
(2006.115,42)
(2006.135,59)
(2006.154,60)
(2006.173,82)
(2006.192,78)
(2006.212,33)
(2006.231,87)
(2006.25,49)
(2006.269,59)
(2006.288,52)
(2006.308,71)
(2006.327,76)
(2006.346,65)
(2006.365,68)
(2006.385,54)
(2006.404,69)
(2006.423,51)
(2006.442,85)
(2006.462,70)
(2006.481,66)
(2006.5,63)
(2006.519,118)
(2006.538,24)
(2006.558,29)
(2006.577,23)
(2006.596,12)
(2006.615,4)
(2006.635,2)
(2006.654,2)
(2006.673,2)
(2006.692,1)
(2006.712,1)
(2006.731,3)
(2006.75,7)
(2006.769,5)
(2006.788,9)
(2006.808,7)
(2006.827,8)
(2006.846,11)
(2006.865,9)
(2006.885,12)
(2006.904,13)
(2006.923,31)
(2006.942,31)
(2006.962,35)
(2006.981,41)
(2007.0,35)
(2007.019,65)
(2007.038,59)
(2007.058,56)
(2007.077,67)
(2007.096,66)
(2007.115,58)
(2007.135,74)
(2007.154,87)
(2007.173,71)
(2007.192,116)
(2007.212,63)
(2007.231,128)
(2007.25,107)
(2007.269,90)
(2007.288,92)
(2007.308,106)
(2007.327,94)
(2007.346,60)
(2007.365,110)
(2007.385,123)
(2007.404,108)
(2007.423,113)
(2007.442,56)
(2007.462,57)
(2007.481,61)
(2007.5,49)
(2007.519,24)
(2007.538,27)
(2007.558,7)
(2007.577,5)
(2007.596,4)
(2007.615,33)
(2007.635,7)
(2007.654,4)
(2007.673,6)
(2007.692,2)
(2007.712,3)
(2007.731,3)
(2007.75,5)
(2007.769,4)
(2007.788,13)
(2007.808,22)
(2007.827,12)
(2007.846,6)
(2007.865,30)
(2007.885,22)
(2007.904,32)
(2007.923,27)
(2007.942,35)
(2007.962,42)
(2007.981,86)
(2008.0,43)
(2008.019,32)
(2008.038,68)
(2008.058,137)
(2008.077,33)
(2008.096,43)
(2008.115,47)
(2008.135,41)
(2008.154,49)
(2008.173,35)
(2008.192,69)
(2008.212,53)
(2008.231,150)
(2008.25,87)
(2008.269,160)
(2008.288,102)
(2008.308,73)
(2008.327,106)
(2008.346,40)
(2008.365,97)
(2008.385,49)
(2008.404,98)
(2008.423,68)
(2008.442,74)
(2008.462,85)
(2008.481,40)
(2008.5,43)
(2008.519,28)
(2008.538,16)
(2008.558,23)
(2008.577,12)
(2008.596,17)
(2008.615,12)
(2008.635,8)
(2008.654,4)
(2008.673,5)
(2008.692,1)
(2008.712,0)
(2008.731,6)
(2008.75,0)
(2008.769,2)
(2008.788,9)
(2008.808,13)
(2008.827,17)
(2008.846,19)
(2008.865,33)
(2008.885,22)
(2008.904,35)
(2008.923,41)
(2008.942,55)
(2008.962,54)
(2008.981,78)
(2009.0,29)
(2009.019,67)
(2009.038,136)
(2009.058,68)
(2009.077,51)
(2009.096,56)
(2009.115,44)
(2009.135,73)
(2009.154,45)
(2009.173,79)
(2009.192,50)
(2009.212,68)
(2009.231,68)
(2009.25,85)
(2009.269,35)
(2009.288,124)
(2009.308,34)
(2009.327,129)
(2009.346,42)
(2009.365,90)
(2009.385,69)
(2009.404,48)
(2009.423,51)
(2009.442,53)
(2009.462,42)
(2009.481,57)
(2009.5,48)
(2009.519,52)
(2009.538,17)
(2009.558,11)
(2009.577,12)
(2009.596,7)
(2009.615,5)
(2009.635,0)
(2009.654,6)
(2009.673,1)
(2009.692,3)
(2009.712,0)
(2009.731,4)
(2009.75,5)
(2009.769,6)
(2009.788,9)
(2009.808,6)
(2009.827,6)
(2009.846,14)
(2009.865,8)
(2009.885,10)
(2009.904,16)
(2009.923,28)
(2009.942,40)
(2009.962,33)
(2009.981,28)
(2010.0,6)
(2010.019,18)
(2010.038,31)
(2010.058,21)
(2010.077,21)
(2010.096,38)
(2010.115,43)
(2010.135,31)
(2010.154,46)
(2010.173,62)
(2010.192,38)
(2010.212,95)
(2010.231,64)
(2010.25,72)
(2010.269,93)
(2010.288,75)
(2010.308,78)
(2010.327,84)
(2010.346,70)
(2010.365,81)
(2010.385,59)
(2010.404,85)
(2010.423,59)
(2010.442,97)
(2010.462,70)
(2010.481,104)
(2010.5,42)
(2010.519,35)
(2010.538,54)
(2010.558,48)
(2010.577,24)
(2010.596,21)
(2010.615,10)
(2010.635,5)
(2010.654,1)
(2010.673,0)
(2010.692,1)
(2010.712,0)
(2010.731,0)
(2010.75,2)
(2010.769,2)
(2010.788,1)
(2010.808,2)
(2010.827,4)
(2010.846,0)
(2010.865,3)
(2010.885,3)
(2010.904,5)
(2010.923,8)
(2010.942,4)
(2010.962,17)
(2010.981,26)
(2011.0,8)
(2011.019,20)
(2011.038,38)
(2011.058,52)
(2011.077,27)
(2011.096,86)
(2011.115,57)
(2011.135,76)
(2011.154,42)
(2011.173,67)
(2011.192,41)
(2011.212,63)
(2011.231,51)
(2011.25,48)
(2011.269,38)
(2011.288,60)
(2011.308,55)
(2011.327,60)
(2011.346,50)
(2011.365,51)
(2011.385,60)
(2011.404,52)
(2011.423,50)
(2011.442,34)
(2011.462,25)
(2011.481,19)
(2011.5,23)
(2011.519,21)
(2011.538,28)
(2011.558,23)
(2011.577,14)
(2011.596,4)
(2011.615,6)
(2011.635,4)
(2011.654,3)
(2011.673,3)
(2011.692,3)
(2011.712,4)
(2011.731,1)
(2011.75,3)
(2011.769,1)
(2011.788,13)
(2011.808,12)
(2011.827,18)
(2011.846,22)
(2011.865,24)
(2011.885,22)
(2011.904,68)
(2011.923,36)
(2011.942,44)
(2011.962,31)
(2011.981,50)
(2012.0,27)
(2012.019,27)
(2012.038,73)
(2012.058,64)
(2012.077,124)
(2012.096,75)
(2012.115,38)
(2012.135,69)
(2012.154,53)
(2012.173,68)
(2012.192,44)
(2012.212,68)
(2012.231,63)
(2012.25,69)
(2012.269,86)
(2012.288,46)
(2012.308,140)
(2012.327,76)
(2012.346,48)
(2012.365,68)
(2012.385,81)
(2012.404,79)
(2012.423,102)
(2012.442,82)
(2012.462,84)
(2012.481,65)
(2012.5,68)
(2012.519,64)
(2012.538,45)
(2012.558,34)
(2012.577,16)
(2012.596,6)
(2012.615,6)
(2012.635,2)
(2012.654,1)
(2012.673,5)
(2012.692,3)
(2012.712,2)
(2012.731,1)
(2012.75,6)
(2012.769,8)
(2012.788,4)
(2012.808,6)
(2012.827,21)
(2012.846,13)
(2012.865,13)
(2012.885,21)
(2012.904,30)
(2012.923,12)
(2012.942,39)
(2012.962,19)
(2012.981,55)
(2013.0,20)
(2013.019,22)
(2013.038,47)
(2013.058,93)
(2013.077,40)
(2013.096,56)
(2013.115,36)
(2013.135,40)
(2013.154,31)
(2013.173,42)
(2013.192,47)
(2013.212,31)
(2013.231,41)
(2013.25,55)
(2013.269,38)
(2013.288,49)
(2013.308,47)
(2013.327,42)
(2013.346,86)
(2013.365,54)
(2013.385,54)
(2013.404,45)
(2013.423,46)
(2013.442,48)
(2013.462,24)
(2013.481,21)
(2013.5,40)
(2013.519,27)
(2013.538,10)
(2013.558,30)
(2013.577,8)
(2013.596,30)
(2013.615,10)
(2013.635,1)
(2013.654,0)
(2013.673,1)
(2013.692,6)
(2013.712,3)
(2013.731,20)
(2013.75,1)
(2013.769,8)
(2013.788,18)
(2013.808,13)
(2013.827,8)
(2013.846,26)
(2013.865,10)
(2013.885,82)
(2013.904,40)
(2013.923,21)
(2013.942,53)
(2013.962,39)
(2013.981,40)
(2014.0,42)
(2014.019,0)
(2014.038,43)
(2014.058,81)
(2014.077,59)
(2014.096,14)
(2014.115,34)
(2014.135,9)
(2014.154,66)
(2014.173,56)
(2014.192,52)
(2014.212,52)
(2014.231,55)
(2014.25,85)
(2014.269,50)
(2014.288,35)
(2014.308,62)
(2014.327,15)
(2014.346,23)
(2014.365,7)
(2014.385,34)
(2014.404,87)
(2014.423,14)
(2014.442,41)
(2014.462,33)
(2014.481,37)
(2014.5,21)
(2014.519,9)
(2014.538,28)
(2014.558,7)
(2014.577,5)
(2014.596,7)
(2014.615,6)
(2014.635,4)
(2014.654,5)
(2014.673,3)
(2014.692,5)
(2014.712,0)
(2014.731,1)
(2014.75,3)
(2014.769,5)
(2014.788,3)
(2014.808,1)
(2014.827,17)
(2014.846,3)
(2014.865,11)
(2014.885,11)
(2014.904,7)
(2014.923,13)
(2014.942,23)
(2014.962,15)
(2014.981,30)
(2015.0,7)
(2015.019,9)
(2015.038,98)
}

\def\HAJDUDATA{
(2005.019,162)
(2005.038,84)
(2005.058,191)
(2005.077,107)
(2005.096,172)
(2005.115,157)
(2005.135,129)
(2005.154,138)
(2005.173,194)
(2005.192,119)
(2005.212,117)
(2005.231,171)
(2005.25,76)
(2005.269,207)
(2005.288,123)
(2005.308,134)
(2005.327,157)
(2005.346,111)
(2005.365,65)
(2005.385,99)
(2005.404,252)
(2005.423,135)
(2005.442,95)
(2005.462,161)
(2005.481,150)
(2005.5,40)
(2005.519,106)
(2005.538,49)
(2005.558,28)
(2005.577,18)
(2005.596,18)
(2005.615,5)
(2005.635,8)
(2005.654,3)
(2005.673,6)
(2005.692,1)
(2005.712,5)
(2005.731,5)
(2005.75,3)
(2005.769,21)
(2005.788,5)
(2005.808,32)
(2005.827,36)
(2005.846,20)
(2005.865,25)
(2005.885,14)
(2005.904,12)
(2005.923,27)
(2005.942,27)
(2005.962,11)
(2005.981,36)
(2006.0,55)
(2006.019,58)
(2006.038,51)
(2006.058,39)
(2006.077,64)
(2006.096,43)
(2006.115,81)
(2006.135,40)
(2006.154,69)
(2006.173,95)
(2006.192,44)
(2006.212,50)
(2006.231,48)
(2006.25,56)
(2006.269,35)
(2006.288,35)
(2006.308,18)
(2006.327,34)
(2006.346,25)
(2006.365,32)
(2006.385,37)
(2006.404,34)
(2006.423,25)
(2006.442,44)
(2006.462,61)
(2006.481,54)
(2006.5,30)
(2006.519,72)
(2006.538,17)
(2006.558,22)
(2006.577,4)
(2006.596,14)
(2006.615,13)
(2006.635,1)
(2006.654,2)
(2006.673,2)
(2006.692,1)
(2006.712,3)
(2006.731,3)
(2006.75,8)
(2006.769,7)
(2006.788,16)
(2006.808,16)
(2006.827,31)
(2006.846,46)
(2006.865,22)
(2006.885,42)
(2006.904,16)
(2006.923,57)
(2006.942,33)
(2006.962,44)
(2006.981,36)
(2007.0,32)
(2007.019,71)
(2007.038,65)
(2007.058,55)
(2007.077,40)
(2007.096,45)
(2007.115,63)
(2007.135,26)
(2007.154,85)
(2007.173,50)
(2007.192,39)
(2007.212,56)
(2007.231,66)
(2007.25,94)
(2007.269,83)
(2007.288,44)
(2007.308,80)
(2007.327,69)
(2007.346,15)
(2007.365,129)
(2007.385,76)
(2007.404,119)
(2007.423,113)
(2007.442,150)
(2007.462,120)
(2007.481,75)
(2007.5,73)
(2007.519,47)
(2007.538,33)
(2007.558,14)
(2007.577,9)
(2007.596,13)
(2007.615,2)
(2007.635,4)
(2007.654,4)
(2007.673,4)
(2007.692,1)
(2007.712,1)
(2007.731,3)
(2007.75,6)
(2007.769,11)
(2007.788,15)
(2007.808,33)
(2007.827,42)
(2007.846,37)
(2007.865,58)
(2007.885,63)
(2007.904,34)
(2007.923,67)
(2007.942,82)
(2007.962,155)
(2007.981,96)
(2008.0,106)
(2008.019,22)
(2008.038,22)
(2008.058,214)
(2008.077,80)
(2008.096,73)
(2008.115,38)
(2008.135,39)
(2008.154,114)
(2008.173,48)
(2008.192,71)
(2008.212,40)
(2008.231,85)
(2008.25,52)
(2008.269,93)
(2008.288,47)
(2008.308,75)
(2008.327,62)
(2008.346,49)
(2008.365,69)
(2008.385,59)
(2008.404,86)
(2008.423,130)
(2008.442,92)
(2008.462,112)
(2008.481,68)
(2008.5,53)
(2008.519,51)
(2008.538,53)
(2008.558,41)
(2008.577,4)
(2008.596,5)
(2008.615,4)
(2008.635,5)
(2008.654,1)
(2008.673,0)
(2008.692,2)
(2008.712,3)
(2008.731,1)
(2008.75,5)
(2008.769,1)
(2008.788,8)
(2008.808,9)
(2008.827,5)
(2008.846,13)
(2008.865,15)
(2008.885,24)
(2008.904,16)
(2008.923,14)
(2008.942,28)
(2008.962,26)
(2008.981,40)
(2009.0,14)
(2009.019,28)
(2009.038,81)
(2009.058,62)
(2009.077,21)
(2009.096,67)
(2009.115,83)
(2009.135,78)
(2009.154,97)
(2009.173,117)
(2009.192,119)
(2009.212,50)
(2009.231,92)
(2009.25,102)
(2009.269,88)
(2009.288,89)
(2009.308,71)
(2009.327,127)
(2009.346,40)
(2009.365,135)
(2009.385,118)
(2009.404,149)
(2009.423,162)
(2009.442,148)
(2009.462,262)
(2009.481,127)
(2009.5,146)
(2009.519,84)
(2009.538,117)
(2009.558,86)
(2009.577,17)
(2009.596,12)
(2009.615,6)
(2009.635,32)
(2009.654,0)
(2009.673,0)
(2009.692,0)
(2009.712,0)
(2009.731,1)
(2009.75,3)
(2009.769,3)
(2009.788,6)
(2009.808,10)
(2009.827,8)
(2009.846,32)
(2009.865,40)
(2009.885,43)
(2009.904,21)
(2009.923,26)
(2009.942,48)
(2009.962,39)
(2009.981,64)
(2010.0,19)
(2010.019,38)
(2010.038,95)
(2010.058,44)
(2010.077,46)
(2010.096,77)
(2010.115,31)
(2010.135,46)
(2010.154,63)
(2010.173,76)
(2010.192,40)
(2010.212,109)
(2010.231,113)
(2010.25,86)
(2010.269,108)
(2010.288,54)
(2010.308,94)
(2010.327,81)
(2010.346,74)
(2010.365,71)
(2010.385,93)
(2010.404,68)
(2010.423,46)
(2010.442,68)
(2010.462,49)
(2010.481,21)
(2010.5,61)
(2010.519,23)
(2010.538,21)
(2010.558,7)
(2010.577,12)
(2010.596,9)
(2010.615,4)
(2010.635,3)
(2010.654,3)
(2010.673,0)
(2010.692,1)
(2010.712,3)
(2010.731,3)
(2010.75,0)
(2010.769,2)
(2010.788,3)
(2010.808,1)
(2010.827,1)
(2010.846,3)
(2010.865,15)
(2010.885,36)
(2010.904,28)
(2010.923,10)
(2010.942,26)
(2010.962,20)
(2010.981,45)
(2011.0,15)
(2011.019,26)
(2011.038,25)
(2011.058,35)
(2011.077,25)
(2011.096,34)
(2011.115,28)
(2011.135,43)
(2011.154,46)
(2011.173,55)
(2011.192,24)
(2011.212,47)
(2011.231,59)
(2011.25,56)
(2011.269,58)
(2011.288,62)
(2011.308,83)
(2011.327,92)
(2011.346,49)
(2011.365,153)
(2011.385,119)
(2011.404,69)
(2011.423,138)
(2011.442,124)
(2011.462,108)
(2011.481,38)
(2011.5,92)
(2011.519,36)
(2011.538,21)
(2011.558,24)
(2011.577,20)
(2011.596,6)
(2011.615,15)
(2011.635,19)
(2011.654,6)
(2011.673,12)
(2011.692,1)
(2011.712,12)
(2011.731,2)
(2011.75,0)
(2011.769,4)
(2011.788,8)
(2011.808,4)
(2011.827,5)
(2011.846,24)
(2011.865,12)
(2011.885,24)
(2011.904,35)
(2011.923,41)
(2011.942,22)
(2011.962,19)
(2011.981,17)
(2012.0,32)
(2012.019,39)
(2012.038,41)
(2012.058,72)
(2012.077,20)
(2012.096,30)
(2012.115,32)
(2012.135,43)
(2012.154,25)
(2012.173,58)
(2012.192,46)
(2012.212,49)
(2012.231,19)
(2012.25,95)
(2012.269,60)
(2012.288,45)
(2012.308,50)
(2012.327,41)
(2012.346,41)
(2012.365,35)
(2012.385,65)
(2012.404,87)
(2012.423,53)
(2012.442,51)
(2012.462,105)
(2012.481,49)
(2012.5,52)
(2012.519,67)
(2012.538,37)
(2012.558,20)
(2012.577,8)
(2012.596,24)
(2012.615,11)
(2012.635,3)
(2012.654,0)
(2012.673,0)
(2012.692,1)
(2012.712,21)
(2012.731,0)
(2012.75,1)
(2012.769,1)
(2012.788,0)
(2012.808,10)
(2012.827,10)
(2012.846,11)
(2012.865,7)
(2012.885,7)
(2012.904,10)
(2012.923,8)
(2012.942,43)
(2012.962,31)
(2012.981,66)
(2013.0,58)
(2013.019,41)
(2013.038,120)
(2013.058,115)
(2013.077,113)
(2013.096,74)
(2013.115,107)
(2013.135,39)
(2013.154,113)
(2013.173,51)
(2013.192,89)
(2013.212,69)
(2013.231,62)
(2013.25,70)
(2013.269,180)
(2013.288,39)
(2013.308,114)
(2013.327,54)
(2013.346,34)
(2013.365,108)
(2013.385,79)
(2013.404,47)
(2013.423,48)
(2013.442,96)
(2013.462,57)
(2013.481,39)
(2013.5,59)
(2013.519,65)
(2013.538,9)
(2013.558,25)
(2013.577,61)
(2013.596,15)
(2013.615,4)
(2013.635,5)
(2013.654,10)
(2013.673,1)
(2013.692,0)
(2013.712,0)
(2013.731,8)
(2013.75,2)
(2013.769,2)
(2013.788,1)
(2013.808,12)
(2013.827,2)
(2013.846,20)
(2013.865,33)
(2013.885,19)
(2013.904,32)
(2013.923,32)
(2013.942,39)
(2013.962,26)
(2013.981,37)
(2014.0,19)
(2014.019,3)
(2014.038,29)
(2014.058,112)
(2014.077,54)
(2014.096,34)
(2014.115,40)
(2014.135,7)
(2014.154,38)
(2014.173,26)
(2014.192,55)
(2014.212,39)
(2014.231,59)
(2014.25,64)
(2014.269,97)
(2014.288,48)
(2014.308,87)
(2014.327,92)
(2014.346,9)
(2014.365,0)
(2014.385,93)
(2014.404,165)
(2014.423,176)
(2014.442,57)
(2014.462,36)
(2014.481,30)
(2014.5,43)
(2014.519,40)
(2014.538,12)
(2014.558,15)
(2014.577,27)
(2014.596,9)
(2014.615,2)
(2014.635,17)
(2014.654,9)
(2014.673,2)
(2014.692,4)
(2014.712,1)
(2014.731,1)
(2014.75,4)
(2014.769,3)
(2014.788,4)
(2014.808,1)
(2014.827,3)
(2014.846,1)
(2014.865,3)
(2014.885,11)
(2014.904,3)
(2014.923,9)
(2014.942,6)
(2014.962,14)
(2014.981,25)
(2015.0,4)
(2015.019,10)
(2015.038,61)
}

\def\HEVESDATA{
(2005.019,36)
(2005.038,28)
(2005.058,51)
(2005.077,42)
(2005.096,40)
(2005.115,44)
(2005.135,40)
(2005.154,60)
(2005.173,60)
(2005.192,34)
(2005.212,57)
(2005.231,21)
(2005.25,41)
(2005.269,42)
(2005.288,35)
(2005.308,70)
(2005.327,50)
(2005.346,51)
(2005.365,104)
(2005.385,96)
(2005.404,104)
(2005.423,103)
(2005.442,62)
(2005.462,83)
(2005.481,47)
(2005.5,42)
(2005.519,40)
(2005.538,18)
(2005.558,25)
(2005.577,12)
(2005.596,5)
(2005.615,5)
(2005.635,5)
(2005.654,5)
(2005.673,2)
(2005.692,2)
(2005.712,5)
(2005.731,1)
(2005.75,3)
(2005.769,7)
(2005.788,7)
(2005.808,27)
(2005.827,11)
(2005.846,21)
(2005.865,23)
(2005.885,20)
(2005.904,21)
(2005.923,33)
(2005.942,15)
(2005.962,25)
(2005.981,35)
(2006.0,25)
(2006.019,32)
(2006.038,28)
(2006.058,13)
(2006.077,33)
(2006.096,19)
(2006.115,21)
(2006.135,10)
(2006.154,34)
(2006.173,19)
(2006.192,24)
(2006.212,30)
(2006.231,50)
(2006.25,12)
(2006.269,18)
(2006.288,28)
(2006.308,16)
(2006.327,23)
(2006.346,37)
(2006.365,25)
(2006.385,30)
(2006.404,22)
(2006.423,34)
(2006.442,28)
(2006.462,44)
(2006.481,26)
(2006.5,43)
(2006.519,19)
(2006.538,11)
(2006.558,6)
(2006.577,6)
(2006.596,2)
(2006.615,3)
(2006.635,4)
(2006.654,1)
(2006.673,4)
(2006.692,3)
(2006.712,4)
(2006.731,2)
(2006.75,9)
(2006.769,2)
(2006.788,11)
(2006.808,13)
(2006.827,7)
(2006.846,25)
(2006.865,28)
(2006.885,36)
(2006.904,56)
(2006.923,54)
(2006.942,50)
(2006.962,68)
(2006.981,105)
(2007.0,82)
(2007.019,138)
(2007.038,210)
(2007.058,103)
(2007.077,151)
(2007.096,110)
(2007.115,126)
(2007.135,116)
(2007.154,147)
(2007.173,150)
(2007.192,173)
(2007.212,74)
(2007.231,184)
(2007.25,111)
(2007.269,86)
(2007.288,111)
(2007.308,146)
(2007.327,85)
(2007.346,18)
(2007.365,142)
(2007.385,67)
(2007.404,101)
(2007.423,40)
(2007.442,92)
(2007.462,40)
(2007.481,25)
(2007.5,15)
(2007.519,15)
(2007.538,14)
(2007.558,2)
(2007.577,9)
(2007.596,2)
(2007.615,1)
(2007.635,6)
(2007.654,0)
(2007.673,0)
(2007.692,3)
(2007.712,3)
(2007.731,3)
(2007.75,1)
(2007.769,2)
(2007.788,1)
(2007.808,3)
(2007.827,1)
(2007.846,4)
(2007.865,1)
(2007.885,1)
(2007.904,7)
(2007.923,37)
(2007.942,23)
(2007.962,27)
(2007.981,12)
(2008.0,23)
(2008.019,11)
(2008.038,46)
(2008.058,73)
(2008.077,15)
(2008.096,16)
(2008.115,7)
(2008.135,35)
(2008.154,37)
(2008.173,37)
(2008.192,18)
(2008.212,20)
(2008.231,17)
(2008.25,27)
(2008.269,44)
(2008.288,51)
(2008.308,23)
(2008.327,55)
(2008.346,20)
(2008.365,97)
(2008.385,26)
(2008.404,110)
(2008.423,76)
(2008.442,49)
(2008.462,24)
(2008.481,21)
(2008.5,29)
(2008.519,5)
(2008.538,14)
(2008.558,8)
(2008.577,7)
(2008.596,7)
(2008.615,3)
(2008.635,2)
(2008.654,2)
(2008.673,1)
(2008.692,2)
(2008.712,0)
(2008.731,1)
(2008.75,1)
(2008.769,4)
(2008.788,9)
(2008.808,14)
(2008.827,9)
(2008.846,6)
(2008.865,9)
(2008.885,6)
(2008.904,3)
(2008.923,38)
(2008.942,13)
(2008.962,29)
(2008.981,27)
(2009.0,13)
(2009.019,13)
(2009.038,36)
(2009.058,36)
(2009.077,22)
(2009.096,48)
(2009.115,30)
(2009.135,51)
(2009.154,49)
(2009.173,79)
(2009.192,52)
(2009.212,51)
(2009.231,49)
(2009.25,49)
(2009.269,95)
(2009.288,19)
(2009.308,53)
(2009.327,119)
(2009.346,53)
(2009.365,54)
(2009.385,69)
(2009.404,32)
(2009.423,38)
(2009.442,7)
(2009.462,86)
(2009.481,53)
(2009.5,46)
(2009.519,53)
(2009.538,36)
(2009.558,13)
(2009.577,21)
(2009.596,8)
(2009.615,9)
(2009.635,3)
(2009.654,4)
(2009.673,5)
(2009.692,1)
(2009.712,4)
(2009.731,1)
(2009.75,2)
(2009.769,0)
(2009.788,3)
(2009.808,0)
(2009.827,6)
(2009.846,12)
(2009.865,11)
(2009.885,16)
(2009.904,2)
(2009.923,43)
(2009.942,10)
(2009.962,26)
(2009.981,10)
(2010.0,8)
(2010.019,31)
(2010.038,30)
(2010.058,24)
(2010.077,18)
(2010.096,33)
(2010.115,40)
(2010.135,35)
(2010.154,47)
(2010.173,20)
(2010.192,41)
(2010.212,26)
(2010.231,28)
(2010.25,25)
(2010.269,24)
(2010.288,15)
(2010.308,17)
(2010.327,12)
(2010.346,5)
(2010.365,42)
(2010.385,11)
(2010.404,42)
(2010.423,18)
(2010.442,35)
(2010.462,28)
(2010.481,24)
(2010.5,50)
(2010.519,7)
(2010.538,20)
(2010.558,12)
(2010.577,1)
(2010.596,4)
(2010.615,0)
(2010.635,0)
(2010.654,1)
(2010.673,2)
(2010.692,0)
(2010.712,0)
(2010.731,2)
(2010.75,1)
(2010.769,2)
(2010.788,8)
(2010.808,10)
(2010.827,3)
(2010.846,2)
(2010.865,20)
(2010.885,26)
(2010.904,12)
(2010.923,32)
(2010.942,10)
(2010.962,18)
(2010.981,13)
(2011.0,10)
(2011.019,10)
(2011.038,7)
(2011.058,22)
(2011.077,5)
(2011.096,6)
(2011.115,17)
(2011.135,8)
(2011.154,41)
(2011.173,22)
(2011.192,30)
(2011.212,38)
(2011.231,25)
(2011.25,82)
(2011.269,57)
(2011.288,63)
(2011.308,57)
(2011.327,30)
(2011.346,38)
(2011.365,47)
(2011.385,90)
(2011.404,60)
(2011.423,95)
(2011.442,120)
(2011.462,97)
(2011.481,82)
(2011.5,57)
(2011.519,37)
(2011.538,39)
(2011.558,21)
(2011.577,43)
(2011.596,24)
(2011.615,12)
(2011.635,21)
(2011.654,8)
(2011.673,6)
(2011.692,6)
(2011.712,2)
(2011.731,4)
(2011.75,5)
(2011.769,8)
(2011.788,9)
(2011.808,6)
(2011.827,19)
(2011.846,7)
(2011.865,42)
(2011.885,7)
(2011.904,15)
(2011.923,37)
(2011.942,47)
(2011.962,33)
(2011.981,29)
(2012.0,58)
(2012.019,37)
(2012.038,87)
(2012.058,66)
(2012.077,43)
(2012.096,41)
(2012.115,23)
(2012.135,43)
(2012.154,104)
(2012.173,66)
(2012.192,50)
(2012.212,27)
(2012.231,23)
(2012.25,87)
(2012.269,39)
(2012.288,55)
(2012.308,25)
(2012.327,76)
(2012.346,71)
(2012.365,17)
(2012.385,19)
(2012.404,32)
(2012.423,44)
(2012.442,12)
(2012.462,19)
(2012.481,52)
(2012.5,54)
(2012.519,22)
(2012.538,12)
(2012.558,9)
(2012.577,4)
(2012.596,7)
(2012.615,1)
(2012.635,2)
(2012.654,3)
(2012.673,0)
(2012.692,0)
(2012.712,1)
(2012.731,0)
(2012.75,8)
(2012.769,17)
(2012.788,1)
(2012.808,1)
(2012.827,8)
(2012.846,0)
(2012.865,4)
(2012.885,0)
(2012.904,20)
(2012.923,2)
(2012.942,24)
(2012.962,20)
(2012.981,2)
(2013.0,3)
(2013.019,0)
(2013.038,28)
(2013.058,17)
(2013.077,28)
(2013.096,9)
(2013.115,44)
(2013.135,25)
(2013.154,22)
(2013.173,32)
(2013.192,29)
(2013.212,36)
(2013.231,31)
(2013.25,28)
(2013.269,42)
(2013.288,19)
(2013.308,69)
(2013.327,38)
(2013.346,45)
(2013.365,37)
(2013.385,16)
(2013.404,9)
(2013.423,11)
(2013.442,8)
(2013.462,7)
(2013.481,10)
(2013.5,24)
(2013.519,15)
(2013.538,4)
(2013.558,26)
(2013.577,10)
(2013.596,14)
(2013.615,7)
(2013.635,5)
(2013.654,9)
(2013.673,2)
(2013.692,2)
(2013.712,0)
(2013.731,0)
(2013.75,4)
(2013.769,2)
(2013.788,2)
(2013.808,1)
(2013.827,11)
(2013.846,20)
(2013.865,35)
(2013.885,66)
(2013.904,49)
(2013.923,20)
(2013.942,18)
(2013.962,64)
(2013.981,21)
(2014.0,82)
(2014.019,1)
(2014.038,18)
(2014.058,22)
(2014.077,84)
(2014.096,40)
(2014.115,27)
(2014.135,23)
(2014.154,30)
(2014.173,50)
(2014.192,14)
(2014.212,6)
(2014.231,53)
(2014.25,32)
(2014.269,31)
(2014.288,66)
(2014.308,43)
(2014.327,32)
(2014.346,18)
(2014.365,4)
(2014.385,31)
(2014.404,77)
(2014.423,96)
(2014.442,100)
(2014.462,38)
(2014.481,70)
(2014.5,76)
(2014.519,48)
(2014.538,11)
(2014.558,40)
(2014.577,33)
(2014.596,12)
(2014.615,8)
(2014.635,1)
(2014.654,12)
(2014.673,1)
(2014.692,0)
(2014.712,2)
(2014.731,0)
(2014.75,3)
(2014.769,0)
(2014.788,1)
(2014.808,0)
(2014.827,0)
(2014.846,3)
(2014.865,1)
(2014.885,1)
(2014.904,1)
(2014.923,3)
(2014.942,0)
(2014.962,10)
(2014.981,19)
(2015.0,2)
(2015.019,17)
(2015.038,38)
}

\def\JASZDATA{
(2005.019,130)
(2005.038,80)
(2005.058,64)
(2005.077,63)
(2005.096,61)
(2005.115,95)
(2005.135,88)
(2005.154,112)
(2005.173,67)
(2005.192,118)
(2005.212,72)
(2005.231,114)
(2005.25,82)
(2005.269,138)
(2005.288,91)
(2005.308,124)
(2005.327,141)
(2005.346,97)
(2005.365,224)
(2005.385,97)
(2005.404,174)
(2005.423,99)
(2005.442,163)
(2005.462,99)
(2005.481,123)
(2005.5,82)
(2005.519,84)
(2005.538,58)
(2005.558,25)
(2005.577,26)
(2005.596,18)
(2005.615,9)
(2005.635,2)
(2005.654,6)
(2005.673,5)
(2005.692,4)
(2005.712,7)
(2005.731,5)
(2005.75,9)
(2005.769,8)
(2005.788,12)
(2005.808,30)
(2005.827,20)
(2005.846,15)
(2005.865,23)
(2005.885,37)
(2005.904,25)
(2005.923,37)
(2005.942,45)
(2005.962,51)
(2005.981,49)
(2006.0,78)
(2006.019,45)
(2006.038,65)
(2006.058,30)
(2006.077,64)
(2006.096,30)
(2006.115,57)
(2006.135,47)
(2006.154,81)
(2006.173,74)
(2006.192,74)
(2006.212,102)
(2006.231,137)
(2006.25,82)
(2006.269,80)
(2006.288,104)
(2006.308,113)
(2006.327,104)
(2006.346,108)
(2006.365,89)
(2006.385,45)
(2006.404,120)
(2006.423,82)
(2006.442,34)
(2006.462,68)
(2006.481,64)
(2006.5,52)
(2006.519,23)
(2006.538,24)
(2006.558,2)
(2006.577,3)
(2006.596,8)
(2006.615,6)
(2006.635,1)
(2006.654,1)
(2006.673,0)
(2006.692,1)
(2006.712,2)
(2006.731,7)
(2006.75,10)
(2006.769,12)
(2006.788,15)
(2006.808,27)
(2006.827,17)
(2006.846,33)
(2006.865,53)
(2006.885,44)
(2006.904,41)
(2006.923,55)
(2006.942,59)
(2006.962,42)
(2006.981,59)
(2007.0,20)
(2007.019,65)
(2007.038,72)
(2007.058,36)
(2007.077,50)
(2007.096,45)
(2007.115,78)
(2007.135,126)
(2007.154,81)
(2007.173,93)
(2007.192,111)
(2007.212,56)
(2007.231,109)
(2007.25,80)
(2007.269,54)
(2007.288,58)
(2007.308,53)
(2007.327,27)
(2007.346,12)
(2007.365,83)
(2007.385,54)
(2007.404,65)
(2007.423,26)
(2007.442,45)
(2007.462,11)
(2007.481,9)
(2007.5,20)
(2007.519,6)
(2007.538,8)
(2007.558,7)
(2007.577,12)
(2007.596,8)
(2007.615,2)
(2007.635,3)
(2007.654,3)
(2007.673,4)
(2007.692,2)
(2007.712,0)
(2007.731,0)
(2007.75,1)
(2007.769,4)
(2007.788,3)
(2007.808,6)
(2007.827,5)
(2007.846,1)
(2007.865,2)
(2007.885,6)
(2007.904,6)
(2007.923,14)
(2007.942,14)
(2007.962,26)
(2007.981,19)
(2008.0,35)
(2008.019,35)
(2008.038,38)
(2008.058,38)
(2008.077,57)
(2008.096,26)
(2008.115,31)
(2008.135,30)
(2008.154,38)
(2008.173,50)
(2008.192,45)
(2008.212,56)
(2008.231,31)
(2008.25,51)
(2008.269,86)
(2008.288,39)
(2008.308,104)
(2008.327,80)
(2008.346,9)
(2008.365,114)
(2008.385,69)
(2008.404,58)
(2008.423,50)
(2008.442,83)
(2008.462,39)
(2008.481,46)
(2008.5,43)
(2008.519,19)
(2008.538,46)
(2008.558,17)
(2008.577,91)
(2008.596,10)
(2008.615,5)
(2008.635,4)
(2008.654,1)
(2008.673,1)
(2008.692,3)
(2008.712,1)
(2008.731,0)
(2008.75,4)
(2008.769,6)
(2008.788,9)
(2008.808,18)
(2008.827,16)
(2008.846,23)
(2008.865,35)
(2008.885,31)
(2008.904,24)
(2008.923,39)
(2008.942,18)
(2008.962,94)
(2008.981,65)
(2009.0,35)
(2009.019,55)
(2009.038,144)
(2009.058,97)
(2009.077,51)
(2009.096,65)
(2009.115,42)
(2009.135,91)
(2009.154,86)
(2009.173,89)
(2009.192,85)
(2009.212,104)
(2009.231,67)
(2009.25,72)
(2009.269,76)
(2009.288,107)
(2009.308,43)
(2009.327,92)
(2009.346,44)
(2009.365,51)
(2009.385,42)
(2009.404,109)
(2009.423,68)
(2009.442,48)
(2009.462,67)
(2009.481,48)
(2009.5,23)
(2009.519,42)
(2009.538,30)
(2009.558,23)
(2009.577,7)
(2009.596,6)
(2009.615,3)
(2009.635,3)
(2009.654,0)
(2009.673,1)
(2009.692,3)
(2009.712,6)
(2009.731,1)
(2009.75,1)
(2009.769,4)
(2009.788,2)
(2009.808,6)
(2009.827,3)
(2009.846,15)
(2009.865,13)
(2009.885,16)
(2009.904,35)
(2009.923,33)
(2009.942,25)
(2009.962,36)
(2009.981,24)
(2010.0,32)
(2010.019,16)
(2010.038,69)
(2010.058,29)
(2010.077,27)
(2010.096,54)
(2010.115,54)
(2010.135,67)
(2010.154,112)
(2010.173,83)
(2010.192,59)
(2010.212,98)
(2010.231,68)
(2010.25,92)
(2010.269,58)
(2010.288,52)
(2010.308,53)
(2010.327,47)
(2010.346,43)
(2010.365,34)
(2010.385,28)
(2010.404,37)
(2010.423,87)
(2010.442,78)
(2010.462,120)
(2010.481,84)
(2010.5,103)
(2010.519,44)
(2010.538,71)
(2010.558,19)
(2010.577,16)
(2010.596,10)
(2010.615,7)
(2010.635,8)
(2010.654,5)
(2010.673,8)
(2010.692,3)
(2010.712,2)
(2010.731,2)
(2010.75,6)
(2010.769,16)
(2010.788,6)
(2010.808,42)
(2010.827,53)
(2010.846,45)
(2010.865,43)
(2010.885,65)
(2010.904,65)
(2010.923,79)
(2010.942,55)
(2010.962,98)
(2010.981,97)
(2011.0,94)
(2011.019,105)
(2011.038,68)
(2011.058,90)
(2011.077,152)
(2011.096,107)
(2011.115,67)
(2011.135,101)
(2011.154,136)
(2011.173,82)
(2011.192,135)
(2011.212,161)
(2011.231,112)
(2011.25,106)
(2011.269,89)
(2011.288,87)
(2011.308,79)
(2011.327,58)
(2011.346,47)
(2011.365,79)
(2011.385,62)
(2011.404,57)
(2011.423,42)
(2011.442,41)
(2011.462,46)
(2011.481,43)
(2011.5,57)
(2011.519,12)
(2011.538,11)
(2011.558,8)
(2011.577,9)
(2011.596,14)
(2011.615,1)
(2011.635,2)
(2011.654,5)
(2011.673,3)
(2011.692,0)
(2011.712,1)
(2011.731,2)
(2011.75,3)
(2011.769,1)
(2011.788,4)
(2011.808,2)
(2011.827,1)
(2011.846,1)
(2011.865,5)
(2011.885,1)
(2011.904,16)
(2011.923,0)
(2011.942,12)
(2011.962,21)
(2011.981,25)
(2012.0,28)
(2012.019,27)
(2012.038,39)
(2012.058,39)
(2012.077,19)
(2012.096,18)
(2012.115,26)
(2012.135,13)
(2012.154,13)
(2012.173,29)
(2012.192,12)
(2012.212,13)
(2012.231,22)
(2012.25,10)
(2012.269,22)
(2012.288,13)
(2012.308,14)
(2012.327,27)
(2012.346,10)
(2012.365,14)
(2012.385,58)
(2012.404,15)
(2012.423,45)
(2012.442,14)
(2012.462,48)
(2012.481,40)
(2012.5,38)
(2012.519,10)
(2012.538,13)
(2012.558,14)
(2012.577,36)
(2012.596,1)
(2012.615,4)
(2012.635,5)
(2012.654,6)
(2012.673,3)
(2012.692,4)
(2012.712,3)
(2012.731,1)
(2012.75,2)
(2012.769,0)
(2012.788,3)
(2012.808,2)
(2012.827,10)
(2012.846,1)
(2012.865,1)
(2012.885,28)
(2012.904,43)
(2012.923,15)
(2012.942,46)
(2012.962,15)
(2012.981,47)
(2013.0,13)
(2013.019,20)
(2013.038,25)
(2013.058,35)
(2013.077,74)
(2013.096,60)
(2013.115,48)
(2013.135,18)
(2013.154,35)
(2013.173,38)
(2013.192,48)
(2013.212,51)
(2013.231,59)
(2013.25,59)
(2013.269,90)
(2013.288,61)
(2013.308,55)
(2013.327,78)
(2013.346,56)
(2013.365,32)
(2013.385,64)
(2013.404,20)
(2013.423,37)
(2013.442,30)
(2013.462,28)
(2013.481,24)
(2013.5,50)
(2013.519,10)
(2013.538,38)
(2013.558,11)
(2013.577,10)
(2013.596,12)
(2013.615,3)
(2013.635,1)
(2013.654,2)
(2013.673,3)
(2013.692,4)
(2013.712,2)
(2013.731,2)
(2013.75,1)
(2013.769,8)
(2013.788,2)
(2013.808,31)
(2013.827,6)
(2013.846,46)
(2013.865,18)
(2013.885,39)
(2013.904,38)
(2013.923,26)
(2013.942,39)
(2013.962,22)
(2013.981,12)
(2014.0,4)
(2014.019,0)
(2014.038,18)
(2014.058,27)
(2014.077,19)
(2014.096,4)
(2014.115,17)
(2014.135,16)
(2014.154,14)
(2014.173,30)
(2014.192,9)
(2014.212,23)
(2014.231,32)
(2014.25,29)
(2014.269,52)
(2014.288,39)
(2014.308,66)
(2014.327,69)
(2014.346,18)
(2014.365,11)
(2014.385,118)
(2014.404,136)
(2014.423,109)
(2014.442,105)
(2014.462,86)
(2014.481,14)
(2014.5,132)
(2014.519,44)
(2014.538,46)
(2014.558,33)
(2014.577,7)
(2014.596,51)
(2014.615,27)
(2014.635,15)
(2014.654,7)
(2014.673,10)
(2014.692,1)
(2014.712,12)
(2014.731,2)
(2014.75,7)
(2014.769,6)
(2014.788,9)
(2014.808,26)
(2014.827,17)
(2014.846,16)
(2014.865,11)
(2014.885,11)
(2014.904,14)
(2014.923,14)
(2014.942,24)
(2014.962,56)
(2014.981,34)
(2015.0,30)
(2015.019,27)
(2015.038,112)
}

\def\KOMAROMDATA{
(2005.019,57)
(2005.038,50)
(2005.058,46)
(2005.077,54)
(2005.096,49)
(2005.115,97)
(2005.135,56)
(2005.154,70)
(2005.173,46)
(2005.192,73)
(2005.212,91)
(2005.231,82)
(2005.25,72)
(2005.269,66)
(2005.288,51)
(2005.308,63)
(2005.327,59)
(2005.346,64)
(2005.365,124)
(2005.385,83)
(2005.404,105)
(2005.423,74)
(2005.442,146)
(2005.462,82)
(2005.481,83)
(2005.5,81)
(2005.519,72)
(2005.538,43)
(2005.558,33)
(2005.577,20)
(2005.596,17)
(2005.615,4)
(2005.635,4)
(2005.654,3)
(2005.673,5)
(2005.692,1)
(2005.712,4)
(2005.731,4)
(2005.75,7)
(2005.769,21)
(2005.788,9)
(2005.808,22)
(2005.827,43)
(2005.846,48)
(2005.865,51)
(2005.885,52)
(2005.904,17)
(2005.923,49)
(2005.942,50)
(2005.962,31)
(2005.981,37)
(2006.0,20)
(2006.019,72)
(2006.038,55)
(2006.058,39)
(2006.077,46)
(2006.096,24)
(2006.115,38)
(2006.135,28)
(2006.154,23)
(2006.173,53)
(2006.192,29)
(2006.212,34)
(2006.231,34)
(2006.25,36)
(2006.269,61)
(2006.288,45)
(2006.308,59)
(2006.327,54)
(2006.346,44)
(2006.365,30)
(2006.385,70)
(2006.404,38)
(2006.423,87)
(2006.442,25)
(2006.462,88)
(2006.481,51)
(2006.5,46)
(2006.519,41)
(2006.538,15)
(2006.558,8)
(2006.577,1)
(2006.596,7)
(2006.615,12)
(2006.635,2)
(2006.654,4)
(2006.673,0)
(2006.692,4)
(2006.712,6)
(2006.731,4)
(2006.75,8)
(2006.769,3)
(2006.788,13)
(2006.808,4)
(2006.827,5)
(2006.846,9)
(2006.865,11)
(2006.885,4)
(2006.904,7)
(2006.923,14)
(2006.942,6)
(2006.962,22)
(2006.981,11)
(2007.0,4)
(2007.019,13)
(2007.038,42)
(2007.058,10)
(2007.077,24)
(2007.096,34)
(2007.115,34)
(2007.135,28)
(2007.154,29)
(2007.173,59)
(2007.192,48)
(2007.212,25)
(2007.231,35)
(2007.25,90)
(2007.269,70)
(2007.288,74)
(2007.308,32)
(2007.327,50)
(2007.346,18)
(2007.365,46)
(2007.385,75)
(2007.404,59)
(2007.423,55)
(2007.442,54)
(2007.462,56)
(2007.481,21)
(2007.5,26)
(2007.519,11)
(2007.538,10)
(2007.558,11)
(2007.577,18)
(2007.596,2)
(2007.615,2)
(2007.635,1)
(2007.654,0)
(2007.673,2)
(2007.692,3)
(2007.712,1)
(2007.731,2)
(2007.75,13)
(2007.769,2)
(2007.788,15)
(2007.808,16)
(2007.827,10)
(2007.846,2)
(2007.865,11)
(2007.885,4)
(2007.904,24)
(2007.923,15)
(2007.942,55)
(2007.962,22)
(2007.981,48)
(2008.0,15)
(2008.019,78)
(2008.038,70)
(2008.058,160)
(2008.077,28)
(2008.096,34)
(2008.115,47)
(2008.135,27)
(2008.154,27)
(2008.173,15)
(2008.192,26)
(2008.212,39)
(2008.231,47)
(2008.25,26)
(2008.269,60)
(2008.288,28)
(2008.308,31)
(2008.327,14)
(2008.346,82)
(2008.365,23)
(2008.385,31)
(2008.404,72)
(2008.423,60)
(2008.442,30)
(2008.462,65)
(2008.481,55)
(2008.5,58)
(2008.519,34)
(2008.538,23)
(2008.558,18)
(2008.577,18)
(2008.596,8)
(2008.615,10)
(2008.635,2)
(2008.654,4)
(2008.673,2)
(2008.692,2)
(2008.712,0)
(2008.731,3)
(2008.75,7)
(2008.769,6)
(2008.788,12)
(2008.808,14)
(2008.827,2)
(2008.846,1)
(2008.865,24)
(2008.885,12)
(2008.904,6)
(2008.923,11)
(2008.942,23)
(2008.962,12)
(2008.981,52)
(2009.0,9)
(2009.019,9)
(2009.038,63)
(2009.058,26)
(2009.077,50)
(2009.096,27)
(2009.115,19)
(2009.135,39)
(2009.154,46)
(2009.173,53)
(2009.192,58)
(2009.212,48)
(2009.231,24)
(2009.25,55)
(2009.269,75)
(2009.288,28)
(2009.308,32)
(2009.327,50)
(2009.346,31)
(2009.365,36)
(2009.385,55)
(2009.404,42)
(2009.423,65)
(2009.442,19)
(2009.462,59)
(2009.481,33)
(2009.5,38)
(2009.519,21)
(2009.538,22)
(2009.558,13)
(2009.577,6)
(2009.596,6)
(2009.615,1)
(2009.635,9)
(2009.654,1)
(2009.673,0)
(2009.692,3)
(2009.712,0)
(2009.731,2)
(2009.75,2)
(2009.769,4)
(2009.788,3)
(2009.808,7)
(2009.827,5)
(2009.846,6)
(2009.865,21)
(2009.885,24)
(2009.904,6)
(2009.923,50)
(2009.942,11)
(2009.962,35)
(2009.981,27)
(2010.0,16)
(2010.019,22)
(2010.038,26)
(2010.058,30)
(2010.077,23)
(2010.096,33)
(2010.115,22)
(2010.135,22)
(2010.154,34)
(2010.173,61)
(2010.192,22)
(2010.212,83)
(2010.231,43)
(2010.25,85)
(2010.269,55)
(2010.288,52)
(2010.308,39)
(2010.327,81)
(2010.346,49)
(2010.365,52)
(2010.385,43)
(2010.404,95)
(2010.423,37)
(2010.442,52)
(2010.462,52)
(2010.481,80)
(2010.5,54)
(2010.519,20)
(2010.538,20)
(2010.558,18)
(2010.577,2)
(2010.596,1)
(2010.615,3)
(2010.635,2)
(2010.654,2)
(2010.673,3)
(2010.692,0)
(2010.712,2)
(2010.731,0)
(2010.75,1)
(2010.769,3)
(2010.788,6)
(2010.808,4)
(2010.827,19)
(2010.846,11)
(2010.865,18)
(2010.885,23)
(2010.904,23)
(2010.923,8)
(2010.942,27)
(2010.962,30)
(2010.981,23)
(2011.0,19)
(2011.019,42)
(2011.038,18)
(2011.058,63)
(2011.077,55)
(2011.096,39)
(2011.115,24)
(2011.135,34)
(2011.154,41)
(2011.173,61)
(2011.192,28)
(2011.212,51)
(2011.231,55)
(2011.25,74)
(2011.269,28)
(2011.288,52)
(2011.308,39)
(2011.327,37)
(2011.346,20)
(2011.365,52)
(2011.385,18)
(2011.404,47)
(2011.423,64)
(2011.442,42)
(2011.462,38)
(2011.481,23)
(2011.5,33)
(2011.519,21)
(2011.538,14)
(2011.558,14)
(2011.577,11)
(2011.596,6)
(2011.615,2)
(2011.635,2)
(2011.654,2)
(2011.673,3)
(2011.692,1)
(2011.712,2)
(2011.731,3)
(2011.75,3)
(2011.769,2)
(2011.788,14)
(2011.808,13)
(2011.827,19)
(2011.846,18)
(2011.865,19)
(2011.885,18)
(2011.904,22)
(2011.923,25)
(2011.942,27)
(2011.962,12)
(2011.981,20)
(2012.0,12)
(2012.019,19)
(2012.038,34)
(2012.058,27)
(2012.077,31)
(2012.096,8)
(2012.115,24)
(2012.135,39)
(2012.154,18)
(2012.173,42)
(2012.192,35)
(2012.212,34)
(2012.231,24)
(2012.25,22)
(2012.269,59)
(2012.288,18)
(2012.308,16)
(2012.327,23)
(2012.346,12)
(2012.365,13)
(2012.385,28)
(2012.404,29)
(2012.423,33)
(2012.442,7)
(2012.462,45)
(2012.481,12)
(2012.5,25)
(2012.519,5)
(2012.538,19)
(2012.558,9)
(2012.577,3)
(2012.596,2)
(2012.615,1)
(2012.635,3)
(2012.654,1)
(2012.673,0)
(2012.692,0)
(2012.712,1)
(2012.731,2)
(2012.75,0)
(2012.769,3)
(2012.788,5)
(2012.808,1)
(2012.827,1)
(2012.846,12)
(2012.865,6)
(2012.885,18)
(2012.904,10)
(2012.923,14)
(2012.942,12)
(2012.962,9)
(2012.981,12)
(2013.0,1)
(2013.019,8)
(2013.038,0)
(2013.058,21)
(2013.077,14)
(2013.096,3)
(2013.115,11)
(2013.135,19)
(2013.154,9)
(2013.173,2)
(2013.192,19)
(2013.212,1)
(2013.231,0)
(2013.25,18)
(2013.269,46)
(2013.288,11)
(2013.308,35)
(2013.327,13)
(2013.346,34)
(2013.365,17)
(2013.385,48)
(2013.404,28)
(2013.423,19)
(2013.442,39)
(2013.462,9)
(2013.481,28)
(2013.5,22)
(2013.519,8)
(2013.538,19)
(2013.558,12)
(2013.577,13)
(2013.596,21)
(2013.615,34)
(2013.635,5)
(2013.654,0)
(2013.673,0)
(2013.692,0)
(2013.712,0)
(2013.731,0)
(2013.75,4)
(2013.769,1)
(2013.788,0)
(2013.808,0)
(2013.827,1)
(2013.846,0)
(2013.865,6)
(2013.885,0)
(2013.904,2)
(2013.923,1)
(2013.942,1)
(2013.962,3)
(2013.981,11)
(2014.0,2)
(2014.019,0)
(2014.038,2)
(2014.058,10)
(2014.077,7)
(2014.096,10)
(2014.115,21)
(2014.135,4)
(2014.154,17)
(2014.173,22)
(2014.192,13)
(2014.212,0)
(2014.231,21)
(2014.25,8)
(2014.269,14)
(2014.288,11)
(2014.308,5)
(2014.327,7)
(2014.346,1)
(2014.365,5)
(2014.385,17)
(2014.404,37)
(2014.423,50)
(2014.442,4)
(2014.462,27)
(2014.481,18)
(2014.5,26)
(2014.519,27)
(2014.538,2)
(2014.558,15)
(2014.577,19)
(2014.596,14)
(2014.615,7)
(2014.635,11)
(2014.654,5)
(2014.673,0)
(2014.692,0)
(2014.712,0)
(2014.731,0)
(2014.75,6)
(2014.769,3)
(2014.788,1)
(2014.808,11)
(2014.827,0)
(2014.846,4)
(2014.865,3)
(2014.885,1)
(2014.904,10)
(2014.923,8)
(2014.942,2)
(2014.962,7)
(2014.981,20)
(2015.0,36)
(2015.019,17)
(2015.038,61)
}

\def\NOGRADDATA{
(2005.019,2)
(2005.038,29)
(2005.058,4)
(2005.077,14)
(2005.096,11)
(2005.115,26)
(2005.135,10)
(2005.154,21)
(2005.173,12)
(2005.192,6)
(2005.212,9)
(2005.231,10)
(2005.25,25)
(2005.269,6)
(2005.288,26)
(2005.308,7)
(2005.327,26)
(2005.346,27)
(2005.365,22)
(2005.385,52)
(2005.404,44)
(2005.423,55)
(2005.442,39)
(2005.462,29)
(2005.481,30)
(2005.5,10)
(2005.519,10)
(2005.538,12)
(2005.558,7)
(2005.577,2)
(2005.596,2)
(2005.615,5)
(2005.635,2)
(2005.654,0)
(2005.673,0)
(2005.692,3)
(2005.712,6)
(2005.731,4)
(2005.75,0)
(2005.769,4)
(2005.788,8)
(2005.808,15)
(2005.827,15)
(2005.846,13)
(2005.865,26)
(2005.885,10)
(2005.904,21)
(2005.923,39)
(2005.942,17)
(2005.962,28)
(2005.981,7)
(2006.0,44)
(2006.019,51)
(2006.038,48)
(2006.058,23)
(2006.077,53)
(2006.096,39)
(2006.115,33)
(2006.135,23)
(2006.154,24)
(2006.173,45)
(2006.192,39)
(2006.212,47)
(2006.231,52)
(2006.25,49)
(2006.269,53)
(2006.288,37)
(2006.308,40)
(2006.327,37)
(2006.346,32)
(2006.365,50)
(2006.385,22)
(2006.404,58)
(2006.423,31)
(2006.442,30)
(2006.462,57)
(2006.481,50)
(2006.5,27)
(2006.519,14)
(2006.538,5)
(2006.558,7)
(2006.577,4)
(2006.596,5)
(2006.615,3)
(2006.635,1)
(2006.654,0)
(2006.673,2)
(2006.692,2)
(2006.712,2)
(2006.731,2)
(2006.75,9)
(2006.769,12)
(2006.788,19)
(2006.808,4)
(2006.827,17)
(2006.846,7)
(2006.865,15)
(2006.885,18)
(2006.904,62)
(2006.923,41)
(2006.942,59)
(2006.962,8)
(2006.981,46)
(2007.0,1)
(2007.019,30)
(2007.038,106)
(2007.058,56)
(2007.077,64)
(2007.096,46)
(2007.115,36)
(2007.135,60)
(2007.154,96)
(2007.173,38)
(2007.192,112)
(2007.212,47)
(2007.231,108)
(2007.25,77)
(2007.269,30)
(2007.288,43)
(2007.308,71)
(2007.327,106)
(2007.346,76)
(2007.365,107)
(2007.385,88)
(2007.404,73)
(2007.423,68)
(2007.442,100)
(2007.462,59)
(2007.481,25)
(2007.5,21)
(2007.519,18)
(2007.538,11)
(2007.558,6)
(2007.577,8)
(2007.596,3)
(2007.615,1)
(2007.635,1)
(2007.654,0)
(2007.673,0)
(2007.692,0)
(2007.712,4)
(2007.731,3)
(2007.75,2)
(2007.769,0)
(2007.788,7)
(2007.808,12)
(2007.827,9)
(2007.846,19)
(2007.865,29)
(2007.885,27)
(2007.904,11)
(2007.923,32)
(2007.942,33)
(2007.962,25)
(2007.981,14)
(2008.0,0)
(2008.019,21)
(2008.038,39)
(2008.058,74)
(2008.077,34)
(2008.096,8)
(2008.115,25)
(2008.135,4)
(2008.154,21)
(2008.173,26)
(2008.192,49)
(2008.212,25)
(2008.231,69)
(2008.25,19)
(2008.269,55)
(2008.288,26)
(2008.308,37)
(2008.327,39)
(2008.346,4)
(2008.365,33)
(2008.385,17)
(2008.404,42)
(2008.423,11)
(2008.442,23)
(2008.462,15)
(2008.481,7)
(2008.5,8)
(2008.519,10)
(2008.538,13)
(2008.558,3)
(2008.577,11)
(2008.596,4)
(2008.615,3)
(2008.635,4)
(2008.654,0)
(2008.673,0)
(2008.692,0)
(2008.712,0)
(2008.731,1)
(2008.75,2)
(2008.769,4)
(2008.788,6)
(2008.808,9)
(2008.827,14)
(2008.846,15)
(2008.865,19)
(2008.885,17)
(2008.904,60)
(2008.923,44)
(2008.942,34)
(2008.962,52)
(2008.981,53)
(2009.0,3)
(2009.019,44)
(2009.038,86)
(2009.058,35)
(2009.077,58)
(2009.096,65)
(2009.115,49)
(2009.135,43)
(2009.154,45)
(2009.173,59)
(2009.192,58)
(2009.212,38)
(2009.231,87)
(2009.25,34)
(2009.269,64)
(2009.288,40)
(2009.308,23)
(2009.327,69)
(2009.346,52)
(2009.365,30)
(2009.385,20)
(2009.404,21)
(2009.423,17)
(2009.442,16)
(2009.462,25)
(2009.481,25)
(2009.5,50)
(2009.519,21)
(2009.538,12)
(2009.558,22)
(2009.577,7)
(2009.596,2)
(2009.615,1)
(2009.635,2)
(2009.654,1)
(2009.673,0)
(2009.692,2)
(2009.712,3)
(2009.731,0)
(2009.75,9)
(2009.769,2)
(2009.788,1)
(2009.808,4)
(2009.827,2)
(2009.846,0)
(2009.865,1)
(2009.885,1)
(2009.904,1)
(2009.923,4)
(2009.942,12)
(2009.962,17)
(2009.981,9)
(2010.0,6)
(2010.019,4)
(2010.038,16)
(2010.058,14)
(2010.077,13)
(2010.096,23)
(2010.115,12)
(2010.135,38)
(2010.154,11)
(2010.173,20)
(2010.192,9)
(2010.212,13)
(2010.231,11)
(2010.25,29)
(2010.269,26)
(2010.288,11)
(2010.308,26)
(2010.327,16)
(2010.346,19)
(2010.365,37)
(2010.385,20)
(2010.404,20)
(2010.423,26)
(2010.442,41)
(2010.462,17)
(2010.481,21)
(2010.5,8)
(2010.519,15)
(2010.538,8)
(2010.558,3)
(2010.577,4)
(2010.596,2)
(2010.615,3)
(2010.635,7)
(2010.654,1)
(2010.673,1)
(2010.692,2)
(2010.712,2)
(2010.731,2)
(2010.75,1)
(2010.769,0)
(2010.788,2)
(2010.808,3)
(2010.827,9)
(2010.846,1)
(2010.865,8)
(2010.885,22)
(2010.904,5)
(2010.923,13)
(2010.942,12)
(2010.962,16)
(2010.981,17)
(2011.0,11)
(2011.019,8)
(2011.038,29)
(2011.058,13)
(2011.077,19)
(2011.096,9)
(2011.115,11)
(2011.135,18)
(2011.154,5)
(2011.173,14)
(2011.192,4)
(2011.212,6)
(2011.231,6)
(2011.25,29)
(2011.269,2)
(2011.288,7)
(2011.308,10)
(2011.327,2)
(2011.346,3)
(2011.365,2)
(2011.385,9)
(2011.404,35)
(2011.423,3)
(2011.442,44)
(2011.462,27)
(2011.481,20)
(2011.5,4)
(2011.519,12)
(2011.538,3)
(2011.558,5)
(2011.577,4)
(2011.596,3)
(2011.615,2)
(2011.635,2)
(2011.654,5)
(2011.673,1)
(2011.692,0)
(2011.712,9)
(2011.731,1)
(2011.75,10)
(2011.769,2)
(2011.788,16)
(2011.808,10)
(2011.827,3)
(2011.846,25)
(2011.865,3)
(2011.885,25)
(2011.904,32)
(2011.923,16)
(2011.942,24)
(2011.962,35)
(2011.981,58)
(2012.0,43)
(2012.019,62)
(2012.038,45)
(2012.058,38)
(2012.077,45)
(2012.096,88)
(2012.115,30)
(2012.135,76)
(2012.154,30)
(2012.173,52)
(2012.192,65)
(2012.212,41)
(2012.231,15)
(2012.25,76)
(2012.269,22)
(2012.288,27)
(2012.308,22)
(2012.327,20)
(2012.346,21)
(2012.365,2)
(2012.385,37)
(2012.404,14)
(2012.423,22)
(2012.442,14)
(2012.462,20)
(2012.481,36)
(2012.5,8)
(2012.519,36)
(2012.538,9)
(2012.558,4)
(2012.577,13)
(2012.596,1)
(2012.615,3)
(2012.635,0)
(2012.654,2)
(2012.673,1)
(2012.692,0)
(2012.712,0)
(2012.731,2)
(2012.75,11)
(2012.769,1)
(2012.788,0)
(2012.808,0)
(2012.827,0)
(2012.846,37)
(2012.865,26)
(2012.885,68)
(2012.904,30)
(2012.923,26)
(2012.942,47)
(2012.962,21)
(2012.981,13)
(2013.0,13)
(2013.019,0)
(2013.038,22)
(2013.058,28)
(2013.077,58)
(2013.096,35)
(2013.115,33)
(2013.135,76)
(2013.154,20)
(2013.173,50)
(2013.192,38)
(2013.212,23)
(2013.231,32)
(2013.25,51)
(2013.269,23)
(2013.288,44)
(2013.308,28)
(2013.327,34)
(2013.346,61)
(2013.365,30)
(2013.385,54)
(2013.404,44)
(2013.423,49)
(2013.442,40)
(2013.462,27)
(2013.481,53)
(2013.5,35)
(2013.519,62)
(2013.538,32)
(2013.558,29)
(2013.577,25)
(2013.596,25)
(2013.615,7)
(2013.635,7)
(2013.654,0)
(2013.673,2)
(2013.692,1)
(2013.712,0)
(2013.731,1)
(2013.75,2)
(2013.769,1)
(2013.788,2)
(2013.808,0)
(2013.827,5)
(2013.846,4)
(2013.865,1)
(2013.885,5)
(2013.904,2)
(2013.923,4)
(2013.942,10)
(2013.962,14)
(2013.981,6)
(2014.0,22)
(2014.019,0)
(2014.038,13)
(2014.058,21)
(2014.077,18)
(2014.096,14)
(2014.115,10)
(2014.135,13)
(2014.154,27)
(2014.173,30)
(2014.192,26)
(2014.212,28)
(2014.231,32)
(2014.25,39)
(2014.269,26)
(2014.288,40)
(2014.308,46)
(2014.327,44)
(2014.346,26)
(2014.365,0)
(2014.385,13)
(2014.404,84)
(2014.423,23)
(2014.442,2)
(2014.462,16)
(2014.481,8)
(2014.5,8)
(2014.519,8)
(2014.538,8)
(2014.558,1)
(2014.577,6)
(2014.596,4)
(2014.615,0)
(2014.635,1)
(2014.654,1)
(2014.673,2)
(2014.692,0)
(2014.712,1)
(2014.731,2)
(2014.75,0)
(2014.769,0)
(2014.788,2)
(2014.808,0)
(2014.827,0)
(2014.846,3)
(2014.865,6)
(2014.885,5)
(2014.904,6)
(2014.923,25)
(2014.942,9)
(2014.962,13)
(2014.981,18)
(2015.0,4)
(2015.019,21)
(2015.038,53)
}

\def\PESTDATA{
(2005.019,178)
(2005.038,141)
(2005.058,157)
(2005.077,107)
(2005.096,124)
(2005.115,146)
(2005.135,119)
(2005.154,178)
(2005.173,112)
(2005.192,130)
(2005.212,113)
(2005.231,141)
(2005.25,162)
(2005.269,121)
(2005.288,144)
(2005.308,125)
(2005.327,138)
(2005.346,83)
(2005.365,159)
(2005.385,153)
(2005.404,137)
(2005.423,165)
(2005.442,145)
(2005.462,66)
(2005.481,117)
(2005.5,89)
(2005.519,91)
(2005.538,31)
(2005.558,25)
(2005.577,23)
(2005.596,19)
(2005.615,20)
(2005.635,7)
(2005.654,22)
(2005.673,13)
(2005.692,9)
(2005.712,8)
(2005.731,18)
(2005.75,15)
(2005.769,27)
(2005.788,51)
(2005.808,79)
(2005.827,77)
(2005.846,70)
(2005.865,67)
(2005.885,48)
(2005.904,31)
(2005.923,81)
(2005.942,44)
(2005.962,133)
(2005.981,85)
(2006.0,163)
(2006.019,173)
(2006.038,102)
(2006.058,114)
(2006.077,121)
(2006.096,105)
(2006.115,104)
(2006.135,119)
(2006.154,170)
(2006.173,140)
(2006.192,181)
(2006.212,149)
(2006.231,160)
(2006.25,161)
(2006.269,168)
(2006.288,153)
(2006.308,157)
(2006.327,221)
(2006.346,165)
(2006.365,121)
(2006.385,126)
(2006.404,143)
(2006.423,151)
(2006.442,105)
(2006.462,145)
(2006.481,134)
(2006.5,130)
(2006.519,89)
(2006.538,49)
(2006.558,24)
(2006.577,29)
(2006.596,15)
(2006.615,4)
(2006.635,6)
(2006.654,4)
(2006.673,17)
(2006.692,1)
(2006.712,4)
(2006.731,7)
(2006.75,12)
(2006.769,8)
(2006.788,29)
(2006.808,29)
(2006.827,23)
(2006.846,46)
(2006.865,47)
(2006.885,71)
(2006.904,94)
(2006.923,127)
(2006.942,99)
(2006.962,103)
(2006.981,125)
(2007.0,68)
(2007.019,179)
(2007.038,205)
(2007.058,118)
(2007.077,155)
(2007.096,175)
(2007.115,135)
(2007.135,86)
(2007.154,224)
(2007.173,211)
(2007.192,214)
(2007.212,136)
(2007.231,279)
(2007.25,255)
(2007.269,209)
(2007.288,160)
(2007.308,290)
(2007.327,204)
(2007.346,105)
(2007.365,257)
(2007.385,210)
(2007.404,198)
(2007.423,149)
(2007.442,220)
(2007.462,125)
(2007.481,131)
(2007.5,101)
(2007.519,65)
(2007.538,30)
(2007.558,29)
(2007.577,19)
(2007.596,9)
(2007.615,7)
(2007.635,5)
(2007.654,3)
(2007.673,19)
(2007.692,1)
(2007.712,2)
(2007.731,6)
(2007.75,10)
(2007.769,31)
(2007.788,34)
(2007.808,32)
(2007.827,33)
(2007.846,28)
(2007.865,91)
(2007.885,95)
(2007.904,83)
(2007.923,99)
(2007.942,78)
(2007.962,108)
(2007.981,82)
(2008.0,46)
(2008.019,167)
(2008.038,197)
(2008.058,431)
(2008.077,87)
(2008.096,143)
(2008.115,114)
(2008.135,90)
(2008.154,109)
(2008.173,146)
(2008.192,95)
(2008.212,98)
(2008.231,108)
(2008.25,123)
(2008.269,114)
(2008.288,76)
(2008.308,156)
(2008.327,115)
(2008.346,45)
(2008.365,70)
(2008.385,50)
(2008.404,85)
(2008.423,71)
(2008.442,129)
(2008.462,70)
(2008.481,50)
(2008.5,93)
(2008.519,47)
(2008.538,50)
(2008.558,18)
(2008.577,3)
(2008.596,15)
(2008.615,11)
(2008.635,4)
(2008.654,4)
(2008.673,3)
(2008.692,3)
(2008.712,4)
(2008.731,15)
(2008.75,6)
(2008.769,12)
(2008.788,15)
(2008.808,27)
(2008.827,34)
(2008.846,23)
(2008.865,30)
(2008.885,45)
(2008.904,15)
(2008.923,48)
(2008.942,41)
(2008.962,35)
(2008.981,70)
(2009.0,13)
(2009.019,65)
(2009.038,125)
(2009.058,143)
(2009.077,75)
(2009.096,107)
(2009.115,102)
(2009.135,103)
(2009.154,142)
(2009.173,125)
(2009.192,124)
(2009.212,161)
(2009.231,173)
(2009.25,155)
(2009.269,168)
(2009.288,35)
(2009.308,120)
(2009.327,193)
(2009.346,159)
(2009.365,176)
(2009.385,117)
(2009.404,100)
(2009.423,138)
(2009.442,115)
(2009.462,105)
(2009.481,135)
(2009.5,182)
(2009.519,72)
(2009.538,72)
(2009.558,55)
(2009.577,41)
(2009.596,29)
(2009.615,32)
(2009.635,11)
(2009.654,1)
(2009.673,19)
(2009.692,6)
(2009.712,4)
(2009.731,13)
(2009.75,8)
(2009.769,5)
(2009.788,12)
(2009.808,20)
(2009.827,36)
(2009.846,30)
(2009.865,27)
(2009.885,30)
(2009.904,30)
(2009.923,46)
(2009.942,126)
(2009.962,96)
(2009.981,102)
(2010.0,55)
(2010.019,48)
(2010.038,88)
(2010.058,62)
(2010.077,81)
(2010.096,96)
(2010.115,99)
(2010.135,129)
(2010.154,104)
(2010.173,113)
(2010.192,128)
(2010.212,124)
(2010.231,137)
(2010.25,190)
(2010.269,148)
(2010.288,143)
(2010.308,272)
(2010.327,214)
(2010.346,191)
(2010.365,243)
(2010.385,151)
(2010.404,216)
(2010.423,145)
(2010.442,221)
(2010.462,205)
(2010.481,168)
(2010.5,253)
(2010.519,97)
(2010.538,70)
(2010.558,56)
(2010.577,37)
(2010.596,26)
(2010.615,7)
(2010.635,8)
(2010.654,4)
(2010.673,1)
(2010.692,8)
(2010.712,6)
(2010.731,3)
(2010.75,12)
(2010.769,18)
(2010.788,46)
(2010.808,36)
(2010.827,36)
(2010.846,56)
(2010.865,58)
(2010.885,108)
(2010.904,75)
(2010.923,118)
(2010.942,85)
(2010.962,115)
(2010.981,114)
(2011.0,74)
(2011.019,93)
(2011.038,155)
(2011.058,240)
(2011.077,86)
(2011.096,93)
(2011.115,108)
(2011.135,145)
(2011.154,77)
(2011.173,189)
(2011.192,142)
(2011.212,140)
(2011.231,120)
(2011.25,312)
(2011.269,199)
(2011.288,143)
(2011.308,145)
(2011.327,106)
(2011.346,195)
(2011.365,149)
(2011.385,172)
(2011.404,152)
(2011.423,152)
(2011.442,141)
(2011.462,93)
(2011.481,68)
(2011.5,95)
(2011.519,61)
(2011.538,103)
(2011.558,65)
(2011.577,16)
(2011.596,21)
(2011.615,6)
(2011.635,11)
(2011.654,6)
(2011.673,7)
(2011.692,3)
(2011.712,4)
(2011.731,6)
(2011.75,6)
(2011.769,10)
(2011.788,11)
(2011.808,15)
(2011.827,29)
(2011.846,17)
(2011.865,12)
(2011.885,17)
(2011.904,33)
(2011.923,71)
(2011.942,64)
(2011.962,66)
(2011.981,100)
(2012.0,92)
(2012.019,53)
(2012.038,179)
(2012.058,118)
(2012.077,184)
(2012.096,114)
(2012.115,96)
(2012.135,144)
(2012.154,100)
(2012.173,170)
(2012.192,98)
(2012.212,98)
(2012.231,98)
(2012.25,86)
(2012.269,89)
(2012.288,77)
(2012.308,91)
(2012.327,107)
(2012.346,86)
(2012.365,32)
(2012.385,70)
(2012.404,131)
(2012.423,104)
(2012.442,131)
(2012.462,93)
(2012.481,140)
(2012.5,113)
(2012.519,38)
(2012.538,107)
(2012.558,113)
(2012.577,31)
(2012.596,24)
(2012.615,30)
(2012.635,8)
(2012.654,9)
(2012.673,7)
(2012.692,3)
(2012.712,12)
(2012.731,11)
(2012.75,8)
(2012.769,5)
(2012.788,18)
(2012.808,12)
(2012.827,23)
(2012.846,16)
(2012.865,44)
(2012.885,44)
(2012.904,72)
(2012.923,54)
(2012.942,65)
(2012.962,69)
(2012.981,107)
(2013.0,90)
(2013.019,9)
(2013.038,129)
(2013.058,156)
(2013.077,240)
(2013.096,190)
(2013.115,166)
(2013.135,138)
(2013.154,128)
(2013.173,239)
(2013.192,196)
(2013.212,108)
(2013.231,160)
(2013.25,131)
(2013.269,331)
(2013.288,155)
(2013.308,168)
(2013.327,137)
(2013.346,144)
(2013.365,119)
(2013.385,126)
(2013.404,239)
(2013.423,133)
(2013.442,140)
(2013.462,129)
(2013.481,97)
(2013.5,121)
(2013.519,66)
(2013.538,56)
(2013.558,42)
(2013.577,34)
(2013.596,30)
(2013.615,62)
(2013.635,9)
(2013.654,8)
(2013.673,4)
(2013.692,4)
(2013.712,2)
(2013.731,0)
(2013.75,4)
(2013.769,6)
(2013.788,9)
(2013.808,20)
(2013.827,18)
(2013.846,15)
(2013.865,12)
(2013.885,38)
(2013.904,25)
(2013.923,43)
(2013.942,52)
(2013.962,35)
(2013.981,29)
(2014.0,54)
(2014.019,10)
(2014.038,30)
(2014.058,80)
(2014.077,140)
(2014.096,43)
(2014.115,108)
(2014.135,35)
(2014.154,68)
(2014.173,77)
(2014.192,106)
(2014.212,77)
(2014.231,76)
(2014.25,107)
(2014.269,100)
(2014.288,36)
(2014.308,80)
(2014.327,60)
(2014.346,38)
(2014.365,15)
(2014.385,42)
(2014.404,125)
(2014.423,85)
(2014.442,63)
(2014.462,85)
(2014.481,102)
(2014.5,133)
(2014.519,84)
(2014.538,61)
(2014.558,76)
(2014.577,107)
(2014.596,69)
(2014.615,23)
(2014.635,35)
(2014.654,15)
(2014.673,10)
(2014.692,9)
(2014.712,2)
(2014.731,1)
(2014.75,27)
(2014.769,8)
(2014.788,26)
(2014.808,43)
(2014.827,26)
(2014.846,39)
(2014.865,32)
(2014.885,46)
(2014.904,87)
(2014.923,33)
(2014.942,25)
(2014.962,122)
(2014.981,70)
(2015.0,72)
(2015.019,12)
(2015.038,256)
}

\def\SOMOGYDATA{
(2005.019,66)
(2005.038,48)
(2005.058,33)
(2005.077,66)
(2005.096,63)
(2005.115,59)
(2005.135,104)
(2005.154,70)
(2005.173,116)
(2005.192,68)
(2005.212,62)
(2005.231,55)
(2005.25,49)
(2005.269,41)
(2005.288,50)
(2005.308,55)
(2005.327,57)
(2005.346,42)
(2005.365,66)
(2005.385,42)
(2005.404,54)
(2005.423,42)
(2005.442,23)
(2005.462,13)
(2005.481,36)
(2005.5,12)
(2005.519,11)
(2005.538,11)
(2005.558,6)
(2005.577,11)
(2005.596,5)
(2005.615,5)
(2005.635,2)
(2005.654,4)
(2005.673,2)
(2005.692,2)
(2005.712,3)
(2005.731,3)
(2005.75,6)
(2005.769,6)
(2005.788,5)
(2005.808,1)
(2005.827,17)
(2005.846,9)
(2005.865,17)
(2005.885,10)
(2005.904,5)
(2005.923,20)
(2005.942,38)
(2005.962,23)
(2005.981,51)
(2006.0,29)
(2006.019,79)
(2006.038,49)
(2006.058,61)
(2006.077,99)
(2006.096,97)
(2006.115,82)
(2006.135,145)
(2006.154,120)
(2006.173,120)
(2006.192,123)
(2006.212,101)
(2006.231,70)
(2006.25,47)
(2006.269,155)
(2006.288,43)
(2006.308,46)
(2006.327,81)
(2006.346,22)
(2006.365,42)
(2006.385,28)
(2006.404,22)
(2006.423,35)
(2006.442,14)
(2006.462,28)
(2006.481,19)
(2006.5,28)
(2006.519,19)
(2006.538,7)
(2006.558,9)
(2006.577,3)
(2006.596,4)
(2006.615,6)
(2006.635,1)
(2006.654,2)
(2006.673,2)
(2006.692,4)
(2006.712,1)
(2006.731,1)
(2006.75,6)
(2006.769,3)
(2006.788,6)
(2006.808,8)
(2006.827,4)
(2006.846,5)
(2006.865,7)
(2006.885,38)
(2006.904,15)
(2006.923,55)
(2006.942,49)
(2006.962,51)
(2006.981,87)
(2007.0,25)
(2007.019,82)
(2007.038,84)
(2007.058,37)
(2007.077,56)
(2007.096,66)
(2007.115,65)
(2007.135,60)
(2007.154,71)
(2007.173,63)
(2007.192,49)
(2007.212,22)
(2007.231,55)
(2007.25,54)
(2007.269,37)
(2007.288,31)
(2007.308,63)
(2007.327,19)
(2007.346,22)
(2007.365,30)
(2007.385,30)
(2007.404,47)
(2007.423,26)
(2007.442,29)
(2007.462,25)
(2007.481,29)
(2007.5,15)
(2007.519,14)
(2007.538,9)
(2007.558,8)
(2007.577,2)
(2007.596,2)
(2007.615,1)
(2007.635,2)
(2007.654,5)
(2007.673,1)
(2007.692,2)
(2007.712,0)
(2007.731,0)
(2007.75,3)
(2007.769,1)
(2007.788,2)
(2007.808,3)
(2007.827,7)
(2007.846,2)
(2007.865,15)
(2007.885,20)
(2007.904,7)
(2007.923,40)
(2007.942,28)
(2007.962,66)
(2007.981,44)
(2008.0,6)
(2008.019,50)
(2008.038,45)
(2008.058,113)
(2008.077,24)
(2008.096,26)
(2008.115,33)
(2008.135,24)
(2008.154,42)
(2008.173,40)
(2008.192,40)
(2008.212,24)
(2008.231,12)
(2008.25,42)
(2008.269,30)
(2008.288,32)
(2008.308,10)
(2008.327,57)
(2008.346,11)
(2008.365,39)
(2008.385,32)
(2008.404,97)
(2008.423,46)
(2008.442,19)
(2008.462,63)
(2008.481,21)
(2008.5,28)
(2008.519,14)
(2008.538,14)
(2008.558,9)
(2008.577,9)
(2008.596,6)
(2008.615,3)
(2008.635,8)
(2008.654,1)
(2008.673,0)
(2008.692,0)
(2008.712,3)
(2008.731,1)
(2008.75,3)
(2008.769,1)
(2008.788,3)
(2008.808,8)
(2008.827,7)
(2008.846,10)
(2008.865,10)
(2008.885,10)
(2008.904,12)
(2008.923,15)
(2008.942,15)
(2008.962,69)
(2008.981,40)
(2009.0,19)
(2009.019,12)
(2009.038,103)
(2009.058,77)
(2009.077,42)
(2009.096,54)
(2009.115,102)
(2009.135,57)
(2009.154,87)
(2009.173,60)
(2009.192,66)
(2009.212,26)
(2009.231,31)
(2009.25,40)
(2009.269,65)
(2009.288,38)
(2009.308,58)
(2009.327,47)
(2009.346,53)
(2009.365,59)
(2009.385,68)
(2009.404,34)
(2009.423,20)
(2009.442,42)
(2009.462,18)
(2009.481,21)
(2009.5,14)
(2009.519,17)
(2009.538,6)
(2009.558,11)
(2009.577,9)
(2009.596,2)
(2009.615,4)
(2009.635,3)
(2009.654,1)
(2009.673,4)
(2009.692,0)
(2009.712,0)
(2009.731,2)
(2009.75,1)
(2009.769,5)
(2009.788,1)
(2009.808,6)
(2009.827,7)
(2009.846,2)
(2009.865,7)
(2009.885,5)
(2009.904,8)
(2009.923,22)
(2009.942,13)
(2009.962,20)
(2009.981,17)
(2010.0,8)
(2010.019,13)
(2010.038,23)
(2010.058,18)
(2010.077,9)
(2010.096,41)
(2010.115,59)
(2010.135,25)
(2010.154,39)
(2010.173,39)
(2010.192,18)
(2010.212,25)
(2010.231,1)
(2010.25,34)
(2010.269,36)
(2010.288,23)
(2010.308,23)
(2010.327,33)
(2010.346,37)
(2010.365,33)
(2010.385,23)
(2010.404,27)
(2010.423,18)
(2010.442,36)
(2010.462,55)
(2010.481,35)
(2010.5,48)
(2010.519,26)
(2010.538,37)
(2010.558,31)
(2010.577,12)
(2010.596,19)
(2010.615,2)
(2010.635,12)
(2010.654,6)
(2010.673,1)
(2010.692,4)
(2010.712,4)
(2010.731,3)
(2010.75,5)
(2010.769,7)
(2010.788,14)
(2010.808,29)
(2010.827,24)
(2010.846,63)
(2010.865,20)
(2010.885,67)
(2010.904,33)
(2010.923,26)
(2010.942,38)
(2010.962,15)
(2010.981,12)
(2011.0,20)
(2011.019,11)
(2011.038,4)
(2011.058,27)
(2011.077,13)
(2011.096,22)
(2011.115,15)
(2011.135,28)
(2011.154,16)
(2011.173,24)
(2011.192,26)
(2011.212,41)
(2011.231,43)
(2011.25,82)
(2011.269,37)
(2011.288,45)
(2011.308,57)
(2011.327,41)
(2011.346,31)
(2011.365,46)
(2011.385,21)
(2011.404,31)
(2011.423,23)
(2011.442,44)
(2011.462,23)
(2011.481,16)
(2011.5,16)
(2011.519,18)
(2011.538,23)
(2011.558,7)
(2011.577,14)
(2011.596,10)
(2011.615,4)
(2011.635,2)
(2011.654,4)
(2011.673,0)
(2011.692,1)
(2011.712,0)
(2011.731,2)
(2011.75,0)
(2011.769,2)
(2011.788,0)
(2011.808,1)
(2011.827,1)
(2011.846,9)
(2011.865,5)
(2011.885,9)
(2011.904,18)
(2011.923,17)
(2011.942,36)
(2011.962,31)
(2011.981,22)
(2012.0,85)
(2012.019,16)
(2012.038,85)
(2012.058,41)
(2012.077,27)
(2012.096,40)
(2012.115,47)
(2012.135,30)
(2012.154,53)
(2012.173,36)
(2012.192,45)
(2012.212,24)
(2012.231,25)
(2012.25,84)
(2012.269,46)
(2012.288,56)
(2012.308,17)
(2012.327,38)
(2012.346,28)
(2012.365,21)
(2012.385,84)
(2012.404,53)
(2012.423,60)
(2012.442,56)
(2012.462,40)
(2012.481,40)
(2012.5,32)
(2012.519,32)
(2012.538,26)
(2012.558,11)
(2012.577,7)
(2012.596,7)
(2012.615,3)
(2012.635,1)
(2012.654,5)
(2012.673,0)
(2012.692,0)
(2012.712,7)
(2012.731,0)
(2012.75,1)
(2012.769,1)
(2012.788,2)
(2012.808,1)
(2012.827,0)
(2012.846,2)
(2012.865,2)
(2012.885,14)
(2012.904,9)
(2012.923,16)
(2012.942,15)
(2012.962,54)
(2012.981,14)
(2013.0,17)
(2013.019,43)
(2013.038,66)
(2013.058,89)
(2013.077,33)
(2013.096,48)
(2013.115,31)
(2013.135,53)
(2013.154,36)
(2013.173,35)
(2013.192,61)
(2013.212,32)
(2013.231,86)
(2013.25,29)
(2013.269,96)
(2013.288,106)
(2013.308,44)
(2013.327,34)
(2013.346,25)
(2013.365,58)
(2013.385,33)
(2013.404,50)
(2013.423,13)
(2013.442,93)
(2013.462,73)
(2013.481,35)
(2013.5,40)
(2013.519,52)
(2013.538,26)
(2013.558,3)
(2013.577,10)
(2013.596,12)
(2013.615,10)
(2013.635,11)
(2013.654,4)
(2013.673,0)
(2013.692,2)
(2013.712,5)
(2013.731,0)
(2013.75,4)
(2013.769,3)
(2013.788,0)
(2013.808,17)
(2013.827,14)
(2013.846,11)
(2013.865,46)
(2013.885,38)
(2013.904,19)
(2013.923,15)
(2013.942,8)
(2013.962,63)
(2013.981,5)
(2014.0,17)
(2014.019,0)
(2014.038,21)
(2014.058,93)
(2014.077,18)
(2014.096,25)
(2014.115,46)
(2014.135,14)
(2014.154,20)
(2014.173,11)
(2014.192,45)
(2014.212,22)
(2014.231,25)
(2014.25,29)
(2014.269,22)
(2014.288,4)
(2014.308,14)
(2014.327,3)
(2014.346,21)
(2014.365,0)
(2014.385,48)
(2014.404,1)
(2014.423,3)
(2014.442,4)
(2014.462,6)
(2014.481,3)
(2014.5,5)
(2014.519,8)
(2014.538,1)
(2014.558,10)
(2014.577,6)
(2014.596,23)
(2014.615,2)
(2014.635,2)
(2014.654,0)
(2014.673,1)
(2014.692,1)
(2014.712,2)
(2014.731,0)
(2014.75,5)
(2014.769,1)
(2014.788,8)
(2014.808,5)
(2014.827,10)
(2014.846,33)
(2014.865,16)
(2014.885,28)
(2014.904,5)
(2014.923,28)
(2014.942,16)
(2014.962,4)
(2014.981,36)
(2015.0,5)
(2015.019,5)
(2015.038,45)
}

\def\SZABOLCSDATA{
(2005.019,64)
(2005.038,29)
(2005.058,33)
(2005.077,50)
(2005.096,56)
(2005.115,54)
(2005.135,85)
(2005.154,75)
(2005.173,76)
(2005.192,59)
(2005.212,22)
(2005.231,45)
(2005.25,203)
(2005.269,92)
(2005.288,41)
(2005.308,49)
(2005.327,41)
(2005.346,29)
(2005.365,57)
(2005.385,36)
(2005.404,93)
(2005.423,19)
(2005.442,84)
(2005.462,14)
(2005.481,73)
(2005.5,60)
(2005.519,86)
(2005.538,35)
(2005.558,8)
(2005.577,18)
(2005.596,11)
(2005.615,7)
(2005.635,6)
(2005.654,1)
(2005.673,4)
(2005.692,2)
(2005.712,3)
(2005.731,6)
(2005.75,6)
(2005.769,3)
(2005.788,12)
(2005.808,14)
(2005.827,6)
(2005.846,7)
(2005.865,33)
(2005.885,24)
(2005.904,19)
(2005.923,24)
(2005.942,28)
(2005.962,38)
(2005.981,18)
(2006.0,28)
(2006.019,35)
(2006.038,30)
(2006.058,19)
(2006.077,24)
(2006.096,41)
(2006.115,40)
(2006.135,54)
(2006.154,21)
(2006.173,64)
(2006.192,58)
(2006.212,38)
(2006.231,31)
(2006.25,8)
(2006.269,81)
(2006.288,81)
(2006.308,71)
(2006.327,95)
(2006.346,67)
(2006.365,83)
(2006.385,73)
(2006.404,48)
(2006.423,63)
(2006.442,30)
(2006.462,45)
(2006.481,41)
(2006.5,55)
(2006.519,37)
(2006.538,17)
(2006.558,17)
(2006.577,12)
(2006.596,3)
(2006.615,2)
(2006.635,9)
(2006.654,1)
(2006.673,2)
(2006.692,3)
(2006.712,5)
(2006.731,6)
(2006.75,10)
(2006.769,7)
(2006.788,6)
(2006.808,24)
(2006.827,13)
(2006.846,14)
(2006.865,42)
(2006.885,14)
(2006.904,7)
(2006.923,46)
(2006.942,21)
(2006.962,16)
(2006.981,5)
(2007.0,1)
(2007.019,1)
(2007.038,53)
(2007.058,38)
(2007.077,46)
(2007.096,61)
(2007.115,125)
(2007.135,33)
(2007.154,57)
(2007.173,37)
(2007.192,46)
(2007.212,26)
(2007.231,74)
(2007.25,37)
(2007.269,26)
(2007.288,30)
(2007.308,80)
(2007.327,43)
(2007.346,70)
(2007.365,65)
(2007.385,72)
(2007.404,39)
(2007.423,76)
(2007.442,113)
(2007.462,14)
(2007.481,28)
(2007.5,24)
(2007.519,10)
(2007.538,14)
(2007.558,1)
(2007.577,14)
(2007.596,4)
(2007.615,9)
(2007.635,0)
(2007.654,1)
(2007.673,1)
(2007.692,1)
(2007.712,0)
(2007.731,1)
(2007.75,3)
(2007.769,5)
(2007.788,55)
(2007.808,17)
(2007.827,4)
(2007.846,44)
(2007.865,31)
(2007.885,35)
(2007.904,64)
(2007.923,47)
(2007.942,17)
(2007.962,77)
(2007.981,28)
(2008.0,170)
(2008.019,69)
(2008.038,110)
(2008.058,203)
(2008.077,82)
(2008.096,71)
(2008.115,93)
(2008.135,58)
(2008.154,70)
(2008.173,73)
(2008.192,51)
(2008.212,69)
(2008.231,78)
(2008.25,49)
(2008.269,55)
(2008.288,66)
(2008.308,60)
(2008.327,46)
(2008.346,14)
(2008.365,115)
(2008.385,43)
(2008.404,23)
(2008.423,50)
(2008.442,63)
(2008.462,15)
(2008.481,30)
(2008.5,13)
(2008.519,4)
(2008.538,14)
(2008.558,33)
(2008.577,10)
(2008.596,1)
(2008.615,18)
(2008.635,18)
(2008.654,1)
(2008.673,0)
(2008.692,1)
(2008.712,1)
(2008.731,1)
(2008.75,1)
(2008.769,1)
(2008.788,7)
(2008.808,3)
(2008.827,4)
(2008.846,9)
(2008.865,8)
(2008.885,5)
(2008.904,15)
(2008.923,11)
(2008.942,23)
(2008.962,41)
(2008.981,73)
(2009.0,12)
(2009.019,48)
(2009.038,17)
(2009.058,74)
(2009.077,13)
(2009.096,74)
(2009.115,26)
(2009.135,86)
(2009.154,39)
(2009.173,88)
(2009.192,32)
(2009.212,162)
(2009.231,47)
(2009.25,62)
(2009.269,75)
(2009.288,44)
(2009.308,93)
(2009.327,33)
(2009.346,61)
(2009.365,27)
(2009.385,13)
(2009.404,43)
(2009.423,47)
(2009.442,38)
(2009.462,31)
(2009.481,16)
(2009.5,8)
(2009.519,16)
(2009.538,29)
(2009.558,5)
(2009.577,6)
(2009.596,1)
(2009.615,2)
(2009.635,0)
(2009.654,1)
(2009.673,2)
(2009.692,0)
(2009.712,4)
(2009.731,2)
(2009.75,8)
(2009.769,5)
(2009.788,4)
(2009.808,5)
(2009.827,9)
(2009.846,5)
(2009.865,8)
(2009.885,1)
(2009.904,2)
(2009.923,9)
(2009.942,10)
(2009.962,19)
(2009.981,20)
(2010.0,3)
(2010.019,16)
(2010.038,1)
(2010.058,14)
(2010.077,13)
(2010.096,14)
(2010.115,22)
(2010.135,3)
(2010.154,42)
(2010.173,44)
(2010.192,77)
(2010.212,89)
(2010.231,2)
(2010.25,34)
(2010.269,30)
(2010.288,22)
(2010.308,39)
(2010.327,63)
(2010.346,22)
(2010.365,32)
(2010.385,12)
(2010.404,3)
(2010.423,3)
(2010.442,8)
(2010.462,3)
(2010.481,9)
(2010.5,61)
(2010.519,8)
(2010.538,40)
(2010.558,10)
(2010.577,9)
(2010.596,9)
(2010.615,1)
(2010.635,1)
(2010.654,0)
(2010.673,1)
(2010.692,1)
(2010.712,0)
(2010.731,4)
(2010.75,2)
(2010.769,0)
(2010.788,1)
(2010.808,12)
(2010.827,9)
(2010.846,5)
(2010.865,3)
(2010.885,6)
(2010.904,6)
(2010.923,2)
(2010.942,26)
(2010.962,15)
(2010.981,28)
(2011.0,18)
(2011.019,42)
(2011.038,17)
(2011.058,14)
(2011.077,67)
(2011.096,52)
(2011.115,40)
(2011.135,62)
(2011.154,38)
(2011.173,102)
(2011.192,48)
(2011.212,125)
(2011.231,56)
(2011.25,184)
(2011.269,97)
(2011.288,66)
(2011.308,88)
(2011.327,40)
(2011.346,33)
(2011.365,167)
(2011.385,128)
(2011.404,100)
(2011.423,72)
(2011.442,54)
(2011.462,52)
(2011.481,48)
(2011.5,40)
(2011.519,19)
(2011.538,67)
(2011.558,11)
(2011.577,4)
(2011.596,5)
(2011.615,4)
(2011.635,3)
(2011.654,2)
(2011.673,9)
(2011.692,1)
(2011.712,0)
(2011.731,3)
(2011.75,6)
(2011.769,7)
(2011.788,2)
(2011.808,9)
(2011.827,19)
(2011.846,16)
(2011.865,3)
(2011.885,16)
(2011.904,26)
(2011.923,2)
(2011.942,19)
(2011.962,83)
(2011.981,41)
(2012.0,36)
(2012.019,16)
(2012.038,12)
(2012.058,13)
(2012.077,20)
(2012.096,21)
(2012.115,15)
(2012.135,11)
(2012.154,15)
(2012.173,23)
(2012.192,3)
(2012.212,16)
(2012.231,0)
(2012.25,31)
(2012.269,47)
(2012.288,7)
(2012.308,27)
(2012.327,67)
(2012.346,45)
(2012.365,7)
(2012.385,74)
(2012.404,41)
(2012.423,23)
(2012.442,36)
(2012.462,48)
(2012.481,15)
(2012.5,13)
(2012.519,56)
(2012.538,4)
(2012.558,32)
(2012.577,13)
(2012.596,9)
(2012.615,9)
(2012.635,3)
(2012.654,2)
(2012.673,1)
(2012.692,6)
(2012.712,1)
(2012.731,1)
(2012.75,1)
(2012.769,2)
(2012.788,0)
(2012.808,26)
(2012.827,4)
(2012.846,14)
(2012.865,9)
(2012.885,35)
(2012.904,22)
(2012.923,2)
(2012.942,10)
(2012.962,23)
(2012.981,12)
(2013.0,15)
(2013.019,3)
(2013.038,65)
(2013.058,45)
(2013.077,13)
(2013.096,7)
(2013.115,7)
(2013.135,3)
(2013.154,16)
(2013.173,29)
(2013.192,30)
(2013.212,20)
(2013.231,55)
(2013.25,7)
(2013.269,113)
(2013.288,32)
(2013.308,32)
(2013.327,96)
(2013.346,29)
(2013.365,47)
(2013.385,61)
(2013.404,71)
(2013.423,38)
(2013.442,51)
(2013.462,27)
(2013.481,75)
(2013.5,20)
(2013.519,96)
(2013.538,44)
(2013.558,74)
(2013.577,14)
(2013.596,24)
(2013.615,9)
(2013.635,0)
(2013.654,0)
(2013.673,6)
(2013.692,2)
(2013.712,2)
(2013.731,0)
(2013.75,0)
(2013.769,4)
(2013.788,11)
(2013.808,0)
(2013.827,15)
(2013.846,20)
(2013.865,30)
(2013.885,8)
(2013.904,7)
(2013.923,12)
(2013.942,46)
(2013.962,14)
(2013.981,18)
(2014.0,19)
(2014.019,3)
(2014.038,26)
(2014.058,3)
(2014.077,9)
(2014.096,3)
(2014.115,83)
(2014.135,2)
(2014.154,34)
(2014.173,16)
(2014.192,25)
(2014.212,1)
(2014.231,29)
(2014.25,44)
(2014.269,42)
(2014.288,34)
(2014.308,54)
(2014.327,25)
(2014.346,13)
(2014.365,8)
(2014.385,9)
(2014.404,56)
(2014.423,46)
(2014.442,63)
(2014.462,11)
(2014.481,27)
(2014.5,9)
(2014.519,33)
(2014.538,11)
(2014.558,13)
(2014.577,15)
(2014.596,4)
(2014.615,13)
(2014.635,12)
(2014.654,1)
(2014.673,1)
(2014.692,2)
(2014.712,0)
(2014.731,5)
(2014.75,0)
(2014.769,1)
(2014.788,6)
(2014.808,2)
(2014.827,3)
(2014.846,14)
(2014.865,7)
(2014.885,28)
(2014.904,15)
(2014.923,7)
(2014.942,0)
(2014.962,23)
(2014.981,5)
(2015.0,21)
(2015.019,17)
(2015.038,39)
}

\def\TOLNADATA{
(2005.019,11)
(2005.038,58)
(2005.058,24)
(2005.077,25)
(2005.096,7)
(2005.115,27)
(2005.135,20)
(2005.154,5)
(2005.173,22)
(2005.192,31)
(2005.212,26)
(2005.231,23)
(2005.25,40)
(2005.269,8)
(2005.288,25)
(2005.308,13)
(2005.327,6)
(2005.346,2)
(2005.365,15)
(2005.385,28)
(2005.404,28)
(2005.423,33)
(2005.442,46)
(2005.462,23)
(2005.481,21)
(2005.5,22)
(2005.519,12)
(2005.538,8)
(2005.558,11)
(2005.577,8)
(2005.596,4)
(2005.615,3)
(2005.635,2)
(2005.654,2)
(2005.673,1)
(2005.692,2)
(2005.712,2)
(2005.731,6)
(2005.75,5)
(2005.769,1)
(2005.788,3)
(2005.808,5)
(2005.827,10)
(2005.846,7)
(2005.865,25)
(2005.885,16)
(2005.904,22)
(2005.923,10)
(2005.942,32)
(2005.962,35)
(2005.981,77)
(2006.0,57)
(2006.019,71)
(2006.038,53)
(2006.058,37)
(2006.077,53)
(2006.096,51)
(2006.115,48)
(2006.135,93)
(2006.154,66)
(2006.173,55)
(2006.192,66)
(2006.212,47)
(2006.231,96)
(2006.25,85)
(2006.269,73)
(2006.288,86)
(2006.308,58)
(2006.327,82)
(2006.346,58)
(2006.365,85)
(2006.385,62)
(2006.404,92)
(2006.423,58)
(2006.442,51)
(2006.462,25)
(2006.481,118)
(2006.5,84)
(2006.519,26)
(2006.538,11)
(2006.558,23)
(2006.577,9)
(2006.596,5)
(2006.615,5)
(2006.635,0)
(2006.654,0)
(2006.673,0)
(2006.692,1)
(2006.712,0)
(2006.731,1)
(2006.75,3)
(2006.769,2)
(2006.788,5)
(2006.808,2)
(2006.827,0)
(2006.846,1)
(2006.865,1)
(2006.885,4)
(2006.904,8)
(2006.923,10)
(2006.942,7)
(2006.962,11)
(2006.981,18)
(2007.0,13)
(2007.019,17)
(2007.038,17)
(2007.058,12)
(2007.077,36)
(2007.096,14)
(2007.115,37)
(2007.135,29)
(2007.154,39)
(2007.173,23)
(2007.192,36)
(2007.212,28)
(2007.231,6)
(2007.25,38)
(2007.269,21)
(2007.288,12)
(2007.308,73)
(2007.327,59)
(2007.346,10)
(2007.365,79)
(2007.385,35)
(2007.404,73)
(2007.423,56)
(2007.442,51)
(2007.462,22)
(2007.481,20)
(2007.5,13)
(2007.519,11)
(2007.538,11)
(2007.558,10)
(2007.577,4)
(2007.596,5)
(2007.615,2)
(2007.635,0)
(2007.654,0)
(2007.673,4)
(2007.692,4)
(2007.712,0)
(2007.731,4)
(2007.75,7)
(2007.769,13)
(2007.788,8)
(2007.808,8)
(2007.827,13)
(2007.846,8)
(2007.865,19)
(2007.885,38)
(2007.904,14)
(2007.923,24)
(2007.942,14)
(2007.962,10)
(2007.981,8)
(2008.0,9)
(2008.019,10)
(2008.038,10)
(2008.058,27)
(2008.077,4)
(2008.096,6)
(2008.115,12)
(2008.135,11)
(2008.154,11)
(2008.173,10)
(2008.192,9)
(2008.212,15)
(2008.231,8)
(2008.25,10)
(2008.269,24)
(2008.288,12)
(2008.308,15)
(2008.327,52)
(2008.346,24)
(2008.365,28)
(2008.385,34)
(2008.404,34)
(2008.423,40)
(2008.442,37)
(2008.462,90)
(2008.481,25)
(2008.5,31)
(2008.519,7)
(2008.538,22)
(2008.558,29)
(2008.577,6)
(2008.596,13)
(2008.615,1)
(2008.635,7)
(2008.654,1)
(2008.673,2)
(2008.692,5)
(2008.712,0)
(2008.731,8)
(2008.75,5)
(2008.769,4)
(2008.788,3)
(2008.808,4)
(2008.827,5)
(2008.846,1)
(2008.865,18)
(2008.885,10)
(2008.904,25)
(2008.923,17)
(2008.942,16)
(2008.962,12)
(2008.981,16)
(2009.0,3)
(2009.019,16)
(2009.038,56)
(2009.058,14)
(2009.077,13)
(2009.096,13)
(2009.115,29)
(2009.135,44)
(2009.154,44)
(2009.173,66)
(2009.192,43)
(2009.212,70)
(2009.231,105)
(2009.25,21)
(2009.269,50)
(2009.288,39)
(2009.308,51)
(2009.327,40)
(2009.346,81)
(2009.365,35)
(2009.385,71)
(2009.404,21)
(2009.423,52)
(2009.442,56)
(2009.462,49)
(2009.481,21)
(2009.5,34)
(2009.519,12)
(2009.538,9)
(2009.558,15)
(2009.577,4)
(2009.596,14)
(2009.615,4)
(2009.635,4)
(2009.654,0)
(2009.673,0)
(2009.692,0)
(2009.712,0)
(2009.731,1)
(2009.75,0)
(2009.769,3)
(2009.788,4)
(2009.808,9)
(2009.827,8)
(2009.846,10)
(2009.865,8)
(2009.885,15)
(2009.904,6)
(2009.923,19)
(2009.942,14)
(2009.962,24)
(2009.981,10)
(2010.0,21)
(2010.019,0)
(2010.038,42)
(2010.058,21)
(2010.077,8)
(2010.096,9)
(2010.115,15)
(2010.135,21)
(2010.154,38)
(2010.173,18)
(2010.192,29)
(2010.212,13)
(2010.231,14)
(2010.25,26)
(2010.269,59)
(2010.288,35)
(2010.308,73)
(2010.327,26)
(2010.346,55)
(2010.365,17)
(2010.385,47)
(2010.404,15)
(2010.423,42)
(2010.442,46)
(2010.462,39)
(2010.481,61)
(2010.5,39)
(2010.519,19)
(2010.538,23)
(2010.558,4)
(2010.577,17)
(2010.596,7)
(2010.615,7)
(2010.635,0)
(2010.654,0)
(2010.673,2)
(2010.692,3)
(2010.712,4)
(2010.731,0)
(2010.75,2)
(2010.769,0)
(2010.788,8)
(2010.808,14)
(2010.827,0)
(2010.846,16)
(2010.865,3)
(2010.885,3)
(2010.904,55)
(2010.923,13)
(2010.942,39)
(2010.962,12)
(2010.981,17)
(2011.0,34)
(2011.019,9)
(2011.038,1)
(2011.058,32)
(2011.077,7)
(2011.096,38)
(2011.115,8)
(2011.135,41)
(2011.154,43)
(2011.173,43)
(2011.192,47)
(2011.212,39)
(2011.231,11)
(2011.25,53)
(2011.269,36)
(2011.288,14)
(2011.308,13)
(2011.327,33)
(2011.346,8)
(2011.365,15)
(2011.385,22)
(2011.404,7)
(2011.423,21)
(2011.442,44)
(2011.462,39)
(2011.481,34)
(2011.5,12)
(2011.519,5)
(2011.538,26)
(2011.558,0)
(2011.577,5)
(2011.596,0)
(2011.615,1)
(2011.635,1)
(2011.654,0)
(2011.673,1)
(2011.692,0)
(2011.712,0)
(2011.731,4)
(2011.75,3)
(2011.769,0)
(2011.788,4)
(2011.808,2)
(2011.827,3)
(2011.846,2)
(2011.865,3)
(2011.885,2)
(2011.904,12)
(2011.923,5)
(2011.942,12)
(2011.962,2)
(2011.981,8)
(2012.0,4)
(2012.019,1)
(2012.038,4)
(2012.058,3)
(2012.077,2)
(2012.096,2)
(2012.115,6)
(2012.135,8)
(2012.154,2)
(2012.173,1)
(2012.192,0)
(2012.212,6)
(2012.231,0)
(2012.25,2)
(2012.269,1)
(2012.288,15)
(2012.308,10)
(2012.327,13)
(2012.346,4)
(2012.365,2)
(2012.385,19)
(2012.404,2)
(2012.423,25)
(2012.442,10)
(2012.462,12)
(2012.481,6)
(2012.5,8)
(2012.519,13)
(2012.538,11)
(2012.558,1)
(2012.577,2)
(2012.596,0)
(2012.615,0)
(2012.635,1)
(2012.654,1)
(2012.673,0)
(2012.692,0)
(2012.712,1)
(2012.731,2)
(2012.75,3)
(2012.769,2)
(2012.788,1)
(2012.808,8)
(2012.827,2)
(2012.846,11)
(2012.865,4)
(2012.885,7)
(2012.904,20)
(2012.923,8)
(2012.942,19)
(2012.962,26)
(2012.981,12)
(2013.0,20)
(2013.019,27)
(2013.038,32)
(2013.058,48)
(2013.077,31)
(2013.096,54)
(2013.115,50)
(2013.135,40)
(2013.154,13)
(2013.173,54)
(2013.192,29)
(2013.212,90)
(2013.231,71)
(2013.25,38)
(2013.269,119)
(2013.288,60)
(2013.308,102)
(2013.327,131)
(2013.346,73)
(2013.365,55)
(2013.385,99)
(2013.404,88)
(2013.423,59)
(2013.442,31)
(2013.462,50)
(2013.481,27)
(2013.5,15)
(2013.519,46)
(2013.538,12)
(2013.558,8)
(2013.577,32)
(2013.596,13)
(2013.615,2)
(2013.635,0)
(2013.654,0)
(2013.673,5)
(2013.692,7)
(2013.712,1)
(2013.731,1)
(2013.75,0)
(2013.769,0)
(2013.788,2)
(2013.808,1)
(2013.827,1)
(2013.846,5)
(2013.865,1)
(2013.885,4)
(2013.904,2)
(2013.923,0)
(2013.942,0)
(2013.962,4)
(2013.981,0)
(2014.0,1)
(2014.019,4)
(2014.038,0)
(2014.058,0)
(2014.077,2)
(2014.096,0)
(2014.115,66)
(2014.135,17)
(2014.154,3)
(2014.173,2)
(2014.192,23)
(2014.212,25)
(2014.231,8)
(2014.25,11)
(2014.269,12)
(2014.288,9)
(2014.308,12)
(2014.327,18)
(2014.346,16)
(2014.365,0)
(2014.385,19)
(2014.404,56)
(2014.423,8)
(2014.442,37)
(2014.462,19)
(2014.481,3)
(2014.5,6)
(2014.519,3)
(2014.538,20)
(2014.558,31)
(2014.577,9)
(2014.596,14)
(2014.615,0)
(2014.635,11)
(2014.654,0)
(2014.673,0)
(2014.692,0)
(2014.712,8)
(2014.731,0)
(2014.75,1)
(2014.769,0)
(2014.788,2)
(2014.808,4)
(2014.827,0)
(2014.846,0)
(2014.865,21)
(2014.885,4)
(2014.904,4)
(2014.923,25)
(2014.942,9)
(2014.962,4)
(2014.981,23)
(2015.0,14)
(2015.019,1)
(2015.038,27)
}

\def\VASDATA{
(2005.019,29)
(2005.038,53)
(2005.058,18)
(2005.077,21)
(2005.096,47)
(2005.115,54)
(2005.135,32)
(2005.154,66)
(2005.173,45)
(2005.192,85)
(2005.212,19)
(2005.231,83)
(2005.25,47)
(2005.269,52)
(2005.288,64)
(2005.308,46)
(2005.327,98)
(2005.346,42)
(2005.365,111)
(2005.385,56)
(2005.404,40)
(2005.423,49)
(2005.442,39)
(2005.462,29)
(2005.481,53)
(2005.5,25)
(2005.519,54)
(2005.538,18)
(2005.558,29)
(2005.577,9)
(2005.596,6)
(2005.615,3)
(2005.635,3)
(2005.654,24)
(2005.673,0)
(2005.692,1)
(2005.712,2)
(2005.731,4)
(2005.75,4)
(2005.769,17)
(2005.788,16)
(2005.808,35)
(2005.827,42)
(2005.846,70)
(2005.865,41)
(2005.885,35)
(2005.904,48)
(2005.923,54)
(2005.942,70)
(2005.962,89)
(2005.981,66)
(2006.0,56)
(2006.019,79)
(2006.038,72)
(2006.058,59)
(2006.077,76)
(2006.096,55)
(2006.115,56)
(2006.135,53)
(2006.154,75)
(2006.173,82)
(2006.192,102)
(2006.212,71)
(2006.231,110)
(2006.25,81)
(2006.269,40)
(2006.288,34)
(2006.308,53)
(2006.327,71)
(2006.346,24)
(2006.365,21)
(2006.385,42)
(2006.404,24)
(2006.423,36)
(2006.442,18)
(2006.462,35)
(2006.481,34)
(2006.5,24)
(2006.519,24)
(2006.538,9)
(2006.558,13)
(2006.577,5)
(2006.596,1)
(2006.615,3)
(2006.635,1)
(2006.654,0)
(2006.673,0)
(2006.692,2)
(2006.712,1)
(2006.731,1)
(2006.75,2)
(2006.769,0)
(2006.788,7)
(2006.808,1)
(2006.827,4)
(2006.846,5)
(2006.865,4)
(2006.885,9)
(2006.904,29)
(2006.923,12)
(2006.942,23)
(2006.962,30)
(2006.981,9)
(2007.0,1)
(2007.019,13)
(2007.038,59)
(2007.058,34)
(2007.077,5)
(2007.096,21)
(2007.115,15)
(2007.135,36)
(2007.154,56)
(2007.173,60)
(2007.192,36)
(2007.212,7)
(2007.231,55)
(2007.25,59)
(2007.269,39)
(2007.288,21)
(2007.308,54)
(2007.327,22)
(2007.346,11)
(2007.365,29)
(2007.385,21)
(2007.404,10)
(2007.423,8)
(2007.442,17)
(2007.462,7)
(2007.481,8)
(2007.5,4)
(2007.519,3)
(2007.538,3)
(2007.558,1)
(2007.577,2)
(2007.596,1)
(2007.615,0)
(2007.635,1)
(2007.654,0)
(2007.673,3)
(2007.692,0)
(2007.712,1)
(2007.731,0)
(2007.75,1)
(2007.769,1)
(2007.788,2)
(2007.808,3)
(2007.827,0)
(2007.846,0)
(2007.865,7)
(2007.885,3)
(2007.904,2)
(2007.923,5)
(2007.942,3)
(2007.962,0)
(2007.981,5)
(2008.0,3)
(2008.019,3)
(2008.038,1)
(2008.058,8)
(2008.077,0)
(2008.096,7)
(2008.115,0)
(2008.135,6)
(2008.154,7)
(2008.173,14)
(2008.192,3)
(2008.212,5)
(2008.231,6)
(2008.25,3)
(2008.269,2)
(2008.288,10)
(2008.308,7)
(2008.327,10)
(2008.346,10)
(2008.365,31)
(2008.385,1)
(2008.404,9)
(2008.423,42)
(2008.442,10)
(2008.462,41)
(2008.481,18)
(2008.5,20)
(2008.519,18)
(2008.538,18)
(2008.558,8)
(2008.577,12)
(2008.596,7)
(2008.615,9)
(2008.635,2)
(2008.654,3)
(2008.673,2)
(2008.692,0)
(2008.712,9)
(2008.731,0)
(2008.75,3)
(2008.769,6)
(2008.788,24)
(2008.808,14)
(2008.827,32)
(2008.846,15)
(2008.865,3)
(2008.885,15)
(2008.904,1)
(2008.923,41)
(2008.942,13)
(2008.962,8)
(2008.981,40)
(2009.0,36)
(2009.019,0)
(2009.038,141)
(2009.058,65)
(2009.077,52)
(2009.096,42)
(2009.115,77)
(2009.135,31)
(2009.154,45)
(2009.173,135)
(2009.192,99)
(2009.212,75)
(2009.231,54)
(2009.25,87)
(2009.269,108)
(2009.288,54)
(2009.308,53)
(2009.327,78)
(2009.346,81)
(2009.365,78)
(2009.385,100)
(2009.404,135)
(2009.423,29)
(2009.442,21)
(2009.462,55)
(2009.481,72)
(2009.5,53)
(2009.519,40)
(2009.538,17)
(2009.558,16)
(2009.577,5)
(2009.596,4)
(2009.615,3)
(2009.635,1)
(2009.654,0)
(2009.673,2)
(2009.692,0)
(2009.712,0)
(2009.731,0)
(2009.75,3)
(2009.769,3)
(2009.788,5)
(2009.808,4)
(2009.827,25)
(2009.846,4)
(2009.865,12)
(2009.885,16)
(2009.904,6)
(2009.923,18)
(2009.942,16)
(2009.962,13)
(2009.981,15)
(2010.0,1)
(2010.019,10)
(2010.038,20)
(2010.058,5)
(2010.077,13)
(2010.096,40)
(2010.115,35)
(2010.135,10)
(2010.154,26)
(2010.173,11)
(2010.192,4)
(2010.212,10)
(2010.231,3)
(2010.25,14)
(2010.269,13)
(2010.288,17)
(2010.308,32)
(2010.327,18)
(2010.346,18)
(2010.365,14)
(2010.385,9)
(2010.404,29)
(2010.423,7)
(2010.442,8)
(2010.462,9)
(2010.481,4)
(2010.5,5)
(2010.519,7)
(2010.538,7)
(2010.558,4)
(2010.577,10)
(2010.596,2)
(2010.615,0)
(2010.635,0)
(2010.654,1)
(2010.673,0)
(2010.692,2)
(2010.712,0)
(2010.731,6)
(2010.75,2)
(2010.769,15)
(2010.788,10)
(2010.808,3)
(2010.827,12)
(2010.846,9)
(2010.865,23)
(2010.885,9)
(2010.904,51)
(2010.923,20)
(2010.942,30)
(2010.962,41)
(2010.981,20)
(2011.0,7)
(2011.019,7)
(2011.038,27)
(2011.058,115)
(2011.077,3)
(2011.096,50)
(2011.115,27)
(2011.135,44)
(2011.154,9)
(2011.173,30)
(2011.192,10)
(2011.212,34)
(2011.231,31)
(2011.25,17)
(2011.269,25)
(2011.288,14)
(2011.308,27)
(2011.327,2)
(2011.346,11)
(2011.365,11)
(2011.385,14)
(2011.404,2)
(2011.423,5)
(2011.442,6)
(2011.462,8)
(2011.481,9)
(2011.5,10)
(2011.519,3)
(2011.538,24)
(2011.558,0)
(2011.577,4)
(2011.596,1)
(2011.615,1)
(2011.635,1)
(2011.654,1)
(2011.673,5)
(2011.692,1)
(2011.712,0)
(2011.731,5)
(2011.75,5)
(2011.769,2)
(2011.788,9)
(2011.808,6)
(2011.827,7)
(2011.846,72)
(2011.865,4)
(2011.885,72)
(2011.904,4)
(2011.923,5)
(2011.942,21)
(2011.962,30)
(2011.981,30)
(2012.0,23)
(2012.019,28)
(2012.038,63)
(2012.058,50)
(2012.077,5)
(2012.096,55)
(2012.115,31)
(2012.135,36)
(2012.154,44)
(2012.173,39)
(2012.192,51)
(2012.212,59)
(2012.231,47)
(2012.25,39)
(2012.269,48)
(2012.288,27)
(2012.308,35)
(2012.327,43)
(2012.346,23)
(2012.365,14)
(2012.385,56)
(2012.404,26)
(2012.423,18)
(2012.442,18)
(2012.462,37)
(2012.481,58)
(2012.5,44)
(2012.519,5)
(2012.538,29)
(2012.558,38)
(2012.577,4)
(2012.596,27)
(2012.615,0)
(2012.635,11)
(2012.654,2)
(2012.673,0)
(2012.692,8)
(2012.712,2)
(2012.731,1)
(2012.75,4)
(2012.769,1)
(2012.788,0)
(2012.808,0)
(2012.827,1)
(2012.846,0)
(2012.865,2)
(2012.885,3)
(2012.904,11)
(2012.923,3)
(2012.942,10)
(2012.962,1)
(2012.981,3)
(2013.0,3)
(2013.019,4)
(2013.038,5)
(2013.058,19)
(2013.077,30)
(2013.096,9)
(2013.115,35)
(2013.135,28)
(2013.154,58)
(2013.173,27)
(2013.192,27)
(2013.212,16)
(2013.231,13)
(2013.25,48)
(2013.269,32)
(2013.288,21)
(2013.308,38)
(2013.327,24)
(2013.346,73)
(2013.365,15)
(2013.385,33)
(2013.404,46)
(2013.423,38)
(2013.442,17)
(2013.462,39)
(2013.481,21)
(2013.5,34)
(2013.519,21)
(2013.538,16)
(2013.558,5)
(2013.577,11)
(2013.596,10)
(2013.615,1)
(2013.635,0)
(2013.654,0)
(2013.673,0)
(2013.692,11)
(2013.712,2)
(2013.731,1)
(2013.75,0)
(2013.769,2)
(2013.788,2)
(2013.808,5)
(2013.827,3)
(2013.846,23)
(2013.865,0)
(2013.885,5)
(2013.904,22)
(2013.923,2)
(2013.942,20)
(2013.962,15)
(2013.981,44)
(2014.0,19)
(2014.019,0)
(2014.038,20)
(2014.058,28)
(2014.077,31)
(2014.096,21)
(2014.115,22)
(2014.135,13)
(2014.154,24)
(2014.173,18)
(2014.192,32)
(2014.212,21)
(2014.231,11)
(2014.25,28)
(2014.269,30)
(2014.288,9)
(2014.308,54)
(2014.327,27)
(2014.346,12)
(2014.365,5)
(2014.385,92)
(2014.404,37)
(2014.423,21)
(2014.442,22)
(2014.462,22)
(2014.481,20)
(2014.5,12)
(2014.519,13)
(2014.538,10)
(2014.558,6)
(2014.577,32)
(2014.596,49)
(2014.615,13)
(2014.635,1)
(2014.654,4)
(2014.673,0)
(2014.692,2)
(2014.712,2)
(2014.731,1)
(2014.75,0)
(2014.769,4)
(2014.788,1)
(2014.808,3)
(2014.827,0)
(2014.846,0)
(2014.865,0)
(2014.885,0)
(2014.904,31)
(2014.923,23)
(2014.942,3)
(2014.962,11)
(2014.981,22)
(2015.0,0)
(2015.019,1)
(2015.038,11)
}

\def\VESZPREMDATA{
(2005.019,87)
(2005.038,68)
(2005.058,62)
(2005.077,43)
(2005.096,85)
(2005.115,48)
(2005.135,153)
(2005.154,149)
(2005.173,102)
(2005.192,96)
(2005.212,118)
(2005.231,127)
(2005.25,63)
(2005.269,108)
(2005.288,59)
(2005.308,87)
(2005.327,64)
(2005.346,39)
(2005.365,76)
(2005.385,44)
(2005.404,141)
(2005.423,98)
(2005.442,87)
(2005.462,69)
(2005.481,66)
(2005.5,49)
(2005.519,51)
(2005.538,39)
(2005.558,24)
(2005.577,8)
(2005.596,17)
(2005.615,4)
(2005.635,4)
(2005.654,1)
(2005.673,0)
(2005.692,1)
(2005.712,2)
(2005.731,4)
(2005.75,5)
(2005.769,20)
(2005.788,25)
(2005.808,14)
(2005.827,50)
(2005.846,13)
(2005.865,35)
(2005.885,14)
(2005.904,22)
(2005.923,42)
(2005.942,72)
(2005.962,97)
(2005.981,80)
(2006.0,84)
(2006.019,109)
(2006.038,110)
(2006.058,111)
(2006.077,120)
(2006.096,136)
(2006.115,157)
(2006.135,155)
(2006.154,221)
(2006.173,180)
(2006.192,161)
(2006.212,221)
(2006.231,210)
(2006.25,230)
(2006.269,130)
(2006.288,192)
(2006.308,97)
(2006.327,190)
(2006.346,79)
(2006.365,133)
(2006.385,62)
(2006.404,136)
(2006.423,54)
(2006.442,58)
(2006.462,47)
(2006.481,39)
(2006.5,35)
(2006.519,34)
(2006.538,7)
(2006.558,7)
(2006.577,4)
(2006.596,4)
(2006.615,1)
(2006.635,2)
(2006.654,0)
(2006.673,3)
(2006.692,2)
(2006.712,2)
(2006.731,5)
(2006.75,1)
(2006.769,3)
(2006.788,7)
(2006.808,4)
(2006.827,10)
(2006.846,1)
(2006.865,15)
(2006.885,13)
(2006.904,10)
(2006.923,19)
(2006.942,29)
(2006.962,32)
(2006.981,43)
(2007.0,48)
(2007.019,47)
(2007.038,59)
(2007.058,29)
(2007.077,48)
(2007.096,73)
(2007.115,118)
(2007.135,64)
(2007.154,122)
(2007.173,59)
(2007.192,85)
(2007.212,92)
(2007.231,101)
(2007.25,76)
(2007.269,107)
(2007.288,76)
(2007.308,98)
(2007.327,75)
(2007.346,42)
(2007.365,90)
(2007.385,69)
(2007.404,84)
(2007.423,54)
(2007.442,55)
(2007.462,32)
(2007.481,71)
(2007.5,18)
(2007.519,15)
(2007.538,19)
(2007.558,6)
(2007.577,6)
(2007.596,5)
(2007.615,5)
(2007.635,4)
(2007.654,2)
(2007.673,0)
(2007.692,0)
(2007.712,4)
(2007.731,0)
(2007.75,5)
(2007.769,5)
(2007.788,3)
(2007.808,12)
(2007.827,18)
(2007.846,18)
(2007.865,25)
(2007.885,38)
(2007.904,15)
(2007.923,29)
(2007.942,39)
(2007.962,27)
(2007.981,51)
(2008.0,38)
(2008.019,37)
(2008.038,29)
(2008.058,131)
(2008.077,37)
(2008.096,48)
(2008.115,32)
(2008.135,27)
(2008.154,26)
(2008.173,25)
(2008.192,53)
(2008.212,50)
(2008.231,52)
(2008.25,48)
(2008.269,52)
(2008.288,38)
(2008.308,47)
(2008.327,54)
(2008.346,16)
(2008.365,65)
(2008.385,21)
(2008.404,82)
(2008.423,33)
(2008.442,40)
(2008.462,29)
(2008.481,29)
(2008.5,25)
(2008.519,15)
(2008.538,23)
(2008.558,6)
(2008.577,11)
(2008.596,4)
(2008.615,5)
(2008.635,4)
(2008.654,2)
(2008.673,8)
(2008.692,4)
(2008.712,3)
(2008.731,13)
(2008.75,16)
(2008.769,19)
(2008.788,45)
(2008.808,47)
(2008.827,30)
(2008.846,45)
(2008.865,50)
(2008.885,44)
(2008.904,37)
(2008.923,36)
(2008.942,42)
(2008.962,35)
(2008.981,71)
(2009.0,53)
(2009.019,27)
(2009.038,123)
(2009.058,26)
(2009.077,37)
(2009.096,61)
(2009.115,45)
(2009.135,103)
(2009.154,43)
(2009.173,65)
(2009.192,49)
(2009.212,57)
(2009.231,59)
(2009.25,69)
(2009.269,37)
(2009.288,32)
(2009.308,41)
(2009.327,45)
(2009.346,31)
(2009.365,12)
(2009.385,76)
(2009.404,18)
(2009.423,22)
(2009.442,24)
(2009.462,13)
(2009.481,9)
(2009.5,8)
(2009.519,7)
(2009.538,8)
(2009.558,13)
(2009.577,3)
(2009.596,5)
(2009.615,3)
(2009.635,4)
(2009.654,1)
(2009.673,8)
(2009.692,2)
(2009.712,2)
(2009.731,2)
(2009.75,3)
(2009.769,4)
(2009.788,2)
(2009.808,6)
(2009.827,2)
(2009.846,8)
(2009.865,3)
(2009.885,3)
(2009.904,14)
(2009.923,10)
(2009.942,2)
(2009.962,4)
(2009.981,4)
(2010.0,0)
(2010.019,4)
(2010.038,32)
(2010.058,13)
(2010.077,11)
(2010.096,35)
(2010.115,43)
(2010.135,75)
(2010.154,29)
(2010.173,42)
(2010.192,41)
(2010.212,36)
(2010.231,41)
(2010.25,83)
(2010.269,57)
(2010.288,39)
(2010.308,118)
(2010.327,71)
(2010.346,105)
(2010.365,101)
(2010.385,133)
(2010.404,99)
(2010.423,141)
(2010.442,126)
(2010.462,102)
(2010.481,97)
(2010.5,120)
(2010.519,43)
(2010.538,50)
(2010.558,33)
(2010.577,40)
(2010.596,33)
(2010.615,5)
(2010.635,12)
(2010.654,4)
(2010.673,4)
(2010.692,0)
(2010.712,2)
(2010.731,0)
(2010.75,3)
(2010.769,4)
(2010.788,8)
(2010.808,12)
(2010.827,8)
(2010.846,15)
(2010.865,8)
(2010.885,20)
(2010.904,39)
(2010.923,25)
(2010.942,43)
(2010.962,91)
(2010.981,63)
(2011.0,45)
(2011.019,104)
(2011.038,68)
(2011.058,55)
(2011.077,38)
(2011.096,70)
(2011.115,60)
(2011.135,69)
(2011.154,88)
(2011.173,98)
(2011.192,42)
(2011.212,132)
(2011.231,153)
(2011.25,90)
(2011.269,75)
(2011.288,116)
(2011.308,74)
(2011.327,73)
(2011.346,74)
(2011.365,85)
(2011.385,102)
(2011.404,100)
(2011.423,104)
(2011.442,85)
(2011.462,72)
(2011.481,60)
(2011.5,50)
(2011.519,39)
(2011.538,37)
(2011.558,16)
(2011.577,33)
(2011.596,9)
(2011.615,13)
(2011.635,5)
(2011.654,3)
(2011.673,5)
(2011.692,6)
(2011.712,1)
(2011.731,3)
(2011.75,2)
(2011.769,2)
(2011.788,1)
(2011.808,5)
(2011.827,1)
(2011.846,2)
(2011.865,0)
(2011.885,2)
(2011.904,8)
(2011.923,10)
(2011.942,12)
(2011.962,15)
(2011.981,20)
(2012.0,82)
(2012.019,14)
(2012.038,63)
(2012.058,30)
(2012.077,61)
(2012.096,66)
(2012.115,32)
(2012.135,90)
(2012.154,73)
(2012.173,65)
(2012.192,45)
(2012.212,36)
(2012.231,46)
(2012.25,67)
(2012.269,57)
(2012.288,45)
(2012.308,24)
(2012.327,52)
(2012.346,23)
(2012.365,42)
(2012.385,48)
(2012.404,39)
(2012.423,23)
(2012.442,36)
(2012.462,34)
(2012.481,29)
(2012.5,33)
(2012.519,20)
(2012.538,24)
(2012.558,6)
(2012.577,8)
(2012.596,10)
(2012.615,2)
(2012.635,2)
(2012.654,1)
(2012.673,0)
(2012.692,0)
(2012.712,3)
(2012.731,2)
(2012.75,1)
(2012.769,2)
(2012.788,1)
(2012.808,3)
(2012.827,0)
(2012.846,13)
(2012.865,2)
(2012.885,30)
(2012.904,9)
(2012.923,14)
(2012.942,24)
(2012.962,6)
(2012.981,49)
(2013.0,21)
(2013.019,31)
(2013.038,44)
(2013.058,68)
(2013.077,55)
(2013.096,60)
(2013.115,53)
(2013.135,42)
(2013.154,81)
(2013.173,65)
(2013.192,65)
(2013.212,29)
(2013.231,54)
(2013.25,43)
(2013.269,49)
(2013.288,31)
(2013.308,62)
(2013.327,52)
(2013.346,67)
(2013.365,40)
(2013.385,3)
(2013.404,36)
(2013.423,24)
(2013.442,23)
(2013.462,26)
(2013.481,32)
(2013.5,46)
(2013.519,18)
(2013.538,25)
(2013.558,8)
(2013.577,15)
(2013.596,3)
(2013.615,3)
(2013.635,4)
(2013.654,3)
(2013.673,3)
(2013.692,7)
(2013.712,0)
(2013.731,0)
(2013.75,1)
(2013.769,2)
(2013.788,3)
(2013.808,5)
(2013.827,11)
(2013.846,6)
(2013.865,18)
(2013.885,27)
(2013.904,11)
(2013.923,20)
(2013.942,28)
(2013.962,25)
(2013.981,11)
(2014.0,39)
(2014.019,2)
(2014.038,23)
(2014.058,63)
(2014.077,32)
(2014.096,36)
(2014.115,43)
(2014.135,28)
(2014.154,38)
(2014.173,47)
(2014.192,38)
(2014.212,31)
(2014.231,50)
(2014.25,70)
(2014.269,47)
(2014.288,40)
(2014.308,54)
(2014.327,40)
(2014.346,10)
(2014.365,2)
(2014.385,89)
(2014.404,34)
(2014.423,19)
(2014.442,38)
(2014.462,13)
(2014.481,2)
(2014.5,2)
(2014.519,9)
(2014.538,6)
(2014.558,8)
(2014.577,4)
(2014.596,1)
(2014.615,4)
(2014.635,1)
(2014.654,3)
(2014.673,8)
(2014.692,4)
(2014.712,2)
(2014.731,5)
(2014.75,8)
(2014.769,7)
(2014.788,18)
(2014.808,27)
(2014.827,3)
(2014.846,40)
(2014.865,34)
(2014.885,30)
(2014.904,21)
(2014.923,24)
(2014.942,25)
(2014.962,110)
(2014.981,63)
(2015.0,17)
(2015.019,83)
(2015.038,103)
}

\def\ZALADATA{
(2005.019,68)
(2005.038,26)
(2005.058,44)
(2005.077,31)
(2005.096,60)
(2005.115,60)
(2005.135,70)
(2005.154,54)
(2005.173,42)
(2005.192,54)
(2005.212,43)
(2005.231,36)
(2005.25,36)
(2005.269,38)
(2005.288,30)
(2005.308,22)
(2005.327,23)
(2005.346,32)
(2005.365,47)
(2005.385,21)
(2005.404,43)
(2005.423,34)
(2005.442,37)
(2005.462,47)
(2005.481,70)
(2005.5,18)
(2005.519,34)
(2005.538,14)
(2005.558,7)
(2005.577,9)
(2005.596,2)
(2005.615,4)
(2005.635,1)
(2005.654,2)
(2005.673,2)
(2005.692,0)
(2005.712,0)
(2005.731,3)
(2005.75,16)
(2005.769,13)
(2005.788,17)
(2005.808,8)
(2005.827,16)
(2005.846,18)
(2005.865,40)
(2005.885,34)
(2005.904,44)
(2005.923,65)
(2005.942,38)
(2005.962,47)
(2005.981,40)
(2006.0,31)
(2006.019,107)
(2006.038,67)
(2006.058,43)
(2006.077,60)
(2006.096,45)
(2006.115,48)
(2006.135,85)
(2006.154,60)
(2006.173,76)
(2006.192,37)
(2006.212,58)
(2006.231,39)
(2006.25,81)
(2006.269,36)
(2006.288,42)
(2006.308,45)
(2006.327,22)
(2006.346,74)
(2006.365,62)
(2006.385,67)
(2006.404,30)
(2006.423,51)
(2006.442,37)
(2006.462,40)
(2006.481,31)
(2006.5,12)
(2006.519,26)
(2006.538,20)
(2006.558,9)
(2006.577,6)
(2006.596,1)
(2006.615,0)
(2006.635,0)
(2006.654,0)
(2006.673,2)
(2006.692,0)
(2006.712,2)
(2006.731,1)
(2006.75,3)
(2006.769,10)
(2006.788,10)
(2006.808,4)
(2006.827,12)
(2006.846,6)
(2006.865,18)
(2006.885,13)
(2006.904,16)
(2006.923,11)
(2006.942,32)
(2006.962,35)
(2006.981,32)
(2007.0,24)
(2007.019,42)
(2007.038,63)
(2007.058,36)
(2007.077,67)
(2007.096,54)
(2007.115,70)
(2007.135,35)
(2007.154,45)
(2007.173,29)
(2007.192,42)
(2007.212,24)
(2007.231,39)
(2007.25,64)
(2007.269,27)
(2007.288,42)
(2007.308,44)
(2007.327,51)
(2007.346,37)
(2007.365,45)
(2007.385,54)
(2007.404,49)
(2007.423,32)
(2007.442,32)
(2007.462,16)
(2007.481,11)
(2007.5,12)
(2007.519,9)
(2007.538,0)
(2007.558,0)
(2007.577,0)
(2007.596,2)
(2007.615,0)
(2007.635,0)
(2007.654,3)
(2007.673,0)
(2007.692,0)
(2007.712,1)
(2007.731,1)
(2007.75,0)
(2007.769,0)
(2007.788,6)
(2007.808,8)
(2007.827,4)
(2007.846,7)
(2007.865,21)
(2007.885,20)
(2007.904,10)
(2007.923,13)
(2007.942,18)
(2007.962,15)
(2007.981,11)
(2008.0,13)
(2008.019,12)
(2008.038,13)
(2008.058,21)
(2008.077,11)
(2008.096,13)
(2008.115,1)
(2008.135,5)
(2008.154,5)
(2008.173,22)
(2008.192,11)
(2008.212,18)
(2008.231,18)
(2008.25,10)
(2008.269,21)
(2008.288,4)
(2008.308,22)
(2008.327,14)
(2008.346,1)
(2008.365,11)
(2008.385,19)
(2008.404,14)
(2008.423,7)
(2008.442,12)
(2008.462,7)
(2008.481,2)
(2008.5,10)
(2008.519,5)
(2008.538,1)
(2008.558,13)
(2008.577,1)
(2008.596,3)
(2008.615,3)
(2008.635,1)
(2008.654,0)
(2008.673,3)
(2008.692,2)
(2008.712,5)
(2008.731,1)
(2008.75,9)
(2008.769,5)
(2008.788,0)
(2008.808,5)
(2008.827,25)
(2008.846,4)
(2008.865,30)
(2008.885,2)
(2008.904,22)
(2008.923,5)
(2008.942,4)
(2008.962,49)
(2008.981,17)
(2009.0,13)
(2009.019,10)
(2009.038,24)
(2009.058,10)
(2009.077,8)
(2009.096,23)
(2009.115,15)
(2009.135,13)
(2009.154,6)
(2009.173,55)
(2009.192,22)
(2009.212,17)
(2009.231,5)
(2009.25,5)
(2009.269,51)
(2009.288,14)
(2009.308,17)
(2009.327,25)
(2009.346,82)
(2009.365,9)
(2009.385,2)
(2009.404,43)
(2009.423,52)
(2009.442,5)
(2009.462,42)
(2009.481,13)
(2009.5,4)
(2009.519,51)
(2009.538,18)
(2009.558,1)
(2009.577,21)
(2009.596,4)
(2009.615,8)
(2009.635,1)
(2009.654,4)
(2009.673,2)
(2009.692,0)
(2009.712,1)
(2009.731,1)
(2009.75,2)
(2009.769,0)
(2009.788,0)
(2009.808,3)
(2009.827,15)
(2009.846,3)
(2009.865,31)
(2009.885,13)
(2009.904,25)
(2009.923,18)
(2009.942,27)
(2009.962,55)
(2009.981,24)
(2010.0,8)
(2010.019,11)
(2010.038,26)
(2010.058,12)
(2010.077,8)
(2010.096,15)
(2010.115,2)
(2010.135,7)
(2010.154,5)
(2010.173,2)
(2010.192,7)
(2010.212,5)
(2010.231,4)
(2010.25,2)
(2010.269,16)
(2010.288,6)
(2010.308,25)
(2010.327,13)
(2010.346,15)
(2010.365,20)
(2010.385,14)
(2010.404,11)
(2010.423,30)
(2010.442,31)
(2010.462,13)
(2010.481,26)
(2010.5,50)
(2010.519,11)
(2010.538,24)
(2010.558,6)
(2010.577,9)
(2010.596,5)
(2010.615,6)
(2010.635,3)
(2010.654,4)
(2010.673,6)
(2010.692,3)
(2010.712,4)
(2010.731,0)
(2010.75,18)
(2010.769,1)
(2010.788,24)
(2010.808,11)
(2010.827,35)
(2010.846,60)
(2010.865,29)
(2010.885,33)
(2010.904,75)
(2010.923,31)
(2010.942,22)
(2010.962,49)
(2010.981,35)
(2011.0,58)
(2011.019,71)
(2011.038,16)
(2011.058,86)
(2011.077,56)
(2011.096,41)
(2011.115,43)
(2011.135,21)
(2011.154,46)
(2011.173,35)
(2011.192,46)
(2011.212,35)
(2011.231,42)
(2011.25,43)
(2011.269,49)
(2011.288,41)
(2011.308,54)
(2011.327,22)
(2011.346,18)
(2011.365,65)
(2011.385,32)
(2011.404,40)
(2011.423,44)
(2011.442,45)
(2011.462,26)
(2011.481,19)
(2011.5,28)
(2011.519,37)
(2011.538,6)
(2011.558,15)
(2011.577,9)
(2011.596,15)
(2011.615,5)
(2011.635,2)
(2011.654,0)
(2011.673,1)
(2011.692,4)
(2011.712,1)
(2011.731,0)
(2011.75,6)
(2011.769,4)
(2011.788,0)
(2011.808,0)
(2011.827,5)
(2011.846,4)
(2011.865,7)
(2011.885,4)
(2011.904,0)
(2011.923,6)
(2011.942,12)
(2011.962,33)
(2011.981,12)
(2012.0,69)
(2012.019,46)
(2012.038,29)
(2012.058,23)
(2012.077,3)
(2012.096,36)
(2012.115,12)
(2012.135,216)
(2012.154,12)
(2012.173,22)
(2012.192,15)
(2012.212,24)
(2012.231,23)
(2012.25,28)
(2012.269,7)
(2012.288,8)
(2012.308,26)
(2012.327,21)
(2012.346,41)
(2012.365,2)
(2012.385,39)
(2012.404,12)
(2012.423,13)
(2012.442,10)
(2012.462,4)
(2012.481,20)
(2012.5,6)
(2012.519,10)
(2012.538,2)
(2012.558,8)
(2012.577,4)
(2012.596,1)
(2012.615,2)
(2012.635,1)
(2012.654,4)
(2012.673,0)
(2012.692,4)
(2012.712,0)
(2012.731,2)
(2012.75,0)
(2012.769,0)
(2012.788,1)
(2012.808,0)
(2012.827,1)
(2012.846,7)
(2012.865,1)
(2012.885,6)
(2012.904,2)
(2012.923,8)
(2012.942,40)
(2012.962,11)
(2012.981,16)
(2013.0,1)
(2013.019,18)
(2013.038,0)
(2013.058,5)
(2013.077,16)
(2013.096,19)
(2013.115,1)
(2013.135,73)
(2013.154,2)
(2013.173,37)
(2013.192,13)
(2013.212,23)
(2013.231,15)
(2013.25,17)
(2013.269,6)
(2013.288,11)
(2013.308,2)
(2013.327,17)
(2013.346,13)
(2013.365,1)
(2013.385,3)
(2013.404,18)
(2013.423,13)
(2013.442,9)
(2013.462,12)
(2013.481,5)
(2013.5,19)
(2013.519,17)
(2013.538,12)
(2013.558,18)
(2013.577,11)
(2013.596,5)
(2013.615,4)
(2013.635,3)
(2013.654,2)
(2013.673,4)
(2013.692,0)
(2013.712,1)
(2013.731,0)
(2013.75,0)
(2013.769,0)
(2013.788,1)
(2013.808,0)
(2013.827,2)
(2013.846,2)
(2013.865,5)
(2013.885,1)
(2013.904,6)
(2013.923,13)
(2013.942,2)
(2013.962,3)
(2013.981,2)
(2014.0,4)
(2014.019,2)
(2014.038,5)
(2014.058,15)
(2014.077,13)
(2014.096,20)
(2014.115,17)
(2014.135,6)
(2014.154,14)
(2014.173,89)
(2014.192,33)
(2014.212,2)
(2014.231,13)
(2014.25,22)
(2014.269,32)
(2014.288,51)
(2014.308,20)
(2014.327,98)
(2014.346,3)
(2014.365,0)
(2014.385,91)
(2014.404,13)
(2014.423,8)
(2014.442,50)
(2014.462,38)
(2014.481,31)
(2014.5,24)
(2014.519,24)
(2014.538,0)
(2014.558,34)
(2014.577,16)
(2014.596,6)
(2014.615,1)
(2014.635,0)
(2014.654,9)
(2014.673,3)
(2014.692,2)
(2014.712,1)
(2014.731,0)
(2014.75,0)
(2014.769,1)
(2014.788,0)
(2014.808,4)
(2014.827,0)
(2014.846,7)
(2014.865,10)
(2014.885,7)
(2014.904,1)
(2014.923,4)
(2014.942,0)
(2014.962,10)
(2014.981,9)
(2015.0,10)
(2015.019,2)
(2015.038,25)
}

\def\STDBUDAPEST{
(2006.0,52.921)
(2006.019,52.047)
(2006.038,53.051)
(2006.058,53.05)
(2006.077,53.05)
(2006.096,52.924)
(2006.115,52.462)
(2006.135,53.311)
(2006.154,53.598)
(2006.173,55.9)
(2006.192,60.071)
(2006.212,60.258)
(2006.231,61.561)
(2006.25,62.36)
(2006.269,62.477)
(2006.288,63.76)
(2006.308,73.275)
(2006.327,76.786)
(2006.346,77.289)
(2006.365,83.55)
(2006.385,84.211)
(2006.404,86.488)
(2006.423,88.216)
(2006.442,88.376)
(2006.462,90.69)
(2006.481,90.561)
(2006.5,91.267)
(2006.519,92.576)
(2006.538,92.574)
(2006.558,92.514)
(2006.577,92.376)
(2006.596,92.504)
(2006.615,92.504)
(2006.635,91.683)
(2006.654,91.254)
(2006.673,90.732)
(2006.692,90.527)
(2006.712,90.336)
(2006.731,90.36)
(2006.75,90.236)
(2006.769,90.424)
(2006.788,89.909)
(2006.808,90.059)
(2006.827,90.147)
(2006.846,89.773)
(2006.865,89.694)
(2006.885,89.114)
(2006.904,88.607)
(2006.923,88.081)
(2006.942,87.952)
(2006.962,88.121)
(2006.981,88.112)
(2007.0,88.593)
(2007.019,88.604)
(2007.038,88.195)
(2007.058,88.079)
(2007.077,87.914)
(2007.096,87.903)
(2007.115,87.916)
(2007.135,87.52)
(2007.154,87.518)
(2007.173,87.596)
(2007.192,87.016)
(2007.212,87.037)
(2007.231,87.341)
(2007.25,89.399)
(2007.269,89.564)
(2007.288,89.253)
(2007.308,88.802)
(2007.327,87.611)
(2007.346,87.394)
(2007.365,84.879)
(2007.385,84.572)
(2007.404,89.953)
(2007.423,89.709)
(2007.442,92.302)
(2007.462,91.011)
(2007.481,91.0)
(2007.5,90.362)
(2007.519,88.955)
(2007.538,88.99)
(2007.558,89.543)
(2007.577,90.228)
(2007.596,90.156)
(2007.615,90.637)
(2007.635,91.703)
(2007.654,92.477)
(2007.673,93.18)
(2007.692,93.383)
(2007.712,93.574)
(2007.731,93.682)
(2007.75,93.64)
(2007.769,93.416)
(2007.788,93.8)
(2007.808,93.568)
(2007.827,93.549)
(2007.846,93.95)
(2007.865,93.619)
(2007.885,93.659)
(2007.904,93.75)
(2007.923,94.081)
(2007.942,93.949)
(2007.962,94.15)
(2007.981,94.636)
(2008.0,94.525)
(2008.019,94.945)
(2008.038,95.015)
(2008.058,106.987)
(2008.077,107.01)
(2008.096,107.374)
(2008.115,107.26)
(2008.135,107.264)
(2008.154,107.346)
(2008.173,106.774)
(2008.192,105.997)
(2008.212,106.284)
(2008.231,105.63)
(2008.25,103.687)
(2008.269,103.501)
(2008.288,103.377)
(2008.308,98.04)
(2008.327,96.765)
(2008.346,96.751)
(2008.365,93.862)
(2008.385,93.719)
(2008.404,86.43)
(2008.423,84.724)
(2008.442,80.77)
(2008.462,79.178)
(2008.481,79.32)
(2008.5,79.521)
(2008.519,79.301)
(2008.538,79.537)
(2008.558,79.815)
(2008.577,79.693)
(2008.596,79.497)
(2008.615,79.22)
(2008.635,79.141)
(2008.654,79.162)
(2008.673,79.039)
(2008.692,79.162)
(2008.712,79.225)
(2008.731,79.164)
(2008.75,79.126)
(2008.769,79.271)
(2008.788,79.13)
(2008.808,79.001)
(2008.827,78.886)
(2008.846,78.418)
(2008.865,78.441)
(2008.885,78.509)
(2008.904,78.443)
(2008.923,77.873)
(2008.942,77.712)
(2008.962,77.663)
(2008.981,76.664)
(2009.0,76.769)
(2009.019,75.89)
(2009.038,81.082)
(2009.058,66.996)
(2009.077,66.638)
(2009.096,66.144)
(2009.115,67.287)
(2009.135,67.652)
(2009.154,69.094)
(2009.173,68.777)
(2009.192,69.597)
(2009.212,69.725)
(2009.231,71.249)
(2009.25,71.43)
(2009.269,73.222)
(2009.288,73.223)
(2009.308,74.894)
(2009.327,78.357)
(2009.346,78.622)
(2009.365,78.713)
(2009.385,79.011)
(2009.404,78.688)
(2009.423,79.568)
(2009.442,80.867)
(2009.462,81.469)
(2009.481,81.781)
(2009.5,81.406)
(2009.519,81.454)
(2009.538,80.993)
(2009.558,80.039)
(2009.577,79.456)
(2009.596,80.063)
(2009.615,80.194)
(2009.635,80.124)
(2009.654,80.023)
(2009.673,79.646)
(2009.692,79.353)
(2009.712,79.151)
(2009.731,79.33)
(2009.75,79.577)
(2009.769,79.699)
(2009.788,79.855)
(2009.808,80.086)
(2009.827,80.008)
(2009.846,80.651)
(2009.865,80.953)
(2009.885,81.293)
(2009.904,81.328)
(2009.923,81.073)
(2009.942,81.581)
(2009.962,81.613)
(2009.981,81.519)
(2010.0,81.687)
(2010.019,81.567)
(2010.038,77.221)
(2010.058,72.182)
(2010.077,72.338)
(2010.096,71.562)
(2010.115,70.916)
(2010.135,70.685)
(2010.154,69.665)
(2010.173,69.625)
(2010.192,68.858)
(2010.212,68.733)
(2010.231,67.18)
(2010.25,66.95)
(2010.269,64.593)
(2010.288,65.346)
(2010.308,64.884)
(2010.327,60.456)
(2010.346,60.11)
(2010.365,60.888)
(2010.385,60.667)
(2010.404,60.216)
(2010.423,58.337)
(2010.442,56.312)
(2010.462,55.533)
(2010.481,55.564)
(2010.5,56.354)
(2010.519,56.34)
(2010.538,56.271)
(2010.558,56.272)
(2010.577,56.255)
(2010.596,55.943)
(2010.615,55.777)
(2010.635,55.847)
(2010.654,55.27)
(2010.673,55.612)
(2010.692,55.485)
(2010.712,55.558)
(2010.731,55.478)
(2010.75,55.033)
(2010.769,54.583)
(2010.788,54.216)
(2010.808,53.744)
(2010.827,53.381)
(2010.846,52.779)
(2010.865,52.585)
(2010.885,53.452)
(2010.904,55.713)
(2010.923,56.006)
(2010.942,56.218)
(2010.962,57.769)
(2010.981,57.887)
(2011.0,57.93)
(2011.019,57.542)
(2011.038,57.072)
(2011.058,65.528)
(2011.077,65.53)
(2011.096,66.979)
(2011.115,67.224)
(2011.135,66.69)
(2011.154,66.983)
(2011.173,68.804)
(2011.192,68.711)
(2011.212,69.277)
(2011.231,69.3)
(2011.25,70.845)
(2011.269,71.032)
(2011.288,71.125)
(2011.308,70.789)
(2011.327,70.736)
(2011.346,70.734)
(2011.365,70.554)
(2011.385,70.379)
(2011.404,69.798)
(2011.423,69.928)
(2011.442,69.888)
(2011.462,69.757)
(2011.481,69.399)
(2011.5,69.284)
(2011.519,69.219)
(2011.538,69.233)
(2011.558,69.271)
(2011.577,69.189)
(2011.596,69.589)
(2011.615,69.803)
(2011.635,69.803)
(2011.654,70.252)
(2011.673,70.167)
(2011.692,70.411)
(2011.712,70.021)
(2011.731,69.063)
(2011.75,69.61)
(2011.769,69.845)
(2011.788,70.564)
(2011.808,71.84)
(2011.827,71.794)
(2011.846,71.759)
(2011.865,72.465)
(2011.885,72.048)
(2011.904,71.473)
(2011.923,71.361)
(2011.942,71.157)
(2011.962,70.199)
(2011.981,70.18)
(2012.0,70.269)
(2012.019,71.037)
(2012.038,70.758)
(2012.058,68.347)
(2012.077,69.026)
(2012.096,68.726)
(2012.115,68.3)
(2012.135,69.29)
(2012.154,69.29)
(2012.173,67.398)
(2012.192,67.211)
(2012.212,66.641)
(2012.231,66.525)
(2012.25,66.747)
(2012.269,66.574)
(2012.288,66.868)
(2012.308,66.107)
(2012.327,66.818)
(2012.346,66.946)
(2012.365,66.477)
(2012.385,67.674)
(2012.404,72.211)
(2012.423,72.011)
(2012.442,71.78)
(2012.462,72.975)
(2012.481,72.975)
(2012.5,73.094)
(2012.519,72.999)
(2012.538,73.015)
(2012.558,73.053)
(2012.577,73.166)
(2012.596,72.402)
(2012.615,72.652)
(2012.635,71.69)
(2012.654,71.666)
(2012.673,71.953)
(2012.692,72.298)
(2012.712,72.508)
(2012.731,73.203)
(2012.75,73.203)
(2012.769,73.443)
(2012.788,73.293)
(2012.808,73.219)
(2012.827,73.668)
(2012.846,74.281)
(2012.865,74.775)
(2012.885,74.886)
(2012.904,74.53)
(2012.923,75.356)
(2012.942,75.313)
(2012.962,75.303)
(2012.981,75.301)
(2013.0,75.859)
(2013.019,76.429)
(2013.038,76.574)
(2013.058,76.199)
(2013.077,75.442)
(2013.096,74.327)
(2013.115,75.419)
(2013.135,74.673)
(2013.154,73.585)
(2013.173,74.262)
(2013.192,74.138)
(2013.212,74.006)
(2013.231,74.007)
(2013.25,73.834)
(2013.269,73.663)
(2013.288,72.919)
(2013.308,75.253)
(2013.327,74.604)
(2013.346,74.946)
(2013.365,74.964)
(2013.385,74.126)
(2013.404,69.452)
(2013.423,70.436)
(2013.442,71.71)
(2013.462,70.472)
(2013.481,70.513)
(2013.5,70.558)
(2013.519,70.894)
(2013.538,70.881)
(2013.558,71.085)
(2013.577,70.756)
(2013.596,70.586)
(2013.615,69.746)
(2013.635,70.116)
(2013.654,70.235)
(2013.673,69.765)
(2013.692,68.973)
(2013.712,69.076)
(2013.731,68.948)
(2013.75,67.985)
(2013.769,68.013)
(2013.788,67.861)
(2013.808,68.049)
(2013.827,68.269)
(2013.846,68.37)
(2013.865,68.042)
(2013.885,67.994)
(2013.904,68.118)
(2013.923,67.477)
(2013.942,68.521)
(2013.962,68.521)
(2013.981,68.726)
(2014.0,67.814)
(2014.019,67.705)
(2014.038,68.32)
(2014.058,63.725)
(2014.077,63.536)
(2014.096,63.388)
(2014.115,61.728)
(2014.135,61.158)
(2014.154,61.265)
(2014.173,59.913)
(2014.192,59.962)
(2014.212,59.567)
(2014.231,59.641)
(2014.25,58.369)
(2014.269,58.198)
(2014.288,60.456)
(2014.308,56.817)
(2014.327,57.172)
(2014.346,56.033)
(2014.365,56.578)
(2014.385,55.79)
(2014.404,70.095)
(2014.423,69.852)
(2014.442,68.358)
(2014.462,71.203)
(2014.481,70.975)
(2014.5,71.737)
(2014.519,71.594)
(2014.538,71.625)
(2014.558,71.045)
(2014.577,71.016)
(2014.596,71.208)
(2014.615,71.271)
(2014.635,71.283)
(2014.654,71.34)
(2014.673,71.632)
(2014.692,72.059)
(2014.712,72.138)
(2014.731,72.211)
(2014.75,72.713)
(2014.769,72.549)
(2014.788,72.517)
(2014.808,72.315)
(2014.827,72.315)
(2014.846,72.263)
(2014.865,72.158)
(2014.885,72.336)
(2014.904,72.594)
(2014.923,72.515)
(2014.942,72.549)
(2014.962,72.52)
(2014.981,72.663)
(2015.0,72.868)
(2015.019,72.478)
}

\def\STDFEJER{
(2006.0,37.776)
(2006.019,35.708)
(2006.038,35.79)
(2006.058,35.234)
(2006.077,35.434)
(2006.096,34.727)
(2006.115,34.621)
(2006.135,33.697)
(2006.154,32.167)
(2006.173,30.165)
(2006.192,29.109)
(2006.212,28.671)
(2006.231,26.422)
(2006.25,26.529)
(2006.269,25.987)
(2006.288,26.11)
(2006.308,25.168)
(2006.327,26.286)
(2006.346,26.042)
(2006.365,25.049)
(2006.385,23.551)
(2006.404,22.885)
(2006.423,22.008)
(2006.442,21.136)
(2006.462,20.944)
(2006.481,19.807)
(2006.5,19.866)
(2006.519,19.865)
(2006.538,20.035)
(2006.558,20.152)
(2006.577,20.352)
(2006.596,20.538)
(2006.615,20.471)
(2006.635,20.494)
(2006.654,20.538)
(2006.673,20.515)
(2006.692,20.491)
(2006.712,20.539)
(2006.731,20.407)
(2006.75,20.451)
(2006.769,20.52)
(2006.788,20.792)
(2006.808,20.957)
(2006.827,20.957)
(2006.846,20.963)
(2006.865,20.945)
(2006.885,20.917)
(2006.904,21.111)
(2006.923,21.627)
(2006.942,23.851)
(2006.962,24.822)
(2006.981,24.793)
(2007.0,24.791)
(2007.019,26.399)
(2007.038,26.798)
(2007.058,26.802)
(2007.077,29.015)
(2007.096,29.124)
(2007.115,28.987)
(2007.135,28.925)
(2007.154,29.858)
(2007.173,29.889)
(2007.192,30.868)
(2007.212,30.877)
(2007.231,31.632)
(2007.25,32.164)
(2007.269,32.295)
(2007.288,32.206)
(2007.308,32.61)
(2007.327,31.726)
(2007.346,31.785)
(2007.365,31.808)
(2007.385,31.821)
(2007.404,31.775)
(2007.423,31.723)
(2007.442,31.828)
(2007.462,32.031)
(2007.481,32.223)
(2007.5,32.254)
(2007.519,32.417)
(2007.538,32.455)
(2007.558,32.475)
(2007.577,32.453)
(2007.596,32.453)
(2007.615,32.54)
(2007.635,32.518)
(2007.654,32.563)
(2007.673,32.563)
(2007.692,32.541)
(2007.712,32.475)
(2007.731,32.495)
(2007.75,32.357)
(2007.769,32.25)
(2007.788,31.951)
(2007.808,31.845)
(2007.827,32.116)
(2007.846,32.373)
(2007.865,32.419)
(2007.885,32.411)
(2007.904,32.503)
(2007.923,32.355)
(2007.942,31.14)
(2007.962,30.949)
(2007.981,30.949)
(2008.0,31.093)
(2008.019,29.937)
(2008.038,32.054)
(2008.058,32.465)
(2008.077,30.557)
(2008.096,30.568)
(2008.115,30.583)
(2008.135,30.994)
(2008.154,31.429)
(2008.173,32.745)
(2008.192,32.214)
(2008.212,32.81)
(2008.231,32.7)
(2008.25,33.313)
(2008.269,35.395)
(2008.288,37.064)
(2008.308,37.627)
(2008.327,40.999)
(2008.346,40.866)
(2008.365,42.225)
(2008.385,42.244)
(2008.404,42.786)
(2008.423,44.406)
(2008.442,44.417)
(2008.462,44.315)
(2008.481,44.027)
(2008.5,43.932)
(2008.519,43.661)
(2008.538,43.484)
(2008.558,43.425)
(2008.577,43.425)
(2008.596,43.148)
(2008.615,43.06)
(2008.635,42.995)
(2008.654,42.972)
(2008.673,43.018)
(2008.692,43.04)
(2008.712,43.084)
(2008.731,43.171)
(2008.75,43.314)
(2008.769,43.532)
(2008.788,43.749)
(2008.808,43.717)
(2008.827,43.717)
(2008.846,43.637)
(2008.865,43.964)
(2008.885,44.293)
(2008.904,44.44)
(2008.923,44.845)
(2008.942,45.185)
(2008.962,45.28)
(2008.981,45.504)
(2009.0,45.889)
(2009.019,46.284)
(2009.038,45.089)
(2009.058,44.982)
(2009.077,45.154)
(2009.096,45.391)
(2009.115,45.537)
(2009.135,45.472)
(2009.154,44.792)
(2009.173,43.773)
(2009.192,43.683)
(2009.212,43.218)
(2009.231,42.682)
(2009.25,41.468)
(2009.269,39.053)
(2009.288,36.983)
(2009.308,35.543)
(2009.327,30.273)
(2009.346,30.07)
(2009.365,26.838)
(2009.385,26.503)
(2009.404,24.569)
(2009.423,19.185)
(2009.442,18.937)
(2009.462,17.088)
(2009.481,16.729)
(2009.5,16.488)
(2009.519,16.406)
(2009.538,16.426)
(2009.558,16.394)
(2009.577,16.394)
(2009.596,16.398)
(2009.615,16.468)
(2009.635,16.468)
(2009.654,16.487)
(2009.673,16.468)
(2009.692,16.485)
(2009.712,16.451)
(2009.731,16.416)
(2009.75,16.355)
(2009.769,16.304)
(2009.788,16.233)
(2009.808,16.26)
(2009.827,16.217)
(2009.846,16.198)
(2009.865,16.126)
(2009.885,16.015)
(2009.904,15.962)
(2009.923,16.331)
(2009.942,16.275)
(2009.962,16.483)
(2009.981,16.468)
(2010.0,16.3)
(2010.019,16.227)
(2010.038,18.368)
(2010.058,18.428)
(2010.077,18.553)
(2010.096,18.721)
(2010.115,18.935)
(2010.135,18.901)
(2010.154,18.785)
(2010.173,18.819)
(2010.192,18.572)
(2010.212,19.133)
(2010.231,19.382)
(2010.25,20.337)
(2010.269,20.912)
(2010.288,20.613)
(2010.308,22.916)
(2010.327,23.369)
(2010.346,23.613)
(2010.365,24.443)
(2010.385,24.687)
(2010.404,25.009)
(2010.423,24.899)
(2010.442,26.372)
(2010.462,26.44)
(2010.481,27.102)
(2010.5,27.623)
(2010.519,27.455)
(2010.538,27.273)
(2010.558,27.157)
(2010.577,26.963)
(2010.596,27.157)
(2010.615,27.085)
(2010.635,26.986)
(2010.654,26.89)
(2010.673,26.89)
(2010.692,26.866)
(2010.712,26.914)
(2010.731,26.914)
(2010.75,27.004)
(2010.769,27.099)
(2010.788,27.264)
(2010.808,27.452)
(2010.827,27.548)
(2010.846,27.664)
(2010.865,27.882)
(2010.885,28.075)
(2010.904,28.05)
(2010.923,28.166)
(2010.942,28.124)
(2010.962,28.373)
(2010.981,28.504)
(2011.0,28.421)
(2011.019,28.422)
(2011.038,27.552)
(2011.058,27.565)
(2011.077,27.544)
(2011.096,27.481)
(2011.115,27.444)
(2011.135,27.819)
(2011.154,27.851)
(2011.173,28.013)
(2011.192,28.449)
(2011.212,28.377)
(2011.231,28.535)
(2011.25,28.561)
(2011.269,29.573)
(2011.288,29.929)
(2011.308,30.535)
(2011.327,30.131)
(2011.346,30.082)
(2011.365,30.535)
(2011.385,30.564)
(2011.404,30.691)
(2011.423,30.709)
(2011.442,29.854)
(2011.462,29.963)
(2011.481,29.38)
(2011.5,29.015)
(2011.519,29.052)
(2011.538,29.049)
(2011.558,29.006)
(2011.577,28.982)
(2011.596,28.926)
(2011.615,28.907)
(2011.635,28.99)
(2011.654,29.07)
(2011.673,29.05)
(2011.692,29.091)
(2011.712,29.112)
(2011.731,29.173)
(2011.75,29.194)
(2011.769,29.06)
(2011.788,28.92)
(2011.808,28.773)
(2011.827,28.597)
(2011.846,28.41)
(2011.865,28.507)
(2011.885,28.338)
(2011.904,28.338)
(2011.923,28.274)
(2011.942,28.399)
(2011.962,28.3)
(2011.981,28.125)
(2012.0,28.157)
(2012.019,28.547)
(2012.038,28.541)
(2012.058,29.935)
(2012.077,29.991)
(2012.096,29.96)
(2012.115,30.092)
(2012.135,29.975)
(2012.154,30.911)
(2012.173,31.041)
(2012.192,32.066)
(2012.212,32.102)
(2012.231,32.209)
(2012.25,33.594)
(2012.269,33.221)
(2012.288,33.176)
(2012.308,31.496)
(2012.327,31.923)
(2012.346,32.361)
(2012.365,31.509)
(2012.385,32.678)
(2012.404,32.607)
(2012.423,34.409)
(2012.442,34.218)
(2012.462,34.307)
(2012.481,34.344)
(2012.5,34.574)
(2012.519,34.564)
(2012.538,34.662)
(2012.558,34.726)
(2012.577,34.684)
(2012.596,34.767)
(2012.615,34.747)
(2012.635,34.788)
(2012.654,34.722)
(2012.673,34.744)
(2012.692,34.744)
(2012.712,34.657)
(2012.731,34.657)
(2012.75,34.613)
(2012.769,34.76)
(2012.788,34.639)
(2012.808,34.715)
(2012.827,34.585)
(2012.846,34.646)
(2012.865,34.625)
(2012.885,34.443)
(2012.904,34.52)
(2012.923,34.534)
(2012.942,34.363)
(2012.962,34.33)
(2012.981,34.331)
(2013.0,34.344)
(2013.019,34.352)
(2013.038,34.534)
(2013.058,34.253)
(2013.077,34.243)
(2013.096,34.24)
(2013.115,34.129)
(2013.135,34.097)
(2013.154,33.346)
(2013.173,33.147)
(2013.192,31.909)
(2013.212,31.79)
(2013.231,31.463)
(2013.25,29.341)
(2013.269,28.631)
(2013.288,28.385)
(2013.308,28.36)
(2013.327,27.557)
(2013.346,26.648)
(2013.365,26.577)
(2013.385,24.33)
(2013.404,23.647)
(2013.423,19.801)
(2013.442,19.17)
(2013.462,18.302)
(2013.481,17.87)
(2013.5,16.534)
(2013.519,16.604)
(2013.538,15.63)
(2013.558,15.67)
(2013.577,15.785)
(2013.596,15.807)
(2013.615,15.845)
(2013.635,15.807)
(2013.654,15.873)
(2013.673,15.851)
(2013.692,15.851)
(2013.712,15.871)
(2013.731,15.871)
(2013.75,15.914)
(2013.769,15.724)
(2013.788,15.933)
(2013.808,15.84)
(2013.827,15.933)
(2013.846,15.922)
(2013.865,15.837)
(2013.885,15.638)
(2013.904,16.104)
(2013.923,16.122)
(2013.942,16.196)
(2013.962,16.272)
(2013.981,16.089)
(2014.0,16.268)
(2014.019,16.46)
(2014.038,16.62)
(2014.058,13.775)
(2014.077,14.925)
(2014.096,15.131)
(2014.115,15.089)
(2014.135,15.099)
(2014.154,15.09)
(2014.173,16.472)
(2014.192,16.472)
(2014.212,18.264)
(2014.231,18.241)
(2014.25,18.32)
(2014.269,18.767)
(2014.288,18.964)
(2014.308,19.119)
(2014.327,19.361)
(2014.346,19.579)
(2014.365,19.788)
(2014.385,21.696)
(2014.404,21.835)
(2014.423,21.728)
(2014.442,22.769)
(2014.462,22.769)
(2014.481,22.884)
(2014.5,23.276)
(2014.519,23.31)
(2014.538,23.325)
(2014.558,23.277)
(2014.577,23.208)
(2014.596,23.208)
(2014.615,23.208)
(2014.635,23.225)
(2014.654,23.166)
(2014.673,23.095)
(2014.692,23.074)
(2014.712,23.074)
(2014.731,23.015)
(2014.75,22.995)
(2014.769,23.117)
(2014.788,23.01)
(2014.808,22.958)
(2014.827,22.755)
(2014.846,22.76)
(2014.865,22.696)
(2014.885,22.662)
(2014.904,22.464)
(2014.923,22.399)
(2014.942,22.497)
(2014.962,22.39)
(2014.981,22.548)
(2015.0,22.611)
(2015.019,22.611)
}

\def\STDSZABOLCS{
(2006.0,35.368)
(2006.019,35.185)
(2006.038,35.18)
(2006.058,35.268)
(2006.077,35.27)
(2006.096,35.174)
(2006.115,35.093)
(2006.135,34.52)
(2006.154,34.133)
(2006.173,33.893)
(2006.192,33.879)
(2006.212,33.837)
(2006.231,33.81)
(2006.25,24.422)
(2006.269,23.933)
(2006.288,24.882)
(2006.308,25.361)
(2006.327,26.786)
(2006.346,27.189)
(2006.365,27.848)
(2006.385,28.359)
(2006.404,27.23)
(2006.423,27.439)
(2006.442,26.561)
(2006.462,26.461)
(2006.481,25.919)
(2006.5,25.83)
(2006.519,24.77)
(2006.538,24.859)
(2006.558,24.721)
(2006.577,24.802)
(2006.596,24.958)
(2006.615,25.064)
(2006.635,25.007)
(2006.654,25.007)
(2006.673,25.052)
(2006.692,25.029)
(2006.712,24.986)
(2006.731,24.986)
(2006.75,24.911)
(2006.769,24.827)
(2006.788,24.935)
(2006.808,24.832)
(2006.827,24.708)
(2006.846,24.587)
(2006.865,24.621)
(2006.885,24.728)
(2006.904,24.909)
(2006.923,24.95)
(2006.942,24.995)
(2006.962,25.086)
(2006.981,25.291)
(2007.0,25.642)
(2007.019,25.975)
(2007.038,26.153)
(2007.058,26.112)
(2007.077,26.163)
(2007.096,26.437)
(2007.115,29.325)
(2007.135,29.192)
(2007.154,29.311)
(2007.173,29.018)
(2007.192,28.873)
(2007.212,28.884)
(2007.231,29.415)
(2007.25,29.188)
(2007.269,28.48)
(2007.288,27.702)
(2007.308,27.967)
(2007.327,26.626)
(2007.346,26.705)
(2007.365,26.152)
(2007.385,26.122)
(2007.404,26.042)
(2007.423,26.404)
(2007.442,28.668)
(2007.462,28.739)
(2007.481,28.722)
(2007.5,28.568)
(2007.519,28.714)
(2007.538,28.746)
(2007.558,28.981)
(2007.577,28.958)
(2007.596,28.939)
(2007.615,28.82)
(2007.635,28.979)
(2007.654,28.979)
(2007.673,28.999)
(2007.692,29.037)
(2007.712,29.131)
(2007.731,29.22)
(2007.75,29.331)
(2007.769,29.363)
(2007.788,29.348)
(2007.808,29.398)
(2007.827,29.531)
(2007.846,29.483)
(2007.865,29.448)
(2007.885,29.349)
(2007.904,29.462)
(2007.923,29.471)
(2007.942,29.507)
(2007.962,30.018)
(2007.981,29.754)
(2008.0,34.723)
(2008.019,34.592)
(2008.038,35.893)
(2008.058,42.28)
(2008.077,42.611)
(2008.096,42.709)
(2008.115,41.762)
(2008.135,41.78)
(2008.154,41.895)
(2008.173,42.064)
(2008.192,42.071)
(2008.212,42.109)
(2008.231,42.163)
(2008.25,42.146)
(2008.269,42.065)
(2008.288,42.077)
(2008.308,41.87)
(2008.327,41.866)
(2008.346,41.988)
(2008.365,42.969)
(2008.385,42.83)
(2008.404,42.941)
(2008.423,42.743)
(2008.442,41.771)
(2008.462,41.757)
(2008.481,41.742)
(2008.5,41.875)
(2008.519,41.979)
(2008.538,41.979)
(2008.558,41.568)
(2008.577,41.63)
(2008.596,41.689)
(2008.615,41.557)
(2008.635,41.253)
(2008.654,41.253)
(2008.673,41.274)
(2008.692,41.274)
(2008.712,41.253)
(2008.731,41.253)
(2008.75,41.294)
(2008.769,41.373)
(2008.788,41.683)
(2008.808,41.904)
(2008.827,41.904)
(2008.846,42.183)
(2008.865,42.432)
(2008.885,42.741)
(2008.904,42.798)
(2008.923,42.996)
(2008.942,42.939)
(2008.962,42.643)
(2008.981,42.84)
(2009.0,39.03)
(2009.019,38.817)
(2009.038,37.569)
(2009.058,29.943)
(2009.077,29.285)
(2009.096,29.364)
(2009.115,28.112)
(2009.135,28.883)
(2009.154,28.397)
(2009.173,28.895)
(2009.192,28.762)
(2009.212,33.572)
(2009.231,33.031)
(2009.25,33.207)
(2009.269,33.582)
(2009.288,33.296)
(2009.308,34.125)
(2009.327,34.075)
(2009.346,34.193)
(2009.365,32.238)
(2009.385,32.301)
(2009.404,32.32)
(2009.423,32.29)
(2009.442,32.003)
(2009.462,31.924)
(2009.481,31.993)
(2009.5,32.055)
(2009.519,31.902)
(2009.538,31.813)
(2009.558,32.02)
(2009.577,32.075)
(2009.596,32.075)
(2009.615,32.273)
(2009.635,32.502)
(2009.654,32.502)
(2009.673,32.467)
(2009.692,32.485)
(2009.712,32.436)
(2009.731,32.418)
(2009.75,32.311)
(2009.769,32.245)
(2009.788,32.29)
(2009.808,32.258)
(2009.827,32.186)
(2009.846,32.243)
(2009.865,32.243)
(2009.885,32.308)
(2009.904,32.477)
(2009.923,32.501)
(2009.942,32.605)
(2009.962,32.602)
(2009.981,32.05)
(2010.0,32.163)
(2010.019,32.087)
(2010.038,32.264)
(2010.058,31.644)
(2010.077,31.644)
(2010.096,30.969)
(2010.115,30.971)
(2010.135,29.908)
(2010.154,29.941)
(2010.173,28.701)
(2010.192,29.634)
(2010.212,24.286)
(2010.231,24.185)
(2010.25,23.575)
(2010.269,22.36)
(2010.288,22.106)
(2010.308,19.769)
(2010.327,20.608)
(2010.346,19.755)
(2010.365,19.81)
(2010.385,19.816)
(2010.404,19.616)
(2010.423,19.271)
(2010.442,19.072)
(2010.462,19.038)
(2010.481,19.059)
(2010.5,20.028)
(2010.519,20.062)
(2010.538,20.252)
(2010.558,20.209)
(2010.577,20.183)
(2010.596,20.093)
(2010.615,20.108)
(2010.635,20.092)
(2010.654,20.108)
(2010.673,20.123)
(2010.692,20.107)
(2010.712,20.163)
(2010.731,20.137)
(2010.75,20.204)
(2010.769,20.271)
(2010.788,20.311)
(2010.808,20.257)
(2010.827,20.257)
(2010.846,20.257)
(2010.865,20.31)
(2010.885,20.248)
(2010.904,20.2)
(2010.923,20.274)
(2010.942,20.293)
(2010.962,20.293)
(2010.981,20.347)
(2011.0,20.254)
(2011.019,20.536)
(2011.038,20.403)
(2011.058,20.403)
(2011.077,21.468)
(2011.096,21.925)
(2011.115,22.099)
(2011.135,22.696)
(2011.154,22.633)
(2011.173,25.031)
(2011.192,24.113)
(2011.212,26.438)
(2011.231,26.674)
(2011.25,34.592)
(2011.269,35.925)
(2011.288,36.298)
(2011.308,37.184)
(2011.327,36.918)
(2011.346,36.909)
(2011.365,41.478)
(2011.385,43.427)
(2011.404,44.131)
(2011.423,44.159)
(2011.442,44.025)
(2011.462,43.787)
(2011.481,43.6)
(2011.5,43.499)
(2011.519,43.372)
(2011.538,43.535)
(2011.558,43.522)
(2011.577,43.596)
(2011.596,43.654)
(2011.615,43.604)
(2011.635,43.571)
(2011.654,43.536)
(2011.673,43.412)
(2011.692,43.412)
(2011.712,43.412)
(2011.731,43.428)
(2011.75,43.364)
(2011.769,43.249)
(2011.788,43.232)
(2011.808,43.272)
(2011.827,43.154)
(2011.846,43.005)
(2011.865,43.005)
(2011.885,42.871)
(2011.904,42.644)
(2011.923,42.644)
(2011.942,42.704)
(2011.962,42.92)
(2011.981,42.872)
(2012.0,42.743)
(2012.019,42.905)
(2012.038,42.968)
(2012.058,42.981)
(2012.077,42.952)
(2012.096,43.019)
(2012.115,43.168)
(2012.135,43.249)
(2012.154,43.381)
(2012.173,42.528)
(2012.192,42.766)
(2012.212,41.026)
(2012.231,41.186)
(2012.25,35.394)
(2012.269,34.239)
(2012.288,33.994)
(2012.308,32.959)
(2012.327,33.364)
(2012.346,33.44)
(2012.365,27.322)
(2012.385,24.243)
(2012.404,21.931)
(2012.423,20.825)
(2012.442,20.435)
(2012.462,20.328)
(2012.481,20.004)
(2012.5,19.85)
(2012.519,20.443)
(2012.538,19.532)
(2012.558,19.562)
(2012.577,19.456)
(2012.596,19.403)
(2012.615,19.333)
(2012.635,19.333)
(2012.654,19.333)
(2012.673,19.457)
(2012.692,19.372)
(2012.712,19.352)
(2012.731,19.389)
(2012.75,19.473)
(2012.769,19.552)
(2012.788,19.59)
(2012.808,19.54)
(2012.827,19.672)
(2012.846,19.683)
(2012.865,19.598)
(2012.885,19.689)
(2012.904,19.677)
(2012.923,19.677)
(2012.942,19.73)
(2012.962,17.693)
(2012.981,17.458)
(2013.0,17.299)
(2013.019,17.425)
(2013.038,18.548)
(2013.058,18.861)
(2013.077,18.884)
(2013.096,18.963)
(2013.115,19.03)
(2013.135,19.128)
(2013.154,19.125)
(2013.173,19.166)
(2013.192,19.085)
(2013.212,19.077)
(2013.231,19.47)
(2013.25,19.509)
(2013.269,23.026)
(2013.288,22.976)
(2013.308,23.006)
(2013.327,24.393)
(2013.346,24.212)
(2013.365,24.342)
(2013.385,23.882)
(2013.404,24.659)
(2013.423,24.737)
(2013.442,24.963)
(2013.462,24.744)
(2013.481,25.688)
(2013.5,25.643)
(2013.519,27.117)
(2013.538,27.053)
(2013.558,27.818)
(2013.577,27.809)
(2013.596,27.693)
(2013.615,27.693)
(2013.635,27.748)
(2013.654,27.785)
(2013.673,27.701)
(2013.692,27.767)
(2013.712,27.749)
(2013.731,27.767)
(2013.75,27.786)
(2013.769,27.752)
(2013.788,27.582)
(2013.808,27.846)
(2013.827,27.709)
(2013.846,27.664)
(2013.865,27.541)
(2013.885,27.661)
(2013.904,27.796)
(2013.923,27.654)
(2013.942,27.658)
(2013.962,27.718)
(2013.981,27.664)
(2014.0,27.632)
(2014.019,27.632)
(2014.038,27.149)
(2014.058,27.24)
(2014.077,27.284)
(2014.096,27.344)
(2014.115,28.278)
(2014.135,28.295)
(2014.154,28.257)
(2014.173,28.307)
(2014.192,28.309)
(2014.212,28.531)
(2014.231,28.273)
(2014.25,28.224)
(2014.269,25.671)
(2014.288,25.68)
(2014.308,25.948)
(2014.327,24.076)
(2014.346,24.131)
(2014.365,24.048)
(2014.385,23.585)
(2014.404,23.088)
(2014.423,23.213)
(2014.442,23.544)
(2014.462,23.602)
(2014.481,22.471)
(2014.5,22.544)
(2014.519,20.101)
(2014.538,19.881)
(2014.558,18.423)
(2014.577,18.418)
(2014.596,18.52)
(2014.615,18.488)
(2014.635,18.326)
(2014.654,18.307)
(2014.673,18.388)
(2014.692,18.388)
(2014.712,18.425)
(2014.731,18.339)
(2014.75,18.339)
(2014.769,18.391)
(2014.788,18.445)
(2014.808,18.408)
(2014.827,18.53)
(2014.846,18.54)
(2014.865,18.537)
(2014.885,18.53)
(2014.904,18.467)
(2014.923,18.514)
(2014.942,18.28)
(2014.962,18.287)
(2014.981,18.371)
(2015.0,18.376)
(2015.019,18.262)
}

\def\STDZALA{
(2006.0,19.856)
(2006.019,21.888)
(2006.038,22.413)
(2006.058,22.403)
(2006.077,22.727)
(2006.096,22.472)
(2006.115,22.248)
(2006.135,22.827)
(2006.154,22.95)
(2006.173,23.682)
(2006.192,23.51)
(2006.212,23.723)
(2006.231,23.734)
(2006.25,24.623)
(2006.269,24.618)
(2006.288,24.635)
(2006.308,24.617)
(2006.327,24.627)
(2006.346,25.206)
(2006.365,25.421)
(2006.385,25.689)
(2006.404,25.69)
(2006.423,25.765)
(2006.442,25.765)
(2006.462,25.729)
(2006.481,25.311)
(2006.5,25.406)
(2006.519,25.44)
(2006.538,25.355)
(2006.558,25.313)
(2006.577,25.377)
(2006.596,25.403)
(2006.615,25.505)
(2006.635,25.531)
(2006.654,25.584)
(2006.673,25.584)
(2006.692,25.584)
(2006.712,25.531)
(2006.731,25.582)
(2006.75,25.834)
(2006.769,25.887)
(2006.788,26.0)
(2006.808,26.086)
(2006.827,26.147)
(2006.846,26.348)
(2006.865,26.437)
(2006.885,26.598)
(2006.904,26.672)
(2006.923,26.47)
(2006.942,26.458)
(2006.962,26.381)
(2006.981,26.357)
(2007.0,26.379)
(2007.019,24.233)
(2007.038,24.123)
(2007.058,24.072)
(2007.077,24.257)
(2007.096,24.392)
(2007.115,24.877)
(2007.135,23.709)
(2007.154,23.432)
(2007.173,22.518)
(2007.192,22.563)
(2007.212,22.2)
(2007.231,22.2)
(2007.25,21.536)
(2007.269,21.507)
(2007.288,21.507)
(2007.308,21.492)
(2007.327,21.711)
(2007.346,20.788)
(2007.365,20.373)
(2007.385,19.96)
(2007.404,20.184)
(2007.423,19.923)
(2007.442,19.886)
(2007.462,19.856)
(2007.481,19.957)
(2007.5,19.957)
(2007.519,20.094)
(2007.538,20.389)
(2007.558,20.565)
(2007.577,20.689)
(2007.596,20.667)
(2007.615,20.667)
(2007.635,20.667)
(2007.654,20.601)
(2007.673,20.646)
(2007.692,20.646)
(2007.712,20.668)
(2007.731,20.668)
(2007.75,20.733)
(2007.769,20.917)
(2007.788,20.979)
(2007.808,20.91)
(2007.827,21.033)
(2007.846,21.016)
(2007.865,21.002)
(2007.885,20.948)
(2007.904,21.013)
(2007.923,20.99)
(2007.942,20.986)
(2007.962,20.974)
(2007.981,21.021)
(2008.0,21.073)
(2008.019,20.97)
(2008.038,20.246)
(2008.058,20.151)
(2008.077,19.174)
(2008.096,18.621)
(2008.115,17.42)
(2008.135,17.361)
(2008.154,17.028)
(2008.173,16.961)
(2008.192,16.615)
(2008.212,16.583)
(2008.231,16.279)
(2008.25,14.837)
(2008.269,14.766)
(2008.288,14.334)
(2008.308,13.756)
(2008.327,12.722)
(2008.346,12.367)
(2008.365,11.483)
(2008.385,9.891)
(2008.404,8.344)
(2008.423,7.776)
(2008.442,7.133)
(2008.462,7.078)
(2008.481,7.141)
(2008.5,7.131)
(2008.519,7.151)
(2008.538,7.129)
(2008.558,7.039)
(2008.577,7.016)
(2008.596,6.997)
(2008.615,6.933)
(2008.635,6.909)
(2008.654,6.973)
(2008.673,6.909)
(2008.692,6.863)
(2008.712,6.791)
(2008.731,6.791)
(2008.75,6.663)
(2008.769,6.56)
(2008.788,6.675)
(2008.808,6.701)
(2008.827,6.985)
(2008.846,7.021)
(2008.865,7.395)
(2008.885,7.34)
(2008.904,7.534)
(2008.923,7.55)
(2008.942,7.5)
(2008.962,9.235)
(2008.981,9.283)
(2009.0,9.283)
(2009.019,9.28)
(2009.038,9.465)
(2009.058,9.348)
(2009.077,9.352)
(2009.096,9.511)
(2009.115,9.441)
(2009.135,9.413)
(2009.154,9.403)
(2009.173,11.1)
(2009.192,11.194)
(2009.212,11.184)
(2009.231,11.184)
(2009.25,11.216)
(2009.269,12.411)
(2009.288,12.365)
(2009.308,12.307)
(2009.327,12.434)
(2009.346,15.61)
(2009.365,15.619)
(2009.385,15.682)
(2009.404,16.203)
(2009.423,16.994)
(2009.442,17.043)
(2009.462,17.416)
(2009.481,17.32)
(2009.5,17.376)
(2009.519,17.985)
(2009.538,17.859)
(2009.558,17.981)
(2009.577,17.862)
(2009.596,17.847)
(2009.615,17.786)
(2009.635,17.786)
(2009.654,17.722)
(2009.673,17.737)
(2009.692,17.772)
(2009.712,17.832)
(2009.731,17.832)
(2009.75,17.916)
(2009.769,17.992)
(2009.788,17.992)
(2009.808,18.019)
(2009.827,17.981)
(2009.846,17.994)
(2009.865,18.009)
(2009.885,17.904)
(2009.904,17.926)
(2009.923,17.853)
(2009.942,17.814)
(2009.962,18.037)
(2009.981,18.059)
(2010.0,18.096)
(2010.019,18.089)
(2010.038,18.105)
(2010.058,18.091)
(2010.077,18.091)
(2010.096,18.078)
(2010.115,18.199)
(2010.135,18.244)
(2010.154,18.256)
(2010.173,17.574)
(2010.192,17.596)
(2010.212,17.656)
(2010.231,17.668)
(2010.25,17.707)
(2010.269,16.992)
(2010.288,17.034)
(2010.308,17.092)
(2010.327,17.032)
(2010.346,14.171)
(2010.365,14.189)
(2010.385,14.098)
(2010.404,13.489)
(2010.423,12.575)
(2010.442,12.777)
(2010.462,12.12)
(2010.481,12.26)
(2010.5,13.215)
(2010.519,12.142)
(2010.538,12.218)
(2010.558,12.142)
(2010.577,12.104)
(2010.596,12.091)
(2010.615,12.11)
(2010.635,12.075)
(2010.654,12.075)
(2010.673,12.018)
(2010.692,11.962)
(2010.712,11.911)
(2010.731,11.931)
(2010.75,11.847)
(2010.769,11.826)
(2010.788,11.761)
(2010.808,11.67)
(2010.827,12.02)
(2010.846,13.443)
(2010.865,13.401)
(2010.885,13.61)
(2010.904,15.81)
(2010.923,15.929)
(2010.942,15.884)
(2010.962,15.627)
(2010.981,15.795)
(2011.0,16.709)
(2011.019,18.189)
(2011.038,18.169)
(2011.058,20.334)
(2011.077,20.833)
(2011.096,20.986)
(2011.115,20.995)
(2011.135,20.883)
(2011.154,20.966)
(2011.173,20.798)
(2011.192,20.863)
(2011.212,20.714)
(2011.231,20.607)
(2011.25,20.443)
(2011.269,20.598)
(2011.288,20.446)
(2011.308,20.736)
(2011.327,20.639)
(2011.346,20.604)
(2011.365,21.134)
(2011.385,21.014)
(2011.404,20.874)
(2011.423,20.949)
(2011.442,21.033)
(2011.462,20.888)
(2011.481,20.947)
(2011.5,20.795)
(2011.519,20.613)
(2011.538,20.89)
(2011.558,20.715)
(2011.577,20.715)
(2011.596,20.513)
(2011.615,20.538)
(2011.635,20.565)
(2011.654,20.676)
(2011.673,20.806)
(2011.692,20.78)
(2011.712,20.861)
(2011.731,20.861)
(2011.75,21.075)
(2011.769,20.995)
(2011.788,21.412)
(2011.808,21.66)
(2011.827,21.936)
(2011.846,21.817)
(2011.865,22.025)
(2011.885,22.273)
(2011.904,21.605)
(2011.923,21.78)
(2011.942,21.859)
(2011.962,21.644)
(2011.981,21.683)
(2012.0,22.053)
(2012.019,21.313)
(2012.038,21.284)
(2012.058,19.505)
(2012.077,19.187)
(2012.096,19.108)
(2012.115,18.953)
(2012.135,32.666)
(2012.154,32.599)
(2012.173,32.574)
(2012.192,32.469)
(2012.212,32.436)
(2012.231,32.341)
(2012.25,32.237)
(2012.269,32.117)
(2012.288,32.077)
(2012.308,31.773)
(2012.327,31.773)
(2012.346,31.879)
(2012.365,31.418)
(2012.385,31.481)
(2012.404,31.391)
(2012.423,31.232)
(2012.442,31.06)
(2012.462,31.114)
(2012.481,31.114)
(2012.5,31.136)
(2012.519,31.045)
(2012.538,31.08)
(2012.558,31.107)
(2012.577,31.141)
(2012.596,31.222)
(2012.615,31.247)
(2012.635,31.257)
(2012.654,31.22)
(2012.673,31.23)
(2012.692,31.23)
(2012.712,31.24)
(2012.731,31.22)
(2012.75,31.273)
(2012.769,31.31)
(2012.788,31.3)
(2012.808,31.3)
(2012.827,31.334)
(2012.846,31.313)
(2012.865,31.361)
(2012.885,31.346)
(2012.904,31.326)
(2012.923,31.314)
(2012.942,31.463)
(2012.962,31.401)
(2012.981,31.393)
(2013.0,30.617)
(2013.019,30.329)
(2013.038,30.341)
(2013.058,30.349)
(2013.077,30.307)
(2013.096,30.167)
(2013.115,30.222)
(2013.135,13.76)
(2013.154,13.821)
(2013.173,14.193)
(2013.192,14.186)
(2013.212,14.17)
(2013.231,14.088)
(2013.25,13.918)
(2013.269,13.924)
(2013.288,13.917)
(2013.308,13.816)
(2013.327,13.77)
(2013.346,13.104)
(2013.365,13.117)
(2013.385,12.509)
(2013.404,12.56)
(2013.423,12.56)
(2013.442,12.56)
(2013.462,12.541)
(2013.481,12.472)
(2013.5,12.532)
(2013.519,12.573)
(2013.538,12.529)
(2013.558,12.574)
(2013.577,12.545)
(2013.596,12.501)
(2013.615,12.478)
(2013.635,12.452)
(2013.654,12.475)
(2013.673,12.423)
(2013.692,12.475)
(2013.712,12.46)
(2013.731,12.489)
(2013.75,12.489)
(2013.769,12.489)
(2013.788,12.489)
(2013.808,12.489)
(2013.827,12.475)
(2013.846,12.52)
(2013.865,12.475)
(2013.885,12.527)
(2013.904,12.488)
(2013.923,12.489)
(2013.942,11.833)
(2013.962,11.867)
(2013.981,11.877)
(2014.0,11.843)
(2014.019,11.822)
(2014.038,11.769)
(2014.058,11.781)
(2014.077,11.756)
(2014.096,11.772)
(2014.115,11.761)
(2014.135,7.72)
(2014.154,7.707)
(2014.173,12.935)
(2014.192,13.324)
(2014.212,13.24)
(2014.231,13.227)
(2014.25,13.299)
(2014.269,13.639)
(2014.288,14.752)
(2014.308,14.752)
(2014.327,18.959)
(2014.346,19.006)
(2014.365,19.019)
(2014.385,21.806)
(2014.404,21.8)
(2014.423,21.816)
(2014.442,22.354)
(2014.462,22.574)
(2014.481,22.628)
(2014.5,22.652)
(2014.519,22.679)
(2014.538,22.78)
(2014.558,22.915)
(2014.577,22.903)
(2014.596,22.894)
(2014.615,22.929)
(2014.635,22.966)
(2014.654,22.903)
(2014.673,22.914)
(2014.692,22.888)
(2014.712,22.888)
(2014.731,22.888)
(2014.75,22.888)
(2014.769,22.875)
(2014.788,22.888)
(2014.808,22.84)
(2014.827,22.866)
(2014.846,22.816)
(2014.865,22.778)
(2014.885,22.714)
(2014.904,22.769)
(2014.923,22.83)
(2014.942,22.856)
(2014.962,22.797)
(2014.981,22.732)
(2015.0,22.683)
(2015.019,22.683)
}

\def\MEANBUDAPEST{
(2006.0,87.731)
(2006.019,86.981)
(2006.038,87.596)
(2006.058,87.269)
(2006.077,87.269)
(2006.096,87.038)
(2006.115,86.731)
(2006.135,87.288)
(2006.154,87.692)
(2006.173,89.385)
(2006.192,92.019)
(2006.212,92.288)
(2006.231,93.404)
(2006.25,94.423)
(2006.269,94.596)
(2006.288,95.827)
(2006.308,100.558)
(2006.327,103.519)
(2006.346,104.865)
(2006.365,108.673)
(2006.385,109.942)
(2006.404,111.731)
(2006.423,113.75)
(2006.442,114.115)
(2006.462,116.519)
(2006.481,116.096)
(2006.5,118.077)
(2006.519,120.942)
(2006.538,120.962)
(2006.558,121.077)
(2006.577,121.25)
(2006.596,121.115)
(2006.615,121.115)
(2006.635,122.135)
(2006.654,122.577)
(2006.673,123.077)
(2006.692,123.25)
(2006.712,123.404)
(2006.731,123.385)
(2006.75,123.481)
(2006.769,123.327)
(2006.788,123.788)
(2006.808,123.654)
(2006.827,123.577)
(2006.846,123.981)
(2006.865,124.077)
(2006.885,124.846)
(2006.904,125.423)
(2006.923,126.269)
(2006.942,126.481)
(2006.962,126.019)
(2006.981,126.038)
(2007.0,124.962)
(2007.019,124.673)
(2007.038,123.827)
(2007.058,124.096)
(2007.077,123.442)
(2007.096,123.519)
(2007.115,122.135)
(2007.135,121.077)
(2007.154,120.5)
(2007.173,120.577)
(2007.192,120.154)
(2007.212,120.231)
(2007.231,120.558)
(2007.25,122.442)
(2007.269,122.923)
(2007.288,122.442)
(2007.308,122.269)
(2007.327,121.423)
(2007.346,120.808)
(2007.365,119.654)
(2007.385,119.173)
(2007.404,121.462)
(2007.423,121.269)
(2007.442,123.615)
(2007.462,122.577)
(2007.481,122.423)
(2007.5,120.558)
(2007.519,118.712)
(2007.538,118.481)
(2007.558,117.692)
(2007.577,116.962)
(2007.596,117.038)
(2007.615,116.596)
(2007.635,115.308)
(2007.654,114.519)
(2007.673,113.808)
(2007.692,113.615)
(2007.712,113.442)
(2007.731,113.346)
(2007.75,113.385)
(2007.769,113.596)
(2007.788,113.192)
(2007.808,113.442)
(2007.827,113.462)
(2007.846,112.962)
(2007.865,113.731)
(2007.885,113.635)
(2007.904,113.481)
(2007.923,115.058)
(2007.942,115.981)
(2007.962,115.558)
(2007.981,117.654)
(2008.0,117.846)
(2008.019,119.173)
(2008.038,119.385)
(2008.058,126.808)
(2008.077,127.077)
(2008.096,128.596)
(2008.115,128.962)
(2008.135,129.135)
(2008.154,128.846)
(2008.173,127.673)
(2008.192,124.846)
(2008.212,123.288)
(2008.231,121.173)
(2008.25,117.865)
(2008.269,117.288)
(2008.288,115.596)
(2008.308,111.212)
(2008.327,109.154)
(2008.346,108.154)
(2008.365,105.365)
(2008.385,103.462)
(2008.404,100.115)
(2008.423,98.442)
(2008.442,95.442)
(2008.462,93.423)
(2008.481,93.712)
(2008.5,92.692)
(2008.519,91.346)
(2008.538,90.327)
(2008.558,89.981)
(2008.577,90.115)
(2008.596,90.442)
(2008.615,90.731)
(2008.635,90.808)
(2008.654,90.788)
(2008.673,90.904)
(2008.692,90.788)
(2008.712,90.731)
(2008.731,90.788)
(2008.75,90.827)
(2008.769,90.673)
(2008.788,90.827)
(2008.808,91.019)
(2008.827,91.154)
(2008.846,91.962)
(2008.865,91.673)
(2008.885,91.462)
(2008.904,91.615)
(2008.923,89.019)
(2008.942,88.481)
(2008.962,89.442)
(2008.981,88.462)
(2009.0,88.173)
(2009.019,85.404)
(2009.038,88.25)
(2009.058,84.808)
(2009.077,84.269)
(2009.096,83.942)
(2009.115,85.308)
(2009.135,85.712)
(2009.154,87.577)
(2009.173,87.154)
(2009.192,88.75)
(2009.212,90.0)
(2009.231,91.865)
(2009.25,92.519)
(2009.269,94.077)
(2009.288,94.635)
(2009.308,96.365)
(2009.327,99.212)
(2009.346,100.462)
(2009.365,100.654)
(2009.385,102.635)
(2009.404,102.385)
(2009.423,103.423)
(2009.442,105.019)
(2009.462,106.404)
(2009.481,106.942)
(2009.5,108.077)
(2009.519,107.981)
(2009.538,109.096)
(2009.558,110.404)
(2009.577,111.038)
(2009.596,110.404)
(2009.615,110.288)
(2009.635,110.346)
(2009.654,110.423)
(2009.673,110.75)
(2009.692,110.981)
(2009.712,111.135)
(2009.731,111.0)
(2009.75,110.808)
(2009.769,110.712)
(2009.788,110.577)
(2009.808,110.327)
(2009.827,110.404)
(2009.846,109.596)
(2009.865,109.058)
(2009.885,108.596)
(2009.904,108.538)
(2009.923,108.885)
(2009.942,107.808)
(2009.962,107.442)
(2009.981,106.365)
(2010.0,106.135)
(2010.019,106.288)
(2010.038,103.635)
(2010.058,100.038)
(2010.077,99.019)
(2010.096,97.212)
(2010.115,94.904)
(2010.135,92.769)
(2010.154,90.077)
(2010.173,90.0)
(2010.192,88.25)
(2010.212,88.0)
(2010.231,86.519)
(2010.25,85.981)
(2010.269,84.135)
(2010.288,85.115)
(2010.308,84.865)
(2010.327,82.596)
(2010.346,82.231)
(2010.365,82.846)
(2010.385,82.673)
(2010.404,82.423)
(2010.423,81.0)
(2010.442,79.846)
(2010.462,79.327)
(2010.481,79.346)
(2010.5,80.192)
(2010.519,80.25)
(2010.538,80.038)
(2010.558,79.827)
(2010.577,79.865)
(2010.596,80.192)
(2010.615,80.346)
(2010.635,80.288)
(2010.654,80.827)
(2010.673,80.519)
(2010.692,80.635)
(2010.712,80.577)
(2010.731,80.635)
(2010.75,81.0)
(2010.769,81.385)
(2010.788,81.731)
(2010.808,82.577)
(2010.827,83.077)
(2010.846,84.423)
(2010.865,84.865)
(2010.885,87.442)
(2010.904,90.269)
(2010.923,91.923)
(2010.942,94.096)
(2010.962,96.019)
(2010.981,96.769)
(2011.0,99.0)
(2011.019,100.423)
(2011.038,98.365)
(2011.058,102.596)
(2011.077,104.192)
(2011.096,106.615)
(2011.115,108.885)
(2011.135,110.462)
(2011.154,113.192)
(2011.173,115.192)
(2011.192,116.635)
(2011.212,117.885)
(2011.231,118.096)
(2011.25,120.519)
(2011.269,121.231)
(2011.288,121.385)
(2011.308,121.038)
(2011.327,120.865)
(2011.346,120.846)
(2011.365,120.577)
(2011.385,119.904)
(2011.404,118.808)
(2011.423,119.327)
(2011.442,119.231)
(2011.462,118.558)
(2011.481,117.615)
(2011.5,117.346)
(2011.519,117.442)
(2011.538,117.404)
(2011.558,117.346)
(2011.577,117.442)
(2011.596,117.135)
(2011.615,116.981)
(2011.635,116.981)
(2011.654,116.635)
(2011.673,116.692)
(2011.692,116.519)
(2011.712,116.788)
(2011.731,117.5)
(2011.75,117.135)
(2011.769,116.962)
(2011.788,116.442)
(2011.808,115.212)
(2011.827,115.269)
(2011.846,115.423)
(2011.865,114.615)
(2011.885,113.308)
(2011.904,110.192)
(2011.923,109.173)
(2011.942,107.788)
(2011.962,106.25)
(2011.981,105.596)
(2012.0,105.712)
(2012.019,103.923)
(2012.038,104.827)
(2012.058,104.019)
(2012.077,104.788)
(2012.096,104.577)
(2012.115,104.058)
(2012.135,105.288)
(2012.154,105.288)
(2012.173,103.731)
(2012.192,103.385)
(2012.212,102.769)
(2012.231,102.154)
(2012.25,102.269)
(2012.269,102.038)
(2012.288,102.308)
(2012.308,100.135)
(2012.327,100.942)
(2012.346,101.192)
(2012.365,99.635)
(2012.385,101.038)
(2012.404,104.077)
(2012.423,103.673)
(2012.442,102.654)
(2012.462,104.269)
(2012.481,104.269)
(2012.5,104.481)
(2012.519,104.769)
(2012.538,104.692)
(2012.558,104.615)
(2012.577,104.442)
(2012.596,105.327)
(2012.615,105.135)
(2012.635,106.154)
(2012.654,106.173)
(2012.673,105.962)
(2012.692,105.712)
(2012.712,105.538)
(2012.731,104.904)
(2012.75,104.904)
(2012.769,104.712)
(2012.788,104.827)
(2012.808,104.885)
(2012.827,104.327)
(2012.846,103.019)
(2012.865,102.577)
(2012.885,102.0)
(2012.904,102.635)
(2012.923,101.288)
(2012.942,101.692)
(2012.962,101.577)
(2012.981,101.596)
(2013.0,102.154)
(2013.019,101.654)
(2013.038,100.904)
(2013.058,100.75)
(2013.077,99.788)
(2013.096,98.481)
(2013.115,99.558)
(2013.135,98.788)
(2013.154,97.192)
(2013.173,97.923)
(2013.192,97.519)
(2013.212,97.308)
(2013.231,97.327)
(2013.25,97.231)
(2013.269,96.923)
(2013.288,96.058)
(2013.308,99.269)
(2013.327,98.462)
(2013.346,98.923)
(2013.365,99.731)
(2013.385,98.865)
(2013.404,95.75)
(2013.423,96.865)
(2013.442,98.827)
(2013.462,97.5)
(2013.481,97.615)
(2013.5,97.673)
(2013.519,98.846)
(2013.538,99.212)
(2013.558,100.827)
(2013.577,101.75)
(2013.596,102.25)
(2013.615,103.019)
(2013.635,102.519)
(2013.654,102.423)
(2013.673,102.788)
(2013.692,103.423)
(2013.712,103.346)
(2013.731,103.442)
(2013.75,104.288)
(2013.769,104.269)
(2013.788,104.385)
(2013.808,104.25)
(2013.827,104.058)
(2013.846,103.962)
(2013.865,104.212)
(2013.885,104.327)
(2013.904,104.077)
(2013.923,104.731)
(2013.942,103.077)
(2013.962,103.0)
(2013.981,102.442)
(2014.0,99.885)
(2014.019,99.962)
(2014.038,99.173)
(2014.058,96.615)
(2014.077,95.558)
(2014.096,94.654)
(2014.115,91.865)
(2014.135,90.558)
(2014.154,89.635)
(2014.173,87.231)
(2014.192,87.346)
(2014.212,86.596)
(2014.231,86.827)
(2014.25,86.231)
(2014.269,85.962)
(2014.288,87.462)
(2014.308,84.519)
(2014.327,84.846)
(2014.346,83.192)
(2014.365,81.096)
(2014.385,79.577)
(2014.404,84.712)
(2014.423,84.538)
(2014.442,83.173)
(2014.462,85.192)
(2014.481,84.462)
(2014.5,85.096)
(2014.519,84.923)
(2014.538,85.019)
(2014.558,83.5)
(2014.577,83.019)
(2014.596,82.115)
(2014.615,82.0)
(2014.635,81.981)
(2014.654,81.923)
(2014.673,81.615)
(2014.692,81.115)
(2014.712,81.038)
(2014.731,80.962)
(2014.75,80.25)
(2014.769,80.423)
(2014.788,80.462)
(2014.808,80.673)
(2014.827,80.673)
(2014.846,80.75)
(2014.865,80.885)
(2014.885,80.212)
(2014.904,79.615)
(2014.923,80.192)
(2014.942,80.154)
(2014.962,80.038)
(2014.981,79.654)
(2015.0,79.192)
(2015.019,79.654)
}

\def\MEANFEJER{
(2006.0,47.462)
(2006.019,45.538)
(2006.038,45.077)
(2006.058,43.808)
(2006.077,43.096)
(2006.096,41.827)
(2006.115,40.692)
(2006.135,39.192)
(2006.154,37.865)
(2006.173,36.442)
(2006.192,35.385)
(2006.212,34.808)
(2006.231,33.442)
(2006.25,33.577)
(2006.269,33.154)
(2006.288,33.269)
(2006.308,32.231)
(2006.327,32.731)
(2006.346,32.096)
(2006.365,31.481)
(2006.385,30.442)
(2006.404,30.096)
(2006.423,29.154)
(2006.442,28.327)
(2006.462,28.154)
(2006.481,27.288)
(2006.5,27.442)
(2006.519,27.231)
(2006.538,26.923)
(2006.558,26.788)
(2006.577,26.577)
(2006.596,26.404)
(2006.615,26.462)
(2006.635,26.442)
(2006.654,26.404)
(2006.673,26.423)
(2006.692,26.442)
(2006.712,26.404)
(2006.731,26.519)
(2006.75,26.481)
(2006.769,26.404)
(2006.788,26.096)
(2006.808,25.75)
(2006.827,25.75)
(2006.846,25.558)
(2006.865,26.019)
(2006.885,26.135)
(2006.904,26.808)
(2006.923,27.442)
(2006.942,28.962)
(2006.962,29.904)
(2006.981,30.346)
(2007.0,30.827)
(2007.019,32.0)
(2007.038,32.75)
(2007.058,32.731)
(2007.077,34.673)
(2007.096,35.154)
(2007.115,35.558)
(2007.135,35.827)
(2007.154,36.75)
(2007.173,36.846)
(2007.192,37.904)
(2007.212,37.692)
(2007.231,38.538)
(2007.25,39.058)
(2007.269,39.25)
(2007.288,39.115)
(2007.308,40.058)
(2007.327,39.462)
(2007.346,39.346)
(2007.365,39.385)
(2007.385,39.519)
(2007.404,39.462)
(2007.423,39.654)
(2007.442,40.25)
(2007.462,39.596)
(2007.481,39.231)
(2007.5,39.038)
(2007.519,38.808)
(2007.538,38.769)
(2007.558,38.75)
(2007.577,38.769)
(2007.596,38.769)
(2007.615,38.692)
(2007.635,38.712)
(2007.654,38.673)
(2007.673,38.673)
(2007.692,38.692)
(2007.712,38.75)
(2007.731,38.731)
(2007.75,38.865)
(2007.769,38.981)
(2007.788,39.288)
(2007.808,39.404)
(2007.827,39.038)
(2007.846,38.731)
(2007.865,38.519)
(2007.885,38.538)
(2007.904,37.904)
(2007.923,37.577)
(2007.942,36.519)
(2007.962,36.365)
(2007.981,36.365)
(2008.0,36.673)
(2008.019,35.615)
(2008.038,36.731)
(2008.058,37.692)
(2008.077,36.173)
(2008.096,36.192)
(2008.115,36.327)
(2008.135,37.038)
(2008.154,37.269)
(2008.173,38.346)
(2008.192,37.962)
(2008.212,38.981)
(2008.231,38.904)
(2008.25,39.269)
(2008.269,40.5)
(2008.288,41.75)
(2008.308,42.192)
(2008.327,43.962)
(2008.346,44.519)
(2008.365,45.75)
(2008.385,45.923)
(2008.404,46.596)
(2008.423,48.481)
(2008.442,48.519)
(2008.462,49.769)
(2008.481,50.346)
(2008.5,50.654)
(2008.519,51.038)
(2008.538,51.231)
(2008.558,51.288)
(2008.577,51.288)
(2008.596,51.558)
(2008.615,51.635)
(2008.635,51.692)
(2008.654,51.712)
(2008.673,51.673)
(2008.692,51.654)
(2008.712,51.615)
(2008.731,51.538)
(2008.75,51.404)
(2008.769,51.192)
(2008.788,50.962)
(2008.808,51.0)
(2008.827,51.0)
(2008.846,51.077)
(2008.865,50.673)
(2008.885,50.269)
(2008.904,50.096)
(2008.923,49.365)
(2008.942,48.654)
(2008.962,47.615)
(2008.981,47.115)
(2009.0,46.019)
(2009.019,45.25)
(2009.038,43.212)
(2009.058,42.096)
(2009.077,41.635)
(2009.096,40.673)
(2009.115,40.115)
(2009.135,38.923)
(2009.154,37.115)
(2009.173,35.75)
(2009.192,34.404)
(2009.212,33.038)
(2009.231,31.731)
(2009.25,30.385)
(2009.269,28.481)
(2009.288,27.173)
(2009.308,25.942)
(2009.327,23.962)
(2009.346,23.442)
(2009.365,21.942)
(2009.385,21.558)
(2009.404,20.615)
(2009.423,18.962)
(2009.442,18.846)
(2009.462,17.885)
(2009.481,17.462)
(2009.5,17.288)
(2009.519,16.942)
(2009.538,16.846)
(2009.558,16.923)
(2009.577,16.923)
(2009.596,16.904)
(2009.615,16.827)
(2009.635,16.827)
(2009.654,16.808)
(2009.673,16.827)
(2009.692,16.808)
(2009.712,16.846)
(2009.731,16.885)
(2009.75,16.962)
(2009.769,17.019)
(2009.788,17.135)
(2009.808,17.038)
(2009.827,17.115)
(2009.846,17.154)
(2009.865,17.5)
(2009.885,17.75)
(2009.904,17.885)
(2009.923,18.654)
(2009.942,18.846)
(2009.962,19.25)
(2009.981,19.385)
(2010.0,19.635)
(2010.019,20.25)
(2010.038,21.5)
(2010.058,21.788)
(2010.077,22.308)
(2010.096,23.135)
(2010.115,23.75)
(2010.135,24.192)
(2010.154,24.635)
(2010.173,24.692)
(2010.192,25.096)
(2010.212,26.019)
(2010.231,26.692)
(2010.25,27.423)
(2010.269,28.038)
(2010.288,27.654)
(2010.308,28.885)
(2010.327,29.154)
(2010.346,29.788)
(2010.365,30.423)
(2010.385,30.827)
(2010.404,31.135)
(2010.423,30.846)
(2010.442,31.654)
(2010.462,32.0)
(2010.481,33.077)
(2010.5,33.769)
(2010.519,34.231)
(2010.538,34.577)
(2010.558,35.288)
(2010.577,35.481)
(2010.596,35.288)
(2010.615,35.346)
(2010.635,35.442)
(2010.654,35.519)
(2010.673,35.519)
(2010.692,35.538)
(2010.712,35.5)
(2010.731,35.5)
(2010.75,35.423)
(2010.769,35.346)
(2010.788,35.192)
(2010.808,35.019)
(2010.827,34.923)
(2010.846,34.808)
(2010.865,34.5)
(2010.885,34.269)
(2010.904,34.308)
(2010.923,33.654)
(2010.942,33.75)
(2010.962,33.135)
(2010.981,32.923)
(2011.0,33.096)
(2011.019,33.115)
(2011.038,32.096)
(2011.058,31.962)
(2011.077,31.75)
(2011.096,31.25)
(2011.115,30.75)
(2011.135,31.327)
(2011.154,31.519)
(2011.173,31.846)
(2011.192,32.712)
(2011.212,32.635)
(2011.231,32.827)
(2011.25,32.846)
(2011.269,33.481)
(2011.288,34.115)
(2011.308,34.365)
(2011.327,33.962)
(2011.346,33.885)
(2011.365,34.154)
(2011.385,34.192)
(2011.404,34.308)
(2011.423,34.481)
(2011.442,33.981)
(2011.462,34.173)
(2011.481,33.5)
(2011.5,33.135)
(2011.519,32.865)
(2011.538,32.981)
(2011.558,32.442)
(2011.577,32.481)
(2011.596,32.538)
(2011.615,32.558)
(2011.635,32.462)
(2011.654,32.385)
(2011.673,32.404)
(2011.692,32.365)
(2011.712,32.346)
(2011.731,32.288)
(2011.75,32.269)
(2011.769,32.404)
(2011.788,32.558)
(2011.808,32.712)
(2011.827,32.942)
(2011.846,33.173)
(2011.865,33.077)
(2011.885,33.288)
(2011.904,33.288)
(2011.923,33.385)
(2011.942,33.154)
(2011.962,33.269)
(2011.981,33.846)
(2012.0,33.769)
(2012.019,34.404)
(2012.038,34.519)
(2012.058,35.962)
(2012.077,35.808)
(2012.096,36.269)
(2012.115,37.096)
(2012.135,36.942)
(2012.154,37.923)
(2012.173,38.115)
(2012.192,38.788)
(2012.212,38.846)
(2012.231,38.981)
(2012.25,39.808)
(2012.269,39.558)
(2012.288,39.5)
(2012.308,38.096)
(2012.327,38.577)
(2012.346,39.115)
(2012.365,38.173)
(2012.385,39.058)
(2012.404,38.981)
(2012.423,40.462)
(2012.442,40.212)
(2012.462,40.365)
(2012.481,40.558)
(2012.5,40.923)
(2012.519,40.942)
(2012.538,41.442)
(2012.558,41.346)
(2012.577,41.404)
(2012.596,41.327)
(2012.615,41.346)
(2012.635,41.308)
(2012.654,41.365)
(2012.673,41.346)
(2012.692,41.346)
(2012.712,41.423)
(2012.731,41.423)
(2012.75,41.462)
(2012.769,41.327)
(2012.788,41.481)
(2012.808,41.404)
(2012.827,41.654)
(2012.846,41.577)
(2012.865,41.596)
(2012.885,42.019)
(2012.904,41.923)
(2012.923,41.904)
(2012.942,42.154)
(2012.962,42.192)
(2012.981,42.288)
(2013.0,42.269)
(2013.019,41.25)
(2013.038,40.846)
(2013.058,40.673)
(2013.077,40.692)
(2013.096,40.5)
(2013.115,39.846)
(2013.135,39.288)
(2013.154,37.981)
(2013.173,37.481)
(2013.192,35.827)
(2013.212,35.077)
(2013.231,34.327)
(2013.25,32.538)
(2013.269,31.135)
(2013.288,30.115)
(2013.308,29.769)
(2013.327,28.577)
(2013.346,27.692)
(2013.365,27.519)
(2013.385,26.096)
(2013.404,25.269)
(2013.423,23.75)
(2013.442,23.192)
(2013.462,22.346)
(2013.481,21.942)
(2013.5,20.712)
(2013.519,20.885)
(2013.538,20.038)
(2013.558,19.942)
(2013.577,19.712)
(2013.596,19.692)
(2013.615,19.654)
(2013.635,19.692)
(2013.654,19.635)
(2013.673,19.654)
(2013.692,19.654)
(2013.712,19.635)
(2013.731,19.635)
(2013.75,19.596)
(2013.769,19.808)
(2013.788,19.481)
(2013.808,19.615)
(2013.827,19.096)
(2013.846,19.346)
(2013.865,19.442)
(2013.885,18.981)
(2013.904,19.673)
(2013.923,19.615)
(2013.942,19.75)
(2013.962,19.615)
(2013.981,19.481)
(2014.0,19.827)
(2014.019,19.558)
(2014.038,19.981)
(2014.058,19.115)
(2014.077,19.865)
(2014.096,20.019)
(2014.115,19.923)
(2014.135,19.942)
(2014.154,19.827)
(2014.173,20.462)
(2014.192,20.558)
(2014.212,21.654)
(2014.231,21.577)
(2014.25,21.788)
(2014.269,22.635)
(2014.288,23.308)
(2014.308,23.654)
(2014.327,24.231)
(2014.346,23.615)
(2014.365,22.962)
(2014.385,24.077)
(2014.404,24.423)
(2014.423,24.077)
(2014.442,24.904)
(2014.462,24.904)
(2014.481,24.442)
(2014.5,25.442)
(2014.519,25.096)
(2014.538,25.058)
(2014.558,25.154)
(2014.577,25.25)
(2014.596,25.25)
(2014.615,25.25)
(2014.635,25.231)
(2014.654,25.288)
(2014.673,25.365)
(2014.692,25.385)
(2014.712,25.385)
(2014.731,25.442)
(2014.75,25.462)
(2014.769,25.308)
(2014.788,25.423)
(2014.808,25.538)
(2014.827,25.865)
(2014.846,25.808)
(2014.865,25.904)
(2014.885,25.981)
(2014.904,25.442)
(2014.923,25.731)
(2014.942,25.288)
(2014.962,25.462)
(2014.981,24.769)
(2015.0,24.192)
(2015.019,24.192)
}

\def\MEANSZABOLCS{
(2006.0,38.192)
(2006.019,37.635)
(2006.038,37.654)
(2006.058,37.385)
(2006.077,36.885)
(2006.096,36.596)
(2006.115,36.327)
(2006.135,35.731)
(2006.154,34.692)
(2006.173,34.462)
(2006.192,34.442)
(2006.212,34.75)
(2006.231,34.481)
(2006.25,30.731)
(2006.269,30.519)
(2006.288,31.288)
(2006.308,31.712)
(2006.327,32.75)
(2006.346,33.481)
(2006.365,33.981)
(2006.385,34.692)
(2006.404,33.827)
(2006.423,34.673)
(2006.442,33.635)
(2006.462,34.231)
(2006.481,33.615)
(2006.5,33.519)
(2006.519,32.577)
(2006.538,32.231)
(2006.558,32.404)
(2006.577,32.288)
(2006.596,32.135)
(2006.615,32.038)
(2006.635,32.096)
(2006.654,32.096)
(2006.673,32.058)
(2006.692,32.077)
(2006.712,32.115)
(2006.731,32.115)
(2006.75,32.192)
(2006.769,32.269)
(2006.788,32.154)
(2006.808,32.346)
(2006.827,32.481)
(2006.846,32.615)
(2006.865,32.788)
(2006.885,32.596)
(2006.904,32.365)
(2006.923,32.788)
(2006.942,32.654)
(2006.962,32.231)
(2006.981,31.981)
(2007.0,31.462)
(2007.019,30.808)
(2007.038,31.25)
(2007.058,31.615)
(2007.077,32.038)
(2007.096,32.423)
(2007.115,34.058)
(2007.135,33.654)
(2007.154,34.346)
(2007.173,33.827)
(2007.192,33.596)
(2007.212,33.365)
(2007.231,34.192)
(2007.25,34.75)
(2007.269,33.692)
(2007.288,32.712)
(2007.308,32.885)
(2007.327,31.885)
(2007.346,31.942)
(2007.365,31.596)
(2007.385,31.577)
(2007.404,31.404)
(2007.423,31.654)
(2007.442,33.25)
(2007.462,32.654)
(2007.481,32.404)
(2007.5,31.808)
(2007.519,31.288)
(2007.538,31.231)
(2007.558,30.923)
(2007.577,30.962)
(2007.596,30.981)
(2007.615,31.115)
(2007.635,30.942)
(2007.654,30.942)
(2007.673,30.923)
(2007.692,30.885)
(2007.712,30.788)
(2007.731,30.692)
(2007.75,30.558)
(2007.769,30.519)
(2007.788,31.462)
(2007.808,31.327)
(2007.827,31.154)
(2007.846,31.731)
(2007.865,31.519)
(2007.885,31.923)
(2007.904,33.019)
(2007.923,33.038)
(2007.942,32.962)
(2007.962,34.135)
(2007.981,34.577)
(2008.0,37.827)
(2008.019,39.135)
(2008.038,40.231)
(2008.058,43.404)
(2008.077,44.096)
(2008.096,44.288)
(2008.115,43.673)
(2008.135,44.154)
(2008.154,44.404)
(2008.173,45.096)
(2008.192,45.192)
(2008.212,46.019)
(2008.231,46.096)
(2008.25,46.327)
(2008.269,46.885)
(2008.288,47.577)
(2008.308,47.192)
(2008.327,47.25)
(2008.346,46.173)
(2008.365,47.135)
(2008.385,46.577)
(2008.404,46.269)
(2008.423,45.769)
(2008.442,44.808)
(2008.462,44.827)
(2008.481,44.865)
(2008.5,44.654)
(2008.519,44.538)
(2008.538,44.538)
(2008.558,45.154)
(2008.577,45.077)
(2008.596,45.019)
(2008.615,45.192)
(2008.635,45.538)
(2008.654,45.538)
(2008.673,45.519)
(2008.692,45.519)
(2008.712,45.538)
(2008.731,45.538)
(2008.75,45.5)
(2008.769,45.423)
(2008.788,44.5)
(2008.808,44.231)
(2008.827,44.231)
(2008.846,43.558)
(2008.865,43.115)
(2008.885,42.538)
(2008.904,41.596)
(2008.923,40.904)
(2008.942,41.019)
(2008.962,40.327)
(2008.981,41.192)
(2009.0,38.154)
(2009.019,37.75)
(2009.038,35.962)
(2009.058,33.481)
(2009.077,32.154)
(2009.096,32.212)
(2009.115,30.923)
(2009.135,31.462)
(2009.154,30.865)
(2009.173,31.154)
(2009.192,30.788)
(2009.212,32.577)
(2009.231,31.981)
(2009.25,32.231)
(2009.269,32.615)
(2009.288,32.192)
(2009.308,32.827)
(2009.327,32.577)
(2009.346,33.481)
(2009.365,31.788)
(2009.385,31.212)
(2009.404,31.596)
(2009.423,31.538)
(2009.442,31.058)
(2009.462,31.365)
(2009.481,31.096)
(2009.5,31.0)
(2009.519,31.231)
(2009.538,31.519)
(2009.558,30.981)
(2009.577,30.904)
(2009.596,30.904)
(2009.615,30.596)
(2009.635,30.25)
(2009.654,30.25)
(2009.673,30.288)
(2009.692,30.269)
(2009.712,30.327)
(2009.731,30.346)
(2009.75,30.481)
(2009.769,30.558)
(2009.788,30.5)
(2009.808,30.538)
(2009.827,30.635)
(2009.846,30.558)
(2009.865,30.558)
(2009.885,30.481)
(2009.904,30.231)
(2009.923,30.192)
(2009.942,29.942)
(2009.962,29.519)
(2009.981,28.5)
(2010.0,28.327)
(2010.019,27.712)
(2010.038,27.404)
(2010.058,26.25)
(2010.077,26.25)
(2010.096,25.096)
(2010.115,25.019)
(2010.135,23.423)
(2010.154,23.481)
(2010.173,22.635)
(2010.192,23.5)
(2010.212,22.096)
(2010.231,21.231)
(2010.25,20.692)
(2010.269,19.827)
(2010.288,19.404)
(2010.308,18.365)
(2010.327,18.942)
(2010.346,18.192)
(2010.365,18.288)
(2010.385,18.269)
(2010.404,17.5)
(2010.423,16.654)
(2010.442,16.077)
(2010.462,15.538)
(2010.481,15.404)
(2010.5,16.423)
(2010.519,16.269)
(2010.538,16.481)
(2010.558,16.577)
(2010.577,16.635)
(2010.596,16.788)
(2010.615,16.769)
(2010.635,16.788)
(2010.654,16.769)
(2010.673,16.75)
(2010.692,16.769)
(2010.712,16.692)
(2010.731,16.731)
(2010.75,16.615)
(2010.769,16.519)
(2010.788,16.462)
(2010.808,16.596)
(2010.827,16.596)
(2010.846,16.596)
(2010.865,16.5)
(2010.885,16.596)
(2010.904,16.673)
(2010.923,16.538)
(2010.942,16.846)
(2010.962,16.769)
(2010.981,16.923)
(2011.0,17.212)
(2011.019,17.712)
(2011.038,18.019)
(2011.058,18.019)
(2011.077,19.058)
(2011.096,19.788)
(2011.115,20.135)
(2011.135,21.269)
(2011.154,21.192)
(2011.173,22.308)
(2011.192,21.75)
(2011.212,22.442)
(2011.231,23.481)
(2011.25,26.365)
(2011.269,27.654)
(2011.288,28.5)
(2011.308,29.442)
(2011.327,29.0)
(2011.346,29.212)
(2011.365,31.808)
(2011.385,34.038)
(2011.404,35.904)
(2011.423,37.231)
(2011.442,38.115)
(2011.462,39.058)
(2011.481,39.808)
(2011.5,39.404)
(2011.519,39.615)
(2011.538,40.135)
(2011.558,40.154)
(2011.577,40.058)
(2011.596,39.981)
(2011.615,40.038)
(2011.635,40.077)
(2011.654,40.115)
(2011.673,40.269)
(2011.692,40.269)
(2011.712,40.269)
(2011.731,40.25)
(2011.75,40.327)
(2011.769,40.462)
(2011.788,40.481)
(2011.808,40.423)
(2011.827,40.615)
(2011.846,40.827)
(2011.865,40.827)
(2011.885,41.019)
(2011.904,41.404)
(2011.923,41.404)
(2011.942,41.269)
(2011.962,42.577)
(2011.981,42.827)
(2012.0,43.173)
(2012.019,42.673)
(2012.038,42.577)
(2012.058,42.558)
(2012.077,41.654)
(2012.096,41.058)
(2012.115,40.577)
(2012.135,39.596)
(2012.154,39.154)
(2012.173,37.635)
(2012.192,36.769)
(2012.212,34.673)
(2012.231,33.596)
(2012.25,30.654)
(2012.269,29.692)
(2012.288,28.558)
(2012.308,27.385)
(2012.327,27.904)
(2012.346,28.135)
(2012.365,25.058)
(2012.385,24.019)
(2012.404,22.885)
(2012.423,21.942)
(2012.442,21.596)
(2012.462,21.519)
(2012.481,20.885)
(2012.5,20.365)
(2012.519,21.077)
(2012.538,19.865)
(2012.558,20.269)
(2012.577,20.442)
(2012.596,20.519)
(2012.615,20.615)
(2012.635,20.615)
(2012.654,20.615)
(2012.673,20.462)
(2012.692,20.558)
(2012.712,20.577)
(2012.731,20.538)
(2012.75,20.442)
(2012.769,20.346)
(2012.788,20.308)
(2012.808,20.635)
(2012.827,20.346)
(2012.846,20.308)
(2012.865,20.423)
(2012.885,20.788)
(2012.904,20.712)
(2012.923,20.712)
(2012.942,20.538)
(2012.962,19.385)
(2012.981,18.827)
(2013.0,18.423)
(2013.019,18.173)
(2013.038,19.192)
(2013.058,19.808)
(2013.077,19.673)
(2013.096,19.404)
(2013.115,19.25)
(2013.135,19.096)
(2013.154,19.115)
(2013.173,19.231)
(2013.192,19.75)
(2013.212,19.827)
(2013.231,20.885)
(2013.25,20.423)
(2013.269,21.692)
(2013.288,22.173)
(2013.308,22.269)
(2013.327,22.827)
(2013.346,22.519)
(2013.365,23.288)
(2013.385,23.038)
(2013.404,23.615)
(2013.423,23.904)
(2013.442,24.192)
(2013.462,23.788)
(2013.481,24.942)
(2013.5,25.077)
(2013.519,25.846)
(2013.538,26.615)
(2013.558,27.423)
(2013.577,27.442)
(2013.596,27.731)
(2013.615,27.731)
(2013.635,27.673)
(2013.654,27.635)
(2013.673,27.731)
(2013.692,27.654)
(2013.712,27.673)
(2013.731,27.654)
(2013.75,27.635)
(2013.769,27.673)
(2013.788,27.885)
(2013.808,27.385)
(2013.827,27.596)
(2013.846,27.712)
(2013.865,28.115)
(2013.885,27.596)
(2013.904,27.308)
(2013.923,27.5)
(2013.942,28.192)
(2013.962,28.019)
(2013.981,28.135)
(2014.0,28.212)
(2014.019,28.212)
(2014.038,27.462)
(2014.058,26.654)
(2014.077,26.577)
(2014.096,26.5)
(2014.115,27.962)
(2014.135,27.942)
(2014.154,28.288)
(2014.173,28.038)
(2014.192,27.942)
(2014.212,27.577)
(2014.231,27.077)
(2014.25,27.788)
(2014.269,26.423)
(2014.288,26.462)
(2014.308,26.885)
(2014.327,25.519)
(2014.346,25.212)
(2014.365,24.462)
(2014.385,23.462)
(2014.404,23.173)
(2014.423,23.327)
(2014.442,23.558)
(2014.462,23.25)
(2014.481,22.327)
(2014.5,22.115)
(2014.519,20.904)
(2014.538,20.269)
(2014.558,19.096)
(2014.577,19.115)
(2014.596,18.731)
(2014.615,18.808)
(2014.635,19.038)
(2014.654,19.058)
(2014.673,18.962)
(2014.692,18.962)
(2014.712,18.923)
(2014.731,19.019)
(2014.75,19.019)
(2014.769,18.962)
(2014.788,18.865)
(2014.808,18.904)
(2014.827,18.673)
(2014.846,18.558)
(2014.865,18.115)
(2014.885,18.5)
(2014.904,18.654)
(2014.923,18.558)
(2014.942,17.673)
(2014.962,17.846)
(2014.981,17.596)
(2015.0,17.635)
(2015.019,17.904)
}

\def\MEANZALA{
(2006.0,30.596)
(2006.019,31.346)
(2006.038,32.135)
(2006.058,32.115)
(2006.077,32.673)
(2006.096,32.385)
(2006.115,32.154)
(2006.135,32.442)
(2006.154,32.558)
(2006.173,33.212)
(2006.192,32.885)
(2006.212,33.173)
(2006.231,33.231)
(2006.25,34.096)
(2006.269,34.058)
(2006.288,34.288)
(2006.308,34.731)
(2006.327,34.712)
(2006.346,35.519)
(2006.365,35.808)
(2006.385,36.692)
(2006.404,36.442)
(2006.423,36.769)
(2006.442,36.769)
(2006.462,36.635)
(2006.481,35.885)
(2006.5,35.769)
(2006.519,35.615)
(2006.538,35.731)
(2006.558,35.769)
(2006.577,35.712)
(2006.596,35.692)
(2006.615,35.615)
(2006.635,35.596)
(2006.654,35.558)
(2006.673,35.558)
(2006.692,35.558)
(2006.712,35.596)
(2006.731,35.558)
(2006.75,35.308)
(2006.769,35.25)
(2006.788,35.115)
(2006.808,35.038)
(2006.827,34.962)
(2006.846,34.731)
(2006.865,34.308)
(2006.885,33.904)
(2006.904,33.365)
(2006.923,32.327)
(2006.942,32.212)
(2006.962,31.981)
(2006.981,31.827)
(2007.0,31.692)
(2007.019,30.442)
(2007.038,30.365)
(2007.058,30.231)
(2007.077,30.365)
(2007.096,30.538)
(2007.115,30.962)
(2007.135,30.0)
(2007.154,29.712)
(2007.173,28.808)
(2007.192,28.904)
(2007.212,28.25)
(2007.231,28.25)
(2007.25,27.923)
(2007.269,27.75)
(2007.288,27.75)
(2007.308,27.731)
(2007.327,28.288)
(2007.346,27.577)
(2007.365,27.25)
(2007.385,27.0)
(2007.404,27.365)
(2007.423,27.0)
(2007.442,26.904)
(2007.462,26.442)
(2007.481,26.058)
(2007.5,26.058)
(2007.519,25.731)
(2007.538,25.346)
(2007.558,25.173)
(2007.577,25.058)
(2007.596,25.077)
(2007.615,25.077)
(2007.635,25.077)
(2007.654,25.135)
(2007.673,25.096)
(2007.692,25.096)
(2007.712,25.077)
(2007.731,25.077)
(2007.75,25.019)
(2007.769,24.827)
(2007.788,24.75)
(2007.808,24.827)
(2007.827,24.673)
(2007.846,24.692)
(2007.865,24.75)
(2007.885,24.885)
(2007.904,24.769)
(2007.923,24.808)
(2007.942,24.538)
(2007.962,24.154)
(2007.981,23.75)
(2008.0,23.538)
(2008.019,22.962)
(2008.038,22.0)
(2008.058,21.712)
(2008.077,20.635)
(2008.096,19.846)
(2008.115,18.519)
(2008.135,17.942)
(2008.154,17.173)
(2008.173,17.038)
(2008.192,16.442)
(2008.212,16.327)
(2008.231,15.923)
(2008.25,14.885)
(2008.269,14.769)
(2008.288,14.038)
(2008.308,13.615)
(2008.327,12.904)
(2008.346,12.212)
(2008.365,11.558)
(2008.385,10.885)
(2008.404,10.212)
(2008.423,9.731)
(2008.442,9.346)
(2008.462,9.173)
(2008.481,9.0)
(2008.5,8.962)
(2008.519,8.885)
(2008.538,8.904)
(2008.558,9.154)
(2008.577,9.173)
(2008.596,9.192)
(2008.615,9.25)
(2008.635,9.269)
(2008.654,9.212)
(2008.673,9.269)
(2008.692,9.308)
(2008.712,9.385)
(2008.731,9.385)
(2008.75,9.558)
(2008.769,9.654)
(2008.788,9.538)
(2008.808,9.481)
(2008.827,9.885)
(2008.846,9.827)
(2008.865,10.0)
(2008.885,9.654)
(2008.904,9.885)
(2008.923,9.731)
(2008.942,9.462)
(2008.962,10.115)
(2008.981,10.231)
(2009.0,10.231)
(2009.019,10.192)
(2009.038,10.404)
(2009.058,10.192)
(2009.077,10.135)
(2009.096,10.327)
(2009.115,10.596)
(2009.135,10.75)
(2009.154,10.769)
(2009.173,11.404)
(2009.192,11.615)
(2009.212,11.596)
(2009.231,11.346)
(2009.25,11.25)
(2009.269,11.827)
(2009.288,12.019)
(2009.308,11.923)
(2009.327,12.135)
(2009.346,13.692)
(2009.365,13.654)
(2009.385,13.327)
(2009.404,13.885)
(2009.423,14.75)
(2009.442,14.615)
(2009.462,15.288)
(2009.481,15.5)
(2009.5,15.385)
(2009.519,16.269)
(2009.538,16.596)
(2009.558,16.365)
(2009.577,16.75)
(2009.596,16.769)
(2009.615,16.865)
(2009.635,16.865)
(2009.654,16.942)
(2009.673,16.923)
(2009.692,16.885)
(2009.712,16.808)
(2009.731,16.808)
(2009.75,16.673)
(2009.769,16.577)
(2009.788,16.577)
(2009.808,16.538)
(2009.827,16.346)
(2009.846,16.327)
(2009.865,16.346)
(2009.885,16.558)
(2009.904,16.615)
(2009.923,16.865)
(2009.942,17.308)
(2009.962,17.423)
(2009.981,17.558)
(2010.0,17.462)
(2010.019,17.481)
(2010.038,17.519)
(2010.058,17.558)
(2010.077,17.558)
(2010.096,17.404)
(2010.115,17.154)
(2010.135,17.038)
(2010.154,17.019)
(2010.173,16.0)
(2010.192,15.712)
(2010.212,15.481)
(2010.231,15.462)
(2010.25,15.404)
(2010.269,14.731)
(2010.288,14.577)
(2010.308,14.731)
(2010.327,14.5)
(2010.346,13.212)
(2010.365,13.423)
(2010.385,13.654)
(2010.404,13.038)
(2010.423,12.615)
(2010.442,13.115)
(2010.462,12.558)
(2010.481,12.808)
(2010.5,13.692)
(2010.519,12.923)
(2010.538,13.038)
(2010.558,13.135)
(2010.577,12.904)
(2010.596,12.923)
(2010.615,12.885)
(2010.635,12.923)
(2010.654,12.923)
(2010.673,13.0)
(2010.692,13.058)
(2010.712,13.115)
(2010.731,13.096)
(2010.75,13.404)
(2010.769,13.423)
(2010.788,13.885)
(2010.808,14.038)
(2010.827,14.423)
(2010.846,15.519)
(2010.865,15.481)
(2010.885,15.865)
(2010.904,16.827)
(2010.923,17.077)
(2010.942,16.981)
(2010.962,16.865)
(2010.981,17.077)
(2011.0,18.038)
(2011.019,19.192)
(2011.038,19.0)
(2011.058,20.423)
(2011.077,21.346)
(2011.096,21.846)
(2011.115,22.635)
(2011.135,22.904)
(2011.154,23.692)
(2011.173,24.327)
(2011.192,25.077)
(2011.212,25.654)
(2011.231,26.385)
(2011.25,27.173)
(2011.269,27.808)
(2011.288,28.481)
(2011.308,29.038)
(2011.327,29.212)
(2011.346,29.269)
(2011.365,30.135)
(2011.385,30.481)
(2011.404,31.038)
(2011.423,31.308)
(2011.442,31.577)
(2011.462,31.827)
(2011.481,31.692)
(2011.5,31.269)
(2011.519,31.769)
(2011.538,31.423)
(2011.558,31.596)
(2011.577,31.596)
(2011.596,31.788)
(2011.615,31.769)
(2011.635,31.75)
(2011.654,31.673)
(2011.673,31.577)
(2011.692,31.596)
(2011.712,31.538)
(2011.731,31.538)
(2011.75,31.308)
(2011.769,31.365)
(2011.788,30.904)
(2011.808,30.692)
(2011.827,30.115)
(2011.846,29.038)
(2011.865,28.615)
(2011.885,28.058)
(2011.904,26.615)
(2011.923,26.135)
(2011.942,25.942)
(2011.962,25.635)
(2011.981,25.192)
(2012.0,25.404)
(2012.019,24.923)
(2012.038,25.173)
(2012.058,23.962)
(2012.077,22.942)
(2012.096,22.846)
(2012.115,22.25)
(2012.135,26.0)
(2012.154,25.346)
(2012.173,25.096)
(2012.192,24.5)
(2012.212,24.288)
(2012.231,23.923)
(2012.25,23.635)
(2012.269,22.827)
(2012.288,22.192)
(2012.308,21.654)
(2012.327,21.635)
(2012.346,22.077)
(2012.365,20.865)
(2012.385,21.0)
(2012.404,20.462)
(2012.423,19.865)
(2012.442,19.192)
(2012.462,18.769)
(2012.481,18.788)
(2012.5,18.365)
(2012.519,17.846)
(2012.538,17.769)
(2012.558,17.635)
(2012.577,17.538)
(2012.596,17.269)
(2012.615,17.212)
(2012.635,17.192)
(2012.654,17.269)
(2012.673,17.25)
(2012.692,17.25)
(2012.712,17.231)
(2012.731,17.269)
(2012.75,17.154)
(2012.769,17.077)
(2012.788,17.096)
(2012.808,17.096)
(2012.827,17.019)
(2012.846,17.077)
(2012.865,16.962)
(2012.885,17.0)
(2012.904,17.038)
(2012.923,17.077)
(2012.942,17.615)
(2012.962,17.192)
(2012.981,17.269)
(2013.0,15.962)
(2013.019,15.423)
(2013.038,14.865)
(2013.058,14.519)
(2013.077,14.769)
(2013.096,14.442)
(2013.115,14.231)
(2013.135,11.481)
(2013.154,11.288)
(2013.173,11.577)
(2013.192,11.538)
(2013.212,11.519)
(2013.231,11.365)
(2013.25,11.154)
(2013.269,11.135)
(2013.288,11.192)
(2013.308,10.731)
(2013.327,10.654)
(2013.346,10.115)
(2013.365,10.096)
(2013.385,9.404)
(2013.404,9.519)
(2013.423,9.519)
(2013.442,9.5)
(2013.462,9.654)
(2013.481,9.365)
(2013.5,9.615)
(2013.519,9.75)
(2013.538,9.942)
(2013.558,10.135)
(2013.577,10.269)
(2013.596,10.346)
(2013.615,10.385)
(2013.635,10.423)
(2013.654,10.385)
(2013.673,10.462)
(2013.692,10.385)
(2013.712,10.404)
(2013.731,10.365)
(2013.75,10.365)
(2013.769,10.365)
(2013.788,10.365)
(2013.808,10.365)
(2013.827,10.385)
(2013.846,10.288)
(2013.865,10.365)
(2013.885,10.269)
(2013.904,10.346)
(2013.923,10.442)
(2013.942,9.712)
(2013.962,9.558)
(2013.981,9.288)
(2014.0,9.346)
(2014.019,9.038)
(2014.038,9.135)
(2014.058,9.327)
(2014.077,9.269)
(2014.096,9.288)
(2014.115,9.596)
(2014.135,8.308)
(2014.154,8.538)
(2014.173,9.538)
(2014.192,9.923)
(2014.212,9.519)
(2014.231,9.481)
(2014.25,9.577)
(2014.269,10.077)
(2014.288,10.846)
(2014.308,11.192)
(2014.327,12.75)
(2014.346,12.558)
(2014.365,12.538)
(2014.385,14.231)
(2014.404,14.135)
(2014.423,14.038)
(2014.442,14.827)
(2014.462,15.327)
(2014.481,15.827)
(2014.5,15.923)
(2014.519,16.058)
(2014.538,15.827)
(2014.558,16.135)
(2014.577,16.231)
(2014.596,16.25)
(2014.615,16.192)
(2014.635,16.135)
(2014.654,16.269)
(2014.673,16.25)
(2014.692,16.288)
(2014.712,16.288)
(2014.731,16.288)
(2014.75,16.288)
(2014.769,16.308)
(2014.788,16.288)
(2014.808,16.365)
(2014.827,16.327)
(2014.846,16.423)
(2014.865,16.519)
(2014.885,16.635)
(2014.904,16.538)
(2014.923,16.365)
(2014.942,16.327)
(2014.962,16.462)
(2014.981,16.596)
(2015.0,16.712)
(2015.019,16.712)
}
\def\LEVELONE{

(2006.0,0.232)
(2006.019,0.237)
(2006.038,0.242)
(2006.058,0.247)
(2006.077,0.247)
(2006.096,0.25)
(2006.115,0.25)
(2006.135,0.25)
(2006.154,0.249)
(2006.173,0.251)
(2006.192,0.249)
(2006.212,0.248)
(2006.231,0.247)
(2006.25,0.249)
(2006.269,0.244)
(2006.288,0.24)
(2006.308,0.24)
(2006.327,0.24)
(2006.346,0.238)
(2006.365,0.237)
(2006.385,0.235)
(2006.404,0.232)
(2006.423,0.234)
(2006.442,0.233)
(2006.462,0.237)
(2006.481,0.235)
(2006.5,0.239)
(2006.519,0.238)
(2006.538,0.241)
(2006.558,0.243)
(2006.577,0.245)
(2006.596,0.248)
(2006.615,0.248)
(2006.635,0.246)
(2006.654,0.244)
(2006.673,0.242)
(2006.692,0.238)
(2006.712,0.237)
(2006.731,0.237)
(2006.75,0.237)
(2006.769,0.238)
(2006.788,0.237)
(2006.808,0.244)
(2006.827,0.245)
(2006.846,0.245)
(2006.865,0.246)
(2006.885,0.25)
(2006.904,0.25)
(2006.923,0.247)
(2006.942,0.24)
(2006.962,0.236)
(2006.981,0.237)
(2007.0,0.237)
(2007.019,0.231)
(2007.038,0.233)
(2007.058,0.232)
(2007.077,0.229)
(2007.096,0.224)
(2007.115,0.224)
(2007.135,0.221)
(2007.154,0.223)
(2007.173,0.219)
(2007.192,0.223)
(2007.212,0.222)
(2007.231,0.223)
(2007.25,0.224)
(2007.269,0.224)
(2007.288,0.227)
(2007.308,0.227)
(2007.327,0.227)
(2007.346,0.229)
(2007.365,0.23)
(2007.385,0.231)
(2007.404,0.235)
(2007.423,0.233)
(2007.442,0.241)
(2007.462,0.24)
(2007.481,0.239)
(2007.5,0.236)
(2007.519,0.235)
(2007.538,0.235)
(2007.558,0.231)
(2007.577,0.233)
(2007.596,0.237)
(2007.615,0.24)
(2007.635,0.244)
(2007.654,0.252)
(2007.673,0.255)
(2007.692,0.257)
(2007.712,0.26)
(2007.731,0.26)
(2007.75,0.261)
(2007.769,0.259)
(2007.788,0.262)
(2007.808,0.257)
(2007.827,0.255)
(2007.846,0.259)
(2007.865,0.256)
(2007.885,0.254)
(2007.904,0.253)
(2007.923,0.26)
(2007.942,0.261)
(2007.962,0.264)
(2007.981,0.262)
(2008.0,0.269)
(2008.019,0.267)
(2008.038,0.265)
(2008.058,0.266)
(2008.077,0.271)
(2008.096,0.275)
(2008.115,0.277)
(2008.135,0.279)
(2008.154,0.277)
(2008.173,0.276)
(2008.192,0.27)
(2008.212,0.268)
(2008.231,0.267)
(2008.25,0.262)
(2008.269,0.259)
(2008.288,0.257)
(2008.308,0.254)
(2008.327,0.253)
(2008.346,0.255)
(2008.365,0.252)
(2008.385,0.254)
(2008.404,0.252)
(2008.423,0.251)
(2008.442,0.244)
(2008.462,0.248)
(2008.481,0.252)
(2008.5,0.253)
(2008.519,0.249)
(2008.538,0.246)
(2008.558,0.249)
(2008.577,0.244)
(2008.596,0.239)
(2008.615,0.238)
(2008.635,0.237)
(2008.654,0.233)
(2008.673,0.234)
(2008.692,0.235)
(2008.712,0.237)
(2008.731,0.235)
(2008.75,0.234)
(2008.769,0.235)
(2008.788,0.231)
(2008.808,0.228)
(2008.827,0.234)
(2008.846,0.226)
(2008.865,0.226)
(2008.885,0.225)
(2008.904,0.221)
(2008.923,0.218)
(2008.942,0.217)
(2008.962,0.215)
(2008.981,0.214)
(2009.0,0.215)
(2009.019,0.216)
(2009.038,0.215)
(2009.058,0.21)
(2009.077,0.205)
(2009.096,0.206)
(2009.115,0.2)
(2009.135,0.195)
(2009.154,0.19)
(2009.173,0.191)
(2009.192,0.188)
(2009.212,0.189)
(2009.231,0.187)
(2009.25,0.189)
(2009.269,0.188)
(2009.288,0.189)
(2009.308,0.189)
(2009.327,0.191)
(2009.346,0.19)
(2009.365,0.188)
(2009.385,0.183)
(2009.404,0.183)
(2009.423,0.184)
(2009.442,0.186)
(2009.462,0.18)
(2009.481,0.179)
(2009.5,0.176)
(2009.519,0.177)
(2009.538,0.181)
(2009.558,0.18)
(2009.577,0.182)
(2009.596,0.184)
(2009.615,0.183)
(2009.635,0.182)
(2009.654,0.181)
(2009.673,0.177)
(2009.692,0.176)
(2009.712,0.174)
(2009.731,0.178)
(2009.75,0.178)
(2009.769,0.181)
(2009.788,0.185)
(2009.808,0.187)
(2009.827,0.182)
(2009.846,0.188)
(2009.865,0.19)
(2009.885,0.189)
(2009.904,0.194)
(2009.923,0.19)
(2009.942,0.185)
(2009.962,0.183)
(2009.981,0.182)
(2010.0,0.178)
(2010.019,0.176)
(2010.038,0.175)
(2010.058,0.178)
(2010.077,0.176)
(2010.096,0.172)
(2010.115,0.175)
(2010.135,0.18)
(2010.154,0.186)
(2010.173,0.187)
(2010.192,0.193)
(2010.212,0.195)
(2010.231,0.198)
(2010.25,0.2)
(2010.269,0.206)
(2010.288,0.211)
(2010.308,0.212)
(2010.327,0.212)
(2010.346,0.216)
(2010.365,0.221)
(2010.385,0.224)
(2010.404,0.227)
(2010.423,0.226)
(2010.442,0.227)
(2010.462,0.23)
(2010.481,0.236)
(2010.5,0.238)
(2010.519,0.243)
(2010.538,0.239)
(2010.558,0.241)
(2010.577,0.244)
(2010.596,0.246)
(2010.615,0.247)
(2010.635,0.249)
(2010.654,0.249)
(2010.673,0.253)
(2010.692,0.252)
(2010.712,0.253)
(2010.731,0.252)
(2010.75,0.252)
(2010.769,0.249)
(2010.788,0.249)
(2010.808,0.248)
(2010.827,0.246)
(2010.846,0.247)
(2010.865,0.248)
(2010.885,0.252)
(2010.904,0.252)
(2010.923,0.257)
(2010.942,0.264)
(2010.962,0.267)
(2010.981,0.269)
(2011.0,0.267)
(2011.019,0.268)
(2011.038,0.266)
(2011.058,0.268)
(2011.077,0.269)
(2011.096,0.27)
(2011.115,0.273)
(2011.135,0.269)
(2011.154,0.264)
(2011.173,0.262)
(2011.192,0.258)
(2011.212,0.254)
(2011.231,0.253)
(2011.25,0.254)
(2011.269,0.249)
(2011.288,0.243)
(2011.308,0.241)
(2011.327,0.241)
(2011.346,0.233)
(2011.365,0.235)
(2011.385,0.238)
(2011.404,0.235)
(2011.423,0.237)
(2011.442,0.234)
(2011.462,0.234)
(2011.481,0.227)
(2011.5,0.23)
(2011.519,0.224)
(2011.538,0.225)
(2011.558,0.221)
(2011.577,0.216)
(2011.596,0.215)
(2011.615,0.215)
(2011.635,0.212)
(2011.654,0.21)
(2011.673,0.21)
(2011.692,0.211)
(2011.712,0.209)
(2011.731,0.202)
(2011.75,0.202)
(2011.769,0.202)
(2011.788,0.199)
(2011.808,0.201)
(2011.827,0.198)
(2011.846,0.194)
(2011.865,0.191)
(2011.885,0.188)
(2011.904,0.184)
(2011.923,0.179)
(2011.942,0.177)
(2011.962,0.176)
(2011.981,0.175)
(2012.0,0.18)
(2012.019,0.182)
(2012.038,0.183)
(2012.058,0.178)
(2012.077,0.183)
(2012.096,0.182)
(2012.115,0.179)
(2012.135,0.184)
(2012.154,0.184)
(2012.173,0.189)
(2012.192,0.191)
(2012.212,0.196)
(2012.231,0.202)
(2012.25,0.197)
(2012.269,0.201)
(2012.288,0.2)
(2012.308,0.203)
(2012.327,0.2)
(2012.346,0.2)
(2012.365,0.198)
(2012.385,0.195)
(2012.404,0.197)
(2012.423,0.196)
(2012.442,0.192)
(2012.462,0.189)
(2012.481,0.191)
(2012.5,0.188)
(2012.519,0.193)
(2012.538,0.196)
(2012.558,0.195)
(2012.577,0.196)
(2012.596,0.194)
(2012.615,0.195)
(2012.635,0.194)
(2012.654,0.196)
(2012.673,0.197)
(2012.692,0.2)
(2012.712,0.203)
(2012.731,0.208)
(2012.75,0.208)
(2012.769,0.207)
(2012.788,0.209)
(2012.808,0.211)
(2012.827,0.211)
(2012.846,0.211)
(2012.865,0.215)
(2012.885,0.214)
(2012.904,0.218)
(2012.923,0.219)
(2012.942,0.221)
(2012.962,0.22)
(2012.981,0.222)
(2013.0,0.221)
(2013.019,0.227)
(2013.038,0.23)
(2013.058,0.231)
(2013.077,0.227)
(2013.096,0.223)
(2013.115,0.224)
(2013.135,0.223)
(2013.154,0.221)
(2013.173,0.217)
(2013.192,0.218)
(2013.212,0.215)
(2013.231,0.212)
(2013.25,0.217)
(2013.269,0.215)
(2013.288,0.215)
(2013.308,0.214)
(2013.327,0.214)
(2013.346,0.218)
(2013.365,0.217)
(2013.385,0.221)
(2013.404,0.219)
(2013.423,0.22)
(2013.442,0.224)
(2013.462,0.228)
(2013.481,0.229)
(2013.5,0.23)
(2013.519,0.228)
(2013.538,0.224)
(2013.558,0.228)
(2013.577,0.23)
(2013.596,0.23)
(2013.615,0.228)
(2013.635,0.226)
(2013.654,0.225)
(2013.673,0.222)
(2013.692,0.217)
(2013.712,0.217)
(2013.731,0.22)
(2013.75,0.213)
(2013.769,0.217)
(2013.788,0.216)
(2013.808,0.214)
(2013.827,0.218)
(2013.846,0.219)
(2013.865,0.217)
(2013.885,0.217)
(2013.904,0.212)
(2013.923,0.212)
(2013.942,0.216)
(2013.962,0.214)
(2013.981,0.211)
(2014.0,0.207)
(2014.019,0.204)
(2014.038,0.203)
(2014.058,0.201)
(2014.077,0.198)
(2014.096,0.199)
(2014.115,0.199)
(2014.135,0.197)
(2014.154,0.2)
(2014.173,0.198)
(2014.192,0.195)
(2014.212,0.193)
(2014.231,0.189)
(2014.25,0.186)
(2014.269,0.187)
(2014.288,0.189)
(2014.308,0.19)
(2014.327,0.191)
(2014.346,0.191)
(2014.365,0.19)
(2014.385,0.192)
(2014.404,0.192)
(2014.423,0.197)
(2014.442,0.193)
(2014.462,0.192)
(2014.481,0.191)
(2014.5,0.193)
(2014.519,0.191)
(2014.538,0.198)
(2014.558,0.193)
(2014.577,0.191)
(2014.596,0.189)
(2014.615,0.192)
(2014.635,0.195)
(2014.654,0.195)
(2014.673,0.198)
(2014.692,0.202)
(2014.712,0.199)
(2014.731,0.198)
(2014.75,0.202)
(2014.769,0.204)
(2014.788,0.203)
(2014.808,0.2)
(2014.827,0.196)
(2014.846,0.197)
(2014.865,0.199)
(2014.885,0.202)
(2014.904,0.202)
(2014.923,0.203)
(2014.942,0.201)
(2014.962,0.199)
(2014.981,0.202)
(2015.0,0.201)
(2015.019,0.195)
}

\def\LEVELTWO{
(2006.0,0.187)
(2006.019,0.192)
(2006.038,0.195)
(2006.058,0.201)
(2006.077,0.203)
(2006.096,0.205)
(2006.115,0.207)
(2006.135,0.208)
(2006.154,0.208)
(2006.173,0.212)
(2006.192,0.212)
(2006.212,0.213)
(2006.231,0.214)
(2006.25,0.216)
(2006.269,0.215)
(2006.288,0.213)
(2006.308,0.214)
(2006.327,0.214)
(2006.346,0.213)
(2006.365,0.214)
(2006.385,0.213)
(2006.404,0.213)
(2006.423,0.215)
(2006.442,0.215)
(2006.462,0.218)
(2006.481,0.217)
(2006.5,0.22)
(2006.519,0.218)
(2006.538,0.219)
(2006.558,0.218)
(2006.577,0.217)
(2006.596,0.218)
(2006.615,0.218)
(2006.635,0.217)
(2006.654,0.215)
(2006.673,0.213)
(2006.692,0.211)
(2006.712,0.21)
(2006.731,0.211)
(2006.75,0.211)
(2006.769,0.212)
(2006.788,0.211)
(2006.808,0.217)
(2006.827,0.217)
(2006.846,0.217)
(2006.865,0.219)
(2006.885,0.222)
(2006.904,0.222)
(2006.923,0.22)
(2006.942,0.213)
(2006.962,0.21)
(2006.981,0.209)
(2007.0,0.207)
(2007.019,0.202)
(2007.038,0.202)
(2007.058,0.2)
(2007.077,0.197)
(2007.096,0.191)
(2007.115,0.19)
(2007.135,0.188)
(2007.154,0.186)
(2007.173,0.182)
(2007.192,0.183)
(2007.212,0.182)
(2007.231,0.181)
(2007.25,0.179)
(2007.269,0.178)
(2007.288,0.179)
(2007.308,0.18)
(2007.327,0.179)
(2007.346,0.179)
(2007.365,0.18)
(2007.385,0.179)
(2007.404,0.181)
(2007.423,0.179)
(2007.442,0.184)
(2007.462,0.184)
(2007.481,0.183)
(2007.5,0.182)
(2007.519,0.183)
(2007.538,0.183)
(2007.558,0.182)
(2007.577,0.184)
(2007.596,0.187)
(2007.615,0.188)
(2007.635,0.191)
(2007.654,0.195)
(2007.673,0.198)
(2007.692,0.199)
(2007.712,0.2)
(2007.731,0.199)
(2007.75,0.199)
(2007.769,0.198)
(2007.788,0.199)
(2007.808,0.194)
(2007.827,0.194)
(2007.846,0.195)
(2007.865,0.192)
(2007.885,0.19)
(2007.904,0.19)
(2007.923,0.194)
(2007.942,0.194)
(2007.962,0.195)
(2007.981,0.194)
(2008.0,0.198)
(2008.019,0.199)
(2008.038,0.198)
(2008.058,0.197)
(2008.077,0.202)
(2008.096,0.205)
(2008.115,0.206)
(2008.135,0.206)
(2008.154,0.205)
(2008.173,0.205)
(2008.192,0.201)
(2008.212,0.2)
(2008.231,0.199)
(2008.25,0.198)
(2008.269,0.197)
(2008.288,0.196)
(2008.308,0.195)
(2008.327,0.195)
(2008.346,0.196)
(2008.365,0.194)
(2008.385,0.197)
(2008.404,0.195)
(2008.423,0.196)
(2008.442,0.192)
(2008.462,0.194)
(2008.481,0.195)
(2008.5,0.195)
(2008.519,0.192)
(2008.538,0.192)
(2008.558,0.193)
(2008.577,0.191)
(2008.596,0.188)
(2008.615,0.189)
(2008.635,0.188)
(2008.654,0.186)
(2008.673,0.187)
(2008.692,0.187)
(2008.712,0.189)
(2008.731,0.189)
(2008.75,0.189)
(2008.769,0.19)
(2008.788,0.188)
(2008.808,0.186)
(2008.827,0.19)
(2008.846,0.185)
(2008.865,0.185)
(2008.885,0.185)
(2008.904,0.182)
(2008.923,0.179)
(2008.942,0.179)
(2008.962,0.178)
(2008.981,0.179)
(2009.0,0.181)
(2009.019,0.182)
(2009.038,0.183)
(2009.058,0.181)
(2009.077,0.176)
(2009.096,0.177)
(2009.115,0.174)
(2009.135,0.172)
(2009.154,0.172)
(2009.173,0.172)
(2009.192,0.171)
(2009.212,0.172)
(2009.231,0.171)
(2009.25,0.172)
(2009.269,0.171)
(2009.288,0.171)
(2009.308,0.17)
(2009.327,0.171)
(2009.346,0.171)
(2009.365,0.17)
(2009.385,0.167)
(2009.404,0.168)
(2009.423,0.168)
(2009.442,0.168)
(2009.462,0.166)
(2009.481,0.166)
(2009.5,0.165)
(2009.519,0.165)
(2009.538,0.17)
(2009.558,0.17)
(2009.577,0.17)
(2009.596,0.171)
(2009.615,0.169)
(2009.635,0.168)
(2009.654,0.168)
(2009.673,0.164)
(2009.692,0.164)
(2009.712,0.162)
(2009.731,0.164)
(2009.75,0.163)
(2009.769,0.164)
(2009.788,0.166)
(2009.808,0.167)
(2009.827,0.163)
(2009.846,0.166)
(2009.865,0.167)
(2009.885,0.166)
(2009.904,0.169)
(2009.923,0.168)
(2009.942,0.166)
(2009.962,0.165)
(2009.981,0.163)
(2010.0,0.16)
(2010.019,0.158)
(2010.038,0.157)
(2010.058,0.159)
(2010.077,0.158)
(2010.096,0.156)
(2010.115,0.157)
(2010.135,0.159)
(2010.154,0.161)
(2010.173,0.16)
(2010.192,0.163)
(2010.212,0.162)
(2010.231,0.164)
(2010.25,0.163)
(2010.269,0.167)
(2010.288,0.169)
(2010.308,0.171)
(2010.327,0.17)
(2010.346,0.173)
(2010.365,0.176)
(2010.385,0.178)
(2010.404,0.18)
(2010.423,0.179)
(2010.442,0.18)
(2010.462,0.181)
(2010.481,0.186)
(2010.5,0.186)
(2010.519,0.19)
(2010.538,0.185)
(2010.558,0.186)
(2010.577,0.191)
(2010.596,0.192)
(2010.615,0.193)
(2010.635,0.194)
(2010.654,0.194)
(2010.673,0.196)
(2010.692,0.196)
(2010.712,0.196)
(2010.731,0.196)
(2010.75,0.196)
(2010.769,0.194)
(2010.788,0.195)
(2010.808,0.195)
(2010.827,0.194)
(2010.846,0.194)
(2010.865,0.194)
(2010.885,0.196)
(2010.904,0.196)
(2010.923,0.198)
(2010.942,0.201)
(2010.962,0.203)
(2010.981,0.205)
(2011.0,0.203)
(2011.019,0.203)
(2011.038,0.202)
(2011.058,0.202)
(2011.077,0.202)
(2011.096,0.203)
(2011.115,0.204)
(2011.135,0.203)
(2011.154,0.201)
(2011.173,0.202)
(2011.192,0.2)
(2011.212,0.199)
(2011.231,0.199)
(2011.25,0.2)
(2011.269,0.198)
(2011.288,0.195)
(2011.308,0.191)
(2011.327,0.19)
(2011.346,0.186)
(2011.365,0.186)
(2011.385,0.188)
(2011.404,0.185)
(2011.423,0.186)
(2011.442,0.184)
(2011.462,0.185)
(2011.481,0.181)
(2011.5,0.183)
(2011.519,0.178)
(2011.538,0.177)
(2011.558,0.174)
(2011.577,0.168)
(2011.596,0.166)
(2011.615,0.166)
(2011.635,0.165)
(2011.654,0.165)
(2011.673,0.165)
(2011.692,0.166)
(2011.712,0.164)
(2011.731,0.161)
(2011.75,0.161)
(2011.769,0.16)
(2011.788,0.158)
(2011.808,0.158)
(2011.827,0.156)
(2011.846,0.154)
(2011.865,0.153)
(2011.885,0.153)
(2011.904,0.15)
(2011.923,0.149)
(2011.942,0.148)
(2011.962,0.147)
(2011.981,0.147)
(2012.0,0.149)
(2012.019,0.15)
(2012.038,0.151)
(2012.058,0.149)
(2012.077,0.151)
(2012.096,0.149)
(2012.115,0.148)
(2012.135,0.15)
(2012.154,0.151)
(2012.173,0.154)
(2012.192,0.155)
(2012.212,0.159)
(2012.231,0.164)
(2012.25,0.162)
(2012.269,0.164)
(2012.288,0.164)
(2012.308,0.167)
(2012.327,0.166)
(2012.346,0.166)
(2012.365,0.167)
(2012.385,0.164)
(2012.404,0.166)
(2012.423,0.166)
(2012.442,0.164)
(2012.462,0.161)
(2012.481,0.161)
(2012.5,0.16)
(2012.519,0.164)
(2012.538,0.167)
(2012.558,0.166)
(2012.577,0.168)
(2012.596,0.168)
(2012.615,0.168)
(2012.635,0.167)
(2012.654,0.168)
(2012.673,0.169)
(2012.692,0.171)
(2012.712,0.173)
(2012.731,0.175)
(2012.75,0.175)
(2012.769,0.174)
(2012.788,0.176)
(2012.808,0.178)
(2012.827,0.179)
(2012.846,0.178)
(2012.865,0.181)
(2012.885,0.18)
(2012.904,0.184)
(2012.923,0.183)
(2012.942,0.183)
(2012.962,0.183)
(2012.981,0.184)
(2013.0,0.185)
(2013.019,0.189)
(2013.038,0.19)
(2013.058,0.189)
(2013.077,0.187)
(2013.096,0.187)
(2013.115,0.188)
(2013.135,0.187)
(2013.154,0.186)
(2013.173,0.183)
(2013.192,0.183)
(2013.212,0.18)
(2013.231,0.177)
(2013.25,0.178)
(2013.269,0.176)
(2013.288,0.176)
(2013.308,0.175)
(2013.327,0.175)
(2013.346,0.176)
(2013.365,0.175)
(2013.385,0.176)
(2013.404,0.176)
(2013.423,0.175)
(2013.442,0.177)
(2013.462,0.18)
(2013.481,0.18)
(2013.5,0.18)
(2013.519,0.178)
(2013.538,0.175)
(2013.558,0.178)
(2013.577,0.178)
(2013.596,0.177)
(2013.615,0.175)
(2013.635,0.175)
(2013.654,0.174)
(2013.673,0.172)
(2013.692,0.169)
(2013.712,0.169)
(2013.731,0.17)
(2013.75,0.168)
(2013.769,0.171)
(2013.788,0.17)
(2013.808,0.169)
(2013.827,0.17)
(2013.846,0.172)
(2013.865,0.17)
(2013.885,0.17)
(2013.904,0.165)
(2013.923,0.166)
(2013.942,0.169)
(2013.962,0.168)
(2013.981,0.166)
(2014.0,0.164)
(2014.019,0.161)
(2014.038,0.161)
(2014.058,0.161)
(2014.077,0.16)
(2014.096,0.16)
(2014.115,0.159)
(2014.135,0.158)
(2014.154,0.16)
(2014.173,0.158)
(2014.192,0.157)
(2014.212,0.157)
(2014.231,0.155)
(2014.25,0.155)
(2014.269,0.157)
(2014.288,0.158)
(2014.308,0.158)
(2014.327,0.158)
(2014.346,0.158)
(2014.365,0.157)
(2014.385,0.159)
(2014.404,0.16)
(2014.423,0.165)
(2014.442,0.163)
(2014.462,0.163)
(2014.481,0.162)
(2014.5,0.164)
(2014.519,0.164)
(2014.538,0.168)
(2014.558,0.164)
(2014.577,0.163)
(2014.596,0.163)
(2014.615,0.164)
(2014.635,0.166)
(2014.654,0.166)
(2014.673,0.167)
(2014.692,0.17)
(2014.712,0.169)
(2014.731,0.168)
(2014.75,0.17)
(2014.769,0.17)
(2014.788,0.17)
(2014.808,0.168)
(2014.827,0.167)
(2014.846,0.167)
(2014.865,0.169)
(2014.885,0.171)
(2014.904,0.172)
(2014.923,0.173)
(2014.942,0.172)
(2014.962,0.172)
(2014.981,0.175)
(2015.0,0.175)
(2015.019,0.172)

}

\def\LEVELTHREE{
(2006.0,0.118)
(2006.019,0.123)
(2006.038,0.125)
(2006.058,0.131)
(2006.077,0.133)
(2006.096,0.136)
(2006.115,0.137)
(2006.135,0.139)
(2006.154,0.14)
(2006.173,0.143)
(2006.192,0.144)
(2006.212,0.146)
(2006.231,0.147)
(2006.25,0.15)
(2006.269,0.149)
(2006.288,0.148)
(2006.308,0.148)
(2006.327,0.149)
(2006.346,0.149)
(2006.365,0.149)
(2006.385,0.15)
(2006.404,0.149)
(2006.423,0.15)
(2006.442,0.151)
(2006.462,0.153)
(2006.481,0.152)
(2006.5,0.155)
(2006.519,0.154)
(2006.538,0.154)
(2006.558,0.154)
(2006.577,0.153)
(2006.596,0.154)
(2006.615,0.155)
(2006.635,0.153)
(2006.654,0.152)
(2006.673,0.15)
(2006.692,0.148)
(2006.712,0.148)
(2006.731,0.148)
(2006.75,0.149)
(2006.769,0.148)
(2006.788,0.148)
(2006.808,0.154)
(2006.827,0.154)
(2006.846,0.154)
(2006.865,0.156)
(2006.885,0.157)
(2006.904,0.157)
(2006.923,0.156)
(2006.942,0.149)
(2006.962,0.147)
(2006.981,0.144)
(2007.0,0.143)
(2007.019,0.138)
(2007.038,0.137)
(2007.058,0.135)
(2007.077,0.132)
(2007.096,0.128)
(2007.115,0.126)
(2007.135,0.124)
(2007.154,0.122)
(2007.173,0.117)
(2007.192,0.118)
(2007.212,0.116)
(2007.231,0.115)
(2007.25,0.113)
(2007.269,0.111)
(2007.288,0.112)
(2007.308,0.113)
(2007.327,0.112)
(2007.346,0.112)
(2007.365,0.113)
(2007.385,0.111)
(2007.404,0.113)
(2007.423,0.112)
(2007.442,0.117)
(2007.462,0.117)
(2007.481,0.117)
(2007.5,0.116)
(2007.519,0.117)
(2007.538,0.117)
(2007.558,0.117)
(2007.577,0.118)
(2007.596,0.121)
(2007.615,0.122)
(2007.635,0.124)
(2007.654,0.127)
(2007.673,0.129)
(2007.692,0.13)
(2007.712,0.131)
(2007.731,0.131)
(2007.75,0.131)
(2007.769,0.13)
(2007.788,0.131)
(2007.808,0.127)
(2007.827,0.126)
(2007.846,0.127)
(2007.865,0.124)
(2007.885,0.122)
(2007.904,0.122)
(2007.923,0.126)
(2007.942,0.126)
(2007.962,0.126)
(2007.981,0.126)
(2008.0,0.128)
(2008.019,0.129)
(2008.038,0.128)
(2008.058,0.127)
(2008.077,0.132)
(2008.096,0.134)
(2008.115,0.135)
(2008.135,0.135)
(2008.154,0.135)
(2008.173,0.134)
(2008.192,0.132)
(2008.212,0.131)
(2008.231,0.13)
(2008.25,0.128)
(2008.269,0.127)
(2008.288,0.126)
(2008.308,0.126)
(2008.327,0.125)
(2008.346,0.126)
(2008.365,0.125)
(2008.385,0.127)
(2008.404,0.125)
(2008.423,0.125)
(2008.442,0.121)
(2008.462,0.123)
(2008.481,0.124)
(2008.5,0.123)
(2008.519,0.121)
(2008.538,0.12)
(2008.558,0.121)
(2008.577,0.119)
(2008.596,0.118)
(2008.615,0.117)
(2008.635,0.117)
(2008.654,0.116)
(2008.673,0.117)
(2008.692,0.117)
(2008.712,0.118)
(2008.731,0.118)
(2008.75,0.119)
(2008.769,0.119)
(2008.788,0.118)
(2008.808,0.116)
(2008.827,0.119)
(2008.846,0.115)
(2008.865,0.115)
(2008.885,0.115)
(2008.904,0.113)
(2008.923,0.11)
(2008.942,0.11)
(2008.962,0.109)
(2008.981,0.11)
(2009.0,0.112)
(2009.019,0.113)
(2009.038,0.114)
(2009.058,0.114)
(2009.077,0.109)
(2009.096,0.11)
(2009.115,0.107)
(2009.135,0.106)
(2009.154,0.106)
(2009.173,0.106)
(2009.192,0.106)
(2009.212,0.106)
(2009.231,0.106)
(2009.25,0.107)
(2009.269,0.107)
(2009.288,0.107)
(2009.308,0.107)
(2009.327,0.107)
(2009.346,0.108)
(2009.365,0.107)
(2009.385,0.105)
(2009.404,0.105)
(2009.423,0.106)
(2009.442,0.106)
(2009.462,0.104)
(2009.481,0.104)
(2009.5,0.104)
(2009.519,0.104)
(2009.538,0.108)
(2009.558,0.109)
(2009.577,0.109)
(2009.596,0.109)
(2009.615,0.108)
(2009.635,0.107)
(2009.654,0.107)
(2009.673,0.104)
(2009.692,0.104)
(2009.712,0.103)
(2009.731,0.104)
(2009.75,0.103)
(2009.769,0.104)
(2009.788,0.105)
(2009.808,0.106)
(2009.827,0.103)
(2009.846,0.105)
(2009.865,0.105)
(2009.885,0.105)
(2009.904,0.107)
(2009.923,0.106)
(2009.942,0.105)
(2009.962,0.104)
(2009.981,0.103)
(2010.0,0.1)
(2010.019,0.099)
(2010.038,0.098)
(2010.058,0.099)
(2010.077,0.099)
(2010.096,0.097)
(2010.115,0.098)
(2010.135,0.1)
(2010.154,0.101)
(2010.173,0.1)
(2010.192,0.102)
(2010.212,0.102)
(2010.231,0.102)
(2010.25,0.102)
(2010.269,0.105)
(2010.288,0.107)
(2010.308,0.107)
(2010.327,0.106)
(2010.346,0.11)
(2010.365,0.112)
(2010.385,0.113)
(2010.404,0.115)
(2010.423,0.115)
(2010.442,0.116)
(2010.462,0.116)
(2010.481,0.12)
(2010.5,0.12)
(2010.519,0.123)
(2010.538,0.119)
(2010.558,0.12)
(2010.577,0.125)
(2010.596,0.127)
(2010.615,0.128)
(2010.635,0.128)
(2010.654,0.128)
(2010.673,0.129)
(2010.692,0.129)
(2010.712,0.129)
(2010.731,0.129)
(2010.75,0.129)
(2010.769,0.128)
(2010.788,0.128)
(2010.808,0.128)
(2010.827,0.127)
(2010.846,0.127)
(2010.865,0.127)
(2010.885,0.128)
(2010.904,0.128)
(2010.923,0.129)
(2010.942,0.132)
(2010.962,0.133)
(2010.981,0.134)
(2011.0,0.133)
(2011.019,0.133)
(2011.038,0.132)
(2011.058,0.131)
(2011.077,0.132)
(2011.096,0.132)
(2011.115,0.132)
(2011.135,0.131)
(2011.154,0.13)
(2011.173,0.13)
(2011.192,0.129)
(2011.212,0.129)
(2011.231,0.129)
(2011.25,0.13)
(2011.269,0.128)
(2011.288,0.126)
(2011.308,0.123)
(2011.327,0.122)
(2011.346,0.118)
(2011.365,0.118)
(2011.385,0.119)
(2011.404,0.117)
(2011.423,0.118)
(2011.442,0.117)
(2011.462,0.118)
(2011.481,0.115)
(2011.5,0.117)
(2011.519,0.113)
(2011.538,0.112)
(2011.558,0.109)
(2011.577,0.104)
(2011.596,0.101)
(2011.615,0.101)
(2011.635,0.1)
(2011.654,0.1)
(2011.673,0.101)
(2011.692,0.101)
(2011.712,0.1)
(2011.731,0.098)
(2011.75,0.097)
(2011.769,0.097)
(2011.788,0.095)
(2011.808,0.095)
(2011.827,0.094)
(2011.846,0.093)
(2011.865,0.092)
(2011.885,0.092)
(2011.904,0.089)
(2011.923,0.089)
(2011.942,0.088)
(2011.962,0.088)
(2011.981,0.087)
(2012.0,0.089)
(2012.019,0.089)
(2012.038,0.09)
(2012.058,0.089)
(2012.077,0.09)
(2012.096,0.089)
(2012.115,0.088)
(2012.135,0.091)
(2012.154,0.091)
(2012.173,0.093)
(2012.192,0.094)
(2012.212,0.098)
(2012.231,0.103)
(2012.25,0.101)
(2012.269,0.103)
(2012.288,0.103)
(2012.308,0.106)
(2012.327,0.106)
(2012.346,0.106)
(2012.365,0.107)
(2012.385,0.104)
(2012.404,0.106)
(2012.423,0.105)
(2012.442,0.103)
(2012.462,0.1)
(2012.481,0.1)
(2012.5,0.098)
(2012.519,0.101)
(2012.538,0.104)
(2012.558,0.103)
(2012.577,0.105)
(2012.596,0.105)
(2012.615,0.106)
(2012.635,0.104)
(2012.654,0.105)
(2012.673,0.105)
(2012.692,0.107)
(2012.712,0.108)
(2012.731,0.11)
(2012.75,0.109)
(2012.769,0.109)
(2012.788,0.111)
(2012.808,0.112)
(2012.827,0.113)
(2012.846,0.112)
(2012.865,0.115)
(2012.885,0.114)
(2012.904,0.118)
(2012.923,0.117)
(2012.942,0.118)
(2012.962,0.118)
(2012.981,0.118)
(2013.0,0.119)
(2013.019,0.123)
(2013.038,0.124)
(2013.058,0.123)
(2013.077,0.121)
(2013.096,0.121)
(2013.115,0.122)
(2013.135,0.121)
(2013.154,0.12)
(2013.173,0.118)
(2013.192,0.117)
(2013.212,0.115)
(2013.231,0.111)
(2013.25,0.112)
(2013.269,0.11)
(2013.288,0.111)
(2013.308,0.109)
(2013.327,0.109)
(2013.346,0.111)
(2013.365,0.109)
(2013.385,0.109)
(2013.404,0.109)
(2013.423,0.109)
(2013.442,0.11)
(2013.462,0.112)
(2013.481,0.112)
(2013.5,0.112)
(2013.519,0.11)
(2013.538,0.108)
(2013.558,0.11)
(2013.577,0.11)
(2013.596,0.11)
(2013.615,0.108)
(2013.635,0.108)
(2013.654,0.108)
(2013.673,0.106)
(2013.692,0.104)
(2013.712,0.104)
(2013.731,0.105)
(2013.75,0.103)
(2013.769,0.105)
(2013.788,0.105)
(2013.808,0.104)
(2013.827,0.105)
(2013.846,0.106)
(2013.865,0.105)
(2013.885,0.104)
(2013.904,0.1)
(2013.923,0.101)
(2013.942,0.104)
(2013.962,0.103)
(2013.981,0.102)
(2014.0,0.1)
(2014.019,0.097)
(2014.038,0.098)
(2014.058,0.097)
(2014.077,0.097)
(2014.096,0.097)
(2014.115,0.096)
(2014.135,0.095)
(2014.154,0.097)
(2014.173,0.095)
(2014.192,0.094)
(2014.212,0.094)
(2014.231,0.093)
(2014.25,0.093)
(2014.269,0.095)
(2014.288,0.096)
(2014.308,0.096)
(2014.327,0.096)
(2014.346,0.095)
(2014.365,0.095)
(2014.385,0.097)
(2014.404,0.097)
(2014.423,0.103)
(2014.442,0.102)
(2014.462,0.102)
(2014.481,0.102)
(2014.5,0.104)
(2014.519,0.104)
(2014.538,0.107)
(2014.558,0.104)
(2014.577,0.103)
(2014.596,0.103)
(2014.615,0.104)
(2014.635,0.105)
(2014.654,0.105)
(2014.673,0.106)
(2014.692,0.108)
(2014.712,0.107)
(2014.731,0.107)
(2014.75,0.108)
(2014.769,0.109)
(2014.788,0.109)
(2014.808,0.108)
(2014.827,0.107)
(2014.846,0.107)
(2014.865,0.109)
(2014.885,0.11)
(2014.904,0.11)
(2014.923,0.112)
(2014.942,0.112)
(2014.962,0.112)
(2014.981,0.115)
(2015.0,0.114)
(2015.019,0.112)
}

\def\LEVELFOUR{
(2006.0,0.088)
(2006.019,0.093)
(2006.038,0.095)
(2006.058,0.1)
(2006.077,0.102)
(2006.096,0.105)
(2006.115,0.106)
(2006.135,0.108)
(2006.154,0.109)
(2006.173,0.113)
(2006.192,0.114)
(2006.212,0.116)
(2006.231,0.118)
(2006.25,0.121)
(2006.269,0.121)
(2006.288,0.12)
(2006.308,0.12)
(2006.327,0.122)
(2006.346,0.121)
(2006.365,0.122)
(2006.385,0.123)
(2006.404,0.122)
(2006.423,0.123)
(2006.442,0.124)
(2006.462,0.125)
(2006.481,0.125)
(2006.5,0.127)
(2006.519,0.126)
(2006.538,0.126)
(2006.558,0.126)
(2006.577,0.125)
(2006.596,0.126)
(2006.615,0.126)
(2006.635,0.125)
(2006.654,0.123)
(2006.673,0.122)
(2006.692,0.121)
(2006.712,0.121)
(2006.731,0.121)
(2006.75,0.121)
(2006.769,0.121)
(2006.788,0.121)
(2006.808,0.126)
(2006.827,0.126)
(2006.846,0.126)
(2006.865,0.128)
(2006.885,0.13)
(2006.904,0.129)
(2006.923,0.128)
(2006.942,0.122)
(2006.962,0.12)
(2006.981,0.117)
(2007.0,0.116)
(2007.019,0.111)
(2007.038,0.11)
(2007.058,0.108)
(2007.077,0.105)
(2007.096,0.101)
(2007.115,0.1)
(2007.135,0.097)
(2007.154,0.095)
(2007.173,0.091)
(2007.192,0.09)
(2007.212,0.088)
(2007.231,0.087)
(2007.25,0.085)
(2007.269,0.083)
(2007.288,0.083)
(2007.308,0.084)
(2007.327,0.084)
(2007.346,0.084)
(2007.365,0.084)
(2007.385,0.082)
(2007.404,0.084)
(2007.423,0.083)
(2007.442,0.087)
(2007.462,0.088)
(2007.481,0.088)
(2007.5,0.087)
(2007.519,0.089)
(2007.538,0.089)
(2007.558,0.089)
(2007.577,0.09)
(2007.596,0.092)
(2007.615,0.093)
(2007.635,0.094)
(2007.654,0.097)
(2007.673,0.098)
(2007.692,0.099)
(2007.712,0.099)
(2007.731,0.099)
(2007.75,0.099)
(2007.769,0.098)
(2007.788,0.099)
(2007.808,0.095)
(2007.827,0.095)
(2007.846,0.095)
(2007.865,0.092)
(2007.885,0.09)
(2007.904,0.091)
(2007.923,0.093)
(2007.942,0.093)
(2007.962,0.093)
(2007.981,0.093)
(2008.0,0.095)
(2008.019,0.095)
(2008.038,0.095)
(2008.058,0.093)
(2008.077,0.098)
(2008.096,0.1)
(2008.115,0.101)
(2008.135,0.101)
(2008.154,0.1)
(2008.173,0.1)
(2008.192,0.098)
(2008.212,0.097)
(2008.231,0.096)
(2008.25,0.096)
(2008.269,0.094)
(2008.288,0.094)
(2008.308,0.094)
(2008.327,0.094)
(2008.346,0.094)
(2008.365,0.093)
(2008.385,0.095)
(2008.404,0.094)
(2008.423,0.094)
(2008.442,0.09)
(2008.462,0.091)
(2008.481,0.092)
(2008.5,0.091)
(2008.519,0.089)
(2008.538,0.088)
(2008.558,0.089)
(2008.577,0.088)
(2008.596,0.087)
(2008.615,0.087)
(2008.635,0.086)
(2008.654,0.086)
(2008.673,0.087)
(2008.692,0.087)
(2008.712,0.088)
(2008.731,0.088)
(2008.75,0.088)
(2008.769,0.088)
(2008.788,0.088)
(2008.808,0.086)
(2008.827,0.089)
(2008.846,0.085)
(2008.865,0.086)
(2008.885,0.086)
(2008.904,0.084)
(2008.923,0.081)
(2008.942,0.082)
(2008.962,0.081)
(2008.981,0.082)
(2009.0,0.084)
(2009.019,0.084)
(2009.038,0.086)
(2009.058,0.086)
(2009.077,0.082)
(2009.096,0.083)
(2009.115,0.081)
(2009.135,0.08)
(2009.154,0.08)
(2009.173,0.081)
(2009.192,0.08)
(2009.212,0.081)
(2009.231,0.081)
(2009.25,0.082)
(2009.269,0.082)
(2009.288,0.082)
(2009.308,0.081)
(2009.327,0.082)
(2009.346,0.082)
(2009.365,0.082)
(2009.385,0.08)
(2009.404,0.08)
(2009.423,0.081)
(2009.442,0.081)
(2009.462,0.08)
(2009.481,0.08)
(2009.5,0.08)
(2009.519,0.08)
(2009.538,0.084)
(2009.558,0.085)
(2009.577,0.085)
(2009.596,0.085)
(2009.615,0.084)
(2009.635,0.083)
(2009.654,0.083)
(2009.673,0.081)
(2009.692,0.08)
(2009.712,0.08)
(2009.731,0.08)
(2009.75,0.079)
(2009.769,0.08)
(2009.788,0.081)
(2009.808,0.081)
(2009.827,0.079)
(2009.846,0.081)
(2009.865,0.08)
(2009.885,0.08)
(2009.904,0.081)
(2009.923,0.081)
(2009.942,0.08)
(2009.962,0.08)
(2009.981,0.079)
(2010.0,0.076)
(2010.019,0.076)
(2010.038,0.075)
(2010.058,0.074)
(2010.077,0.074)
(2010.096,0.073)
(2010.115,0.074)
(2010.135,0.075)
(2010.154,0.075)
(2010.173,0.075)
(2010.192,0.076)
(2010.212,0.075)
(2010.231,0.076)
(2010.25,0.076)
(2010.269,0.078)
(2010.288,0.079)
(2010.308,0.079)
(2010.327,0.078)
(2010.346,0.081)
(2010.365,0.083)
(2010.385,0.084)
(2010.404,0.086)
(2010.423,0.085)
(2010.442,0.086)
(2010.462,0.086)
(2010.481,0.09)
(2010.5,0.09)
(2010.519,0.092)
(2010.538,0.088)
(2010.558,0.089)
(2010.577,0.094)
(2010.596,0.096)
(2010.615,0.096)
(2010.635,0.097)
(2010.654,0.096)
(2010.673,0.097)
(2010.692,0.097)
(2010.712,0.097)
(2010.731,0.097)
(2010.75,0.097)
(2010.769,0.097)
(2010.788,0.097)
(2010.808,0.097)
(2010.827,0.096)
(2010.846,0.096)
(2010.865,0.096)
(2010.885,0.097)
(2010.904,0.098)
(2010.923,0.098)
(2010.942,0.1)
(2010.962,0.1)
(2010.981,0.101)
(2011.0,0.1)
(2011.019,0.1)
(2011.038,0.099)
(2011.058,0.098)
(2011.077,0.099)
(2011.096,0.099)
(2011.115,0.099)
(2011.135,0.098)
(2011.154,0.097)
(2011.173,0.098)
(2011.192,0.097)
(2011.212,0.097)
(2011.231,0.097)
(2011.25,0.098)
(2011.269,0.097)
(2011.288,0.096)
(2011.308,0.093)
(2011.327,0.093)
(2011.346,0.089)
(2011.365,0.089)
(2011.385,0.09)
(2011.404,0.089)
(2011.423,0.09)
(2011.442,0.088)
(2011.462,0.09)
(2011.481,0.087)
(2011.5,0.089)
(2011.519,0.086)
(2011.538,0.085)
(2011.558,0.082)
(2011.577,0.077)
(2011.596,0.075)
(2011.615,0.075)
(2011.635,0.074)
(2011.654,0.074)
(2011.673,0.075)
(2011.692,0.075)
(2011.712,0.075)
(2011.731,0.073)
(2011.75,0.072)
(2011.769,0.072)
(2011.788,0.07)
(2011.808,0.07)
(2011.827,0.07)
(2011.846,0.068)
(2011.865,0.068)
(2011.885,0.068)
(2011.904,0.066)
(2011.923,0.066)
(2011.942,0.066)
(2011.962,0.065)
(2011.981,0.065)
(2012.0,0.066)
(2012.019,0.066)
(2012.038,0.067)
(2012.058,0.066)
(2012.077,0.067)
(2012.096,0.066)
(2012.115,0.066)
(2012.135,0.068)
(2012.154,0.068)
(2012.173,0.069)
(2012.192,0.07)
(2012.212,0.073)
(2012.231,0.078)
(2012.25,0.076)
(2012.269,0.078)
(2012.288,0.078)
(2012.308,0.08)
(2012.327,0.08)
(2012.346,0.08)
(2012.365,0.081)
(2012.385,0.079)
(2012.404,0.08)
(2012.423,0.079)
(2012.442,0.078)
(2012.462,0.075)
(2012.481,0.074)
(2012.5,0.073)
(2012.519,0.075)
(2012.538,0.078)
(2012.558,0.077)
(2012.577,0.079)
(2012.596,0.079)
(2012.615,0.08)
(2012.635,0.079)
(2012.654,0.079)
(2012.673,0.079)
(2012.692,0.08)
(2012.712,0.081)
(2012.731,0.082)
(2012.75,0.082)
(2012.769,0.081)
(2012.788,0.083)
(2012.808,0.084)
(2012.827,0.084)
(2012.846,0.084)
(2012.865,0.086)
(2012.885,0.086)
(2012.904,0.09)
(2012.923,0.089)
(2012.942,0.089)
(2012.962,0.089)
(2012.981,0.09)
(2013.0,0.09)
(2013.019,0.093)
(2013.038,0.094)
(2013.058,0.093)
(2013.077,0.092)
(2013.096,0.092)
(2013.115,0.093)
(2013.135,0.092)
(2013.154,0.091)
(2013.173,0.09)
(2013.192,0.089)
(2013.212,0.087)
(2013.231,0.083)
(2013.25,0.084)
(2013.269,0.082)
(2013.288,0.082)
(2013.308,0.081)
(2013.327,0.081)
(2013.346,0.082)
(2013.365,0.081)
(2013.385,0.081)
(2013.404,0.081)
(2013.423,0.08)
(2013.442,0.081)
(2013.462,0.083)
(2013.481,0.083)
(2013.5,0.082)
(2013.519,0.081)
(2013.538,0.079)
(2013.558,0.081)
(2013.577,0.081)
(2013.596,0.08)
(2013.615,0.079)
(2013.635,0.079)
(2013.654,0.079)
(2013.673,0.078)
(2013.692,0.076)
(2013.712,0.076)
(2013.731,0.077)
(2013.75,0.076)
(2013.769,0.078)
(2013.788,0.077)
(2013.808,0.077)
(2013.827,0.077)
(2013.846,0.078)
(2013.865,0.077)
(2013.885,0.076)
(2013.904,0.073)
(2013.923,0.073)
(2013.942,0.075)
(2013.962,0.075)
(2013.981,0.074)
(2014.0,0.073)
(2014.019,0.071)
(2014.038,0.071)
(2014.058,0.07)
(2014.077,0.07)
(2014.096,0.07)
(2014.115,0.069)
(2014.135,0.069)
(2014.154,0.07)
(2014.173,0.069)
(2014.192,0.068)
(2014.212,0.068)
(2014.231,0.067)
(2014.25,0.068)
(2014.269,0.07)
(2014.288,0.07)
(2014.308,0.071)
(2014.327,0.071)
(2014.346,0.07)
(2014.365,0.07)
(2014.385,0.072)
(2014.404,0.073)
(2014.423,0.078)
(2014.442,0.078)
(2014.462,0.078)
(2014.481,0.078)
(2014.5,0.079)
(2014.519,0.08)
(2014.538,0.082)
(2014.558,0.079)
(2014.577,0.078)
(2014.596,0.078)
(2014.615,0.079)
(2014.635,0.08)
(2014.654,0.08)
(2014.673,0.081)
(2014.692,0.082)
(2014.712,0.081)
(2014.731,0.081)
(2014.75,0.082)
(2014.769,0.083)
(2014.788,0.083)
(2014.808,0.082)
(2014.827,0.082)
(2014.846,0.082)
(2014.865,0.084)
(2014.885,0.085)
(2014.904,0.085)
(2014.923,0.087)
(2014.942,0.086)
(2014.962,0.087)
(2014.981,0.089)
(2015.0,0.089)
(2015.019,0.088)
}

\def\LEVELFIVE{
(2006.0,0.066)
(2006.019,0.07)
(2006.038,0.072)
(2006.058,0.077)
(2006.077,0.079)
(2006.096,0.081)
(2006.115,0.083)
(2006.135,0.085)
(2006.154,0.086)
(2006.173,0.09)
(2006.192,0.091)
(2006.212,0.094)
(2006.231,0.095)
(2006.25,0.098)
(2006.269,0.099)
(2006.288,0.098)
(2006.308,0.098)
(2006.327,0.1)
(2006.346,0.099)
(2006.365,0.1)
(2006.385,0.101)
(2006.404,0.1)
(2006.423,0.101)
(2006.442,0.102)
(2006.462,0.103)
(2006.481,0.103)
(2006.5,0.104)
(2006.519,0.104)
(2006.538,0.104)
(2006.558,0.103)
(2006.577,0.103)
(2006.596,0.103)
(2006.615,0.104)
(2006.635,0.103)
(2006.654,0.101)
(2006.673,0.1)
(2006.692,0.1)
(2006.712,0.1)
(2006.731,0.1)
(2006.75,0.1)
(2006.769,0.1)
(2006.788,0.1)
(2006.808,0.104)
(2006.827,0.104)
(2006.846,0.105)
(2006.865,0.107)
(2006.885,0.108)
(2006.904,0.107)
(2006.923,0.106)
(2006.942,0.1)
(2006.962,0.099)
(2006.981,0.096)
(2007.0,0.095)
(2007.019,0.091)
(2007.038,0.09)
(2007.058,0.087)
(2007.077,0.085)
(2007.096,0.081)
(2007.115,0.08)
(2007.135,0.077)
(2007.154,0.075)
(2007.173,0.071)
(2007.192,0.07)
(2007.212,0.068)
(2007.231,0.067)
(2007.25,0.064)
(2007.269,0.063)
(2007.288,0.063)
(2007.308,0.064)
(2007.327,0.063)
(2007.346,0.063)
(2007.365,0.063)
(2007.385,0.062)
(2007.404,0.063)
(2007.423,0.063)
(2007.442,0.066)
(2007.462,0.067)
(2007.481,0.067)
(2007.5,0.067)
(2007.519,0.068)
(2007.538,0.069)
(2007.558,0.069)
(2007.577,0.07)
(2007.596,0.071)
(2007.615,0.072)
(2007.635,0.073)
(2007.654,0.075)
(2007.673,0.076)
(2007.692,0.076)
(2007.712,0.076)
(2007.731,0.076)
(2007.75,0.076)
(2007.769,0.076)
(2007.788,0.076)
(2007.808,0.073)
(2007.827,0.072)
(2007.846,0.072)
(2007.865,0.069)
(2007.885,0.068)
(2007.904,0.069)
(2007.923,0.07)
(2007.942,0.07)
(2007.962,0.07)
(2007.981,0.07)
(2008.0,0.072)
(2008.019,0.072)
(2008.038,0.072)
(2008.058,0.07)
(2008.077,0.075)
(2008.096,0.076)
(2008.115,0.077)
(2008.135,0.077)
(2008.154,0.076)
(2008.173,0.076)
(2008.192,0.075)
(2008.212,0.074)
(2008.231,0.073)
(2008.25,0.073)
(2008.269,0.072)
(2008.288,0.072)
(2008.308,0.072)
(2008.327,0.072)
(2008.346,0.072)
(2008.365,0.071)
(2008.385,0.072)
(2008.404,0.072)
(2008.423,0.071)
(2008.442,0.068)
(2008.462,0.069)
(2008.481,0.069)
(2008.5,0.068)
(2008.519,0.067)
(2008.538,0.066)
(2008.558,0.066)
(2008.577,0.065)
(2008.596,0.065)
(2008.615,0.065)
(2008.635,0.065)
(2008.654,0.064)
(2008.673,0.065)
(2008.692,0.065)
(2008.712,0.066)
(2008.731,0.066)
(2008.75,0.067)
(2008.769,0.067)
(2008.788,0.066)
(2008.808,0.065)
(2008.827,0.067)
(2008.846,0.064)
(2008.865,0.065)
(2008.885,0.065)
(2008.904,0.063)
(2008.923,0.061)
(2008.942,0.061)
(2008.962,0.061)
(2008.981,0.061)
(2009.0,0.063)
(2009.019,0.064)
(2009.038,0.065)
(2009.058,0.066)
(2009.077,0.061)
(2009.096,0.062)
(2009.115,0.061)
(2009.135,0.06)
(2009.154,0.061)
(2009.173,0.061)
(2009.192,0.061)
(2009.212,0.061)
(2009.231,0.062)
(2009.25,0.062)
(2009.269,0.062)
(2009.288,0.062)
(2009.308,0.062)
(2009.327,0.063)
(2009.346,0.063)
(2009.365,0.062)
(2009.385,0.061)
(2009.404,0.061)
(2009.423,0.062)
(2009.442,0.062)
(2009.462,0.061)
(2009.481,0.061)
(2009.5,0.062)
(2009.519,0.061)
(2009.538,0.065)
(2009.558,0.067)
(2009.577,0.066)
(2009.596,0.066)
(2009.615,0.065)
(2009.635,0.065)
(2009.654,0.065)
(2009.673,0.063)
(2009.692,0.063)
(2009.712,0.062)
(2009.731,0.062)
(2009.75,0.061)
(2009.769,0.062)
(2009.788,0.062)
(2009.808,0.063)
(2009.827,0.06)
(2009.846,0.062)
(2009.865,0.062)
(2009.885,0.061)
(2009.904,0.062)
(2009.923,0.062)
(2009.942,0.061)
(2009.962,0.061)
(2009.981,0.06)
(2010.0,0.058)
(2010.019,0.058)
(2010.038,0.057)
(2010.058,0.056)
(2010.077,0.056)
(2010.096,0.055)
(2010.115,0.056)
(2010.135,0.057)
(2010.154,0.057)
(2010.173,0.056)
(2010.192,0.057)
(2010.212,0.056)
(2010.231,0.057)
(2010.25,0.056)
(2010.269,0.058)
(2010.288,0.059)
(2010.308,0.059)
(2010.327,0.058)
(2010.346,0.061)
(2010.365,0.062)
(2010.385,0.063)
(2010.404,0.064)
(2010.423,0.064)
(2010.442,0.065)
(2010.462,0.065)
(2010.481,0.068)
(2010.5,0.067)
(2010.519,0.07)
(2010.538,0.066)
(2010.558,0.067)
(2010.577,0.071)
(2010.596,0.073)
(2010.615,0.074)
(2010.635,0.074)
(2010.654,0.074)
(2010.673,0.074)
(2010.692,0.074)
(2010.712,0.074)
(2010.731,0.074)
(2010.75,0.074)
(2010.769,0.074)
(2010.788,0.075)
(2010.808,0.075)
(2010.827,0.074)
(2010.846,0.074)
(2010.865,0.074)
(2010.885,0.074)
(2010.904,0.075)
(2010.923,0.076)
(2010.942,0.077)
(2010.962,0.077)
(2010.981,0.077)
(2011.0,0.077)
(2011.019,0.077)
(2011.038,0.076)
(2011.058,0.075)
(2011.077,0.076)
(2011.096,0.076)
(2011.115,0.076)
(2011.135,0.075)
(2011.154,0.074)
(2011.173,0.075)
(2011.192,0.074)
(2011.212,0.074)
(2011.231,0.075)
(2011.25,0.075)
(2011.269,0.075)
(2011.288,0.074)
(2011.308,0.072)
(2011.327,0.071)
(2011.346,0.068)
(2011.365,0.069)
(2011.385,0.07)
(2011.404,0.069)
(2011.423,0.069)
(2011.442,0.068)
(2011.462,0.069)
(2011.481,0.068)
(2011.5,0.069)
(2011.519,0.067)
(2011.538,0.066)
(2011.558,0.063)
(2011.577,0.059)
(2011.596,0.056)
(2011.615,0.056)
(2011.635,0.056)
(2011.654,0.056)
(2011.673,0.056)
(2011.692,0.057)
(2011.712,0.057)
(2011.731,0.055)
(2011.75,0.055)
(2011.769,0.055)
(2011.788,0.053)
(2011.808,0.053)
(2011.827,0.052)
(2011.846,0.051)
(2011.865,0.051)
(2011.885,0.051)
(2011.904,0.049)
(2011.923,0.049)
(2011.942,0.049)
(2011.962,0.049)
(2011.981,0.048)
(2012.0,0.049)
(2012.019,0.049)
(2012.038,0.05)
(2012.058,0.05)
(2012.077,0.05)
(2012.096,0.049)
(2012.115,0.049)
(2012.135,0.051)
(2012.154,0.051)
(2012.173,0.052)
(2012.192,0.053)
(2012.212,0.055)
(2012.231,0.059)
(2012.25,0.058)
(2012.269,0.059)
(2012.288,0.059)
(2012.308,0.061)
(2012.327,0.061)
(2012.346,0.061)
(2012.365,0.062)
(2012.385,0.061)
(2012.404,0.061)
(2012.423,0.061)
(2012.442,0.059)
(2012.462,0.056)
(2012.481,0.056)
(2012.5,0.054)
(2012.519,0.056)
(2012.538,0.058)
(2012.558,0.058)
(2012.577,0.059)
(2012.596,0.059)
(2012.615,0.06)
(2012.635,0.059)
(2012.654,0.059)
(2012.673,0.059)
(2012.692,0.06)
(2012.712,0.06)
(2012.731,0.061)
(2012.75,0.061)
(2012.769,0.061)
(2012.788,0.062)
(2012.808,0.063)
(2012.827,0.063)
(2012.846,0.063)
(2012.865,0.065)
(2012.885,0.065)
(2012.904,0.068)
(2012.923,0.067)
(2012.942,0.068)
(2012.962,0.068)
(2012.981,0.068)
(2013.0,0.069)
(2013.019,0.072)
(2013.038,0.072)
(2013.058,0.071)
(2013.077,0.071)
(2013.096,0.07)
(2013.115,0.071)
(2013.135,0.07)
(2013.154,0.07)
(2013.173,0.069)
(2013.192,0.068)
(2013.212,0.066)
(2013.231,0.062)
(2013.25,0.063)
(2013.269,0.062)
(2013.288,0.062)
(2013.308,0.06)
(2013.327,0.06)
(2013.346,0.061)
(2013.365,0.06)
(2013.385,0.06)
(2013.404,0.06)
(2013.423,0.059)
(2013.442,0.06)
(2013.462,0.061)
(2013.481,0.061)
(2013.5,0.061)
(2013.519,0.06)
(2013.538,0.058)
(2013.558,0.06)
(2013.577,0.059)
(2013.596,0.059)
(2013.615,0.058)
(2013.635,0.058)
(2013.654,0.058)
(2013.673,0.057)
(2013.692,0.056)
(2013.712,0.056)
(2013.731,0.057)
(2013.75,0.056)
(2013.769,0.057)
(2013.788,0.057)
(2013.808,0.057)
(2013.827,0.057)
(2013.846,0.058)
(2013.865,0.056)
(2013.885,0.056)
(2013.904,0.053)
(2013.923,0.053)
(2013.942,0.055)
(2013.962,0.054)
(2013.981,0.054)
(2014.0,0.053)
(2014.019,0.051)
(2014.038,0.051)
(2014.058,0.051)
(2014.077,0.05)
(2014.096,0.05)
(2014.115,0.049)
(2014.135,0.049)
(2014.154,0.05)
(2014.173,0.049)
(2014.192,0.049)
(2014.212,0.049)
(2014.231,0.048)
(2014.25,0.049)
(2014.269,0.051)
(2014.288,0.052)
(2014.308,0.052)
(2014.327,0.052)
(2014.346,0.051)
(2014.365,0.051)
(2014.385,0.053)
(2014.404,0.054)
(2014.423,0.059)
(2014.442,0.059)
(2014.462,0.059)
(2014.481,0.059)
(2014.5,0.06)
(2014.519,0.061)
(2014.538,0.062)
(2014.558,0.06)
(2014.577,0.06)
(2014.596,0.06)
(2014.615,0.06)
(2014.635,0.061)
(2014.654,0.061)
(2014.673,0.061)
(2014.692,0.062)
(2014.712,0.062)
(2014.731,0.061)
(2014.75,0.062)
(2014.769,0.063)
(2014.788,0.063)
(2014.808,0.063)
(2014.827,0.063)
(2014.846,0.063)
(2014.865,0.065)
(2014.885,0.065)
(2014.904,0.066)
(2014.923,0.067)
(2014.942,0.067)
(2014.962,0.067)
(2014.981,0.07)
(2015.0,0.07)
(2015.019,0.069)
}

\def\DIFFERENCEONE{

(2006.0,0.176)
(2006.019,0.179)
(2006.038,0.182)
(2006.058,0.186)
(2006.077,0.185)
(2006.096,0.189)
(2006.115,0.191)
(2006.135,0.184)
(2006.154,0.18)
(2006.173,0.182)
(2006.192,0.18)
(2006.212,0.182)
(2006.231,0.181)
(2006.25,0.182)
(2006.269,0.176)
(2006.288,0.178)
(2006.308,0.181)
(2006.327,0.182)
(2006.346,0.187)
(2006.365,0.185)
(2006.385,0.183)
(2006.404,0.174)
(2006.423,0.171)
(2006.442,0.171)
(2006.462,0.172)
(2006.481,0.17)
(2006.5,0.169)
(2006.519,0.169)
(2006.538,0.173)
(2006.558,0.173)
(2006.577,0.176)
(2006.596,0.173)
(2006.615,0.173)
(2006.635,0.178)
(2006.654,0.176)
(2006.673,0.18)
(2006.692,0.187)
(2006.712,0.188)
(2006.731,0.187)
(2006.75,0.187)
(2006.769,0.194)
(2006.788,0.198)
(2006.808,0.193)
(2006.827,0.194)
(2006.846,0.192)
(2006.865,0.188)
(2006.885,0.189)
(2006.904,0.188)
(2006.923,0.187)
(2006.942,0.187)
(2006.962,0.187)
(2006.981,0.188)
(2007.0,0.185)
(2007.019,0.185)
(2007.038,0.185)
(2007.058,0.183)
(2007.077,0.183)
(2007.096,0.184)
(2007.115,0.183)
(2007.135,0.182)
(2007.154,0.183)
(2007.173,0.181)
(2007.192,0.184)
(2007.212,0.183)
(2007.231,0.184)
(2007.25,0.186)
(2007.269,0.187)
(2007.288,0.189)
(2007.308,0.187)
(2007.327,0.189)
(2007.346,0.186)
(2007.365,0.187)
(2007.385,0.19)
(2007.404,0.195)
(2007.423,0.203)
(2007.442,0.207)
(2007.462,0.207)
(2007.481,0.208)
(2007.5,0.208)
(2007.519,0.203)
(2007.538,0.201)
(2007.558,0.197)
(2007.577,0.199)
(2007.596,0.202)
(2007.615,0.2)
(2007.635,0.197)
(2007.654,0.199)
(2007.673,0.192)
(2007.692,0.19)
(2007.712,0.196)
(2007.731,0.197)
(2007.75,0.197)
(2007.769,0.197)
(2007.788,0.194)
(2007.808,0.198)
(2007.827,0.201)
(2007.846,0.203)
(2007.865,0.205)
(2007.885,0.204)
(2007.904,0.206)
(2007.923,0.209)
(2007.942,0.211)
(2007.962,0.213)
(2007.981,0.214)
(2008.0,0.213)
(2008.019,0.21)
(2008.038,0.212)
(2008.058,0.216)
(2008.077,0.221)
(2008.096,0.219)
(2008.115,0.216)
(2008.135,0.219)
(2008.154,0.219)
(2008.173,0.218)
(2008.192,0.214)
(2008.212,0.215)
(2008.231,0.214)
(2008.25,0.213)
(2008.269,0.215)
(2008.288,0.215)
(2008.308,0.221)
(2008.327,0.22)
(2008.346,0.219)
(2008.365,0.217)
(2008.385,0.215)
(2008.404,0.215)
(2008.423,0.21)
(2008.442,0.209)
(2008.462,0.214)
(2008.481,0.216)
(2008.5,0.214)
(2008.519,0.213)
(2008.538,0.211)
(2008.558,0.209)
(2008.577,0.206)
(2008.596,0.206)
(2008.615,0.21)
(2008.635,0.211)
(2008.654,0.209)
(2008.673,0.217)
(2008.692,0.217)
(2008.712,0.213)
(2008.731,0.21)
(2008.75,0.212)
(2008.769,0.207)
(2008.788,0.208)
(2008.808,0.207)
(2008.827,0.203)
(2008.846,0.201)
(2008.865,0.2)
(2008.885,0.198)
(2008.904,0.193)
(2008.923,0.193)
(2008.942,0.192)
(2008.962,0.19)
(2008.981,0.19)
(2009.0,0.192)
(2009.019,0.195)
(2009.038,0.197)
(2009.058,0.194)
(2009.077,0.193)
(2009.096,0.191)
(2009.115,0.19)
(2009.135,0.189)
(2009.154,0.192)
(2009.173,0.193)
(2009.192,0.194)
(2009.212,0.194)
(2009.231,0.196)
(2009.25,0.198)
(2009.269,0.195)
(2009.288,0.193)
(2009.308,0.192)
(2009.327,0.195)
(2009.346,0.197)
(2009.365,0.196)
(2009.385,0.196)
(2009.404,0.195)
(2009.423,0.197)
(2009.442,0.195)
(2009.462,0.189)
(2009.481,0.186)
(2009.5,0.187)
(2009.519,0.192)
(2009.538,0.193)
(2009.558,0.196)
(2009.577,0.197)
(2009.596,0.195)
(2009.615,0.191)
(2009.635,0.187)
(2009.654,0.191)
(2009.673,0.188)
(2009.692,0.192)
(2009.712,0.192)
(2009.731,0.192)
(2009.75,0.196)
(2009.769,0.196)
(2009.788,0.195)
(2009.808,0.196)
(2009.827,0.198)
(2009.846,0.199)
(2009.865,0.201)
(2009.885,0.204)
(2009.904,0.207)
(2009.923,0.201)
(2009.942,0.198)
(2009.962,0.196)
(2009.981,0.193)
(2010.0,0.191)
(2010.019,0.19)
(2010.038,0.184)
(2010.058,0.177)
(2010.077,0.175)
(2010.096,0.173)
(2010.115,0.174)
(2010.135,0.177)
(2010.154,0.177)
(2010.173,0.176)
(2010.192,0.18)
(2010.212,0.181)
(2010.231,0.179)
(2010.25,0.176)
(2010.269,0.177)
(2010.288,0.171)
(2010.308,0.164)
(2010.327,0.16)
(2010.346,0.161)
(2010.365,0.163)
(2010.385,0.164)
(2010.404,0.163)
(2010.423,0.164)
(2010.442,0.161)
(2010.462,0.162)
(2010.481,0.168)
(2010.5,0.167)
(2010.519,0.165)
(2010.538,0.163)
(2010.558,0.16)
(2010.577,0.161)
(2010.596,0.158)
(2010.615,0.156)
(2010.635,0.159)
(2010.654,0.158)
(2010.673,0.157)
(2010.692,0.155)
(2010.712,0.155)
(2010.731,0.155)
(2010.75,0.147)
(2010.769,0.148)
(2010.788,0.149)
(2010.808,0.147)
(2010.827,0.149)
(2010.846,0.154)
(2010.865,0.155)
(2010.885,0.156)
(2010.904,0.16)
(2010.923,0.164)
(2010.942,0.167)
(2010.962,0.175)
(2010.981,0.174)
(2011.0,0.177)
(2011.019,0.176)
(2011.038,0.182)
(2011.058,0.189)
(2011.077,0.19)
(2011.096,0.191)
(2011.115,0.192)
(2011.135,0.19)
(2011.154,0.186)
(2011.173,0.189)
(2011.192,0.189)
(2011.212,0.189)
(2011.231,0.194)
(2011.25,0.198)
(2011.269,0.198)
(2011.288,0.203)
(2011.308,0.205)
(2011.327,0.206)
(2011.346,0.208)
(2011.365,0.21)
(2011.385,0.212)
(2011.404,0.212)
(2011.423,0.217)
(2011.442,0.224)
(2011.462,0.225)
(2011.481,0.222)
(2011.5,0.224)
(2011.519,0.224)
(2011.538,0.224)
(2011.558,0.224)
(2011.577,0.222)
(2011.596,0.224)
(2011.615,0.228)
(2011.635,0.223)
(2011.654,0.225)
(2011.673,0.225)
(2011.692,0.222)
(2011.712,0.224)
(2011.731,0.229)
(2011.75,0.232)
(2011.769,0.233)
(2011.788,0.233)
(2011.808,0.234)
(2011.827,0.229)
(2011.846,0.224)
(2011.865,0.221)
(2011.885,0.22)
(2011.904,0.212)
(2011.923,0.215)
(2011.942,0.215)
(2011.962,0.212)
(2011.981,0.217)
(2012.0,0.214)
(2012.019,0.21)
(2012.038,0.208)
(2012.058,0.204)
(2012.077,0.202)
(2012.096,0.205)
(2012.115,0.206)
(2012.135,0.207)
(2012.154,0.207)
(2012.173,0.205)
(2012.192,0.204)
(2012.212,0.203)
(2012.231,0.198)
(2012.25,0.197)
(2012.269,0.204)
(2012.288,0.206)
(2012.308,0.209)
(2012.327,0.206)
(2012.346,0.203)
(2012.365,0.201)
(2012.385,0.203)
(2012.404,0.207)
(2012.423,0.2)
(2012.442,0.193)
(2012.462,0.195)
(2012.481,0.193)
(2012.5,0.192)
(2012.519,0.199)
(2012.538,0.204)
(2012.558,0.205)
(2012.577,0.207)
(2012.596,0.207)
(2012.615,0.207)
(2012.635,0.214)
(2012.654,0.213)
(2012.673,0.216)
(2012.692,0.217)
(2012.712,0.215)
(2012.731,0.212)
(2012.75,0.21)
(2012.769,0.21)
(2012.788,0.212)
(2012.808,0.212)
(2012.827,0.211)
(2012.846,0.209)
(2012.865,0.205)
(2012.885,0.203)
(2012.904,0.208)
(2012.923,0.204)
(2012.942,0.204)
(2012.962,0.2)
(2012.981,0.2)
(2013.0,0.206)
(2013.019,0.216)
(2013.038,0.218)
(2013.058,0.219)
(2013.077,0.219)
(2013.096,0.217)
(2013.115,0.218)
(2013.135,0.219)
(2013.154,0.22)
(2013.173,0.221)
(2013.192,0.217)
(2013.212,0.215)
(2013.231,0.219)
(2013.25,0.218)
(2013.269,0.214)
(2013.288,0.212)
(2013.308,0.21)
(2013.327,0.216)
(2013.346,0.219)
(2013.365,0.218)
(2013.385,0.214)
(2013.404,0.212)
(2013.423,0.212)
(2013.442,0.214)
(2013.462,0.214)
(2013.481,0.215)
(2013.5,0.217)
(2013.519,0.212)
(2013.538,0.21)
(2013.558,0.208)
(2013.577,0.206)
(2013.596,0.205)
(2013.615,0.203)
(2013.635,0.199)
(2013.654,0.196)
(2013.673,0.191)
(2013.692,0.19)
(2013.712,0.192)
(2013.731,0.193)
(2013.75,0.198)
(2013.769,0.197)
(2013.788,0.193)
(2013.808,0.192)
(2013.827,0.192)
(2013.846,0.193)
(2013.865,0.2)
(2013.885,0.201)
(2013.904,0.196)
(2013.923,0.2)
(2013.942,0.201)
(2013.962,0.199)
(2013.981,0.197)
(2014.0,0.19)
(2014.019,0.185)
(2014.038,0.179)
(2014.058,0.18)
(2014.077,0.176)
(2014.096,0.175)
(2014.115,0.172)
(2014.135,0.175)
(2014.154,0.173)
(2014.173,0.17)
(2014.192,0.174)
(2014.212,0.176)
(2014.231,0.171)
(2014.25,0.171)
(2014.269,0.171)
(2014.288,0.17)
(2014.308,0.169)
(2014.327,0.166)
(2014.346,0.165)
(2014.365,0.171)
(2014.385,0.175)
(2014.404,0.174)
(2014.423,0.175)
(2014.442,0.174)
(2014.462,0.17)
(2014.481,0.172)
(2014.5,0.172)
(2014.519,0.169)
(2014.538,0.168)
(2014.558,0.17)
(2014.577,0.169)
(2014.596,0.17)
(2014.615,0.174)
(2014.635,0.176)
(2014.654,0.176)
(2014.673,0.177)
(2014.692,0.176)
(2014.712,0.175)
(2014.731,0.172)
(2014.75,0.17)
(2014.769,0.171)
(2014.788,0.172)
(2014.808,0.168)
(2014.827,0.169)
(2014.846,0.168)
(2014.865,0.166)
(2014.885,0.167)
(2014.904,0.168)
(2014.923,0.165)
(2014.942,0.164)
(2014.962,0.166)
(2014.981,0.166)
(2015.0,0.165)
}

\def\DIFFERENCETWO{
(2006.0,0.157)
(2006.019,0.16)
(2006.038,0.161)
(2006.058,0.163)
(2006.077,0.161)
(2006.096,0.164)
(2006.115,0.166)
(2006.135,0.163)
(2006.154,0.16)
(2006.173,0.162)
(2006.192,0.161)
(2006.212,0.16)
(2006.231,0.16)
(2006.25,0.161)
(2006.269,0.159)
(2006.288,0.16)
(2006.308,0.163)
(2006.327,0.163)
(2006.346,0.165)
(2006.365,0.165)
(2006.385,0.164)
(2006.404,0.16)
(2006.423,0.16)
(2006.442,0.162)
(2006.462,0.162)
(2006.481,0.16)
(2006.5,0.159)
(2006.519,0.157)
(2006.538,0.159)
(2006.558,0.158)
(2006.577,0.16)
(2006.596,0.159)
(2006.615,0.159)
(2006.635,0.162)
(2006.654,0.16)
(2006.673,0.164)
(2006.692,0.167)
(2006.712,0.168)
(2006.731,0.169)
(2006.75,0.169)
(2006.769,0.172)
(2006.788,0.175)
(2006.808,0.17)
(2006.827,0.17)
(2006.846,0.168)
(2006.865,0.167)
(2006.885,0.168)
(2006.904,0.167)
(2006.923,0.166)
(2006.942,0.166)
(2006.962,0.167)
(2006.981,0.166)
(2007.0,0.163)
(2007.019,0.161)
(2007.038,0.162)
(2007.058,0.162)
(2007.077,0.162)
(2007.096,0.162)
(2007.115,0.161)
(2007.135,0.162)
(2007.154,0.161)
(2007.173,0.16)
(2007.192,0.163)
(2007.212,0.164)
(2007.231,0.165)
(2007.25,0.165)
(2007.269,0.164)
(2007.288,0.165)
(2007.308,0.163)
(2007.327,0.163)
(2007.346,0.161)
(2007.365,0.161)
(2007.385,0.161)
(2007.404,0.163)
(2007.423,0.167)
(2007.442,0.169)
(2007.462,0.17)
(2007.481,0.171)
(2007.5,0.17)
(2007.519,0.169)
(2007.538,0.168)
(2007.558,0.166)
(2007.577,0.166)
(2007.596,0.168)
(2007.615,0.167)
(2007.635,0.166)
(2007.654,0.166)
(2007.673,0.162)
(2007.692,0.161)
(2007.712,0.163)
(2007.731,0.162)
(2007.75,0.162)
(2007.769,0.16)
(2007.788,0.158)
(2007.808,0.161)
(2007.827,0.164)
(2007.846,0.166)
(2007.865,0.166)
(2007.885,0.166)
(2007.904,0.168)
(2007.923,0.17)
(2007.942,0.171)
(2007.962,0.172)
(2007.981,0.172)
(2008.0,0.173)
(2008.019,0.172)
(2008.038,0.173)
(2008.058,0.175)
(2008.077,0.177)
(2008.096,0.176)
(2008.115,0.175)
(2008.135,0.175)
(2008.154,0.176)
(2008.173,0.175)
(2008.192,0.173)
(2008.212,0.173)
(2008.231,0.173)
(2008.25,0.173)
(2008.269,0.175)
(2008.288,0.176)
(2008.308,0.18)
(2008.327,0.18)
(2008.346,0.181)
(2008.365,0.181)
(2008.385,0.18)
(2008.404,0.179)
(2008.423,0.175)
(2008.442,0.177)
(2008.462,0.179)
(2008.481,0.179)
(2008.5,0.179)
(2008.519,0.177)
(2008.538,0.176)
(2008.558,0.176)
(2008.577,0.176)
(2008.596,0.176)
(2008.615,0.178)
(2008.635,0.179)
(2008.654,0.179)
(2008.673,0.183)
(2008.692,0.184)
(2008.712,0.181)
(2008.731,0.18)
(2008.75,0.18)
(2008.769,0.178)
(2008.788,0.179)
(2008.808,0.178)
(2008.827,0.176)
(2008.846,0.174)
(2008.865,0.174)
(2008.885,0.173)
(2008.904,0.17)
(2008.923,0.169)
(2008.942,0.167)
(2008.962,0.166)
(2008.981,0.166)
(2009.0,0.168)
(2009.019,0.17)
(2009.038,0.173)
(2009.058,0.173)
(2009.077,0.173)
(2009.096,0.174)
(2009.115,0.173)
(2009.135,0.174)
(2009.154,0.176)
(2009.173,0.177)
(2009.192,0.178)
(2009.212,0.178)
(2009.231,0.179)
(2009.25,0.179)
(2009.269,0.177)
(2009.288,0.175)
(2009.308,0.174)
(2009.327,0.176)
(2009.346,0.177)
(2009.365,0.176)
(2009.385,0.177)
(2009.404,0.178)
(2009.423,0.178)
(2009.442,0.175)
(2009.462,0.172)
(2009.481,0.171)
(2009.5,0.172)
(2009.519,0.176)
(2009.538,0.177)
(2009.558,0.179)
(2009.577,0.178)
(2009.596,0.176)
(2009.615,0.174)
(2009.635,0.172)
(2009.654,0.174)
(2009.673,0.171)
(2009.692,0.173)
(2009.712,0.175)
(2009.731,0.176)
(2009.75,0.178)
(2009.769,0.178)
(2009.788,0.177)
(2009.808,0.177)
(2009.827,0.177)
(2009.846,0.177)
(2009.865,0.176)
(2009.885,0.177)
(2009.904,0.179)
(2009.923,0.177)
(2009.942,0.176)
(2009.962,0.175)
(2009.981,0.174)
(2010.0,0.172)
(2010.019,0.171)
(2010.038,0.166)
(2010.058,0.162)
(2010.077,0.161)
(2010.096,0.158)
(2010.115,0.158)
(2010.135,0.159)
(2010.154,0.157)
(2010.173,0.156)
(2010.192,0.158)
(2010.212,0.159)
(2010.231,0.158)
(2010.25,0.157)
(2010.269,0.159)
(2010.288,0.158)
(2010.308,0.155)
(2010.327,0.152)
(2010.346,0.153)
(2010.365,0.153)
(2010.385,0.154)
(2010.404,0.153)
(2010.423,0.155)
(2010.442,0.153)
(2010.462,0.153)
(2010.481,0.156)
(2010.5,0.155)
(2010.519,0.152)
(2010.538,0.15)
(2010.558,0.148)
(2010.577,0.149)
(2010.596,0.148)
(2010.615,0.146)
(2010.635,0.148)
(2010.654,0.147)
(2010.673,0.146)
(2010.692,0.145)
(2010.712,0.143)
(2010.731,0.143)
(2010.75,0.138)
(2010.769,0.139)
(2010.788,0.141)
(2010.808,0.14)
(2010.827,0.143)
(2010.846,0.146)
(2010.865,0.148)
(2010.885,0.149)
(2010.904,0.153)
(2010.923,0.155)
(2010.942,0.156)
(2010.962,0.163)
(2010.981,0.162)
(2011.0,0.163)
(2011.019,0.161)
(2011.038,0.163)
(2011.058,0.167)
(2011.077,0.168)
(2011.096,0.169)
(2011.115,0.168)
(2011.135,0.167)
(2011.154,0.166)
(2011.173,0.167)
(2011.192,0.168)
(2011.212,0.169)
(2011.231,0.172)
(2011.25,0.174)
(2011.269,0.173)
(2011.288,0.174)
(2011.308,0.173)
(2011.327,0.174)
(2011.346,0.174)
(2011.365,0.175)
(2011.385,0.175)
(2011.404,0.175)
(2011.423,0.177)
(2011.442,0.181)
(2011.462,0.182)
(2011.481,0.18)
(2011.5,0.181)
(2011.519,0.18)
(2011.538,0.179)
(2011.558,0.179)
(2011.577,0.178)
(2011.596,0.178)
(2011.615,0.182)
(2011.635,0.18)
(2011.654,0.181)
(2011.673,0.181)
(2011.692,0.18)
(2011.712,0.182)
(2011.731,0.184)
(2011.75,0.185)
(2011.769,0.185)
(2011.788,0.184)
(2011.808,0.183)
(2011.827,0.18)
(2011.846,0.176)
(2011.865,0.174)
(2011.885,0.173)
(2011.904,0.168)
(2011.923,0.17)
(2011.942,0.169)
(2011.962,0.166)
(2011.981,0.168)
(2012.0,0.166)
(2012.019,0.166)
(2012.038,0.166)
(2012.058,0.163)
(2012.077,0.16)
(2012.096,0.161)
(2012.115,0.163)
(2012.135,0.164)
(2012.154,0.164)
(2012.173,0.165)
(2012.192,0.163)
(2012.212,0.16)
(2012.231,0.158)
(2012.25,0.157)
(2012.269,0.161)
(2012.288,0.163)
(2012.308,0.165)
(2012.327,0.164)
(2012.346,0.163)
(2012.365,0.162)
(2012.385,0.163)
(2012.404,0.164)
(2012.423,0.162)
(2012.442,0.159)
(2012.462,0.159)
(2012.481,0.158)
(2012.5,0.158)
(2012.519,0.163)
(2012.538,0.167)
(2012.558,0.169)
(2012.577,0.171)
(2012.596,0.173)
(2012.615,0.172)
(2012.635,0.175)
(2012.654,0.174)
(2012.673,0.175)
(2012.692,0.175)
(2012.712,0.174)
(2012.731,0.172)
(2012.75,0.17)
(2012.769,0.171)
(2012.788,0.173)
(2012.808,0.173)
(2012.827,0.172)
(2012.846,0.173)
(2012.865,0.172)
(2012.885,0.171)
(2012.904,0.174)
(2012.923,0.171)
(2012.942,0.172)
(2012.962,0.171)
(2012.981,0.172)
(2013.0,0.177)
(2013.019,0.181)
(2013.038,0.181)
(2013.058,0.181)
(2013.077,0.181)
(2013.096,0.182)
(2013.115,0.182)
(2013.135,0.183)
(2013.154,0.183)
(2013.173,0.182)
(2013.192,0.179)
(2013.212,0.179)
(2013.231,0.18)
(2013.25,0.18)
(2013.269,0.177)
(2013.288,0.175)
(2013.308,0.174)
(2013.327,0.178)
(2013.346,0.18)
(2013.365,0.179)
(2013.385,0.177)
(2013.404,0.177)
(2013.423,0.176)
(2013.442,0.176)
(2013.462,0.178)
(2013.481,0.179)
(2013.5,0.18)
(2013.519,0.176)
(2013.538,0.174)
(2013.558,0.171)
(2013.577,0.17)
(2013.596,0.169)
(2013.615,0.167)
(2013.635,0.165)
(2013.654,0.165)
(2013.673,0.163)
(2013.692,0.162)
(2013.712,0.163)
(2013.731,0.165)
(2013.75,0.168)
(2013.769,0.167)
(2013.788,0.165)
(2013.808,0.164)
(2013.827,0.165)
(2013.846,0.163)
(2013.865,0.166)
(2013.885,0.166)
(2013.904,0.162)
(2013.923,0.164)
(2013.942,0.165)
(2013.962,0.163)
(2013.981,0.161)
(2014.0,0.156)
(2014.019,0.152)
(2014.038,0.15)
(2014.058,0.151)
(2014.077,0.151)
(2014.096,0.149)
(2014.115,0.147)
(2014.135,0.149)
(2014.154,0.147)
(2014.173,0.146)
(2014.192,0.149)
(2014.212,0.151)
(2014.231,0.148)
(2014.25,0.148)
(2014.269,0.147)
(2014.288,0.147)
(2014.308,0.146)
(2014.327,0.144)
(2014.346,0.142)
(2014.365,0.145)
(2014.385,0.149)
(2014.404,0.148)
(2014.423,0.15)
(2014.442,0.149)
(2014.462,0.146)
(2014.481,0.147)
(2014.5,0.148)
(2014.519,0.147)
(2014.538,0.146)
(2014.558,0.148)
(2014.577,0.147)
(2014.596,0.148)
(2014.615,0.15)
(2014.635,0.151)
(2014.654,0.15)
(2014.673,0.15)
(2014.692,0.149)
(2014.712,0.149)
(2014.731,0.147)
(2014.75,0.145)
(2014.769,0.145)
(2014.788,0.146)
(2014.808,0.144)
(2014.827,0.144)
(2014.846,0.144)
(2014.865,0.143)
(2014.885,0.144)
(2014.904,0.146)
(2014.923,0.144)
(2014.942,0.143)
(2014.962,0.145)
(2014.981,0.145)
(2015.0,0.144)
}

\def\DIFFERENCETHREE{
(2006.0,0.097)
(2006.019,0.1)
(2006.038,0.101)
(2006.058,0.102)
(2006.077,0.101)
(2006.096,0.103)
(2006.115,0.105)
(2006.135,0.103)
(2006.154,0.101)
(2006.173,0.102)
(2006.192,0.101)
(2006.212,0.1)
(2006.231,0.1)
(2006.25,0.101)
(2006.269,0.099)
(2006.288,0.1)
(2006.308,0.102)
(2006.327,0.102)
(2006.346,0.104)
(2006.365,0.103)
(2006.385,0.103)
(2006.404,0.099)
(2006.423,0.099)
(2006.442,0.1)
(2006.462,0.101)
(2006.481,0.099)
(2006.5,0.099)
(2006.519,0.098)
(2006.538,0.099)
(2006.558,0.098)
(2006.577,0.099)
(2006.596,0.099)
(2006.615,0.099)
(2006.635,0.1)
(2006.654,0.099)
(2006.673,0.101)
(2006.692,0.104)
(2006.712,0.105)
(2006.731,0.105)
(2006.75,0.105)
(2006.769,0.108)
(2006.788,0.11)
(2006.808,0.105)
(2006.827,0.105)
(2006.846,0.104)
(2006.865,0.103)
(2006.885,0.103)
(2006.904,0.103)
(2006.923,0.101)
(2006.942,0.101)
(2006.962,0.102)
(2006.981,0.101)
(2007.0,0.097)
(2007.019,0.096)
(2007.038,0.096)
(2007.058,0.096)
(2007.077,0.096)
(2007.096,0.095)
(2007.115,0.094)
(2007.135,0.094)
(2007.154,0.094)
(2007.173,0.093)
(2007.192,0.095)
(2007.212,0.095)
(2007.231,0.097)
(2007.25,0.097)
(2007.269,0.096)
(2007.288,0.097)
(2007.308,0.096)
(2007.327,0.095)
(2007.346,0.094)
(2007.365,0.094)
(2007.385,0.094)
(2007.404,0.096)
(2007.423,0.1)
(2007.442,0.101)
(2007.462,0.103)
(2007.481,0.104)
(2007.5,0.103)
(2007.519,0.102)
(2007.538,0.102)
(2007.558,0.1)
(2007.577,0.1)
(2007.596,0.102)
(2007.615,0.101)
(2007.635,0.1)
(2007.654,0.1)
(2007.673,0.098)
(2007.692,0.096)
(2007.712,0.098)
(2007.731,0.097)
(2007.75,0.098)
(2007.769,0.097)
(2007.788,0.094)
(2007.808,0.097)
(2007.827,0.1)
(2007.846,0.101)
(2007.865,0.102)
(2007.885,0.102)
(2007.904,0.103)
(2007.923,0.105)
(2007.942,0.106)
(2007.962,0.107)
(2007.981,0.107)
(2008.0,0.108)
(2008.019,0.107)
(2008.038,0.107)
(2008.058,0.108)
(2008.077,0.11)
(2008.096,0.11)
(2008.115,0.11)
(2008.135,0.11)
(2008.154,0.111)
(2008.173,0.11)
(2008.192,0.109)
(2008.212,0.109)
(2008.231,0.109)
(2008.25,0.109)
(2008.269,0.11)
(2008.288,0.112)
(2008.308,0.115)
(2008.327,0.116)
(2008.346,0.116)
(2008.365,0.116)
(2008.385,0.116)
(2008.404,0.115)
(2008.423,0.112)
(2008.442,0.114)
(2008.462,0.115)
(2008.481,0.115)
(2008.5,0.114)
(2008.519,0.114)
(2008.538,0.113)
(2008.558,0.113)
(2008.577,0.113)
(2008.596,0.113)
(2008.615,0.115)
(2008.635,0.115)
(2008.654,0.115)
(2008.673,0.119)
(2008.692,0.119)
(2008.712,0.117)
(2008.731,0.116)
(2008.75,0.116)
(2008.769,0.115)
(2008.788,0.115)
(2008.808,0.114)
(2008.827,0.112)
(2008.846,0.111)
(2008.865,0.112)
(2008.885,0.11)
(2008.904,0.108)
(2008.923,0.107)
(2008.942,0.106)
(2008.962,0.104)
(2008.981,0.104)
(2009.0,0.106)
(2009.019,0.109)
(2009.038,0.112)
(2009.058,0.113)
(2009.077,0.113)
(2009.096,0.115)
(2009.115,0.113)
(2009.135,0.114)
(2009.154,0.116)
(2009.173,0.117)
(2009.192,0.118)
(2009.212,0.118)
(2009.231,0.118)
(2009.25,0.119)
(2009.269,0.118)
(2009.288,0.116)
(2009.308,0.114)
(2009.327,0.116)
(2009.346,0.116)
(2009.365,0.116)
(2009.385,0.116)
(2009.404,0.117)
(2009.423,0.117)
(2009.442,0.114)
(2009.462,0.111)
(2009.481,0.11)
(2009.5,0.112)
(2009.519,0.115)
(2009.538,0.116)
(2009.558,0.118)
(2009.577,0.117)
(2009.596,0.116)
(2009.615,0.114)
(2009.635,0.112)
(2009.654,0.113)
(2009.673,0.111)
(2009.692,0.112)
(2009.712,0.114)
(2009.731,0.114)
(2009.75,0.116)
(2009.769,0.116)
(2009.788,0.116)
(2009.808,0.116)
(2009.827,0.116)
(2009.846,0.116)
(2009.865,0.115)
(2009.885,0.115)
(2009.904,0.117)
(2009.923,0.115)
(2009.942,0.114)
(2009.962,0.114)
(2009.981,0.112)
(2010.0,0.111)
(2010.019,0.11)
(2010.038,0.105)
(2010.058,0.101)
(2010.077,0.1)
(2010.096,0.097)
(2010.115,0.098)
(2010.135,0.098)
(2010.154,0.097)
(2010.173,0.096)
(2010.192,0.097)
(2010.212,0.097)
(2010.231,0.096)
(2010.25,0.096)
(2010.269,0.097)
(2010.288,0.095)
(2010.308,0.093)
(2010.327,0.09)
(2010.346,0.091)
(2010.365,0.091)
(2010.385,0.091)
(2010.404,0.091)
(2010.423,0.092)
(2010.442,0.09)
(2010.462,0.09)
(2010.481,0.093)
(2010.5,0.092)
(2010.519,0.089)
(2010.538,0.088)
(2010.558,0.086)
(2010.577,0.087)
(2010.596,0.086)
(2010.615,0.085)
(2010.635,0.086)
(2010.654,0.085)
(2010.673,0.085)
(2010.692,0.084)
(2010.712,0.083)
(2010.731,0.083)
(2010.75,0.079)
(2010.769,0.08)
(2010.788,0.082)
(2010.808,0.082)
(2010.827,0.083)
(2010.846,0.086)
(2010.865,0.087)
(2010.885,0.089)
(2010.904,0.093)
(2010.923,0.095)
(2010.942,0.096)
(2010.962,0.101)
(2010.981,0.101)
(2011.0,0.102)
(2011.019,0.101)
(2011.038,0.102)
(2011.058,0.105)
(2011.077,0.106)
(2011.096,0.107)
(2011.115,0.106)
(2011.135,0.105)
(2011.154,0.104)
(2011.173,0.105)
(2011.192,0.106)
(2011.212,0.107)
(2011.231,0.11)
(2011.25,0.112)
(2011.269,0.111)
(2011.288,0.112)
(2011.308,0.112)
(2011.327,0.113)
(2011.346,0.113)
(2011.365,0.113)
(2011.385,0.114)
(2011.404,0.114)
(2011.423,0.115)
(2011.442,0.119)
(2011.462,0.119)
(2011.481,0.117)
(2011.5,0.118)
(2011.519,0.118)
(2011.538,0.117)
(2011.558,0.117)
(2011.577,0.116)
(2011.596,0.115)
(2011.615,0.118)
(2011.635,0.117)
(2011.654,0.118)
(2011.673,0.118)
(2011.692,0.118)
(2011.712,0.119)
(2011.731,0.12)
(2011.75,0.12)
(2011.769,0.12)
(2011.788,0.119)
(2011.808,0.118)
(2011.827,0.116)
(2011.846,0.113)
(2011.865,0.111)
(2011.885,0.11)
(2011.904,0.105)
(2011.923,0.107)
(2011.942,0.106)
(2011.962,0.102)
(2011.981,0.103)
(2012.0,0.101)
(2012.019,0.101)
(2012.038,0.101)
(2012.058,0.099)
(2012.077,0.097)
(2012.096,0.098)
(2012.115,0.099)
(2012.135,0.1)
(2012.154,0.101)
(2012.173,0.101)
(2012.192,0.1)
(2012.212,0.098)
(2012.231,0.096)
(2012.25,0.095)
(2012.269,0.098)
(2012.288,0.099)
(2012.308,0.102)
(2012.327,0.1)
(2012.346,0.1)
(2012.365,0.099)
(2012.385,0.099)
(2012.404,0.1)
(2012.423,0.099)
(2012.442,0.095)
(2012.462,0.095)
(2012.481,0.094)
(2012.5,0.094)
(2012.519,0.099)
(2012.538,0.102)
(2012.558,0.104)
(2012.577,0.105)
(2012.596,0.106)
(2012.615,0.105)
(2012.635,0.107)
(2012.654,0.107)
(2012.673,0.108)
(2012.692,0.107)
(2012.712,0.106)
(2012.731,0.105)
(2012.75,0.104)
(2012.769,0.105)
(2012.788,0.106)
(2012.808,0.106)
(2012.827,0.106)
(2012.846,0.106)
(2012.865,0.105)
(2012.885,0.104)
(2012.904,0.107)
(2012.923,0.105)
(2012.942,0.106)
(2012.962,0.105)
(2012.981,0.106)
(2013.0,0.11)
(2013.019,0.114)
(2013.038,0.114)
(2013.058,0.113)
(2013.077,0.114)
(2013.096,0.115)
(2013.115,0.115)
(2013.135,0.115)
(2013.154,0.115)
(2013.173,0.114)
(2013.192,0.113)
(2013.212,0.112)
(2013.231,0.113)
(2013.25,0.113)
(2013.269,0.111)
(2013.288,0.109)
(2013.308,0.108)
(2013.327,0.112)
(2013.346,0.113)
(2013.365,0.113)
(2013.385,0.112)
(2013.404,0.112)
(2013.423,0.111)
(2013.442,0.112)
(2013.462,0.114)
(2013.481,0.115)
(2013.5,0.115)
(2013.519,0.112)
(2013.538,0.11)
(2013.558,0.108)
(2013.577,0.107)
(2013.596,0.107)
(2013.615,0.105)
(2013.635,0.104)
(2013.654,0.104)
(2013.673,0.102)
(2013.692,0.102)
(2013.712,0.102)
(2013.731,0.104)
(2013.75,0.106)
(2013.769,0.105)
(2013.788,0.103)
(2013.808,0.103)
(2013.827,0.103)
(2013.846,0.102)
(2013.865,0.104)
(2013.885,0.105)
(2013.904,0.102)
(2013.923,0.104)
(2013.942,0.104)
(2013.962,0.102)
(2013.981,0.101)
(2014.0,0.097)
(2014.019,0.094)
(2014.038,0.092)
(2014.058,0.093)
(2014.077,0.092)
(2014.096,0.09)
(2014.115,0.088)
(2014.135,0.09)
(2014.154,0.089)
(2014.173,0.088)
(2014.192,0.09)
(2014.212,0.091)
(2014.231,0.089)
(2014.25,0.089)
(2014.269,0.088)
(2014.288,0.088)
(2014.308,0.087)
(2014.327,0.084)
(2014.346,0.083)
(2014.365,0.086)
(2014.385,0.088)
(2014.404,0.088)
(2014.423,0.09)
(2014.442,0.089)
(2014.462,0.087)
(2014.481,0.088)
(2014.5,0.088)
(2014.519,0.088)
(2014.538,0.087)
(2014.558,0.088)
(2014.577,0.088)
(2014.596,0.088)
(2014.615,0.089)
(2014.635,0.09)
(2014.654,0.09)
(2014.673,0.09)
(2014.692,0.089)
(2014.712,0.089)
(2014.731,0.088)
(2014.75,0.087)
(2014.769,0.087)
(2014.788,0.087)
(2014.808,0.086)
(2014.827,0.086)
(2014.846,0.086)
(2014.865,0.086)
(2014.885,0.086)
(2014.904,0.087)
(2014.923,0.086)
(2014.942,0.085)
(2014.962,0.087)
(2014.981,0.087)
(2015.0,0.086)
}

\def\DIFFERENCEFOUR{
(2006.0,0.075)
(2006.019,0.077)
(2006.038,0.078)
(2006.058,0.079)
(2006.077,0.078)
(2006.096,0.08)
(2006.115,0.082)
(2006.135,0.08)
(2006.154,0.079)
(2006.173,0.08)
(2006.192,0.079)
(2006.212,0.078)
(2006.231,0.078)
(2006.25,0.079)
(2006.269,0.077)
(2006.288,0.078)
(2006.308,0.079)
(2006.327,0.08)
(2006.346,0.081)
(2006.365,0.08)
(2006.385,0.08)
(2006.404,0.078)
(2006.423,0.078)
(2006.442,0.079)
(2006.462,0.079)
(2006.481,0.078)
(2006.5,0.078)
(2006.519,0.077)
(2006.538,0.078)
(2006.558,0.077)
(2006.577,0.078)
(2006.596,0.078)
(2006.615,0.077)
(2006.635,0.078)
(2006.654,0.077)
(2006.673,0.078)
(2006.692,0.081)
(2006.712,0.081)
(2006.731,0.082)
(2006.75,0.081)
(2006.769,0.083)
(2006.788,0.085)
(2006.808,0.081)
(2006.827,0.081)
(2006.846,0.08)
(2006.865,0.079)
(2006.885,0.079)
(2006.904,0.079)
(2006.923,0.077)
(2006.942,0.078)
(2006.962,0.078)
(2006.981,0.077)
(2007.0,0.073)
(2007.019,0.072)
(2007.038,0.072)
(2007.058,0.072)
(2007.077,0.072)
(2007.096,0.071)
(2007.115,0.069)
(2007.135,0.07)
(2007.154,0.07)
(2007.173,0.069)
(2007.192,0.07)
(2007.212,0.071)
(2007.231,0.072)
(2007.25,0.072)
(2007.269,0.071)
(2007.288,0.072)
(2007.308,0.071)
(2007.327,0.07)
(2007.346,0.07)
(2007.365,0.07)
(2007.385,0.069)
(2007.404,0.07)
(2007.423,0.074)
(2007.442,0.075)
(2007.462,0.077)
(2007.481,0.078)
(2007.5,0.077)
(2007.519,0.076)
(2007.538,0.076)
(2007.558,0.075)
(2007.577,0.075)
(2007.596,0.076)
(2007.615,0.076)
(2007.635,0.075)
(2007.654,0.076)
(2007.673,0.074)
(2007.692,0.072)
(2007.712,0.074)
(2007.731,0.073)
(2007.75,0.073)
(2007.769,0.073)
(2007.788,0.071)
(2007.808,0.072)
(2007.827,0.075)
(2007.846,0.076)
(2007.865,0.076)
(2007.885,0.076)
(2007.904,0.077)
(2007.923,0.079)
(2007.942,0.079)
(2007.962,0.08)
(2007.981,0.081)
(2008.0,0.081)
(2008.019,0.08)
(2008.038,0.081)
(2008.058,0.081)
(2008.077,0.083)
(2008.096,0.082)
(2008.115,0.083)
(2008.135,0.083)
(2008.154,0.083)
(2008.173,0.083)
(2008.192,0.082)
(2008.212,0.082)
(2008.231,0.082)
(2008.25,0.082)
(2008.269,0.082)
(2008.288,0.084)
(2008.308,0.087)
(2008.327,0.087)
(2008.346,0.088)
(2008.365,0.088)
(2008.385,0.088)
(2008.404,0.087)
(2008.423,0.085)
(2008.442,0.087)
(2008.462,0.088)
(2008.481,0.088)
(2008.5,0.087)
(2008.519,0.087)
(2008.538,0.086)
(2008.558,0.086)
(2008.577,0.087)
(2008.596,0.086)
(2008.615,0.088)
(2008.635,0.088)
(2008.654,0.088)
(2008.673,0.091)
(2008.692,0.092)
(2008.712,0.09)
(2008.731,0.089)
(2008.75,0.089)
(2008.769,0.088)
(2008.788,0.088)
(2008.808,0.088)
(2008.827,0.086)
(2008.846,0.086)
(2008.865,0.086)
(2008.885,0.085)
(2008.904,0.084)
(2008.923,0.083)
(2008.942,0.081)
(2008.962,0.08)
(2008.981,0.08)
(2009.0,0.082)
(2009.019,0.085)
(2009.038,0.088)
(2009.058,0.089)
(2009.077,0.09)
(2009.096,0.092)
(2009.115,0.091)
(2009.135,0.091)
(2009.154,0.093)
(2009.173,0.094)
(2009.192,0.095)
(2009.212,0.095)
(2009.231,0.095)
(2009.25,0.096)
(2009.269,0.095)
(2009.288,0.093)
(2009.308,0.091)
(2009.327,0.092)
(2009.346,0.092)
(2009.365,0.092)
(2009.385,0.093)
(2009.404,0.094)
(2009.423,0.093)
(2009.442,0.09)
(2009.462,0.088)
(2009.481,0.087)
(2009.5,0.088)
(2009.519,0.091)
(2009.538,0.092)
(2009.558,0.094)
(2009.577,0.094)
(2009.596,0.092)
(2009.615,0.091)
(2009.635,0.09)
(2009.654,0.09)
(2009.673,0.088)
(2009.692,0.089)
(2009.712,0.091)
(2009.731,0.091)
(2009.75,0.093)
(2009.769,0.092)
(2009.788,0.092)
(2009.808,0.092)
(2009.827,0.092)
(2009.846,0.092)
(2009.865,0.09)
(2009.885,0.091)
(2009.904,0.091)
(2009.923,0.091)
(2009.942,0.09)
(2009.962,0.09)
(2009.981,0.089)
(2010.0,0.088)
(2010.019,0.086)
(2010.038,0.082)
(2010.058,0.079)
(2010.077,0.078)
(2010.096,0.075)
(2010.115,0.076)
(2010.135,0.076)
(2010.154,0.074)
(2010.173,0.073)
(2010.192,0.074)
(2010.212,0.074)
(2010.231,0.073)
(2010.25,0.073)
(2010.269,0.074)
(2010.288,0.073)
(2010.308,0.072)
(2010.327,0.07)
(2010.346,0.07)
(2010.365,0.07)
(2010.385,0.07)
(2010.404,0.07)
(2010.423,0.071)
(2010.442,0.069)
(2010.462,0.07)
(2010.481,0.072)
(2010.5,0.071)
(2010.519,0.068)
(2010.538,0.067)
(2010.558,0.065)
(2010.577,0.066)
(2010.596,0.066)
(2010.615,0.065)
(2010.635,0.065)
(2010.654,0.064)
(2010.673,0.064)
(2010.692,0.064)
(2010.712,0.063)
(2010.731,0.063)
(2010.75,0.06)
(2010.769,0.061)
(2010.788,0.062)
(2010.808,0.062)
(2010.827,0.063)
(2010.846,0.065)
(2010.865,0.066)
(2010.885,0.068)
(2010.904,0.072)
(2010.923,0.073)
(2010.942,0.074)
(2010.962,0.079)
(2010.981,0.078)
(2011.0,0.079)
(2011.019,0.079)
(2011.038,0.079)
(2011.058,0.082)
(2011.077,0.082)
(2011.096,0.083)
(2011.115,0.083)
(2011.135,0.082)
(2011.154,0.082)
(2011.173,0.082)
(2011.192,0.083)
(2011.212,0.084)
(2011.231,0.087)
(2011.25,0.088)
(2011.269,0.087)
(2011.288,0.088)
(2011.308,0.088)
(2011.327,0.088)
(2011.346,0.088)
(2011.365,0.088)
(2011.385,0.089)
(2011.404,0.089)
(2011.423,0.089)
(2011.442,0.092)
(2011.462,0.092)
(2011.481,0.09)
(2011.5,0.091)
(2011.519,0.091)
(2011.538,0.09)
(2011.558,0.09)
(2011.577,0.089)
(2011.596,0.088)
(2011.615,0.091)
(2011.635,0.09)
(2011.654,0.09)
(2011.673,0.091)
(2011.692,0.091)
(2011.712,0.092)
(2011.731,0.092)
(2011.75,0.093)
(2011.769,0.092)
(2011.788,0.09)
(2011.808,0.09)
(2011.827,0.088)
(2011.846,0.086)
(2011.865,0.084)
(2011.885,0.083)
(2011.904,0.078)
(2011.923,0.08)
(2011.942,0.079)
(2011.962,0.075)
(2011.981,0.076)
(2012.0,0.075)
(2012.019,0.075)
(2012.038,0.075)
(2012.058,0.074)
(2012.077,0.072)
(2012.096,0.072)
(2012.115,0.073)
(2012.135,0.074)
(2012.154,0.075)
(2012.173,0.076)
(2012.192,0.075)
(2012.212,0.072)
(2012.231,0.071)
(2012.25,0.071)
(2012.269,0.073)
(2012.288,0.074)
(2012.308,0.076)
(2012.327,0.075)
(2012.346,0.074)
(2012.365,0.074)
(2012.385,0.074)
(2012.404,0.074)
(2012.423,0.074)
(2012.442,0.071)
(2012.462,0.071)
(2012.481,0.07)
(2012.5,0.07)
(2012.519,0.074)
(2012.538,0.077)
(2012.558,0.078)
(2012.577,0.079)
(2012.596,0.08)
(2012.615,0.079)
(2012.635,0.08)
(2012.654,0.08)
(2012.673,0.081)
(2012.692,0.08)
(2012.712,0.079)
(2012.731,0.078)
(2012.75,0.078)
(2012.769,0.079)
(2012.788,0.08)
(2012.808,0.08)
(2012.827,0.08)
(2012.846,0.08)
(2012.865,0.08)
(2012.885,0.079)
(2012.904,0.082)
(2012.923,0.08)
(2012.942,0.081)
(2012.962,0.08)
(2012.981,0.081)
(2013.0,0.084)
(2013.019,0.087)
(2013.038,0.087)
(2013.058,0.086)
(2013.077,0.087)
(2013.096,0.088)
(2013.115,0.088)
(2013.135,0.088)
(2013.154,0.088)
(2013.173,0.087)
(2013.192,0.086)
(2013.212,0.086)
(2013.231,0.086)
(2013.25,0.085)
(2013.269,0.084)
(2013.288,0.083)
(2013.308,0.082)
(2013.327,0.084)
(2013.346,0.086)
(2013.365,0.086)
(2013.385,0.085)
(2013.404,0.085)
(2013.423,0.084)
(2013.442,0.085)
(2013.462,0.087)
(2013.481,0.088)
(2013.5,0.089)
(2013.519,0.085)
(2013.538,0.083)
(2013.558,0.082)
(2013.577,0.082)
(2013.596,0.082)
(2013.615,0.08)
(2013.635,0.08)
(2013.654,0.08)
(2013.673,0.079)
(2013.692,0.078)
(2013.712,0.079)
(2013.731,0.08)
(2013.75,0.081)
(2013.769,0.081)
(2013.788,0.079)
(2013.808,0.079)
(2013.827,0.079)
(2013.846,0.078)
(2013.865,0.08)
(2013.885,0.08)
(2013.904,0.077)
(2013.923,0.079)
(2013.942,0.079)
(2013.962,0.078)
(2013.981,0.077)
(2014.0,0.074)
(2014.019,0.071)
(2014.038,0.07)
(2014.058,0.07)
(2014.077,0.07)
(2014.096,0.068)
(2014.115,0.066)
(2014.135,0.068)
(2014.154,0.067)
(2014.173,0.066)
(2014.192,0.068)
(2014.212,0.069)
(2014.231,0.067)
(2014.25,0.067)
(2014.269,0.067)
(2014.288,0.066)
(2014.308,0.065)
(2014.327,0.063)
(2014.346,0.062)
(2014.365,0.064)
(2014.385,0.067)
(2014.404,0.066)
(2014.423,0.068)
(2014.442,0.068)
(2014.462,0.066)
(2014.481,0.066)
(2014.5,0.066)
(2014.519,0.066)
(2014.538,0.065)
(2014.558,0.066)
(2014.577,0.066)
(2014.596,0.066)
(2014.615,0.067)
(2014.635,0.067)
(2014.654,0.067)
(2014.673,0.067)
(2014.692,0.066)
(2014.712,0.066)
(2014.731,0.065)
(2014.75,0.065)
(2014.769,0.065)
(2014.788,0.065)
(2014.808,0.064)
(2014.827,0.064)
(2014.846,0.064)
(2014.865,0.064)
(2014.885,0.064)
(2014.904,0.065)
(2014.923,0.064)
(2014.942,0.064)
(2014.962,0.065)
(2014.981,0.065)
(2015.0,0.064)
}

\def\DIFFERENCEFIVE{
(2006.0,0.057)
(2006.019,0.059)
(2006.038,0.06)
(2006.058,0.061)
(2006.077,0.06)
(2006.096,0.061)
(2006.115,0.063)
(2006.135,0.062)
(2006.154,0.061)
(2006.173,0.062)
(2006.192,0.061)
(2006.212,0.061)
(2006.231,0.061)
(2006.25,0.061)
(2006.269,0.06)
(2006.288,0.061)
(2006.308,0.062)
(2006.327,0.062)
(2006.346,0.063)
(2006.365,0.062)
(2006.385,0.062)
(2006.404,0.06)
(2006.423,0.06)
(2006.442,0.061)
(2006.462,0.061)
(2006.481,0.061)
(2006.5,0.06)
(2006.519,0.06)
(2006.538,0.06)
(2006.558,0.06)
(2006.577,0.06)
(2006.596,0.061)
(2006.615,0.06)
(2006.635,0.061)
(2006.654,0.059)
(2006.673,0.06)
(2006.692,0.062)
(2006.712,0.063)
(2006.731,0.063)
(2006.75,0.063)
(2006.769,0.064)
(2006.788,0.065)
(2006.808,0.062)
(2006.827,0.062)
(2006.846,0.061)
(2006.865,0.06)
(2006.885,0.06)
(2006.904,0.06)
(2006.923,0.059)
(2006.942,0.059)
(2006.962,0.059)
(2006.981,0.058)
(2007.0,0.055)
(2007.019,0.053)
(2007.038,0.053)
(2007.058,0.054)
(2007.077,0.053)
(2007.096,0.052)
(2007.115,0.05)
(2007.135,0.05)
(2007.154,0.05)
(2007.173,0.05)
(2007.192,0.051)
(2007.212,0.051)
(2007.231,0.052)
(2007.25,0.052)
(2007.269,0.052)
(2007.288,0.053)
(2007.308,0.052)
(2007.327,0.051)
(2007.346,0.051)
(2007.365,0.051)
(2007.385,0.05)
(2007.404,0.051)
(2007.423,0.054)
(2007.442,0.055)
(2007.462,0.057)
(2007.481,0.058)
(2007.5,0.057)
(2007.519,0.057)
(2007.538,0.057)
(2007.558,0.056)
(2007.577,0.055)
(2007.596,0.056)
(2007.615,0.056)
(2007.635,0.056)
(2007.654,0.057)
(2007.673,0.055)
(2007.692,0.054)
(2007.712,0.055)
(2007.731,0.054)
(2007.75,0.055)
(2007.769,0.055)
(2007.788,0.053)
(2007.808,0.054)
(2007.827,0.056)
(2007.846,0.057)
(2007.865,0.057)
(2007.885,0.057)
(2007.904,0.058)
(2007.923,0.059)
(2007.942,0.06)
(2007.962,0.06)
(2007.981,0.061)
(2008.0,0.061)
(2008.019,0.061)
(2008.038,0.061)
(2008.058,0.061)
(2008.077,0.062)
(2008.096,0.062)
(2008.115,0.063)
(2008.135,0.063)
(2008.154,0.063)
(2008.173,0.063)
(2008.192,0.062)
(2008.212,0.062)
(2008.231,0.062)
(2008.25,0.062)
(2008.269,0.062)
(2008.288,0.064)
(2008.308,0.066)
(2008.327,0.067)
(2008.346,0.067)
(2008.365,0.068)
(2008.385,0.067)
(2008.404,0.067)
(2008.423,0.065)
(2008.442,0.067)
(2008.462,0.068)
(2008.481,0.067)
(2008.5,0.067)
(2008.519,0.067)
(2008.538,0.066)
(2008.558,0.066)
(2008.577,0.067)
(2008.596,0.066)
(2008.615,0.067)
(2008.635,0.067)
(2008.654,0.068)
(2008.673,0.07)
(2008.692,0.071)
(2008.712,0.07)
(2008.731,0.069)
(2008.75,0.069)
(2008.769,0.068)
(2008.788,0.068)
(2008.808,0.068)
(2008.827,0.067)
(2008.846,0.066)
(2008.865,0.067)
(2008.885,0.066)
(2008.904,0.065)
(2008.923,0.064)
(2008.942,0.063)
(2008.962,0.062)
(2008.981,0.062)
(2009.0,0.064)
(2009.019,0.066)
(2009.038,0.069)
(2009.058,0.071)
(2009.077,0.072)
(2009.096,0.074)
(2009.115,0.073)
(2009.135,0.073)
(2009.154,0.075)
(2009.173,0.076)
(2009.192,0.076)
(2009.212,0.076)
(2009.231,0.077)
(2009.25,0.077)
(2009.269,0.076)
(2009.288,0.074)
(2009.308,0.072)
(2009.327,0.074)
(2009.346,0.074)
(2009.365,0.074)
(2009.385,0.074)
(2009.404,0.075)
(2009.423,0.075)
(2009.442,0.072)
(2009.462,0.07)
(2009.481,0.069)
(2009.5,0.07)
(2009.519,0.073)
(2009.538,0.074)
(2009.558,0.076)
(2009.577,0.075)
(2009.596,0.074)
(2009.615,0.073)
(2009.635,0.072)
(2009.654,0.072)
(2009.673,0.071)
(2009.692,0.071)
(2009.712,0.073)
(2009.731,0.073)
(2009.75,0.074)
(2009.769,0.074)
(2009.788,0.074)
(2009.808,0.074)
(2009.827,0.074)
(2009.846,0.073)
(2009.865,0.072)
(2009.885,0.072)
(2009.904,0.073)
(2009.923,0.072)
(2009.942,0.072)
(2009.962,0.072)
(2009.981,0.071)
(2010.0,0.069)
(2010.019,0.068)
(2010.038,0.064)
(2010.058,0.061)
(2010.077,0.06)
(2010.096,0.058)
(2010.115,0.058)
(2010.135,0.058)
(2010.154,0.057)
(2010.173,0.056)
(2010.192,0.056)
(2010.212,0.056)
(2010.231,0.056)
(2010.25,0.056)
(2010.269,0.056)
(2010.288,0.055)
(2010.308,0.055)
(2010.327,0.053)
(2010.346,0.053)
(2010.365,0.053)
(2010.385,0.053)
(2010.404,0.053)
(2010.423,0.053)
(2010.442,0.052)
(2010.462,0.052)
(2010.481,0.054)
(2010.5,0.053)
(2010.519,0.05)
(2010.538,0.05)
(2010.558,0.048)
(2010.577,0.049)
(2010.596,0.049)
(2010.615,0.048)
(2010.635,0.048)
(2010.654,0.048)
(2010.673,0.048)
(2010.692,0.048)
(2010.712,0.046)
(2010.731,0.046)
(2010.75,0.044)
(2010.769,0.045)
(2010.788,0.047)
(2010.808,0.046)
(2010.827,0.047)
(2010.846,0.049)
(2010.865,0.05)
(2010.885,0.052)
(2010.904,0.055)
(2010.923,0.056)
(2010.942,0.057)
(2010.962,0.061)
(2010.981,0.061)
(2011.0,0.062)
(2011.019,0.061)
(2011.038,0.061)
(2011.058,0.063)
(2011.077,0.064)
(2011.096,0.065)
(2011.115,0.064)
(2011.135,0.064)
(2011.154,0.064)
(2011.173,0.064)
(2011.192,0.065)
(2011.212,0.066)
(2011.231,0.069)
(2011.25,0.069)
(2011.269,0.069)
(2011.288,0.069)
(2011.308,0.069)
(2011.327,0.07)
(2011.346,0.069)
(2011.365,0.069)
(2011.385,0.07)
(2011.404,0.07)
(2011.423,0.07)
(2011.442,0.072)
(2011.462,0.072)
(2011.481,0.07)
(2011.5,0.071)
(2011.519,0.071)
(2011.538,0.07)
(2011.558,0.07)
(2011.577,0.069)
(2011.596,0.068)
(2011.615,0.071)
(2011.635,0.07)
(2011.654,0.071)
(2011.673,0.071)
(2011.692,0.072)
(2011.712,0.072)
(2011.731,0.072)
(2011.75,0.072)
(2011.769,0.071)
(2011.788,0.07)
(2011.808,0.069)
(2011.827,0.068)
(2011.846,0.066)
(2011.865,0.065)
(2011.885,0.064)
(2011.904,0.06)
(2011.923,0.061)
(2011.942,0.06)
(2011.962,0.057)
(2011.981,0.057)
(2012.0,0.056)
(2012.019,0.056)
(2012.038,0.057)
(2012.058,0.055)
(2012.077,0.053)
(2012.096,0.053)
(2012.115,0.055)
(2012.135,0.056)
(2012.154,0.056)
(2012.173,0.057)
(2012.192,0.056)
(2012.212,0.054)
(2012.231,0.053)
(2012.25,0.053)
(2012.269,0.055)
(2012.288,0.055)
(2012.308,0.057)
(2012.327,0.056)
(2012.346,0.056)
(2012.365,0.056)
(2012.385,0.056)
(2012.404,0.056)
(2012.423,0.056)
(2012.442,0.054)
(2012.462,0.053)
(2012.481,0.053)
(2012.5,0.053)
(2012.519,0.056)
(2012.538,0.059)
(2012.558,0.059)
(2012.577,0.06)
(2012.596,0.061)
(2012.615,0.06)
(2012.635,0.061)
(2012.654,0.061)
(2012.673,0.061)
(2012.692,0.061)
(2012.712,0.059)
(2012.731,0.059)
(2012.75,0.059)
(2012.769,0.059)
(2012.788,0.061)
(2012.808,0.061)
(2012.827,0.06)
(2012.846,0.061)
(2012.865,0.06)
(2012.885,0.06)
(2012.904,0.062)
(2012.923,0.061)
(2012.942,0.062)
(2012.962,0.061)
(2012.981,0.062)
(2013.0,0.065)
(2013.019,0.067)
(2013.038,0.067)
(2013.058,0.066)
(2013.077,0.066)
(2013.096,0.068)
(2013.115,0.068)
(2013.135,0.068)
(2013.154,0.068)
(2013.173,0.067)
(2013.192,0.066)
(2013.212,0.066)
(2013.231,0.065)
(2013.25,0.065)
(2013.269,0.064)
(2013.288,0.063)
(2013.308,0.062)
(2013.327,0.064)
(2013.346,0.065)
(2013.365,0.065)
(2013.385,0.065)
(2013.404,0.065)
(2013.423,0.064)
(2013.442,0.065)
(2013.462,0.067)
(2013.481,0.068)
(2013.5,0.068)
(2013.519,0.065)
(2013.538,0.064)
(2013.558,0.063)
(2013.577,0.062)
(2013.596,0.062)
(2013.615,0.062)
(2013.635,0.061)
(2013.654,0.061)
(2013.673,0.06)
(2013.692,0.06)
(2013.712,0.06)
(2013.731,0.061)
(2013.75,0.062)
(2013.769,0.062)
(2013.788,0.061)
(2013.808,0.061)
(2013.827,0.061)
(2013.846,0.06)
(2013.865,0.061)
(2013.885,0.061)
(2013.904,0.059)
(2013.923,0.06)
(2013.942,0.06)
(2013.962,0.059)
(2013.981,0.059)
(2014.0,0.056)
(2014.019,0.054)
(2014.038,0.053)
(2014.058,0.053)
(2014.077,0.053)
(2014.096,0.051)
(2014.115,0.05)
(2014.135,0.051)
(2014.154,0.05)
(2014.173,0.049)
(2014.192,0.05)
(2014.212,0.051)
(2014.231,0.05)
(2014.25,0.05)
(2014.269,0.05)
(2014.288,0.049)
(2014.308,0.049)
(2014.327,0.047)
(2014.346,0.046)
(2014.365,0.048)
(2014.385,0.05)
(2014.404,0.049)
(2014.423,0.051)
(2014.442,0.051)
(2014.462,0.049)
(2014.481,0.049)
(2014.5,0.049)
(2014.519,0.049)
(2014.538,0.048)
(2014.558,0.049)
(2014.577,0.048)
(2014.596,0.049)
(2014.615,0.049)
(2014.635,0.05)
(2014.654,0.05)
(2014.673,0.049)
(2014.692,0.049)
(2014.712,0.049)
(2014.731,0.048)
(2014.75,0.048)
(2014.769,0.048)
(2014.788,0.048)
(2014.808,0.047)
(2014.827,0.048)
(2014.846,0.048)
(2014.865,0.048)
(2014.885,0.047)
(2014.904,0.048)
(2014.923,0.047)
(2014.942,0.047)
(2014.962,0.048)
(2014.981,0.048)
(2015.0,0.048)
}

\title{Chickenpox Cases in Hungary: a Benchmark Dataset for Spatiotemporal Signal Processing with Graph Neural Networks}
\author{Benedek Rozemberczki}
\affiliation{
  \institution{The University of Edinburgh}
  \country{United Kingdom}
}
\email{benedek.rozemberczki@ed.ac.uk}

\author{Paul Scherer}
\affiliation{
  \institution{University of Cambridge}
  \country{United Kingdom}
}
\email{pms69@cam.ac.uk}

\author{Oliver Kiss}
\affiliation{
  \institution{Central European University}
  \country{Hungary}
}
\email{kiss_oliver@phd.ceu.edu}

\author{Rik Sarkar}
\affiliation{
  \institution{The University of Edinburgh}
  \country{United Kingdom}
}
\email{rsarkar@inf.ed.ac.uk}

\author{Tamas Ferenci}
\affiliation{
  \institution{Obuda University; Corvinus University of Budapest}
  \country{Hungary}
}
\email{ferenci.tamas@nik.uni-obuda.hu}

\acmConference[WWW'21]{TheWebConf'21: Graph Learning Benchmarks Workshop}{April 19--23, 2021}{Ljubljana, SL}
\acmBooktitle{WWW'21: Graph Learning Benchmarks Workshop, 2021, Ljubljana, SL}
\begin{document}

\begin{abstract}
Recurrent graph convolutional neural networks are highly effective machine learning techniques for spatiotemporal signal processing. Newly proposed graph neural network architectures are repetitively evaluated on standard tasks such as traffic or weather forecasting. In this paper, we propose the \textit{Chickenpox Cases in Hungary} dataset as a new dataset for comparing graph neural network architectures. Our time series analysis and forecasting experiments demonstrate that the \textit{Chickenpox Cases in Hungary} dataset is adequate for comparing the predictive performance and forecasting capabilities of novel recurrent graph neural network architectures.
\end{abstract}
\maketitle
\section{Introduction}
Forecasting future edge and vertex attributes using spatial structure and historical values of the node and link attributes is a fundamental research problem for spatiotemporal machine learning. Recurrent graph neural networks can elegantly solve such spatiotemporal signal tasks with high predictive performance by training a graph convolutional neural network which is integrated or stacked with a recurrent neural network layer \cite{gconvlstm,evolvegcn}. These machine learning techniques have favorable practical characteristics such as online training and models that are transferable across graphs \cite{bojchevski2020pprgo,pytorch_geometric,kipf2017semi}. Hence, finding real world problems on which the forecasting performance of these architectures can be tested is crucial for fostering temporal graph representation learning research. However, recurrent graph neural networks are often iteratively evaluated and compared using the same overutilized datasets from a restricted number of application domains such as urban traffic and weather forecasting \cite{yu2018spatio,li2018diffusion}. These challenges related to the public availability of suitable and relevant spatiotemporal benchmark datasets are the main motivations of the present work.

\textbf{Present work.} In the pursuit of advancing temporal graph neural network research we publicly release the \textit{Chickenpox Cases in Hungary} dataset: a multivariate time series of weekly reported chickenpox cases in Hungarian counties. By utilizing this novel epidemiological dataset, the forecasting capabilities of newly proposed graph neural network models can be quantified. The intrinsic statistical characteristics of the dataset such as seasonality, spatial and temporal autocorrelation, zero inflation, heteroskedasticity and structural changes make the forecasting a challenging machine learning task. Our contribution also opens up venues to assess the predictive performance of existing spatiotemporal models.

\textbf{Main contributions.} The major results and contributions presented in our work can be summed up as follows:
\begin{enumerate}
    \item We release \textit{Chickenpox Cases in Hungary}, a novel spatiotemporal dataset which can be used to benchmark the forecasting performance of graph neural network architectures.
    \item We conduct a descriptive analysis of the time series and discuss the particular spatiotemporal modeling challenges that the dataset poses.
    \item We assess the performance of existing recurrent graph neural network architectures on multiple forecasting horizons.
\end{enumerate}
The remainder of this paper has the following structure. In Section \ref{sec:related_work} we overview the related literature about chickenpox and parametric spatiotemporal statistical models. In Section \ref{sec:descriptives} we carry out a descriptive analysis of the dataset, while Section \ref{sec:challenges} discusses the potential modeling challenges. We present predictive performance benchmarks on the dataset in Section \ref{sec:benchmarks} and the paper concludes with Section \ref{sec:conclusions}. The spatial adjacency matrix and the county level time series are available at \url{https://github.com/benedekrozemberczki/spatiotemporal_datasets}.
\section{Related Work}\label{sec:related_work}
In this section we give a brief overview of related work about the epidemiology and characteristics of chickenpox and the design of recurrent graph neural network architectures.

\begin{figure*}[h!]
\centering
\begin{tikzpicture}
	\tikzset{font={\fontsize{7pt}{12}\selectfont}}
\begin{groupplot}[group style={
                      group name=myplot,
                      group size= 5 by 4, horizontal sep=0.3cm,vertical sep=0.65cm},height=2.5cm,width=4.6cm, ymin=-10,ymax=110,ytick={0,20,40,60,80,100},title style={at={(0.5,0.85)},anchor=south},every axis x label/.style={at={(axis description cs:0.5,-0.4)},anchor=north},]
\nextgroupplot[
    /pgf/number format/.cd,
        use comma,
        1000 sep={},
 	title = \textbf{Bacs},
	ylabel=Cases,
	xtick={2005,2010,2015},
	xmin=2004.5,
	xmax=2015.5,
	xticklabels={,,},
	ymin=-50,
	ymax=500,
	ytick={0,250,500},
]
\addplot [mark=none, thick,blue]coordinates {
\BACSDATA

};

\nextgroupplot[
    /pgf/number format/.cd,
        use comma,
        1000 sep={},
 	title = \textbf{Baranya},
	xtick={2005,2010,2015},
	xmin=2004.5,
	xmax=2015.5,
	ymin=-50,
	xticklabels={,,},
	ymax=500,
	ytick={0,250,500},
	yticklabels={,,},
]
\addplot [mark=none, thick,blue]coordinates {
\BARANYADATA
};

\nextgroupplot[
    /pgf/number format/.cd,
        use comma,
        1000 sep={},
 	title = \textbf{Bekes},
	xtick={2005,2010,2015},
	xmin=2004.5,
	xmax=2015.5,
	xticklabels={,,},
	ymin=-50,
	ymax=500,
	ytick={0,250,500},
	yticklabels={,,},
]
\addplot [mark=none, thick,blue]coordinates {
\BEKESDATA

};

\nextgroupplot[
    /pgf/number format/.cd,
        use comma,
        1000 sep={},
 	title = \textbf{Borsod},
	xtick={2005,2010,2015},
	xmin=2004.5,
	xmax=2015.5,
	xticklabels={,,},
	ymin=-50,
	ymax=500,
	ytick={0,250,500},
	yticklabels={,,},
]
\addplot [mark=none, thick,blue]coordinates {
\BORSODDATA
};

\nextgroupplot[
    /pgf/number format/.cd,
        use comma,
        1000 sep={},
 	title = \textbf{Budapest},
	xtick={2005,2010,2015},
	xmin=2004.5,
	xticklabels={,,},
	xmax=2015.5,
	ymin=-50,
	ymax=500,
	ytick={0,250,500},
	yticklabels={,,},
]
\addplot [mark=none, thick,blue]coordinates {
\BUDAPESTDATA

};
%%%%%%%%%%%%%%%%%%%%%%%%%%%%%%%%%%%%
\nextgroupplot[
    /pgf/number format/.cd,
        use comma,
        1000 sep={},
 	title = \textbf{Csongrad},
 	ylabel=Cases,
	xtick={2005,2010,2015},
	xticklabels={,,},
	xmin=2004.5,
	xmax=2015.5,
	ymin=-50,
	ymax=500,
	ytick={0,250,500},
]
\addplot [mark=none, thick,blue]coordinates {
\CSONGRADDATA
};

\nextgroupplot[
    /pgf/number format/.cd,
        use comma,
        1000 sep={},
 	title = \textbf{Fejer},
	xtick={2005,2010,2015},
		xticklabels={,,},
	xmin=2004.5,
	xmax=2015.5,
	ymin=-50,
	ymax=500,
	ytick={0,250,500},
	yticklabels={,,},
]
\addplot [mark=none, thick,blue]coordinates {
\FEJERDATA

};

\nextgroupplot[
    /pgf/number format/.cd,
        use comma,
        1000 sep={},
 	title = \textbf{Gyor},
	xtick={2005,2010,2015},
	xmin=2004.5,
	xmax=2015.5,
	xticklabels={,,},
	ymin=-50,
	ymax=500,
	ytick={0,250,500},
	yticklabels={,,},
]
\addplot [mark=none, thick,blue]coordinates {
\GYORDATA
};

\nextgroupplot[
    /pgf/number format/.cd,
        use comma,
        1000 sep={},
 	title = \textbf{Hajdu},
	xtick={2005,2010,2015},
	xmin=2004.5,
	xmax=2015.5,
	xticklabels={,,},
	ymin=-50,
	ymax=500,
	ytick={0,250,500},
	yticklabels={,,},
]
\addplot [mark=none, thick,blue]coordinates {
\HAJDUDATA

};

\nextgroupplot[
    /pgf/number format/.cd,
        use comma,
        1000 sep={},
 	title = \textbf{Heves},
	xtick={2005,2010,2015},
	xmin=2004.5,
	xmax=2015.5,
	ymin=-50,
	ymax=500,
	xticklabels={,,},
	ytick={0,250,500},
	yticklabels={,,},
]
\addplot [mark=none, thick,blue]coordinates {
\HEVESDATA
};
%%%%%%%%%%%%%%%%%%%%%%%%%%%%%%%%%%%%%%%%5
\nextgroupplot[
    /pgf/number format/.cd,
        use comma,
        1000 sep={},
 	title = \textbf{Jasz},
	ylabel=Cases,
	xtick={2005,2010,2015},
	xmin=2004.5,
	xticklabels={,,},
	xmax=2015.5,
	ymin=-50,
	ymax=500,
	ytick={0,250,500},
]
\addplot [mark=none, thick,blue]coordinates {
\JASZDATA

};

\nextgroupplot[
    /pgf/number format/.cd,
        use comma,
        1000 sep={},
 	title = \textbf{Komarom},
	xtick={2005,2010,2015},
	xmin=2004.5,
	xmax=2015.5,
	ymin=-50,
	xticklabels={,,},
	ymax=500,
	ytick={0,250,500},
	yticklabels={,,},
]
\addplot [mark=none, thick,blue]coordinates {
\KOMAROMDATA
};

\nextgroupplot[
    /pgf/number format/.cd,
        use comma,
        1000 sep={},
 	title = \textbf{Nograd},
	ytick={0,250,500},
	yticklabels={,,},
	xtick={2005,2010,2015},
	xmin=2004.5,
	xmax=2015.5,
	xticklabels={,,},
	ymin=-50,
	ymax=500,
]
\addplot [mark=none, thick,blue]coordinates {
\NOGRADDATA

};

\nextgroupplot[
    /pgf/number format/.cd,
        use comma,
        1000 sep={},
 	title = \textbf{Pest},
	xtick={2005,2010,2015},
	xmin=2004.5,
	xticklabels={,,},
	xmax=2015.5,
	ymin=-50,
	ymax=500,
	ytick={0,250,500},
	yticklabels={,,},
]
\addplot [mark=none, thick,blue]coordinates {
\PESTDATA
};

\nextgroupplot[
    /pgf/number format/.cd,
        use comma,
        1000 sep={},
 	title = \textbf{Somogy},
	xticklabels={,,},
	xtick={2005,2010,2015},
	xmin=2004.5,
	xmax=2015.5,
	ymin=-50,
	ymax=500,
	ytick={0,250,500},
	yticklabels={,,},
]
\addplot [mark=none, thick,blue]coordinates {
\SOMOGYDATA

};
%%%%%%%%%%%%%%%%%%%%%%%5
\nextgroupplot[
    /pgf/number format/.cd,
        use comma,
        1000 sep={},
 	title = \textbf{Szabolcs},
	xtick={2005,2010,2015},
	xmin=2004.5,
	ylabel=Cases,
	xmax=2015.5,
	ymin=-50,
	ymax=500,
	xlabel=Year,
	ytick={0,250,500},
]
\addplot [mark=none, thick,blue]coordinates {
\SZABOLCSDATA
};

\nextgroupplot[
    /pgf/number format/.cd,
        use comma,
        1000 sep={},
 	title = \textbf{Tolna},
	xlabel=Year,
	xtick={2005,2010,2015},
	xmin=2004.5,
	xmax=2015.5,
	ymin=-50,
	ymax=500,
	ytick={0,250,500},
	yticklabels={,,},
]
\addplot [mark=none, thick,blue]coordinates {
\TOLNADATA

};

\nextgroupplot[
    /pgf/number format/.cd,
        use comma,
        1000 sep={},
 	title = \textbf{Vas},
	xtick={2005,2010,2015},
	xmin=2004.5,
	xmax=2015.5,
	ymin=-50,
	xlabel=Year,
	ymax=500,
	ytick={0,250,500},
	yticklabels={,,},
]
\addplot [mark=none, thick,blue]coordinates {
\VASDATA
};

\nextgroupplot[
    /pgf/number format/.cd,
        use comma,
        1000 sep={},
 	title = \textbf{Veszprem},
	xlabel=Year,
	xtick={2005,2010,2015},
	xmin=2004.5,
	xmax=2015.5,
	ymin=-50,
	ymax=500,
	ytick={0,250,500},
	yticklabels={,,},
]
 \addplot [mark=none, thick,blue]coordinates {
\VESZPREMDATA
};

\nextgroupplot[
    /pgf/number format/.cd,
        use comma,
        1000 sep={},
 	title = \textbf{Zala},
	xlabel=Year,
	xtick={2005,2010,2015},
	xmin=2004.5,
	xmax=2015.5,
	ymin=-50,
	ymax=500,
	ytick={0,250,500},
	yticklabels={,,},
]
\addplot [mark=none, thick,blue]coordinates {
\ZALADATA

};

\end{groupplot}
\end{tikzpicture}

\caption{The weekly number of chickenpox cases in Hungarian counties and the capital between 2005 and 2015.}\label{fig:time_series}

\end{figure*}
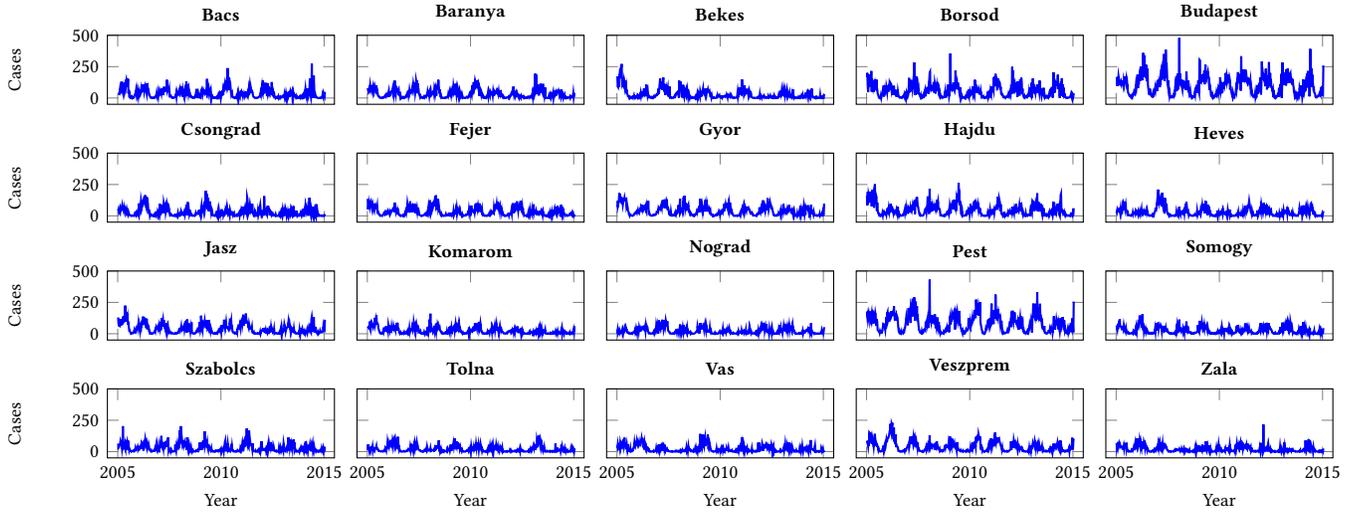

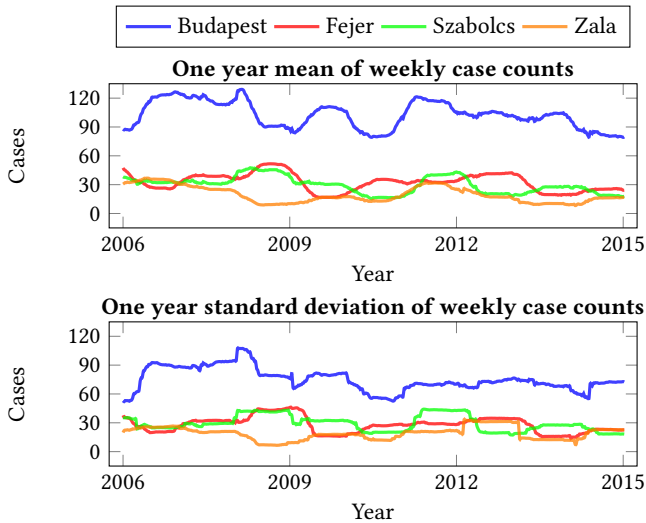
\begin{figure}[h!]
\centering
\begin{tikzpicture}
	\tikzset{font={\fontsize{9pt}{12}\selectfont}}
\begin{groupplot}[group style={
                      group name=myplot,
                      group size= 1 by 2, horizontal sep=0.3cm,vertical sep=1.25cm},height=3.5cm,width=8.6cm, ymin=-10,ymax=110,ytick={0,20,40,60,80,100},title style={at={(0.5,0.85)},anchor=south},every axis x label/.style={at={(axis description cs:0.5,-0.2)},anchor=north},]
\nextgroupplot[
    /pgf/number format/.cd,
        use comma,
        1000 sep={},
     	legend columns=4,
	legend style={at={(0.5,1.25)},anchor=south},
    legend entries={Budapest,Fejer,Szabolcs,Zala},
 	title = \textbf{One year mean of weekly case counts},
	xlabel=Year,
	xtick={2006,2009,2012,2015},
	ylabel=Cases,
	xmin=2005.75,
	xmax=2015.25,
	ymin=-15,
	ymax=135,
	ytick={0,30,60,90,120},
	ylabel=Cases,
	yticklabels={0,30,60,90,120},
]
%\addplot [mark=none, very thick,blue,smooth,opacity=0.5]coordinates {\MEANBACS};
%\addplot [mark=none, very thick,blue,smooth,opacity=0.5]coordinates {\MEANBARANYA};
%\addplot [mark=none, very thick,blue,smooth,opacity=0.5]coordinates {\MEANBEKES};
%\addplot [mark=none, very thick,blue,smooth,opacity=0.5]coordinates {\MEANBORSOD};
\addplot [mark=none, very thick,blue,smooth,opacity=0.75]coordinates {\MEANBUDAPEST};

%\addplot [mark=none, very thick,blue,smooth,opacity=0.5]coordinates {\MEANCSONGRAD};
\addplot [mark=none, very thick,red,smooth,opacity=0.75]coordinates {\MEANFEJER};
%\addplot [mark=none, very thick,blue,smooth,opacity=0.5]coordinates {\MEANGYOR};
%\addplot [mark=none, very thick,blue,smooth,opacity=0.5]coordinates {\MEANHAJDU};
%\addplot [mark=none, very thick,blue,smooth,opacity=0.5]coordinates {\MEANHEVES};

%\addplot [mark=none, very thick,blue,smooth,opacity=0.5]coordinates {\MEANJASZ};
%\addplot [mark=none, very thick,blue,smooth,opacity=0.5]coordinates {\MEANKOMAROM};
%\addplot [mark=none, very thick,blue,smooth,opacity=0.5]coordinates {\MEANNOGRAD};
%\addplot [mark=none, very thick,blue,smooth,opacity=0.5]coordinates {\MEANPEST};
%\addplot [mark=none, very thick,blue,smooth,opacity=0.5]coordinates {\MEANSOMOGY};

\addplot [mark=none, very thick,green,smooth,opacity=0.75]coordinates {\MEANSZABOLCS};
%\addplot [mark=none, very thick,blue,smooth,opacity=0.5]coordinates {\MEANTOLNA};
%\addplot [mark=none, very thick,blue,smooth,opacity=0.5]coordinates {\MEANVAS};
%\addplot [mark=none, very thick,blue,smooth,opacity=0.5]coordinates {\MEANVESZPREM};
\addplot [mark=none, very thick,orange,smooth,opacity=0.75]coordinates {\MEANZALA};

\nextgroupplot[
    /pgf/number format/.cd,
        use comma,
        1000 sep={},
    legend style = { column sep = 10pt, legend columns = 2, legend to name = grouplegend},
 	title = \textbf{One year standard deviation of weekly case counts},
	xlabel=Year,
	xtick={2006,2009,2012,2015},
	xmin=2005.75,
	xmax=2015.25,
	ymin=-15,
	ymax=135,
	ytick={0,30,60,90,120},
	ylabel=Cases,
	yticklabels={0,30,60,90,120},
]

\addplot [mark=none, very thick,blue,smooth,opacity=0.75]coordinates {\STDBUDAPEST};\addlegendentry{Budapest}
\addplot [mark=none, very thick,red,smooth,opacity=0.75]coordinates {\STDFEJER};
\addplot [mark=none, very thick,green,smooth,opacity=0.75]coordinates {\STDSZABOLCS};
\addplot [mark=none, very thick,orange,smooth,opacity=0.75]coordinates {\STDZALA};

\end{groupplot}
\end{tikzpicture}
\caption{One year running mean and standard deviation of weekly chickenpox cases in selected Hungarian counties.}\label{fig:running_average}
\end{figure}

\subsection{Chickenpox}
Chickenpox or varicella is a highly contagious airborne disease caused by the varicella zoster virus (VZV) \cite{Arvin361}. By the age of 20, more than 90 percent of the population is exposed to the VZV in developed countries \cite{Choo}. While chickenpox might produce common early symptoms such as headache or nausea, its onset is characterized by the rapid appearance of an easily distinguishable skin rash \cite{mccrary}. Although vaccines against VZV infection are available \cite{Seward}, only a handful of countries have national immunization programme \cite{flatt}. In Hungary there is no mandatory vaccination against chickenpox but vaccines are available and are routinely recommended to parents. Physicians have to report each case to the local centre of epidemiology which are then aggregated and publicly presented weekly for each of the 20 counties of Hungary, resulting in an ideal data collection environment from the modeling perspective \cite{karsai}.

\subsection{Spatiotemporal neural models}
Spatiotemporal neural models are a family of parametric statistical models which can handle data that has distinct time and spatial dimensions e.g. traffic measurements, regional epidemiological reporting or weather. The specific \textit{recurrent graph neural network models} compared in our work fuse ideas from the design of graph convolutional neural network layers \cite{kipf2017semi, gat_iclr18, rozemberczki2021pathfinder, ppnp_iclr19, bojchevski2020pprgo} and recurrent neural networks \cite{lstm, gru}. These models operate on temporal sequences of spatial data; at each time step a graph neural network layer convolves the input features or hidden states of the recurrent unit. Recurrent and graph convolutional layers are trained jointly on a downstream task and the design of these architectures requires the choice of a graph neural network and a recurrent unit. Popular choices for graph neural networks are spectral graph convolutions \cite{kipf2017semi} and graph attention networks \cite{gat_iclr18} while the most frequently augmented recurrent neural networks include long short-term memory cells \cite{lstm} and gated recurrent units \cite{gru}.

\section{Characteristics of the Dataset}\label{sec:descriptives}
Our main contribution is the release of the \textit{Chickenpox Cases in Hungary} dataset which consists of county level time series and a spatial graph which describes the spatial connectivity of the counties. The county level time series describe the weekly number of chickenpox cases reported by general practitioners in Hungary.

We manually collected the time series by collating the reported case counts from the digital version of the \textit{Hungarian Epidemiological Info}\footnote{\url{http://www.oek.hu/}}, a weekly bulletin of morbidity and mortality of infectious diseases in Hungary. Our data collection covered the weeks between the January of 2005 and January of 2015 and the resulting time series has more than 500 entries for all of the counties without any missingness. The underlying spatial graph has 20 vertices (19 counties and the capital Budapest) and there are 61 edges between the nodes. We plotted the county level reported case count time series on Figure \ref{fig:time_series}.

\textbf{Main characteristics of the time series.} Looking at the time series on Figure \ref{fig:time_series} we can make multiple important observations. These are the following:

\begin{itemize}
    \item The number of reported cases is population dependent; spatial units with more inhabitants such as the capital Budapest report more cases on average.
    \item The time series all exhibit strong seasonality which can be a result of weather conditions or the periodicity of the school year.
    \item A large number of counties report no new cases in the summer months -- these county level time series are zero inflated.
\end{itemize}

\subsection{Structural changes and trends}
The county level time series are noisy and exhibit a strong yearly seasonality, due to this, we cannot theorize whether there are structural changes without correcting the seasonality. We calculated the 52 weeks running average and standard deviation of the weekly case counts for selected counties and the capital and plotted the resulting times series on Figure \ref{fig:running_average}.
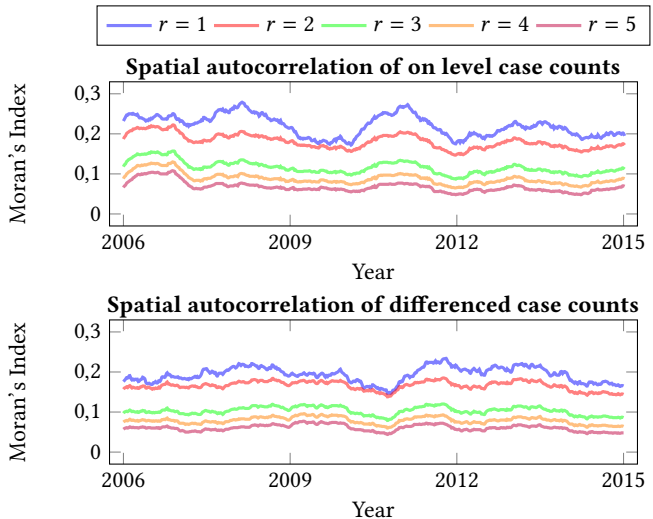
\begin{figure}[h!]
\centering
\begin{tikzpicture}
	\tikzset{font={\fontsize{9pt}{12}\selectfont}}
\begin{groupplot}[group style={
                      group name=myplot,
                      group size= 1 by 2, horizontal sep=0.3cm,vertical sep=1.25cm},height=3.5cm,width=8.6cm, ymin=-10,ymax=110,ytick={0,20,40,60,80,100},title style={at={(0.5,0.85)},anchor=south},every axis x label/.style={at={(axis description cs:0.5,-0.2)},anchor=north},]
\nextgroupplot[
    /pgf/number format/.cd,
        use comma,
        1000 sep={},
     	legend columns=5,
	legend style={at={(0.5,1.25)},anchor=south},
    legend entries={$r=1$,$r=2$,$r=3$,$r=4$,$r=5$},
 	title = \textbf{Spatial autocorrelation of on level case counts},
	xlabel=Year,
	xtick={2006,2009,2012,2015},
	ylabel=Cases,
	xmin=2005.75,
	xmax=2015.25,
	ymin=-0.03,
	ymax=0.33,
	ytick={0,0.1,0.2,0.3},
	ylabel=Moran's Index,
]

\addplot [mark=none, very thick,blue,smooth,opacity=0.5]coordinates {\LEVELONE};
\addplot [mark=none, very thick,red,smooth,opacity=0.5]coordinates {\LEVELTWO};
\addplot [mark=none, very thick,green,smooth,opacity=0.5]coordinates {\LEVELTHREE};
\addplot [mark=none, very thick,orange,smooth,opacity=0.5]coordinates {\LEVELFOUR};
\addplot [mark=none, very thick,purple,smooth,opacity=0.5]coordinates {\LEVELFIVE};

\nextgroupplot[
    /pgf/number format/.cd,
        use comma,
        1000 sep={},
    legend style = { column sep = 10pt, legend columns = 2, legend to name = grouplegend},
 	title = \textbf{Spatial autocorrelation of differenced case counts},
	xlabel=Year,
	xtick={2006,2009,2012,2015},
	ylabel=Cases,
	xmin=2005.75,
	xmax=2015.25,
	ymin=-0.03,
	ymax=0.33,
	ytick={0,0.1,0.2,0.3},
	ylabel=Moran's Index,
]

\addplot [mark=none, very thick,blue,smooth,opacity=0.5]coordinates {\DIFFERENCEONE};
\addplot [mark=none, very thick,red,smooth,opacity=0.5]coordinates {\DIFFERENCETWO};
\addplot [mark=none, very thick,green,smooth,opacity=0.5]coordinates {\DIFFERENCETHREE};
\addplot [mark=none, very thick,orange,smooth,opacity=0.5]coordinates {\DIFFERENCEFOUR};
\addplot [mark=none, very thick,purple,smooth,opacity=0.5]coordinates {\DIFFERENCEFIVE};

\end{groupplot}
\end{tikzpicture}

\caption{The truncated random walk weighted spatial autocorrelation of the on level and first-order differenced chickenpox case count time series.}\label{fig:spatial_autocorrelation}

\end{figure}

\textbf{Main findings.} We observe multiple statistical phenomena which pose modelling challenges. These can be summarized as:
\begin{itemize}
    \item The yearly rolling average of the county level mean case count is not constant. This can be an artifact of population shift or the result of intrinsic epidemiological dynamics.
    \item Standard deviations of the time series change with time which means that the models have to take into consideration the time dependent variation in the outcome variable.
\end{itemize}

\subsection{Measuring spatial autocorrelation}
We quantify the spatial autocorrelation of chickenpox cases by using a truncated random walk weighted variant of Moran's I index. Given an unweighted and undirected graph $G=(V,E)$ let us denote the adjacency matrix by $\textbf{A}$ and the diagonal degree matrix as $\textbf{D}$. The row normalized adjacency matrix is defined as $\widehat{\textbf{A}} =\textbf{D}^{-1}\textbf{A}$ and the transition probability matrix of an $r$-length truncated random walk equals to $\widehat{\textbf{A}}^r$. Correspondingly the truncated random walk weighted spatial autocorrelation index \cite{moran1950notes} of the node feature vector $\textbf{x}\in \mathbb{R}^{|V|}$ at scale $r$ is defined by Equation \eqref{eq:moran}. Here $\overline{\textbf{x}}$ is the average of the generic vertex feature and $u,v \in V$ are vertices.
\begin{align}
    \mathcal{I}&=\frac{|V|}{\sum_{v\in V} \sum_{u \in V} \widehat{\textbf{A}}^r_{u,v}}\frac{\sum_{v\in V} \sum_{u \in V} \widehat{\textbf{A}}^r_{u,v}(\textbf{x}_v- \overline{\textbf{x}}) (\textbf{x}_u-\overline{\textbf{x}})}{\sum_{v \in V} (\textbf{x}_v-\overline{\textbf{x}})^2}\label{eq:moran}
\end{align}

We plotted on Figure \ref{fig:spatial_autocorrelation} the spatial autocorrelation index using the first 5 proximity scales for the on level case count and first-order differenced case number time series. All of the county level time series were centralized to be 0 mean and standardized before the autocorrelation index computation.

\textbf{Main findings.} The most important empirical regularities that we can observe are the following:

\begin{itemize}
    \item Both the county level case count and differenced case count time series exhibit spatial autocorrelation through the years.
    \item The spatial autocorrelation is present at multiple scales but it decreases with increasing the distance being considered.
    \item The strength of spatial autocorrelation is not constant  -- there are visible trends in the spatial autocorrelation time series.
\end{itemize}
\section{The modeling Challenges}\label{sec:challenges}
The \textit{Chickenpox Cases in Hungary} dataset poses a number of machine learning modeling challenges for researchers. Based on our descriptive analysis of the time series these challenges can be briefly summarized as:
\begin{itemize}
    \item \textbf{Temporal autocorrelation.} The weekly number of new chickenpox cases is correlated with the case numbers from earlier weeks.
    \item \textbf{Spatial autocorrelation.} The number of newly infected children and the difference in the number of new cases are correlated across neighboring spatial units.
    \item \textbf{Heteroskedasticity.} The standard deviation of the county level time series is not constant over time.
    \item \textbf{Seasonality.} The county level count of chickenpox cases exhibit strong yearly seasonality. This can be an artifact of weather conditions or caused by the periodicity of the school year.
    \item \textbf{Multiple scales.} The Hungarian county system consist of spatial units which have a heterogeneous size. Budapest, the largest one, has nearly 10 times more inhabitants than Nograd which is the least populated one.
    \item \textbf{Count data.} The target variables describe the weekly count of chickenpox cases. The design of the graph neural network has to take this fact into account: particularly when it comes to the choice of loss function and the activation functions in the output layer.
    \item \textbf{Zero inflation.} Certain smaller counties report no cases during the weeks of the summer in a randomly dispersed manner, which causes challenges for traditional count data modeling.
    \item \textbf{Structural changes and random events.} The one and half decade long time horizon of the dataset gives plenty of space for population shift and years when the winter surge in chickenpox cases did not happen in certain counties.
\end{itemize}
Designing novel graph neural network architectures that can appropriately model a dataset with these statistical characteristics is a challenging task.
\section{Neural Benchmarks}\label{sec:benchmarks}
We tested the predictive performance of recurrent graph neural networks on county level chickenpox time series forecasting. Using the \textit{PyTorch Geometric Temporal} \cite{pytorch,pytorch_geometric,pytorch_geometric_temporal} implementation of the models we trained on the standardized chickenpox time series and predicted it for a fixed number of weeks ahead. The input graph describes the undirected direct adjacency relations of the counties.

All recurrent models used 8 temporal lags as input features and had 32 dimensional graph convolutional filters. The output of the convolutional layer was fed to a feedforward output layer. Each model was trained for 200 epochs with the Adam optimizer \cite{kingma_adam_2014} and we used a learning rate of $10^{-2}$. In Table \ref{tab:chickenpox_performance} we present average mean squared error values for various forecasting horizons calculated from 10 experimental runs.

\begin{figure}[h!]
\caption{The average test mean squared error with standard deviations obtained over 10/20/40 weeks long forecasting horizons calculated from a 10 experimental runs. Bold numbers denote the best performing models.}\label{tab:chickenpox_performance}
{\centering
{\small
\begin{tabular}{lccc}
\cline{2-4}
          & \textbf{10 weeks} & \textbf{20 weeks} &  \textbf{40 weeks}\\ \hline
\textbf{GConvLSTM}  \cite{gconvlstm}  & $0.741\pm0.005$ &  $0.403\pm0.003$ &  $1.221\pm0.010$  \\
\textbf{GConvGRU}   \cite{gconvlstm}   &  $0.757\pm0.001$ &  $0.407\pm0.001$  &  $1.117\pm0.002$   \\
\textbf{Evolve GCN-O}      \cite{evolvegcn}   &$0.775\pm0.007$ &  $0.419\pm0.004$ &  $1.120\pm0.003$  \\
\textbf{Evolve GCN-H}   \cite{evolvegcn}      &$0.766\pm0.009$ &  $0.413\pm0.009$ &  $1.115\pm0.013$   \\
\textbf{DynGRAE}    \cite{dyggnn,dyngrae_1}     &$\mathbf{0.706\pm0.004}$ &  $\mathbf{0.382\pm0.002}$ &  $\mathbf{1.112\pm0.010}$  \\
\textbf{STGCN}    \cite{yu2018spatio}     &$0.763\pm0.008$ &  $0.405\pm0.007$ &  $1.118\pm0.005$   \\
\textbf{DCRNN}    \cite{li2018diffusion}     &$0.753\pm0.003$ &  $0.395\pm0.001$ &  $1.119\pm0.002$   \\ 
\hline
\end{tabular}}}
\end{figure}

\textbf{Main findings.} Based on the forecasting performance of graph neural networks we can make the following observations:
\begin{itemize}
    \item The DynGRAE \cite{dyngrae_1,dyggnn} architecture works best at forecasting horizon and the advantage is significant at $\alpha = 5\%$.
    \item There are considerable performance differences between models; see for example Evolve GCN-O and DynGRAE.
    \item When the forecasting horizon is increased the predictive performance of some models is worse than random.
\end{itemize}
\section{Conclusions}\label{sec:conclusions}

We introduced \textit{Chickenpox Cases in Hungary}, a longitudinal dataset for benchmarking the predictive performance of spatiotemporal graph neural network architectures. Our exploratory analysis highlighted the unique statistical characteristics of the dataset which make predicting the weekly number of cases a challenging task. We evaluated the forecasting capabilities of the state-of-the-art recurrent graph neural networks. Our findings demonstrate that the current design of graph neural networks is moderately well suited for solving this task.

\bibliographystyle{ACM-Reference-Format}
\bibliography{main}
\end{document}